\documentclass[]{fairmeta}
\usepackage{array}
\usepackage{mathtools}
\usepackage{algorithm}
\usepackage{algpseudocode}
\usepackage{makecell}
\usepackage[table]{xcolor} 
\usepackage{arydshln}
\usepackage{amssymb}
\newcolumntype{L}[1]{>{\raggedright\let\newline\\\arraybackslash\hspace{0pt}}m{#1}}
\newcolumntype{C}[1]{>{\centering\let\newline\\\arraybackslash\hspace{0pt}}m{#1}}
\newcolumntype{R}[1]{>{\raggedleft\let\newline\\\arraybackslash\hspace{0pt}}m{#1}}
\newcolumntype{Y}{>{\centering\arraybackslash}X}

\title{HiStream: Efficient High-Resolution Video Generation via Redundancy-Eliminated Streaming}

\author[1,2,*]{Haonan Qiu}
\author[1]{Shikun Liu}
\author[1]{Zijian Zhou}
\author[1]{Zhaochong An}
\author[1]{Weiming Ren}
\author[1]{Zhiheng Liu}
\author[1]{Jonas Schult}
\author[1]{Sen He}
\author[1]{Shoufa Chen}
\author[1]{Yuren Cong}
\author[1]{Tao Xiang}
\author[2,\dagger]{Ziwei Liu}
\author[1,\dagger]{Juan-Manuel Perez-Rua}

\affiliation[1]{Meta AI}
\affiliation[2]{Nanyang Technological University}

\contribution[*]{Work done at Meta}
\contribution[\dagger]{Joint last author}

\abstract{
High-resolution video generation, while crucial for digital media and film, is computationally bottlenecked by the quadratic complexity of diffusion models, making practical inference infeasible. To address this, we introduce \textbf{HiStream}, an efficient autoregressive framework that systematically reduces redundancy across three axes: 
i) \textbf{Spatial Compression}: denoising at low resolution before refining at high resolution with cached features; 
ii) \textbf{Temporal Compression}: a chunk-by-chunk strategy with a fixed-size anchor cache, ensuring stable inference speed; 
and iii) \textbf{Timestep Compression}: applying fewer denoising steps to subsequent, cache-conditioned chunks. 
On 1080p benchmarks, our primary HiStream model (i+ii) achieves state-of-the-art visual quality while demonstrating up to $76.2\times$ faster denoising compared to the Wan2.1 baseline and negligible quality loss. Our faster variant, \textbf{HiStream+}, applies all three optimizations (i+ii+iii), achieving a $\mathbf{107.5\times}$ acceleration over the baseline, offering a compelling trade-off between speed and quality, thereby making high-resolution video generation both practical and scalable.
}

\date{\today}
\metadata[Project page]{\url{http://haonanqiu.com/projects/HiStream.html}}

\begin{document}

\maketitle

\begin{figure}[t!]
\centering
\includegraphics[width=0.99\textwidth]{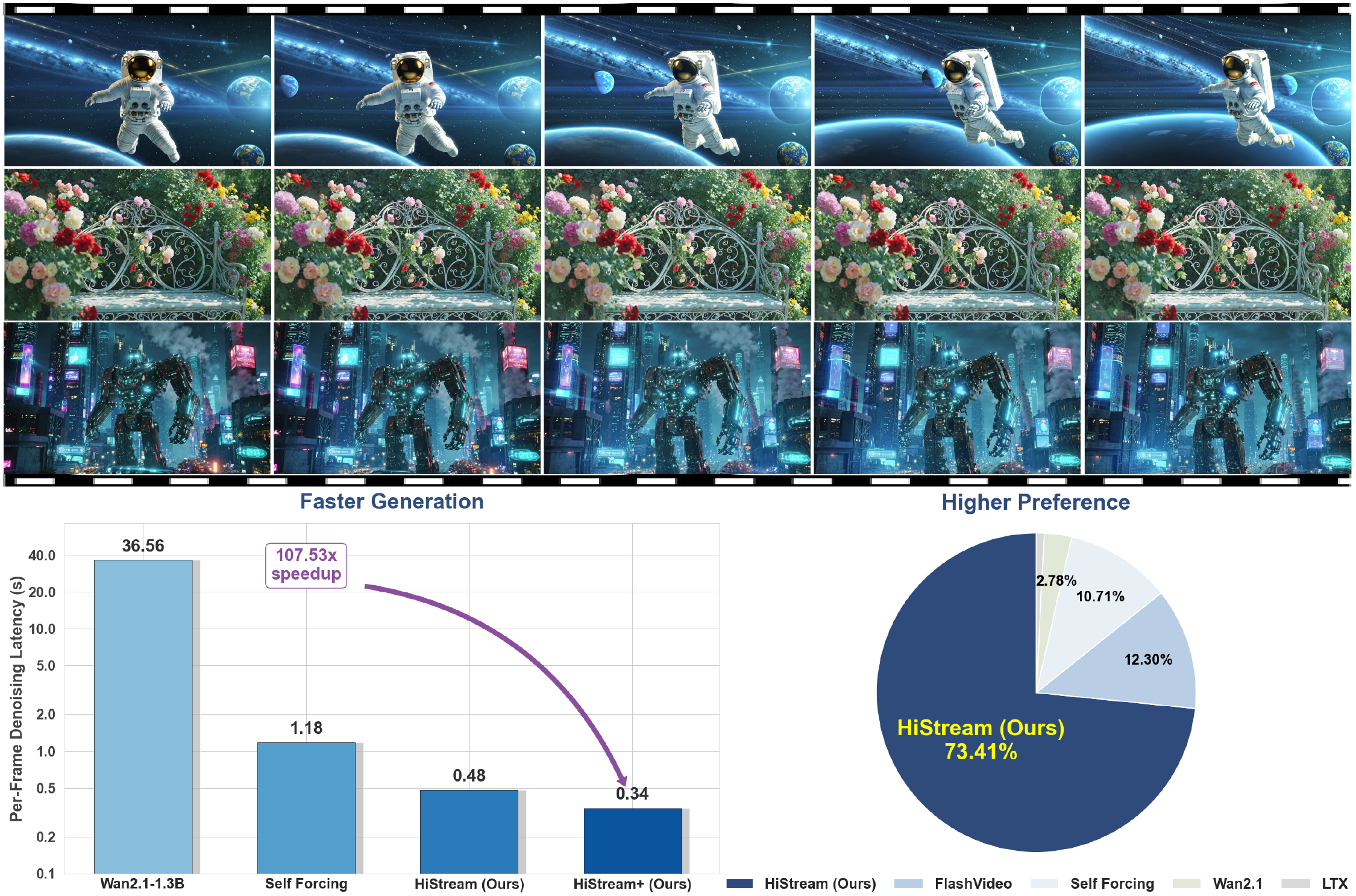}
\vspace{-0.5em}
\caption{
    HiStream delivers high-resolution (1080p) autoregressive video generation, offering up to $76.2\times$ faster denoising than the bi-directional baseline and $2.5\times$ faster than self-forcing, all while preserving the highest preference ratings. Furthermore, our faster variant, HiStream+, achieves a remarkable $107.5\times$ acceleration against the baseline and $3.5\times$ faster than self-forcing.
}
\vspace{-1.0em}
\label{fig_teaser}
\end{figure}

\section{Introduction}
\label{sec:intro}

With the rapid development of video diffusion models~\citep{blattmann2023align,yang2024cogvideox,chen2024gentron,HaCohen2024LTXVideo,kong2024hunyuanvideo,wan2025}, high-resolution video generation is increasingly essential for applications in film production, virtual reality, and digital media, where visual realism and fine-grained details are critical. However, generating high-fidelity and temporally coherent videos remains extremely challenging due to the substantial spatio-temporal complexity. Although recent high-resolution generation methods~\citep{zhang2025flashvideo,ren2025turbo2k,qiu2024freescale,qiu2025cinescale} have demonstrated impressive success, their computational efficiency is far from practical, as the inference cost scales quadratically based on the video spatial resolution and temporal length. As a result, the challenge of generating realistic 1080p videos in a time-efficient manner remains unsolved.

To address these computational bottlenecks, several strategies have been proposed. Sampling acceleration techniques, such as timestep distillation~\citep{wang2023videolcm,sauer2024adversarial,yin2024onestep,yin2024improved}, effectively reduce the number of denoising steps. However, the computational cost within each step remains high when applied to high-resolution videos. Concurrently, sparse or sliding attention mechanisms~\citep{zhang2025training,cai2025mixture,huang2025self} mitigate the full quadratic cost, but their cache and memory requirements still grow linearly with spatiotemporal volume. Thus, while these methods offer partial acceleration in low-resolution or short-video settings, they remain insufficient for scaling to long-duration, high-fidelity 1080p video generation.

We argue that the fundamental source of this inefficiency is inherent computational redundancy across \textbf{spatial, temporal, and timestep axes}. Spatially, significant computation is wasted in early denoising steps, which primarily establish coarse motion and structure~\citep{liu2025faster} and can be simply approximated at a low-resolution. High-resolution details are only recovered in later steps. This insight directly motivates our \textbf{Dual-Resolution Caching} mechanism. Temporally, similar redundancy exists in autoregressive models. We find that only the initial frame (acting as a persistent keyframe, akin to an attention sink~\citep{gu2024attention}) and a small set of neighboring frames are crucial for temporal consistency. This observation leads to our \textbf{Anchor-Guided Sliding Window} strategy. Finally, we observe that the denoising trajectory differs significantly between the initial and subsequent chunks, a phenomenon that motivates our \textbf{Asymmetric Denoising} Strategy.

In this work, we propose \textbf{HiStream}, an efficient diffusion autoregressive framework that integrates core insights to reduce spatio-temporal complexity. HiStream's design is built on two primary optimizations. First, its dual-resolution caching performs low-resolution denoising to capture global structure, then refines high-resolution details using cached feature states, slashing redundant computation. Second, its anchor-guided sliding window generates video chunk-wise, using an anchor keyframe and a small neighbor cache to maintain consistency. This design achieves a fixed-size attention cache, ensuring stable inference speed regardless of video length. As a third, optional optimization, we also introduce an asymmetric denoising strategy, where subsequent chunks can achieve high fidelity with as few as two denoising steps (one low-res, one high-res), enabling unprecedented efficiency.

Extensive experiments on 1080p video benchmarks demonstrate that our primary HiStream model (utilizing the first two optimizations) attains state-of-the-art visual quality with negligible quality loss, while offering up to $76.2\times$ faster denoising compared to the base video foundation model Wan2.1~\citep{wan2025}. Our faster variant, HiStream+, which additionally employs Asymmetric Denoising, pushes this acceleration to a remarkable $107.5\times$ over the baseline. This variant provides a powerful speed-quality trade-off, making HiStream a practical and scalable solution for both high-fidelity and ultra-fast high-resolution video generation.
\section{Related Work}
\label{sec:related}

\paragraph{Video Diffusion Models.}
Diffusion-based methods have recently emerged as the leading approach for video generation. VDM~\citep{vdm} first introduced diffusion models to this domain, establishing a foundation for subsequent research. Later works~\citep{blattmann2023align,lvdm,guo2023animatediff,chen2023videocrafter1,wang2023modelscope,wang2023lavie} built upon this framework with hierarchical latent designs, enabling the creation of longer videos with improved temporal consistency. However, these methods still rely on the UNet~\citep{ronneberger2015u} architecture, which suffers from limited scalability.
A recent shift towards Transformer-based architectures, specifically Diffusion Transformers (DiT)~\citep{peebles2023scalable}, has emerged in video generation. Models like CogVideoX~\citep{yang2024cogvideox} and Pyramid Flow~\citep{jin2024pyramidal} were among the first to demonstrate the superior scalability and strong performance of this approach. Leveraging the scalability of DiT, models like LTX~\citep{HaCohen2024LTXVideo}, Hunyuan~\citep{kong2024hunyuanvideo}, and Wan~\citep{wan2025} now generate visually realistic videos from text, surpassing prior UNet-based approaches.

\paragraph{High-Resolution Visual Generation.}
High resolution visual synthesis remains a long-standing challenge in the generative field due to the high-resolution data scarcity and the substantial computational cost of modeling, particularly for videos. Recent efforts have diverged into two main categories: 
(1) {\it Training-based} methods, which fine-tune on high-resolution data directly with large-size models~\citep{teng2023relay,hoogeboom2023simple,ren2024ultrapixel,liu2024linfusion,guo2024make,zheng2024any,cheng2024resadapter,zhang2025flashvideo,ren2025turbo2k}; and
(2) {\it Tuning-free} methods, which modify architecture to improve resolution without additional training data~\citep{haji2024elasticdiffusion,lin2024cutdiffusion,lee2023syncdiffusion,jin2023training,zhang2024hidiffusion,kim2024diffusehigh}.
While directly training on high-resolution data is a fundamental solution, the scarcity of such data and the immense memory requirements make it impractical. Consequently, tuning-free methods have also become a practical choice.

However, most tuning-free approaches~\citep{he2023scalecrafter,du2024demofusion,huang2024fouriscale,lin2024accdiffusion,liu2024hiprompt,qiu2024freescale} are designed for UNet-based models, focusing on mitigating receptive-field limitations that cause artifacts. DiT-based models face a distinct challenge, where misaligned positional encodings can cause blurriness at high resolutions. As foundational models shift to DiT architectures, tuning-free methods~\citep{du2024max,qiu2025cinescale} for them are crucial. A common alternative to native synthesis is a two-stage approach using external super-resolution models~\citep{wang2021real,zhou2024upscale,wang2025seedvr}. While computationally cheap, this pipeline often fails to render fine details, yielding less faithful visual quality.

\paragraph{Efficient Visual Generation.}
The slow inference speed of iterative diffusion models has spurred multiple efficiency efforts. The most direct approach is reducing denoising steps, evolving from simple caching of redundant timesteps~\citep{lv2024fastercache,liu2025timestep,liu2025reusing} to more powerful distillation techniques~\citep{luo2023latent,wang2023videolcm,sauer2024adversarial,yin2024onestep,yin2024improved}. Concurrently, other works have focused on reducing intra-network redundancy via sparsification~\citep{zhang2025packing,zhang2025training,cai2025mixture}. A more recent and relevant trend is the acceleration of temporal self-attention. By reformulating video diffusion models causally~\citep{teng2025magi,chen2025skyreels,yin2025slow,huang2025self}, these methods enable incremental inference using KV caching. This development mirrors discoveries in LLMs, namely the ``attention sink'' phenomenon~\citep{xiao2023efficient,gu2024attention,qiu2025gated}. This property was famously exploited by StreamingLLM~\citep{xiao2023efficient} to implement a local sliding-window attention with a persistent anchor token, achieving stable inference with a fixed-size KV cache. HiStream builds directly on this lineage of ideas, adapting them to the unique challenges of high-resolution video.
\section{Methodology}
\label{sec:method}

\begin{figure*}[t]
\centering
\begin{subfigure}{0.47\textwidth}
    \centering
    \includegraphics[width=\linewidth]{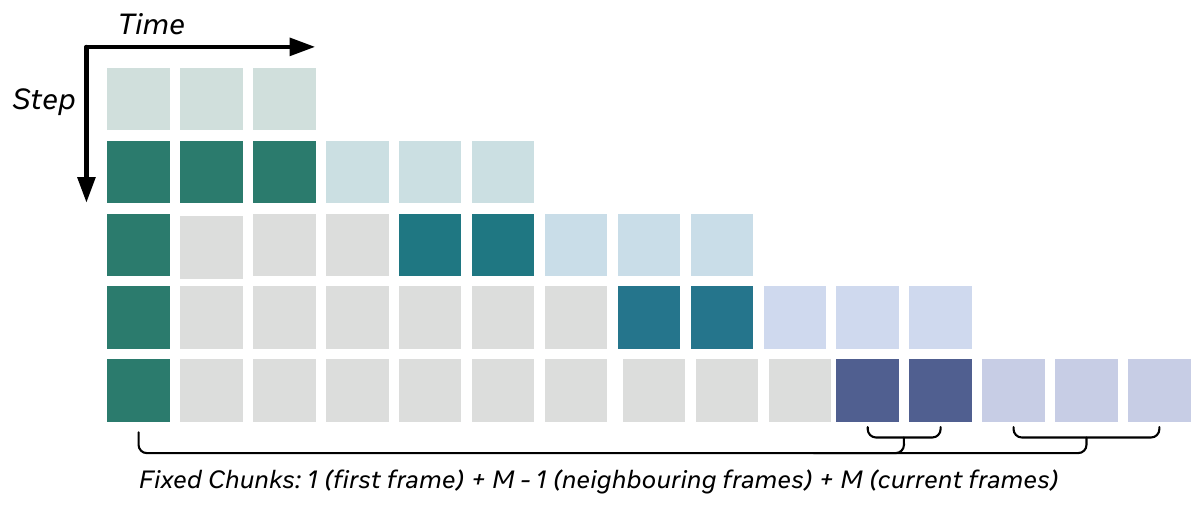}
    \caption{AR Inference with Anchor-Guided Sliding Window}
\end{subfigure}\hfill
\begin{subfigure}{0.53\textwidth}
    \centering
    \includegraphics[width=\linewidth]{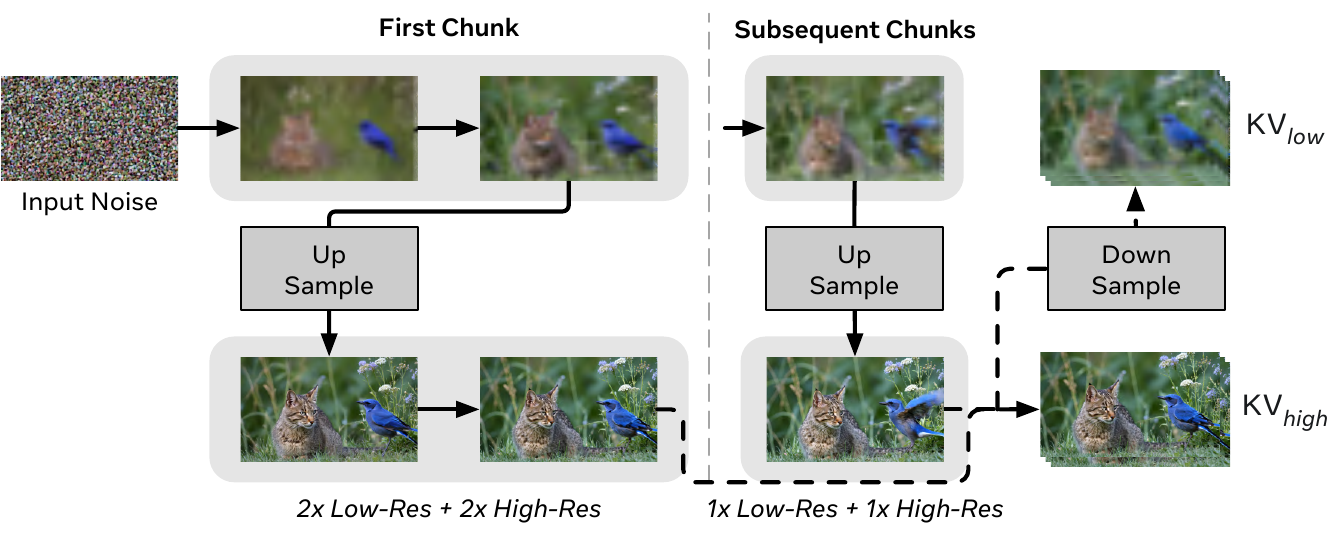}
    \caption{Dual-Resolution Caching with Asymmetric Denoising}
\end{subfigure}
\vspace{-2.0em}
\caption{\textbf{Pipeline details.} Illustration of the synergy among our three core efficiency mechanisms.
(\textbf{Left}) The \textbf{Anchor-Guided Sliding Window} strategy ensures robust temporal scalability by generating the video in fixed-size chunks. Each generation step maintains a constant attention context by combining tokens from the persistent content anchor (the first frame, dark green) and recent historical context (local frames, dark blue), thereby avoiding the growth of the KV cache over time.
(\textbf{Right}) The \textbf{Dual-Resolution Caching} accelerates synthesis by adopting a two-stage process: initial low-resolution denoising is followed by high-resolution refinement. Crucially, the final high-resolution output updates $\text{KV}_{\text{high}}$ and is subsequently downsampled to update $\text{KV}_{\text{low}}$, ensuring spatial consistency for subsequent chunks. With \textbf{Asymmetric Denoising}, subsequent chunks use half as many denoising steps as the first chunk.}
\vspace{-1.0em}
\label{fig:framework}
\end{figure*}

\begin{algorithm}[t]
\small
\caption{Autoregressive Inference with HiStream}
\label{alg:rolling_kv}
\begin{algorithmic}[1]  
\Require $\text{KV}_\text{low}$ and $\text{KV}_\text{high}$ caches
\Require Denoise timesteps $\{t_1, \ldots, t_T\}$
\Require Number of generated chunk $C_{N/M}$ with chunk size $M$
\Require AR diffusion model $G_\theta$ (returns $\text{KV}_\text{low}$ and $\text{KV}_\text{high}$ embeddings via $G_\theta^{\text{KV}}$)
\State \textbf{Initialize:} model output $\mathbf{X}_\theta \gets [\,]$
\State \textbf{Initialize:} KV caches $\text{KV}_\text{low} \gets [\,]$ and $\text{KV}_\text{high} \gets [\,]$
\For{$i = 1, \ldots, C_{N/M}$}
    \State Initialize $\mathbf{x}_T^i \sim \mathcal{N}(0, I)$
    \For{$j = T_\text{low}, \ldots, T_\text{high}+1$}
        \State Set $\hat{\mathbf{x}}_0^i \gets G_\theta(\mathbf{x}_{t_j}^i; t_j, \text{KV}_\text{low})$
        \If{$j = T_\text{high}+1$}
            \State $\hat{\mathbf{x}}_0^i \gets \text{Up}(\hat{\mathbf{x}}_0^i)$
        \EndIf
        \State Sample $\boldsymbol{\epsilon} \sim \mathcal{N}(0, I)$
        \State Set $\mathbf{x}_{t_{j-1}}^i \gets \Psi(\hat{\mathbf{x}}_0^i, \boldsymbol{\epsilon}, t_{j-1})$
    \EndFor
    \For{$j = T_\text{high}, \ldots, 1$}
        \State Set $\hat{\mathbf{x}}_0^i \gets G_\theta(\mathbf{x}_{t_j}^i; t_j, \text{KV}_\text{high})$
        \If{$j = 1$}
            \State $\mathbf{X}_\theta.\text{append}(\hat{\mathbf{x}}_0^i)$
            \State Cache $\mathbf{kv}^i_\text{high} \gets G_\theta^{\text{KV}}(\hat{\mathbf{x}}_0^i; 0, \text{KV}_\text{high})$
            \State Cache $\mathbf{kv}^i_\text{low} \gets G_\theta^{\text{KV}}(\text{Down}(\hat{\mathbf{x}}_0^i); 0, \text{KV}_\text{low})$
            \Comment{Dual-Resolution Caching}
            \If{$i = 1$}
                \State $\text{KV}_\text{high}\text{.append}(\mathbf{kv}^i_\text{high})$
                \State $\text{KV}_\text{low}\text{.append}(\mathbf{kv}^i_\text{low})$
            \Else
                \State $\text{KV}_{\text{high}} \gets \text{Concat} \left( \text{KV}_{\text{high}}[0], \mathbf{kv}_{\text{high}}^{i}[1:S] \right)$
                \State $\text{KV}_{\text{low}} \gets \text{Concat} \left( \text{KV}_{\text{low}}[0], \mathbf{kv}_{\text{low}}^{i}[1:S] \right)$
                \Comment{Anchor-Guided Sliding Window}
            \EndIf
        \Else
            \State Sample $\boldsymbol{\epsilon} \sim \mathcal{N}(0, I)$
            \State Set $\mathbf{x}_{t_{j-1}}^i \gets \Psi(\hat{\mathbf{x}}_0^i, \boldsymbol{\epsilon}, t_{j-1})$
        \EndIf
    \EndFor
\EndFor
\State \Return $\mathbf{X}_\theta$
\end{algorithmic}
\end{algorithm}

We present HiStream, an efficient high-resolution video generation framework designed to overcome the computational bottlenecks of existing diffusion models. Our complete pipeline is illustrated in Figure~\ref{fig:framework}.

\subsection{Preliminaries}
\label{subsec:preliminary}

\paragraph{Autoregressive Video Diffusion Models} Our HiStream framework is an optimization of an autoregressive video diffusion model. This baseline architecture models the joint distribution of $N$ video frames $\mathbf{x}^{1:N}$ by factorizing it autoregressively: $p(\mathbf{x}^{1:N}) = \prod_{i=1}^{N} p(\mathbf{x}^i \mid \mathbf{x}^{<i})$. Generation is a sequential process where each frame $\mathbf{x}^i$ is denoised from Gaussian noise, conditioned on the preceding frames $\mathbf{x}^{<i}$. In this work, our framework generates video in chunks of frames, rather than processing single frames individually. Our baseline WAN-2.1 is a Diffusion Transformer (DiT)~\citep{peebles2023scalable} that operates in the latent space of a causal 3D VAE~\citep{wan2025}. The autoregressive factorization is implemented efficiently within the Transformer using causal attention masks, which restrict the model to only attend to past and current frames. This model is trained using the standard frame-wise diffusion loss, minimizing the MSE between the predicted noise $\hat{\mathbf{\epsilon}}_{\theta}$ and the true sampled noise $\mathbf{\epsilon}$:
\begin{align}
\mathcal{L}_{\text{DM}}({\theta}) = \mathbb{E}_{\mathbf{x}, t, \mathbf{\epsilon}} \left[ \| \hat{\mathbf{\epsilon}}_{\theta}(\mathbf{x}_{t}, t, \mathbf{c}) - \mathbf{\epsilon} \|_2^2 \right].
\end{align}
We use this architecture as the foundation for our further efficiency optimizations.

\paragraph{Consistency Distillation} 
A key strategy for accelerating diffusion models is consistency distillation, which trains a student model to produce high-quality samples in one or a few steps, mimicking a multi-step teacher. While traditional consistency distillation can involve complex multi-step losses, Flow Matching (FM)~\citep{lipman2022flow} provides a more direct and stable training objective. FM trains the student model $f_{\phi}$ to learn a conditional vector field that directly maps a noisy input $\mathbf{x}_t$ to the clean data $\mathbf{x}_0$. This simplifies the objective to a stable $L_2$ loss, making it a robust method for few-step inference with the conditioned information $\mathbf{c}$:
{\small
\begin{align}
\mathcal{L}_{\text{FM}}(\phi) = \mathbb{E}_{p_0(\mathbf{x}_0), p_t(\mathbf{x}_t)} \left[ \| f_{\phi}(\mathbf{x}_t, t, \mathbf{c}) - (\mathbf{x}_0 - \mathbf{x}_t) \|_2^2 \right]
\end{align}}

\subsection{Spatial Compression}

The computational cost of high-resolution video generation is dominated by spatial redundancy due to the quadratic complexity of the bi-directional self-attention. Traditional models operate at a fixed high resolution, wasting significant computation in early denoising steps that only establish coarse layouts. To address this, we introduce \textbf{Dual-Resolution Caching (DRC)}, a mechanism designed to optimize computational efficiency by exploiting this redundancy without sacrificing visual fidelity.

\begin{figure*}[t!]
\centering
\setlength{\tabcolsep}{0.1em}  
\renewcommand{\arraystretch}{0.2}
 \begin{tabular}{C{0.08\linewidth} C{0.145\linewidth} C{0.145\linewidth} C{0.145\linewidth} @{\hspace{0.5em}} C{0.145\linewidth} C{0.145\linewidth} C{0.145\linewidth}}
 \multicolumn{1}{c}{} & 
  & Baseline &  & 
  & Blur Trial&  \\
  \noalign{\vspace{0.2em}}
  Step 1 & 
  \includegraphics[width=\linewidth]{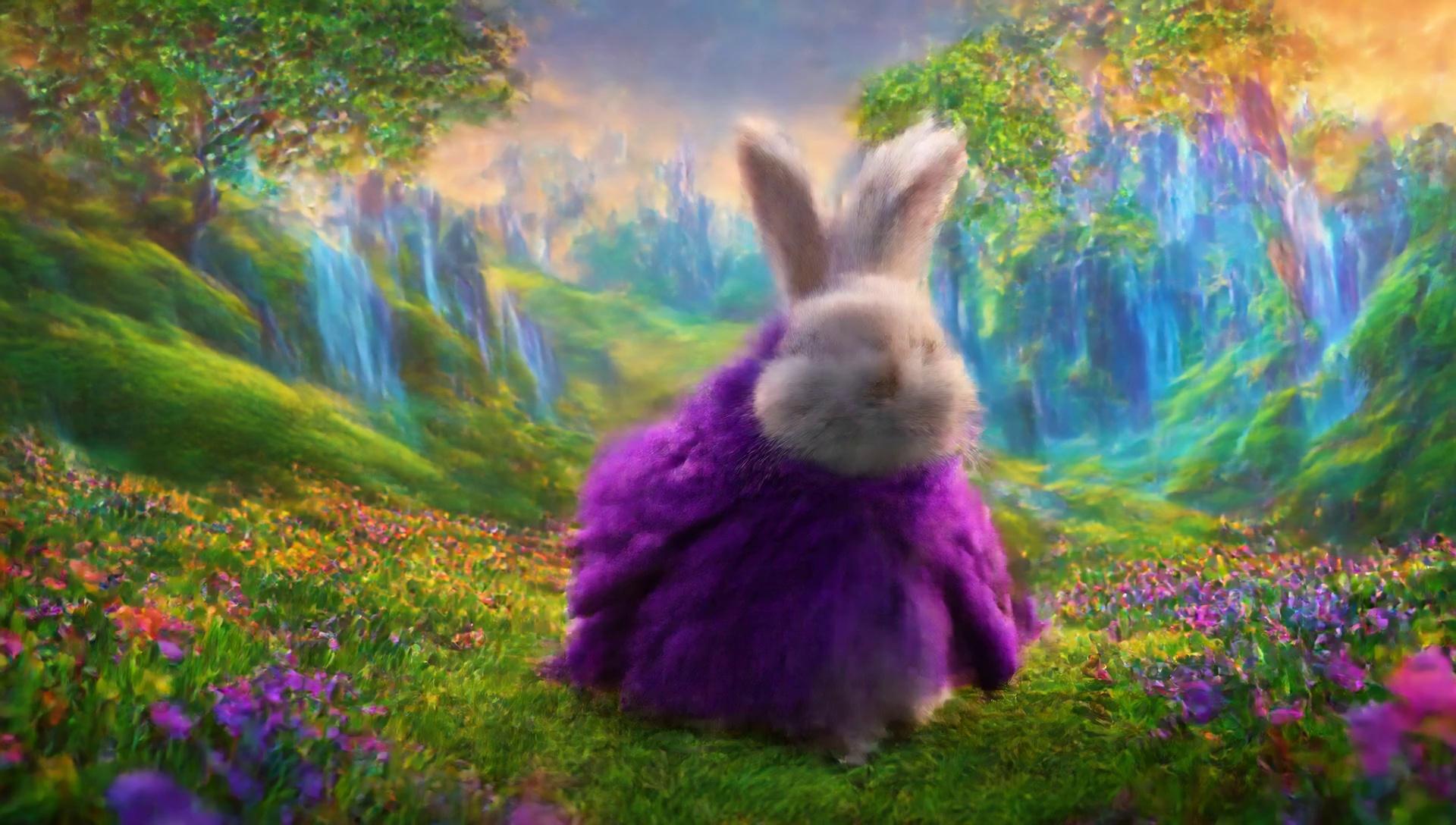} & 
  \includegraphics[width=\linewidth]{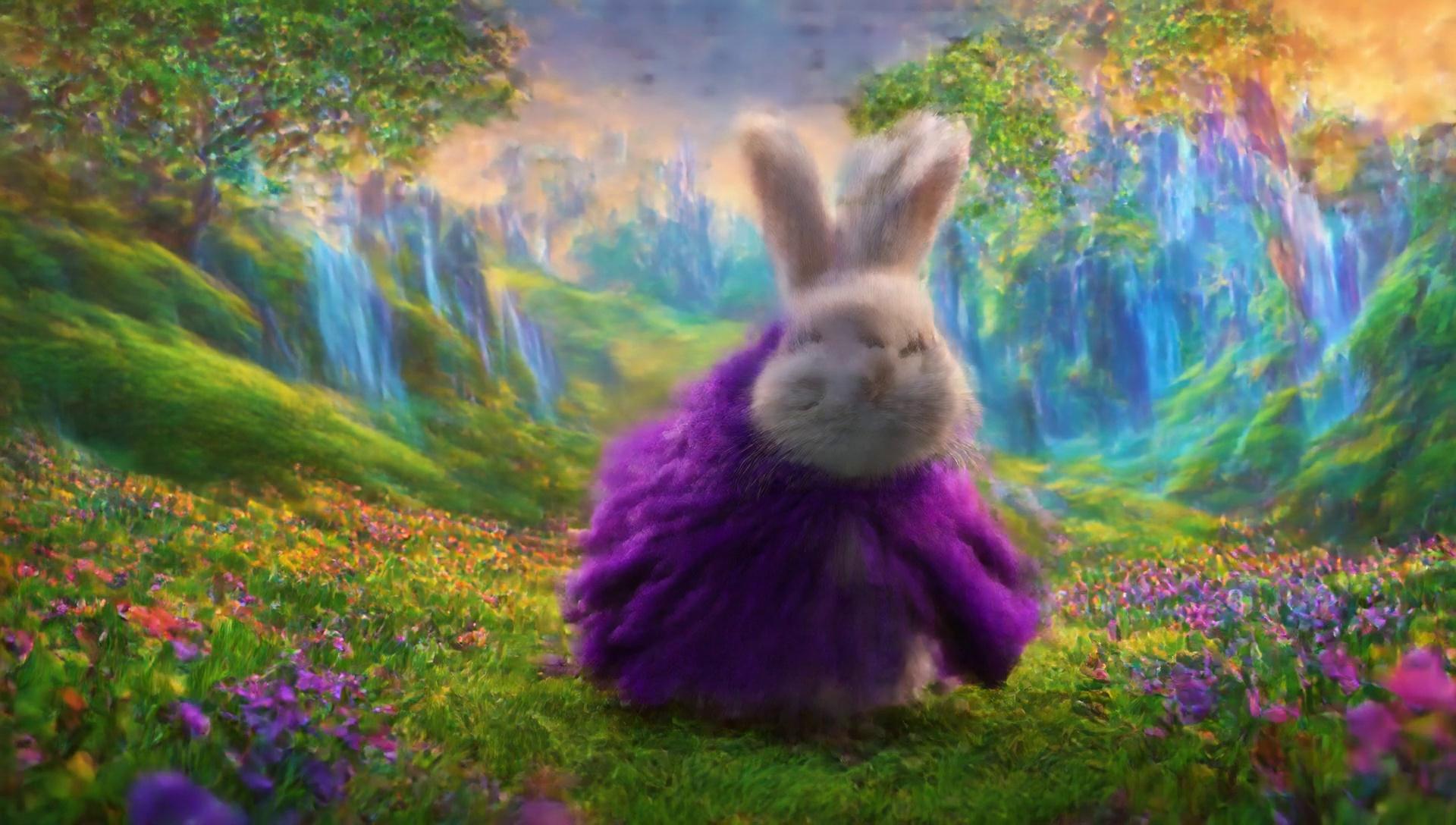} & 
  \includegraphics[width=\linewidth]{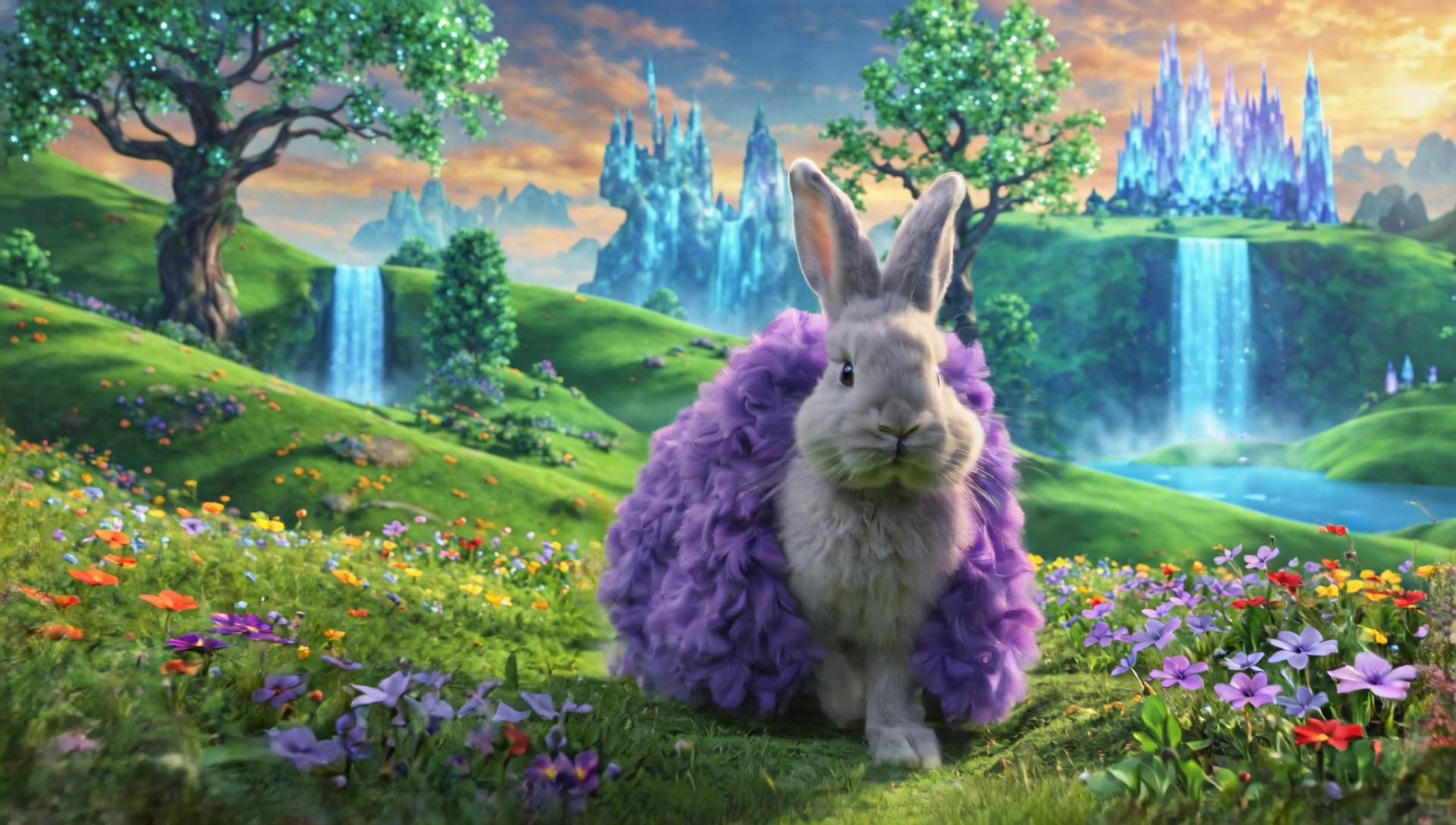} & 
  \includegraphics[width=\linewidth]{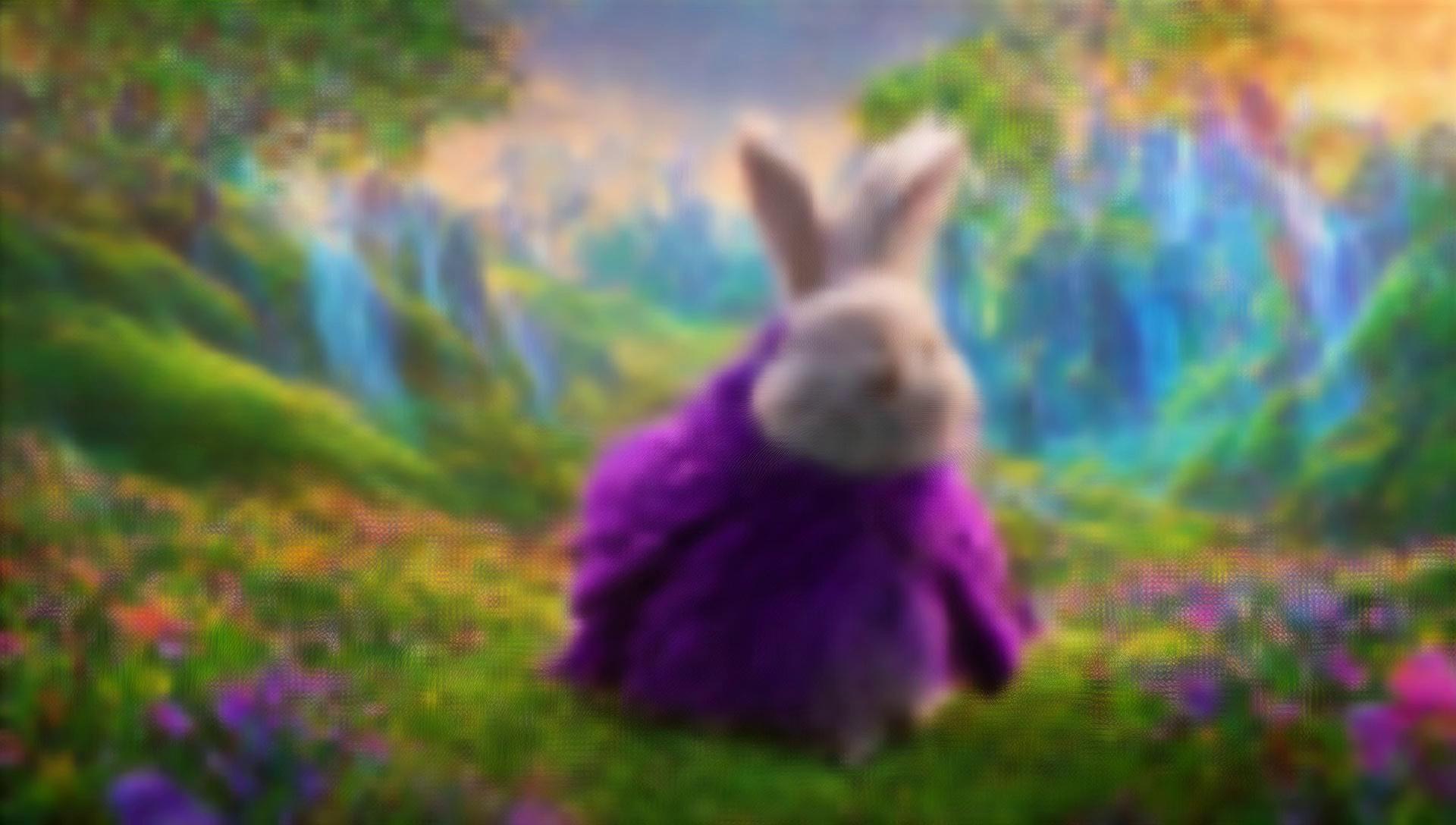} & 
  \includegraphics[width=\linewidth]{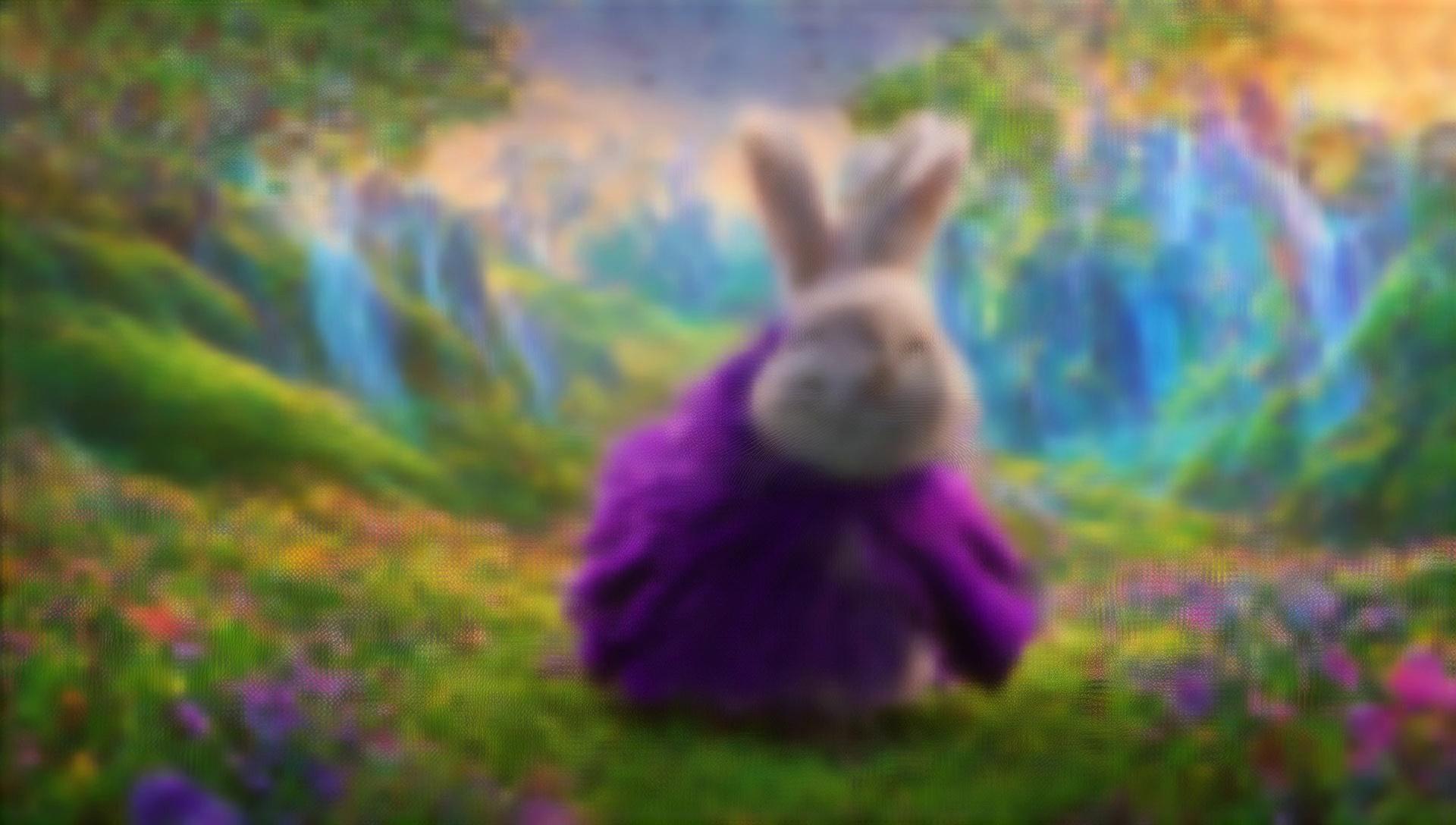} & 
  \includegraphics[width=\linewidth]{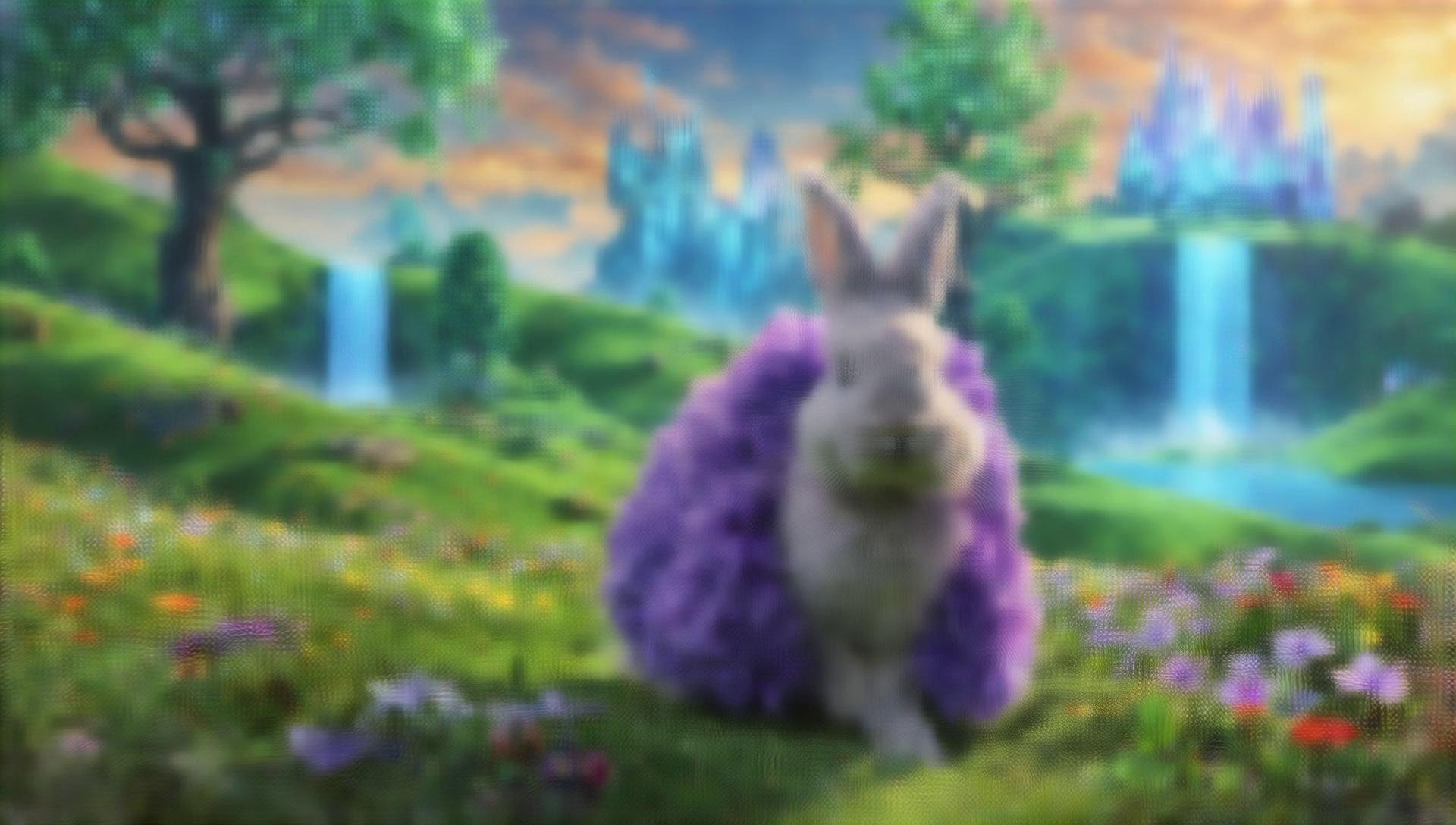} \\
  Step 2 & 
  \includegraphics[width=\linewidth]{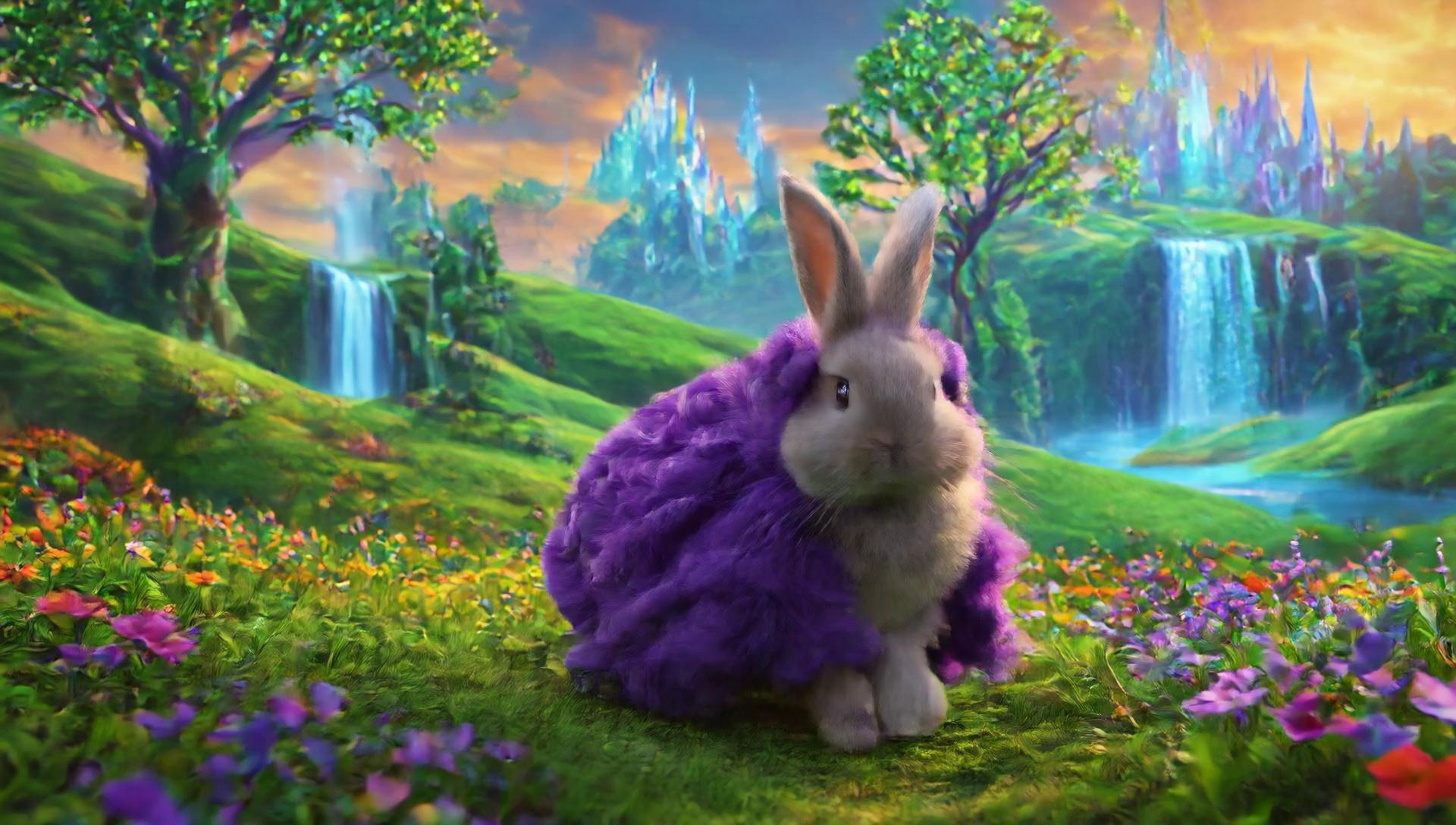} & 
  \includegraphics[width=\linewidth]{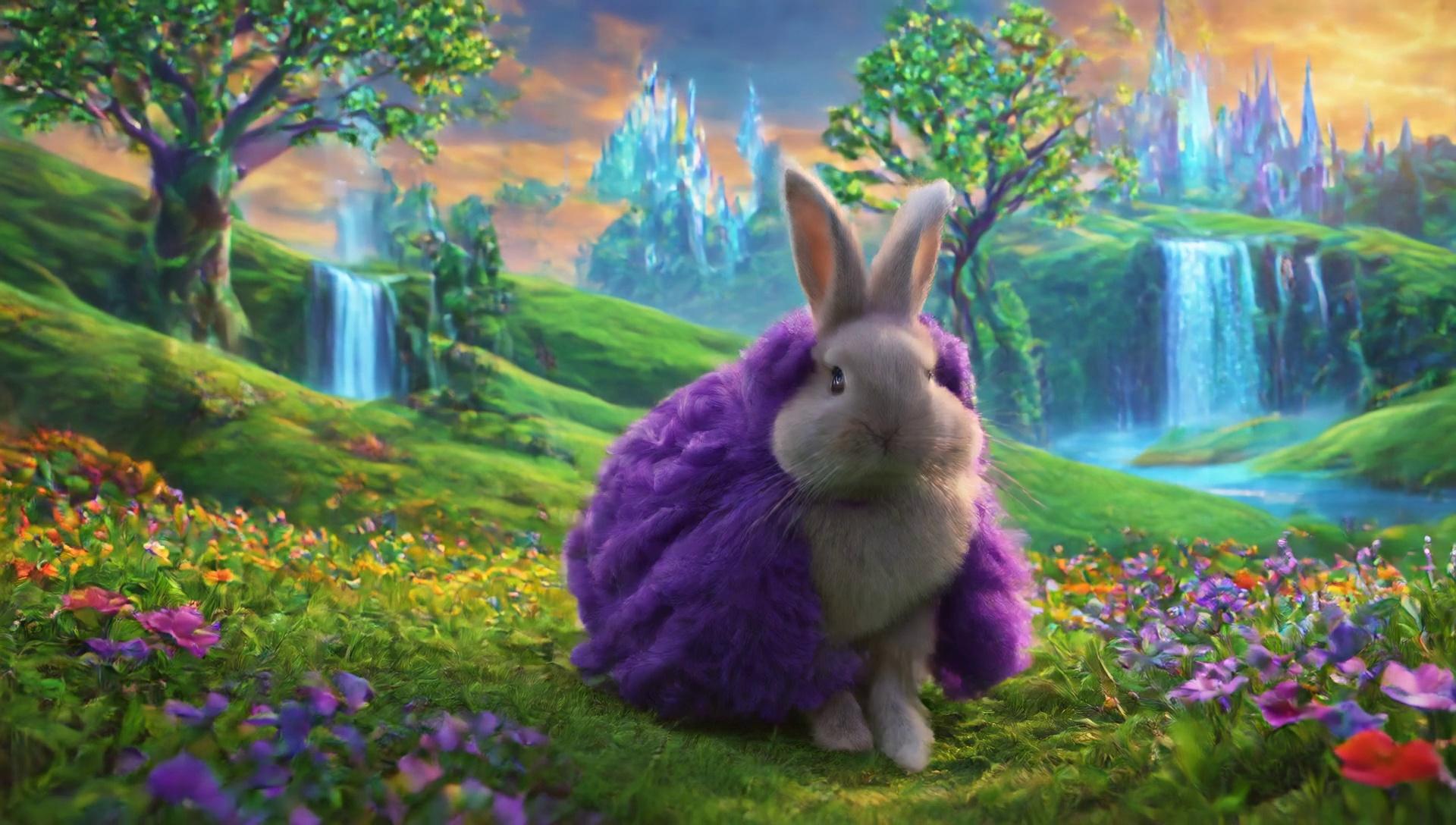} & 
  \includegraphics[width=\linewidth]{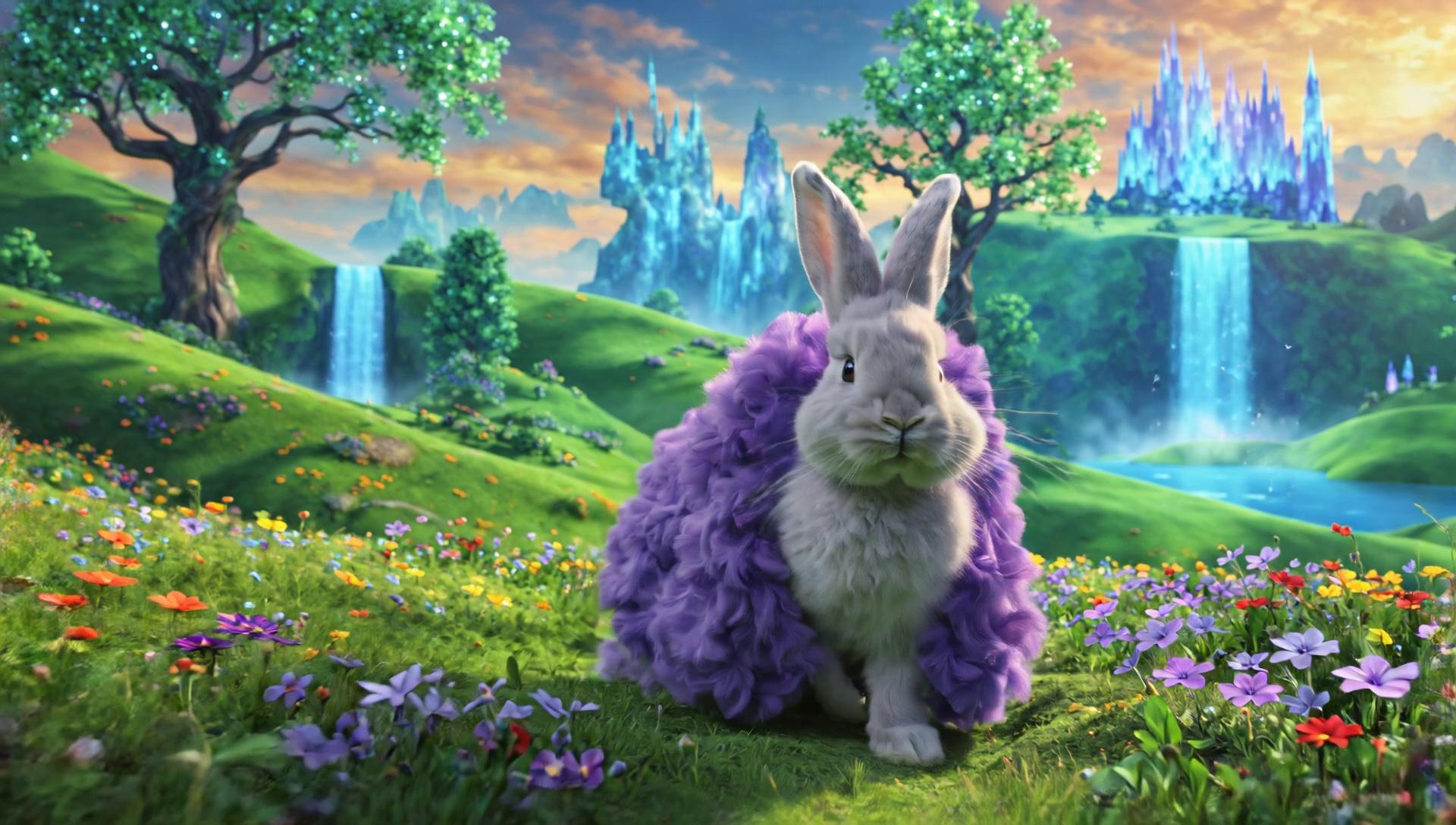} & 
  \includegraphics[width=\linewidth]{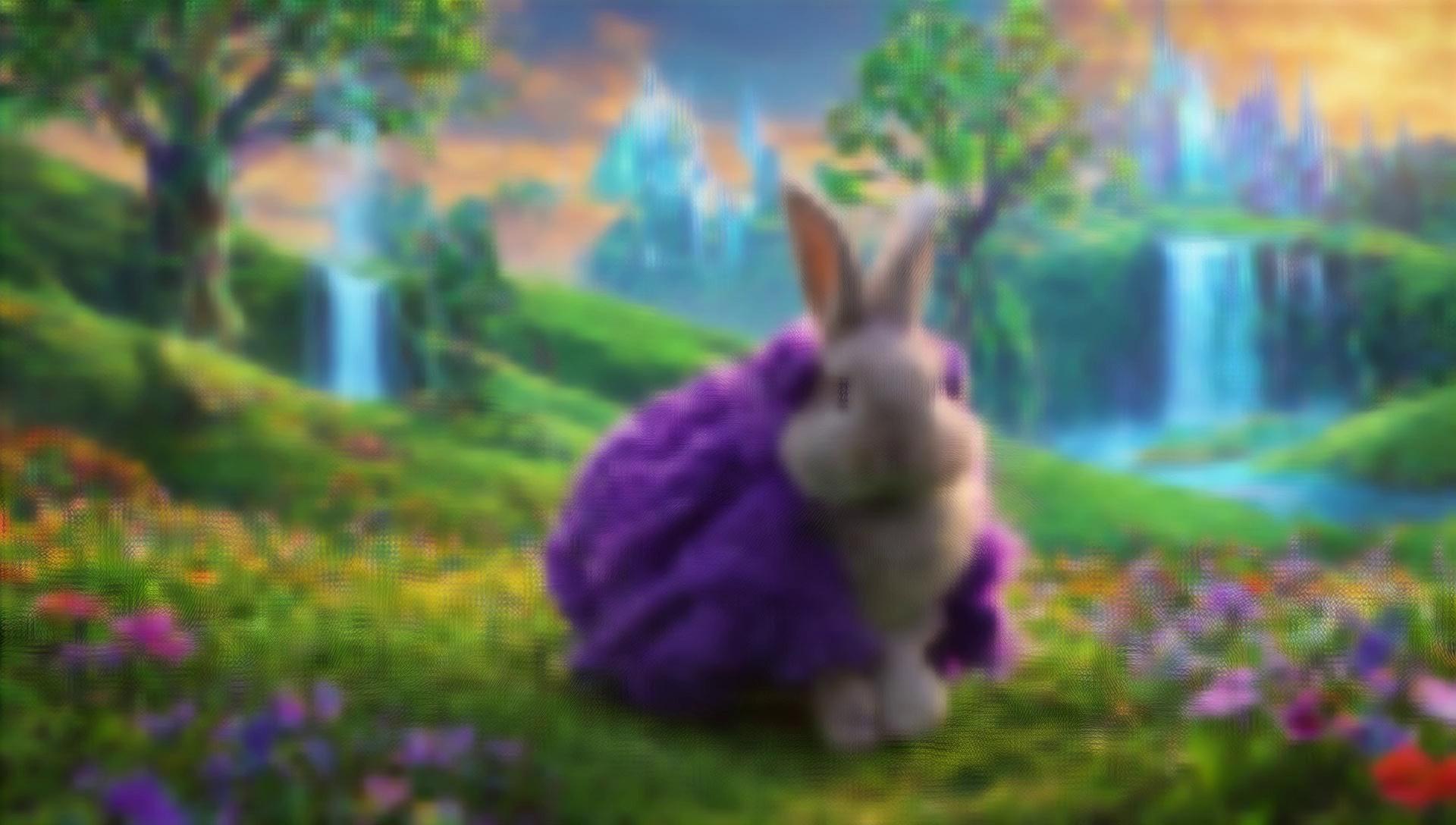} & 
  \includegraphics[width=\linewidth]{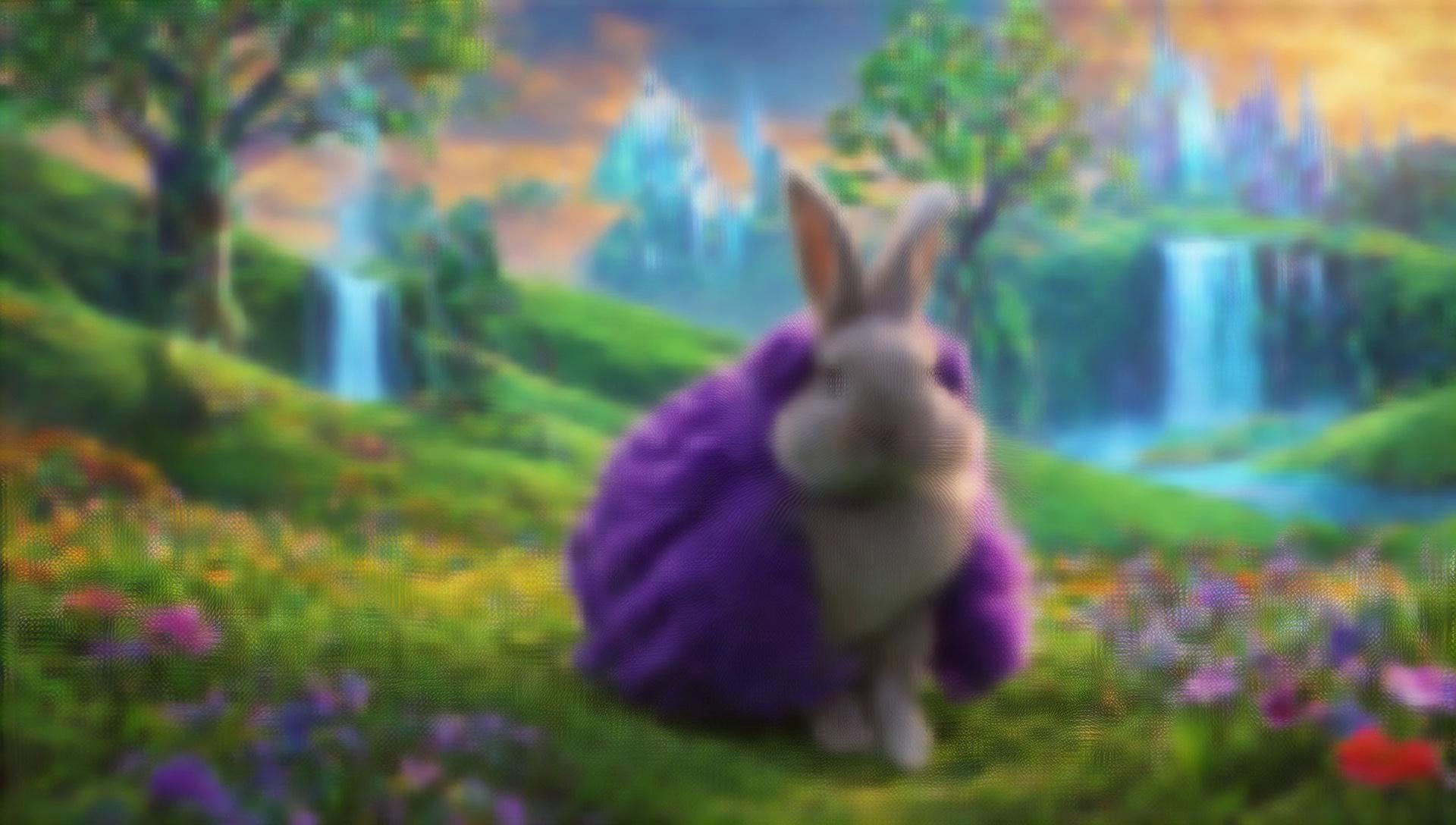} & 
  \includegraphics[width=\linewidth]{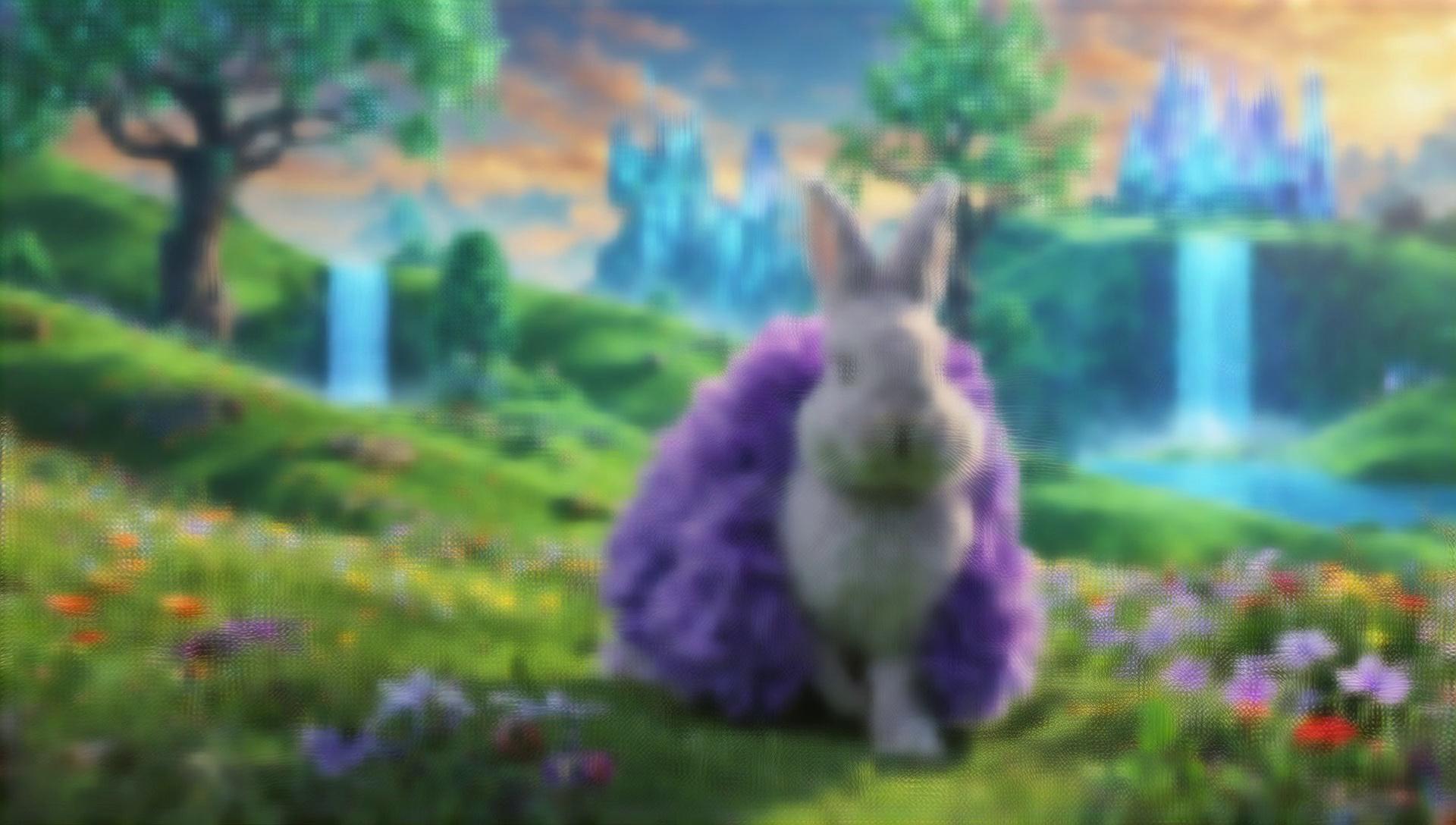} \\
  Step 3 & 
  \includegraphics[width=\linewidth]{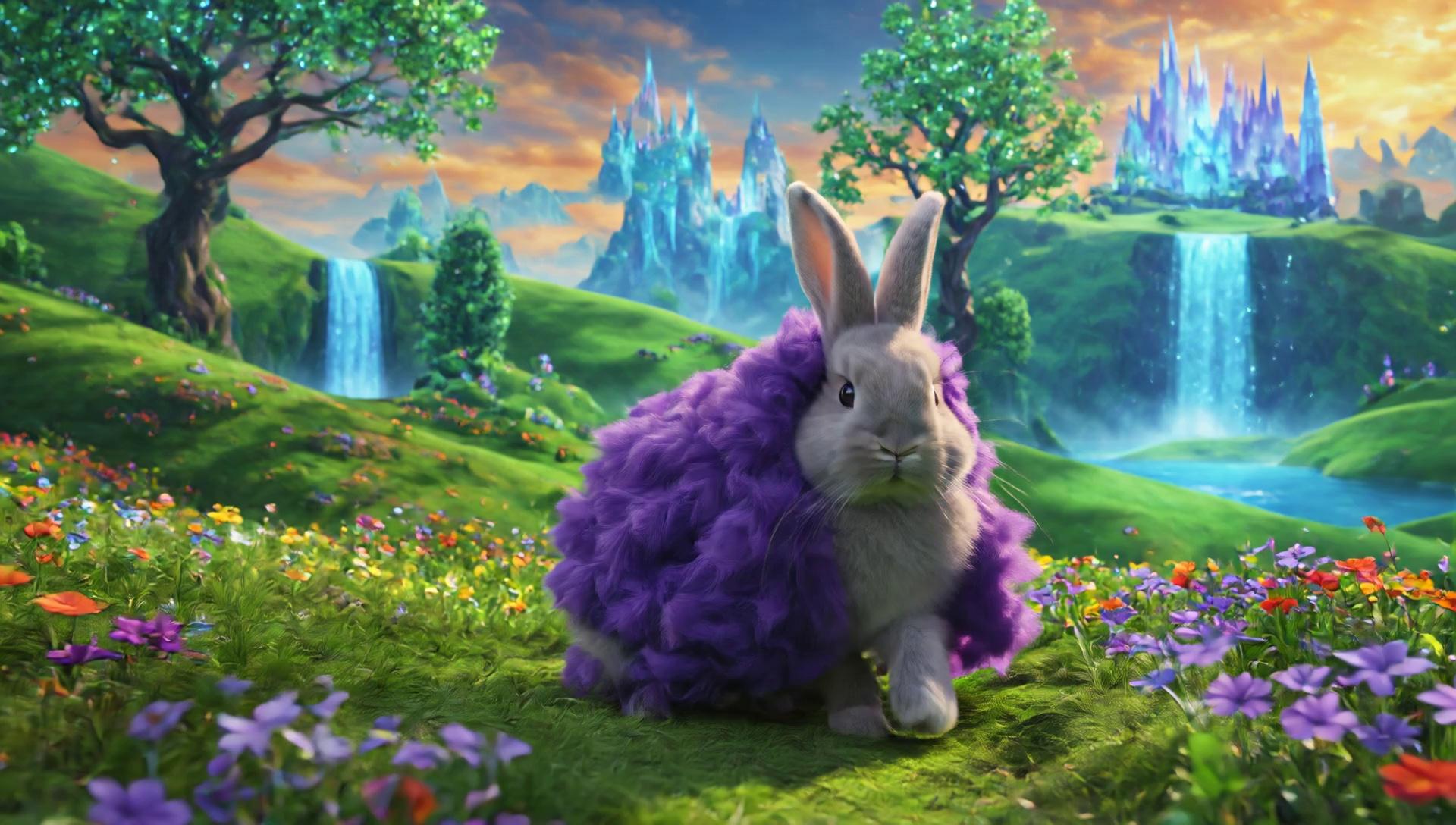} & 
  \includegraphics[width=\linewidth]{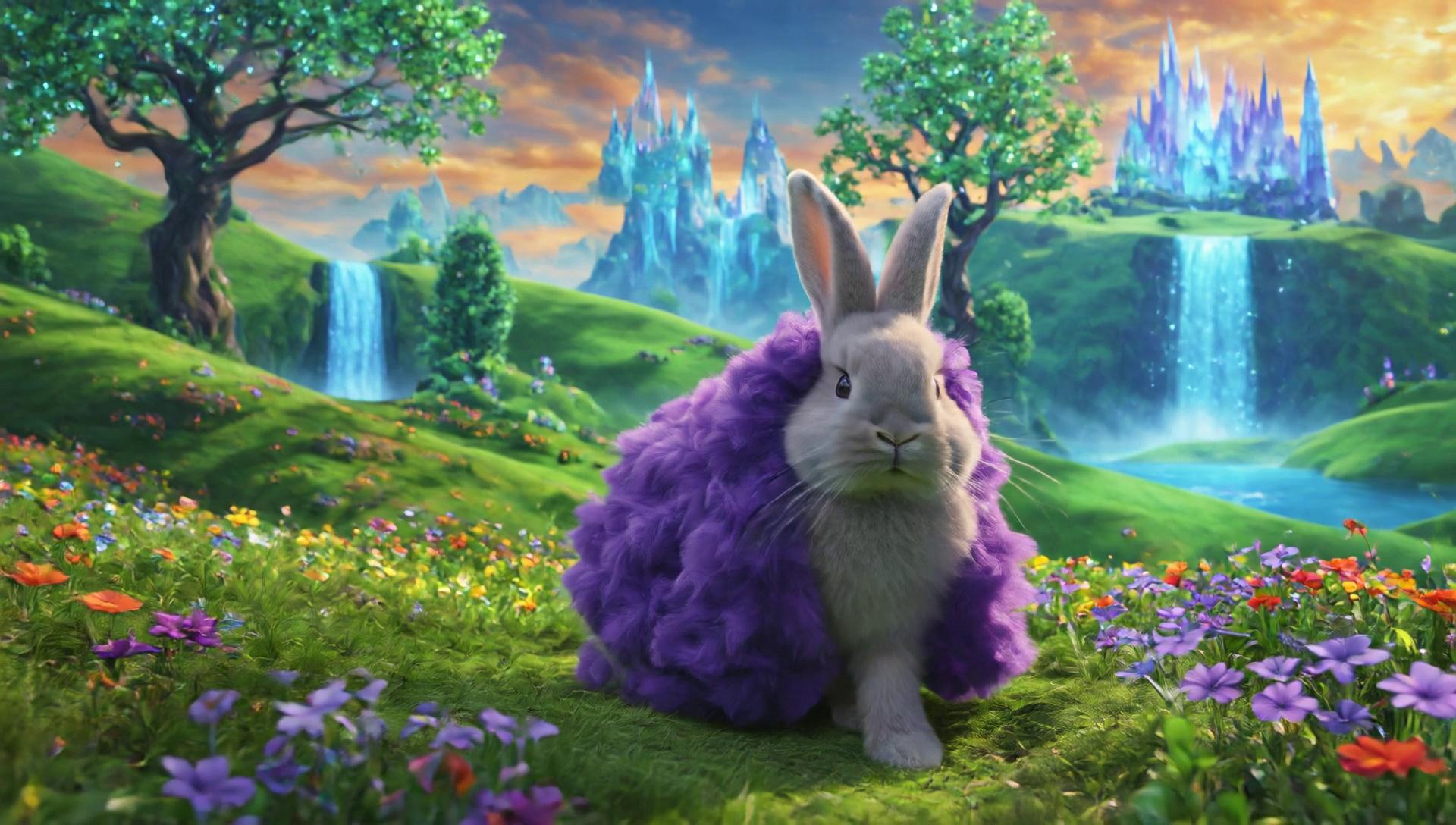} & 
  \includegraphics[width=\linewidth]{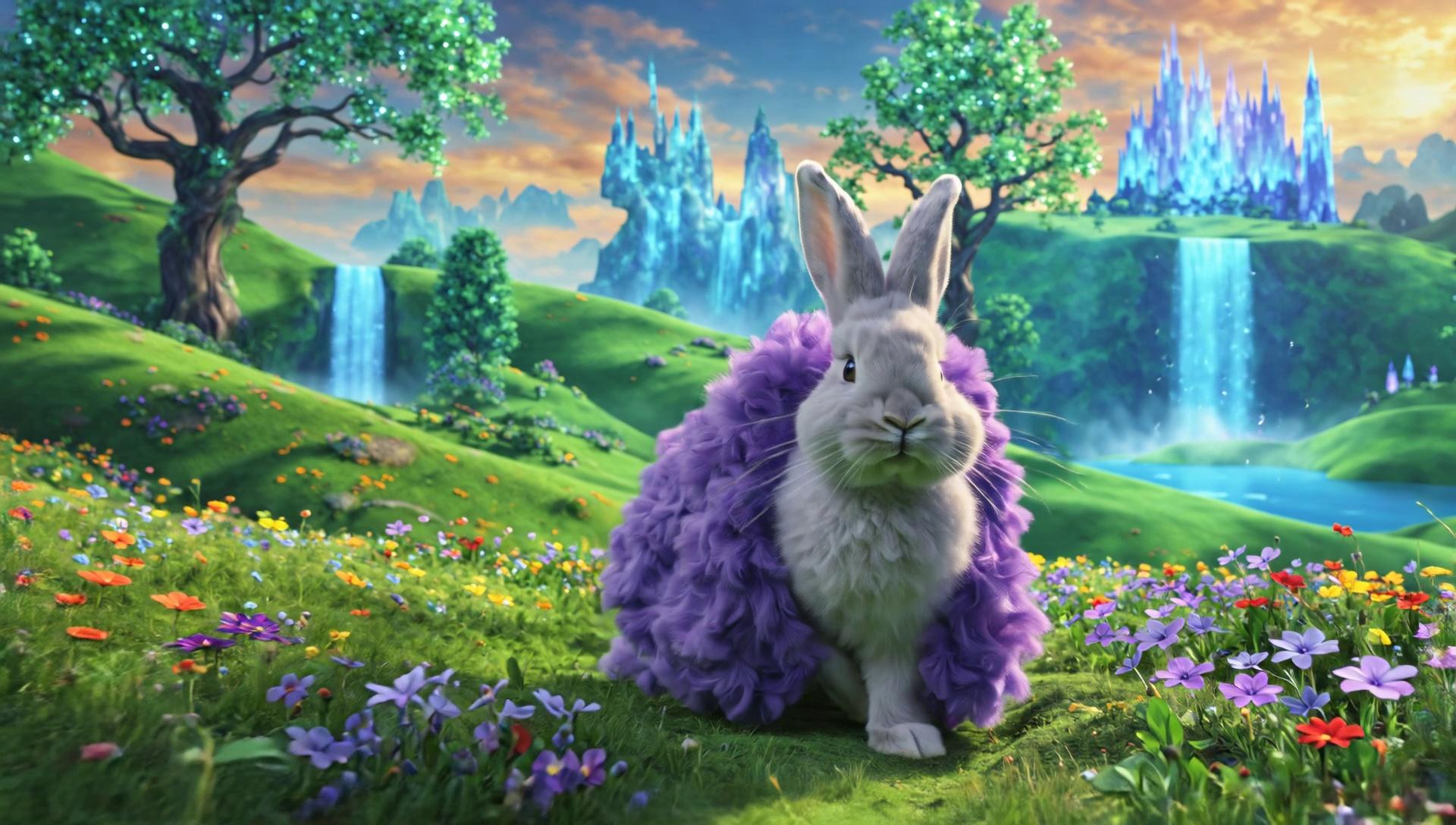} & 
  \includegraphics[width=\linewidth]{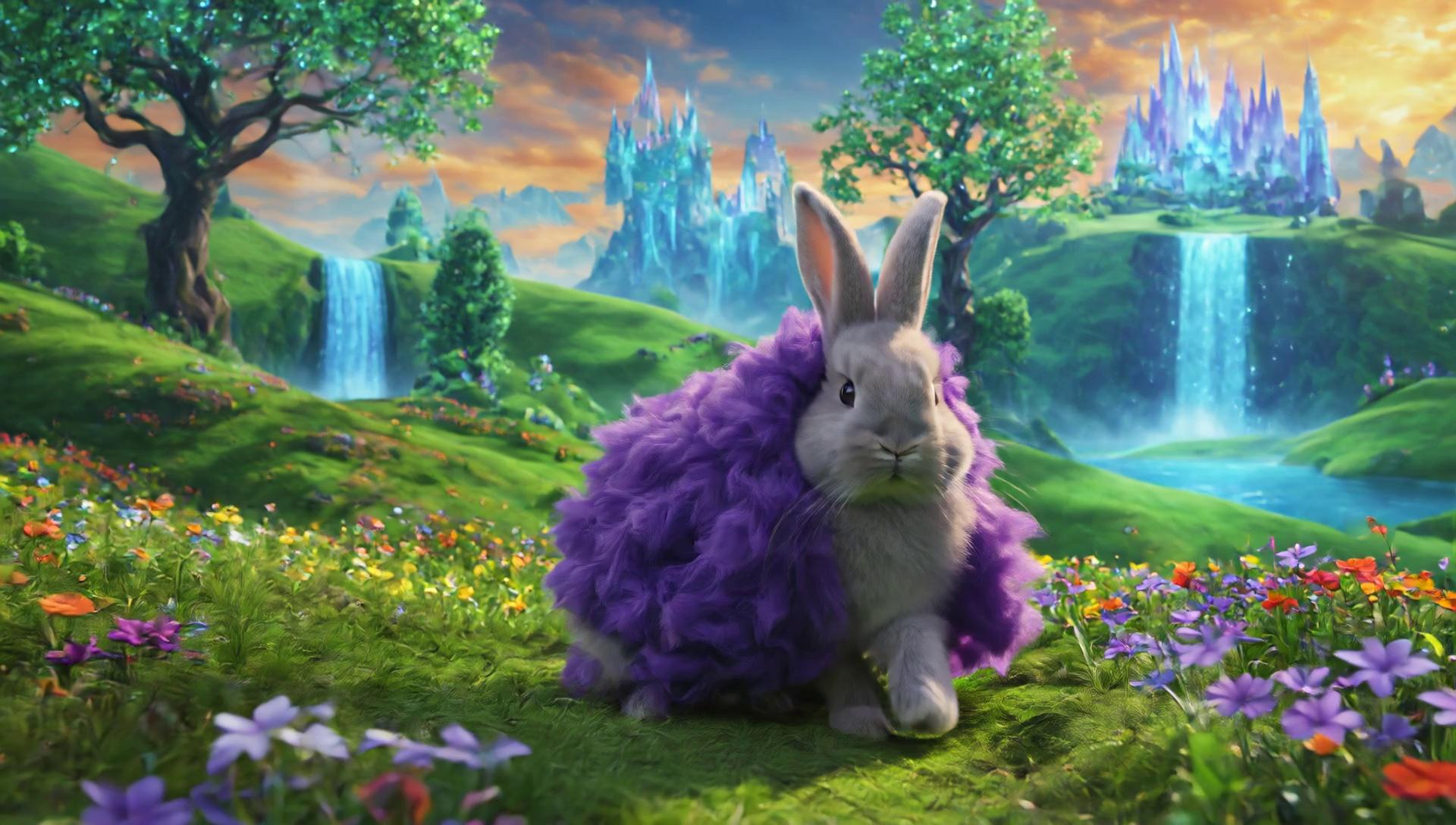} & 
  \includegraphics[width=\linewidth]{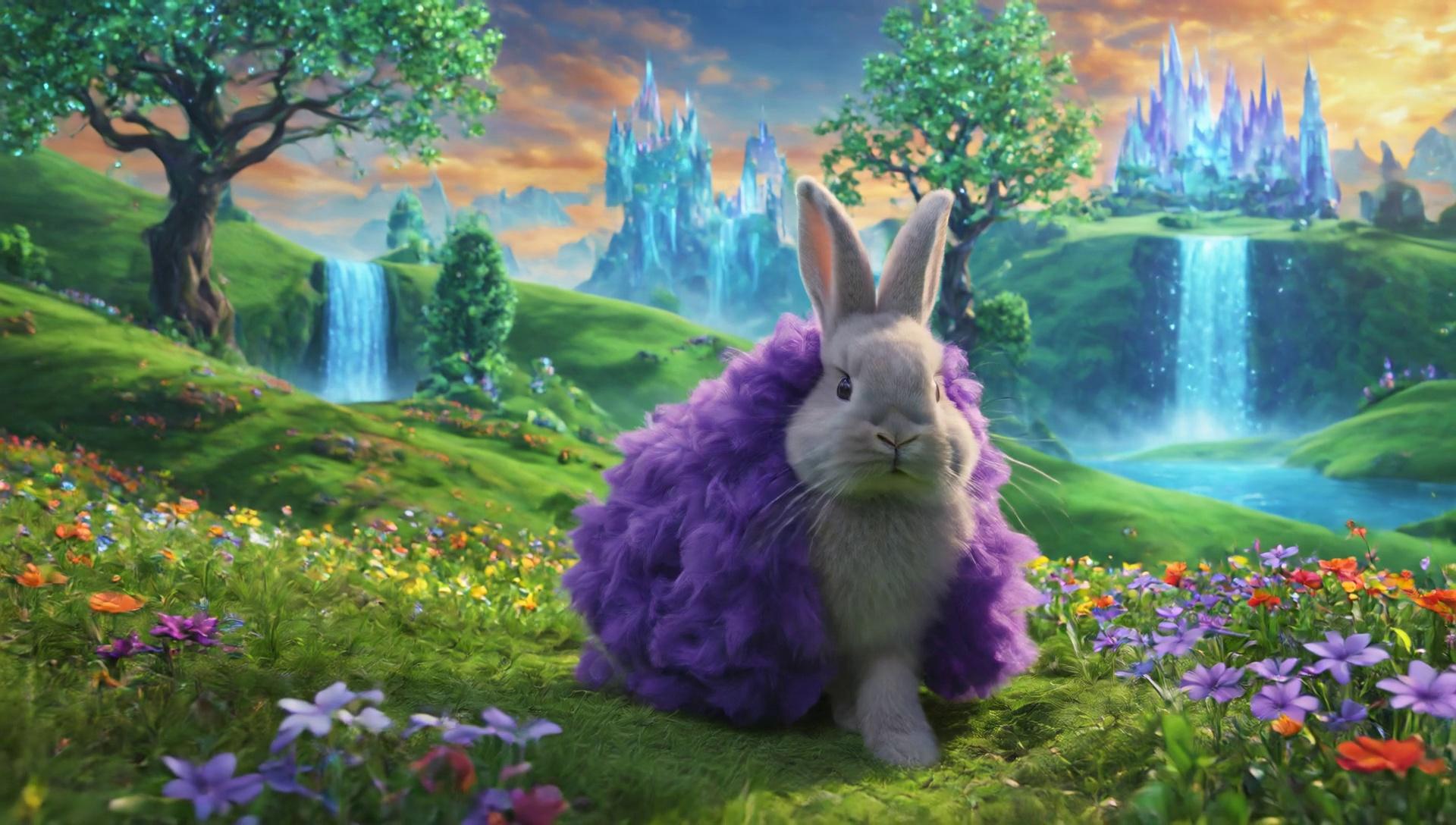} & 
  \includegraphics[width=\linewidth]{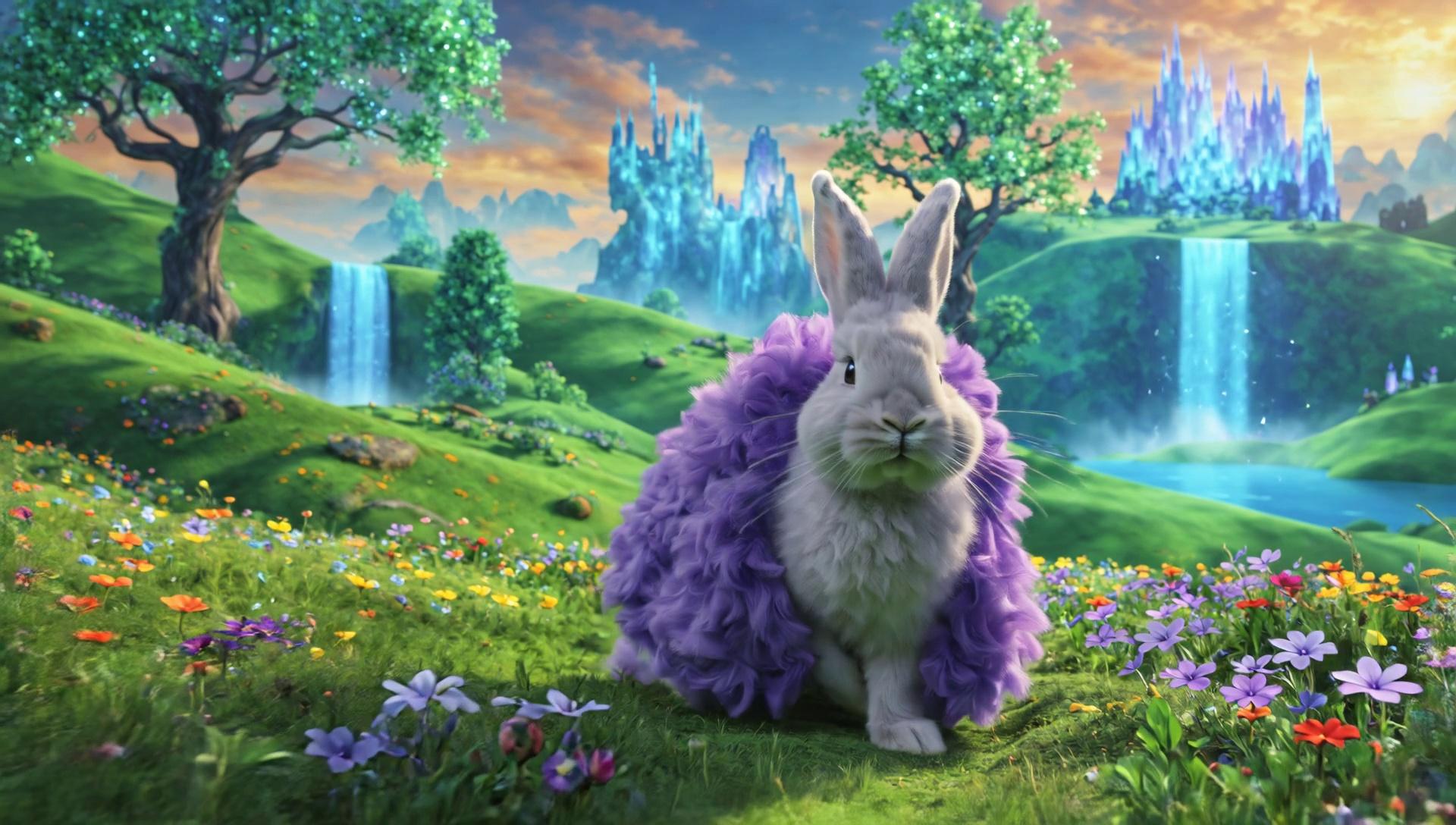} \\  
  Step 4 & 
  \includegraphics[width=\linewidth]{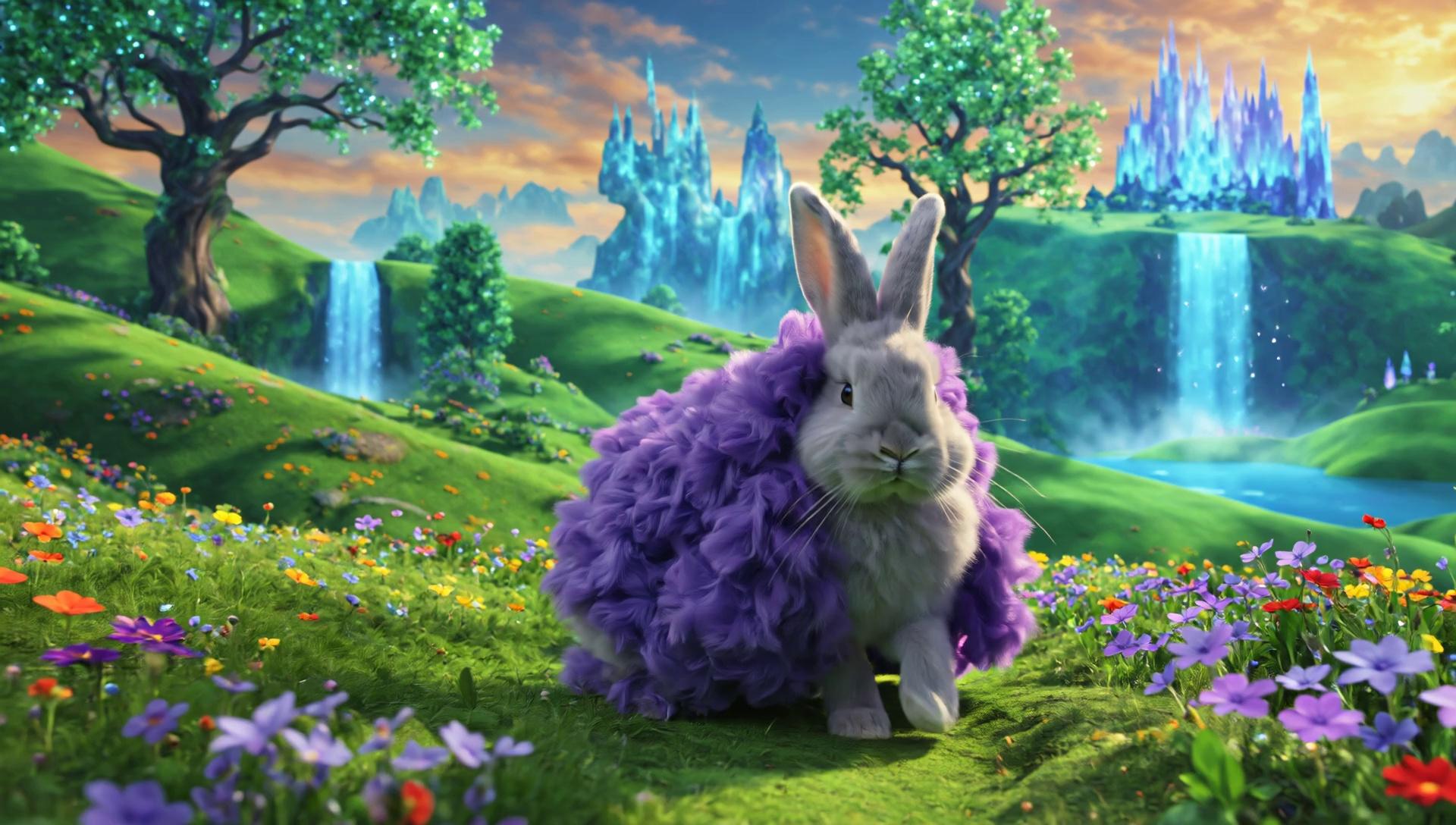} & 
  \includegraphics[width=\linewidth]{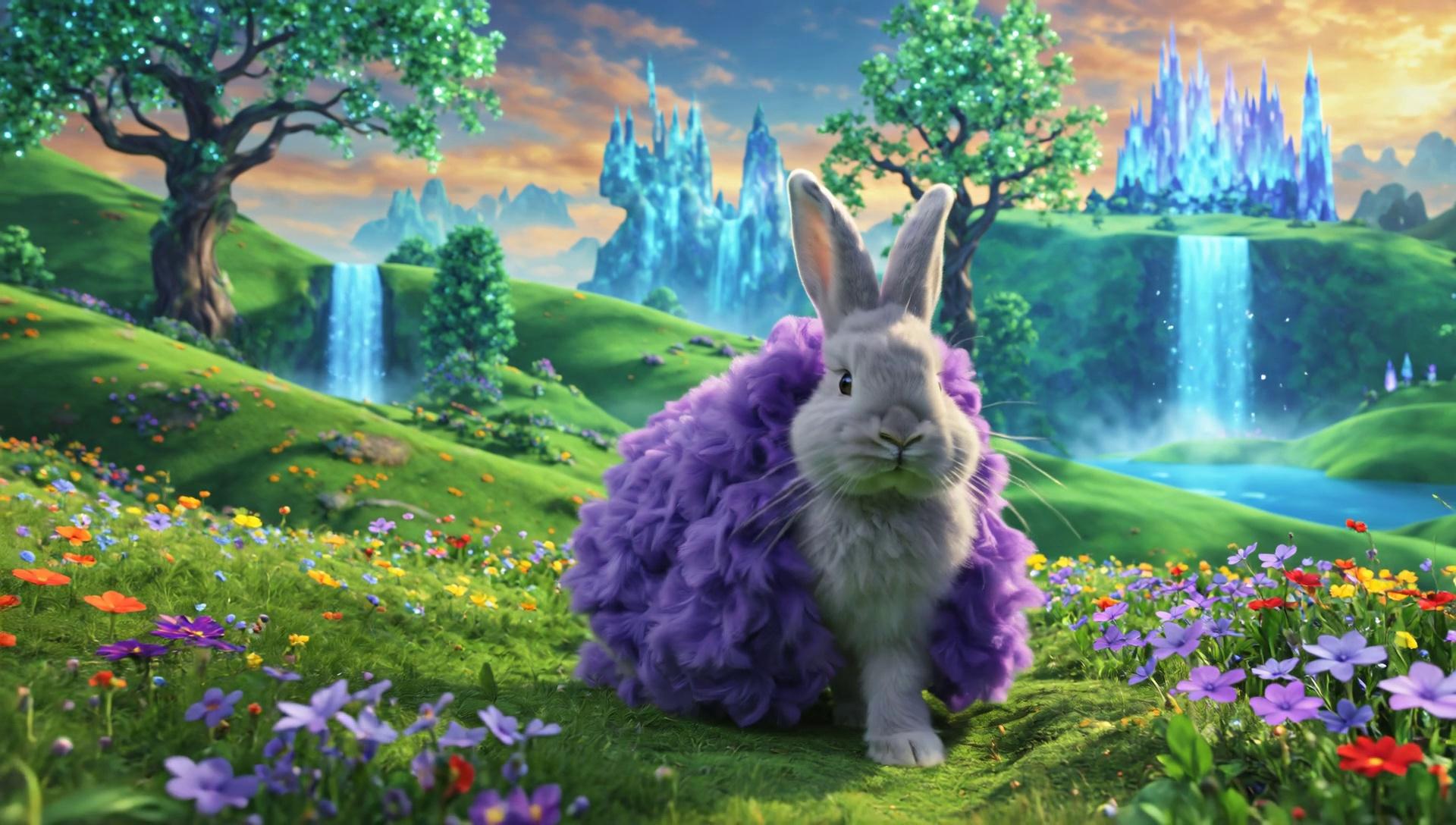} & 
  \includegraphics[width=\linewidth]{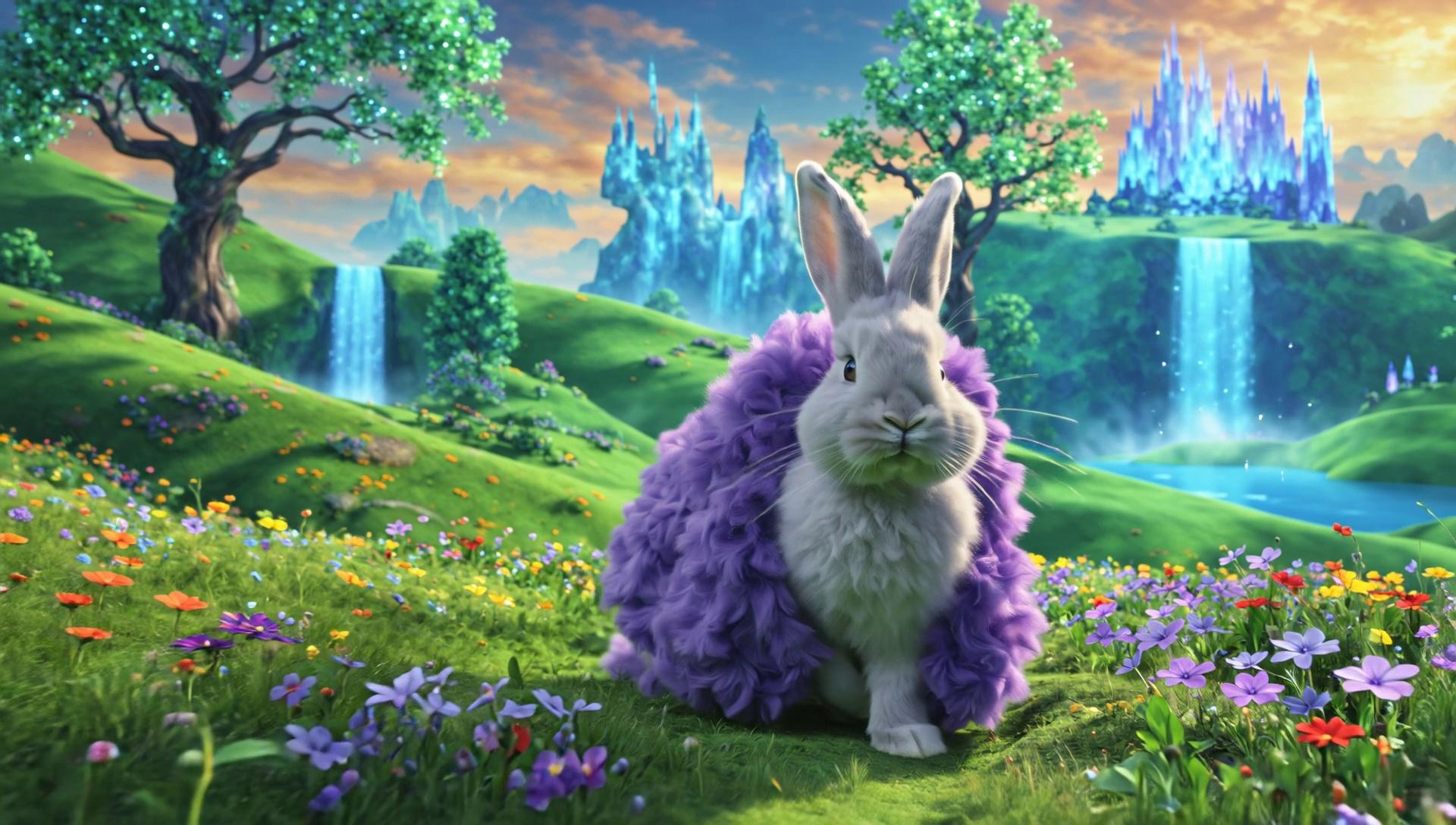} & 
  \includegraphics[width=\linewidth]{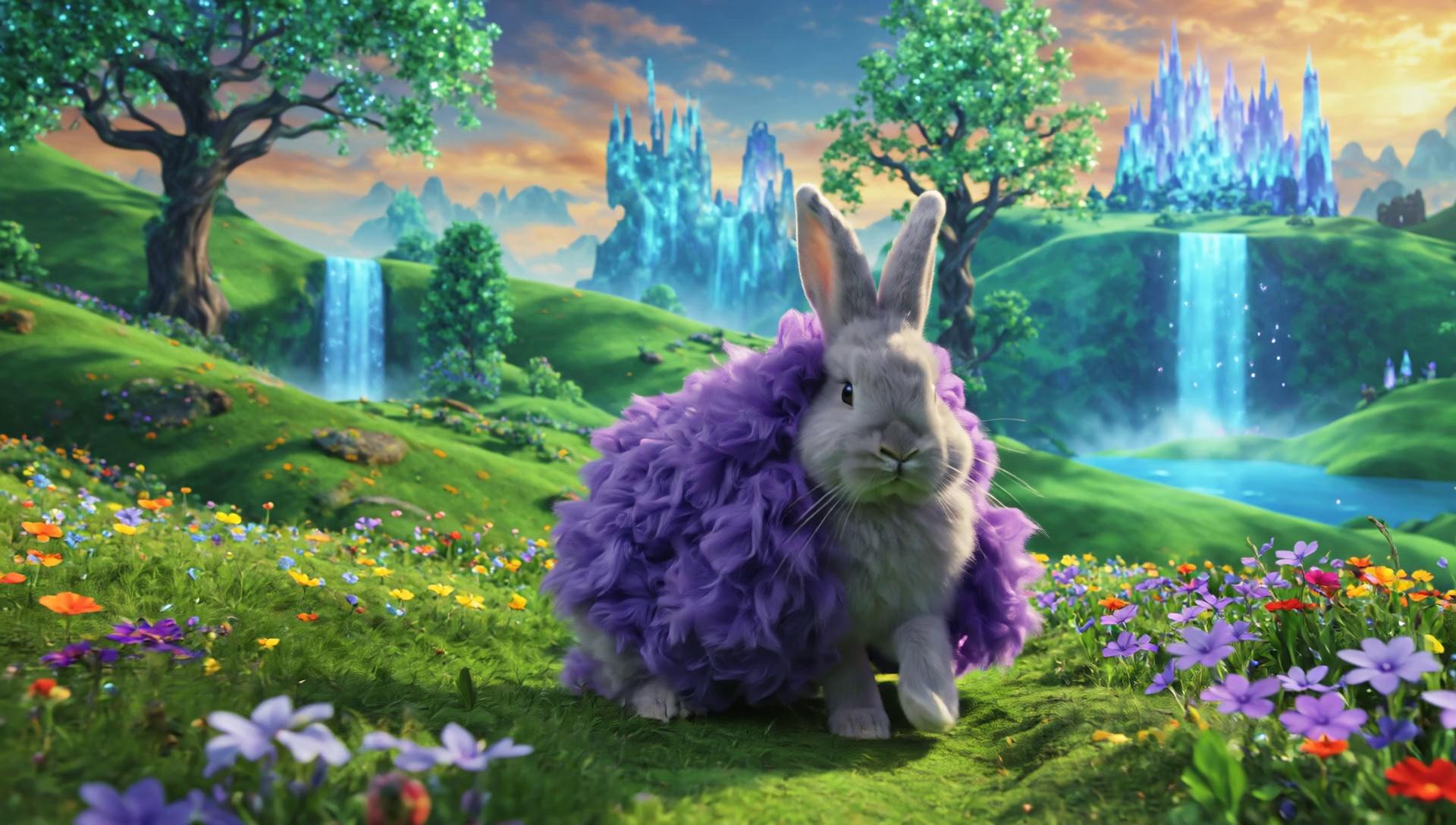} & 
  \includegraphics[width=\linewidth]{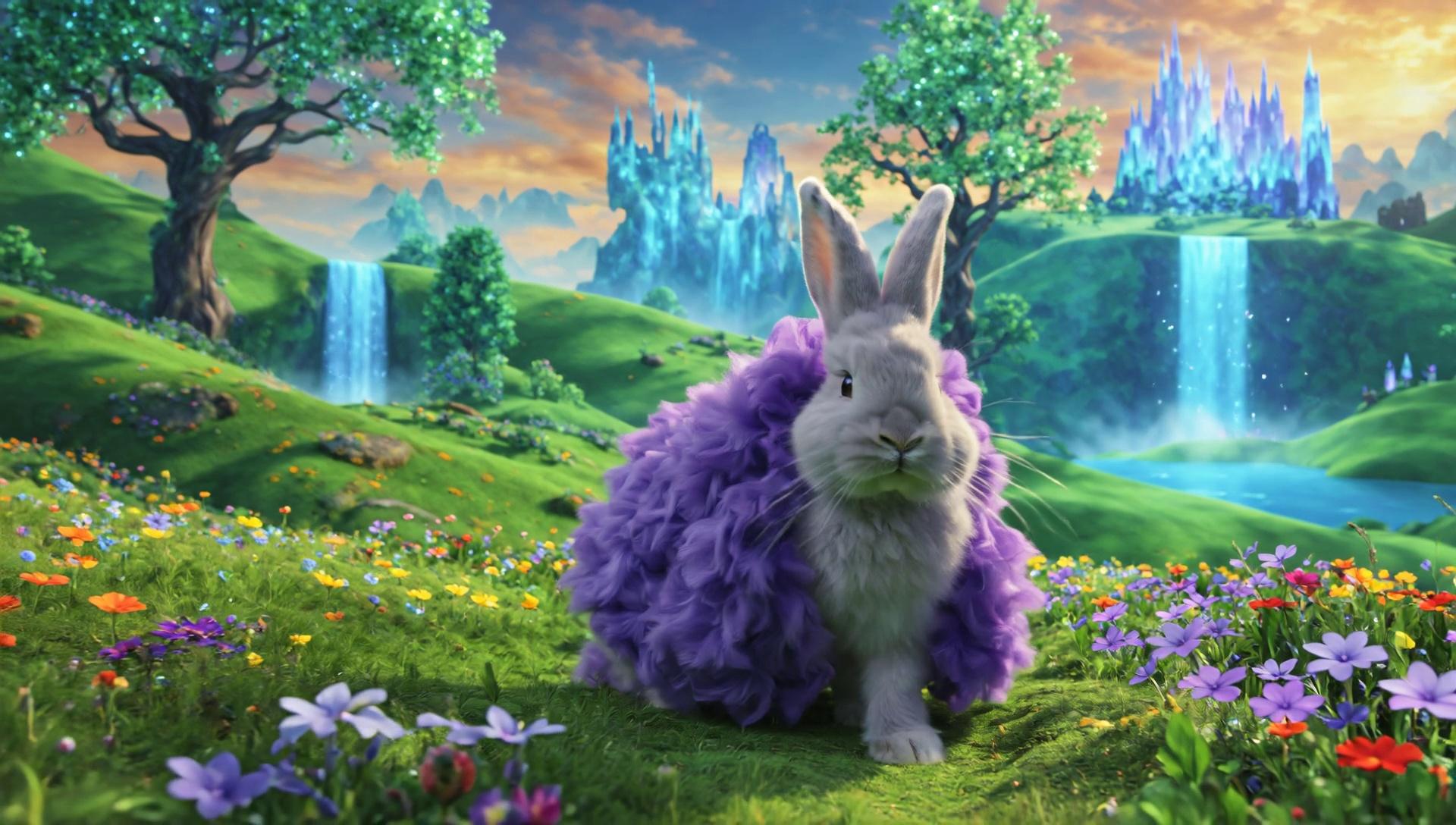} & 
  \includegraphics[width=\linewidth]{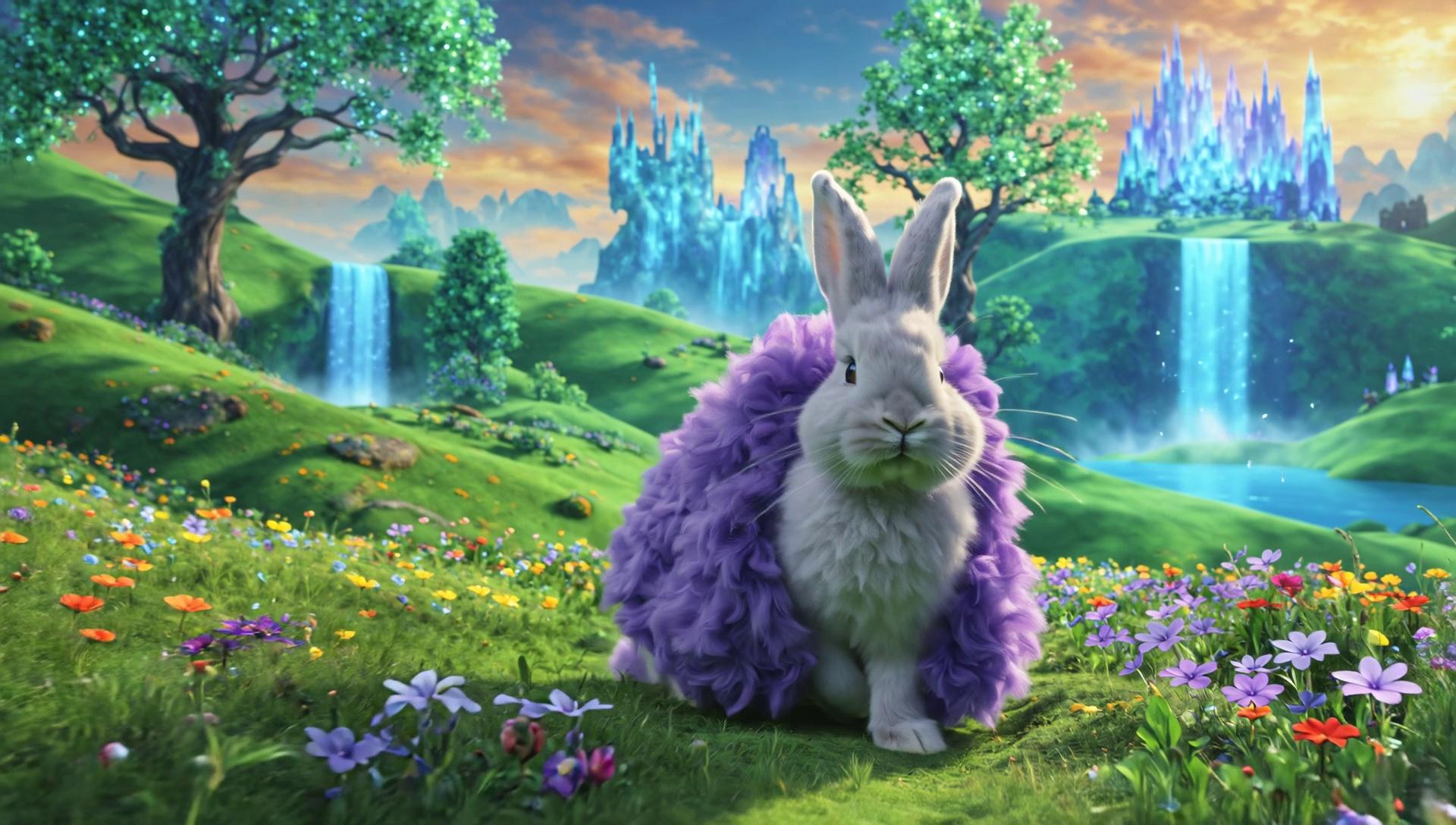} \\
  \noalign{\vspace{0.2em}}
  \multicolumn{1}{c}{} & Frame 1 & Frame 9 & Frame 17 & Frame 1 & Frame 9 & Frame 17 \\ 
\end{tabular}
\vspace{-0.5em}
\caption{\textbf{Results visualization of different timesteps.} (\textbf{Left}) A notable quality gap exists at Step 1 between frame 9 (chunk 1) and frame 17 (chunk 2). The fully denoised cache from chunk 1 provides high-value context, enabling chunk 2 to achieve a near-final quality (comparable to Step 3) in just one step. 
(\textbf{Right}) Blur Trial: We also test robustness by blurring the outputs of Steps 1 and 2 (via down/upsampling) in the 4-step process. The final result remains nearly indistinguishable from the Baseline Setting.
}
\vspace{-1.0em}
\label{fig:steps}
\end{figure*}

\begin{figure}[t!]
\centering
\includegraphics[width=0.99\linewidth]{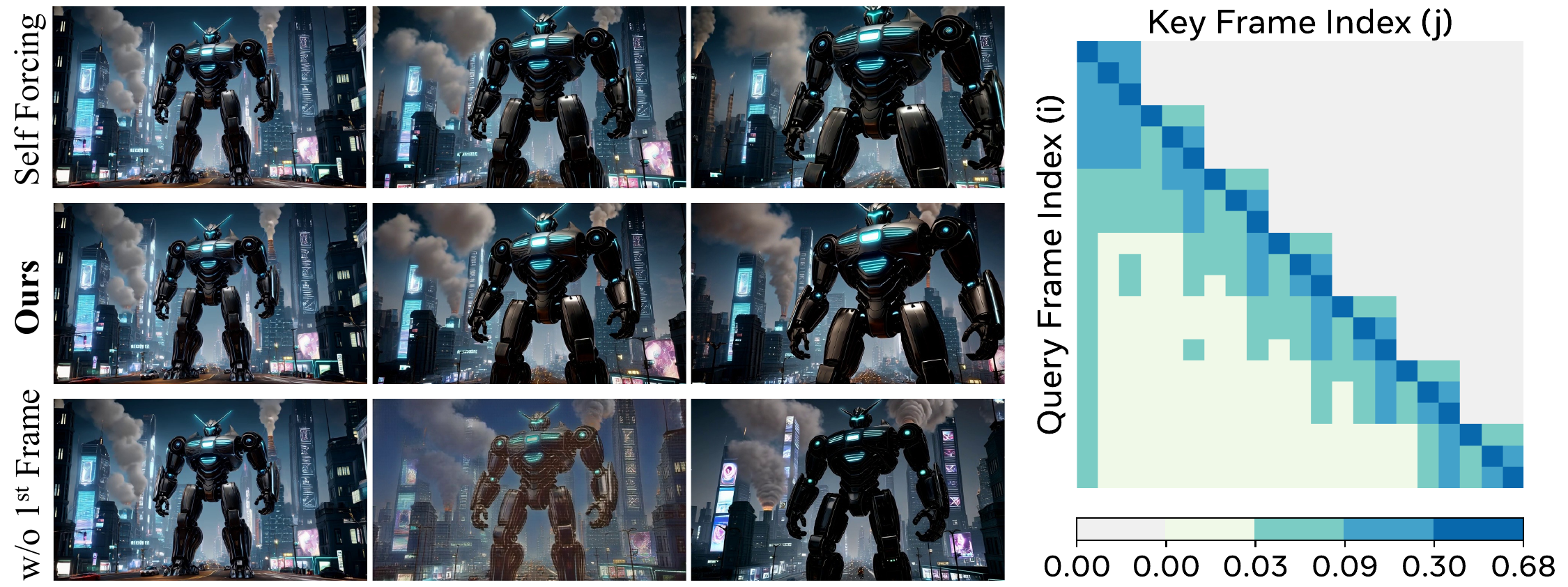}
\vspace{-0.5em}
\caption{\textbf{Visualization of temporal attention.}
(\textbf{Right}) Attention during autoregressive generation naturally focuses on the initial frame and the two most recent neighboring frames. (\textbf{Left}) This visualization demonstrates that dropping intermediate, less important frames has a negligible impact, even without retraining. Conversely, dropping the initial frame causes catastrophic quality degradation.
}
\vspace{-1.0em}
\label{fig:skip}
\end{figure}

\paragraph{Progressive Upscaling.}
Our approach is motivated by the significant spatial redundancy we observe in few-step flow matching models. For a 4-step denoising process, we find the initial two steps (Steps 1-2) primarily establish global composition, while any fine-grained details they generate are largely overridden by the final refinement steps (Steps 3-4). We validated this empirically (Fig~\ref{fig:steps}): by deliberately blurring the outputs of Steps 1-2 (via downsampling and upsampling) before feeding them to the unmodified Steps 3-4, the final video is nearly indistinguishable from the full-resolution baseline. This confirms that the initial, structure-focused steps can be executed entirely at a low resolution without compromising final quality.

Based on this insight, our DRC strategy re-architects the denoising trajectory. Steps 1-2 are performed at a low resolution to efficiently establish the global structure. The latent representation is then upsampled to the target high resolution, and Steps 3-4 perform high-resolution refinement. To enable a single network to handle this resolution shift, we adapt techniques validated in LLMs~\citep{qiu2025cinescale}, integrating Rotary Positional Embeddings (RoPE) with NTK-inspired scaling~\citep{peng2023yarn} for generalizing positional information across scales. This progressive upscaling approach reserves the most expensive high-resolution operations for the final, critical refinement stages.

\paragraph{Dual KV Caching}
Our progressive generation structure necessitates a specialized KV cache mechanism that deviates from the conventional single-resolution designs. If we cache the features from the low-resolution Steps 1-2, they will be misaligned with the final high-resolution output from Steps 3-4 (due to the iterative refinement of flow matching). Using this misaligned low-resolution cache to condition the next video chunk would introduce structural conflicts and degrade temporal consistency.

To solve this, our dual-resolution caching design ensures the cache aligns with the final high-resolution output. After the eventual high-resolution refinement (Step 4) of a chunk is complete, we downsample this final, high-fidelity result back to the low resolution. We then use these downsampled latents to update the low-resolution KV cache. This guarantees the cache passed to the next chunk is spatially consistent with the high-resolution video just generated, maintaining spatio-temporal coherence.

\subsection{Temporal Compression}
Our dual-resolution strategy does not, by itself, solve the problem of temporal complexity. In a naive autoregressive structure, the KV cache still grows indefinitely, leading to memory exhaustion and slowing inference with long video generation. We therefore introduce the \textbf{Anchor-Guided Sliding Window (AGSW)}, a strategy to maintain a fixed computational budget. AGSW generates video in chunks, restricting the attention context for any new chunk to a constant size. This window is composed of the current chunk ($M$ frames), a local history cache ($M-1$ frames), and a global anchor ($1$ frame), fixing the maximum computation length to $2M$ frames.

\paragraph{Temporal Decay Contribution.}
The primary mechanism for maintaining smooth local motion and temporal coherence is the Temporal Decay Contribution. This set consists of the $M-1$ most recent past frames (i.e., the last two frames of the previously generated chunk, $\mathbf{x}^{i-2:i}$, for $M=3$). These neighboring frames provide essential local historical context (Figure~\ref{fig:skip}), offering the necessary motion vectors and fine-grained state information for the generation of the current chunk ($\mathbf{x}^{i:i+M}$). The decay aspect reflects the empirical observation that fine-grained motion context is predominantly derived from immediate history, rapidly diminishing in relevance over longer temporal distances. By limiting this local context to a fixed, small number of frames, we efficiently capture short-range dependencies crucial for visually fluid transitions.

\paragraph{Temporal Attention Sink.}
To ensure long-range consistency, we leverage the ``attention sink'' phenomenon observed in LLMs and autoregressive generation. Efficiency-focused models like StreamingLLM~\citep{xiao2023efficient} strategically leverage this by using fixed anchor tokens to stabilize a sliding window. We observe that this phenomenon also naturally emerges in autoregressive video generation, with attention strongly concentrating on the first frame. This insight is so fundamental that, {\it even without fine-tuning}, dropping intermediate, low-value frames still yields high-quality results closely approximating full-history generation (Figure~\ref{fig:skip}).

Therefore, our AGSW strategy formalizes this behavior by designating the first frame as a {\it Temporal Attention Sink}. The tokens from $\mathbf{x}^1$ are always cached and included in the attention computation for every subsequent chunk. This single-frame anchor provides a stable, long-range reference point that grounds the generation, ensuring global scene and object consistency across the entire video without the computational burden of attending to the full history.

\subsection{Timestep Compression}
Our analysis of the denoising trajectory (visualized in Figure~\ref{fig:steps}) revealed a critical and exploitable redundancy in timestep. We observed that while the first chunk (e.g., frame 9) requires the full 4-step denoising process to establish a high-quality state, the second chunk (e.g., frame 17) achieves a near-final quality, comparable to a Step 3 output, after just \emph{one} denoising step.

This is because the second chunk's generation is conditioned on the fully denoised and cached result of the first, providing it with a high-value starting point. This insight leads to a logical and highly effective optimization: applying the full 4-step denoising process to subsequent, high-quality chunks is computationally redundant.

Therefore, we introduce an \textbf{Asymmetric Denoising} strategy. Only the initial chunk is processed using the complete 4-step denoising path to generate a robust anchor cache. All subsequent chunks (Chunk $2 \dots C_{N/M}$) are processed using an accelerated 2-step path. This simple modification drastically enhances generation efficiency by leveraging the cache's power. However, as this aggressive acceleration is not strictly lossless, we introduce it as an optional, high-speed variant named \textbf{HiStream+}. 
\section{Experiments}
\label{sec:exp}

\begin{table*}[t!]
\centering
\small
\caption{\textbf{Quantitative comparisons with baselines.} HiStream achieves the best or second-best scores for all metrics in high-resolution generation. The best results are marked in \textbf{bold}, and the second-best results are marked by \underline{underline}.}
\vspace{-2mm}
\label{tab:baseline}
\scalebox{0.9}{\begin{tabular}{lccccccc}
\toprule
 Method  &\text{\#Params} & frames & \makecell{Resolution \\ (W×H)} & \makecell{Quality \\ Score $\uparrow$} & \makecell{Semantic \\ Score $\uparrow$} & \makecell{Total \\ Score $\uparrow$} & \makecell{Per-Frame \\ \textbf{Denoising} Latency (s) $\downarrow$} \\ 
\midrule
\rowcolor{lightgray}
\multicolumn{8}{l}{\textit{Low-Resolution Generation (For Reference Only)}} \\
Wan2.1~\citep{wan2025} &  1.3B &  81 & 832×480    &  84.13   &  80.05    &  83.32    &    2.04     \\ 
Self Forcing~\citep{huang2025self} &  1.3B &  81 &  832×480    &  84.85   &  80.91    &  84.06    &  0.10        \\ 
\midrule
\rowcolor{lightgray}
\multicolumn{8}{l}{\textit{High-Resolution Generation}} \\
Wan2.1~\citep{wan2025} &  1.3B &  81 &  1920×1088    &  81.74   &  76.39    &  80.67    &  36.56       \\ 
Self Forcing~\citep{huang2025self} &  1.3B &  81 &  1920×1088    &  \underline{84.65}   &  79.97    &  \underline{83.71}    &  \underline{1.18}     \\ 
LTX~\citep{HaCohen2024LTXVideo} &  2B &  81  &  1920×1088    &  80.31   &  69.54    &  78.16    &  1.60     \\ 
FlashVideo~\citep{zhang2025flashvideo} &  5B &  49  &  1920×1072    &  83.75   &  \textbf{81.62}    &  83.32    &  6.40     \\ 
% LTX-new~\citep{HaCohen2024LTXVideo} &  13B &  81  &  1920×1088    &  84.05   &  77.21    &  82.68    &  0.68     \\ 
HiStream (Ours) &  1.3B  &  81 &   1920×1088     &  \textbf{85.00}    & \underline{80.97}     &  \textbf{84.20} & \textbf{0.48}  \\ 
\bottomrule
\end{tabular}}
% \vspace{-2mm}
\end{table*}

\begin{figure*}[t!]
\centering
\setlength{\tabcolsep}{0.1em}  
\renewcommand{\arraystretch}{0.2}
 \begin{tabular}{C{0.09\linewidth} C{0.145\linewidth} C{0.145\linewidth} C{0.145\linewidth} @{\hspace{0.5em}} C{0.145\linewidth} C{0.145\linewidth} C{0.145\linewidth}}
  Wan2.1 & 
  \includegraphics[width=\linewidth]{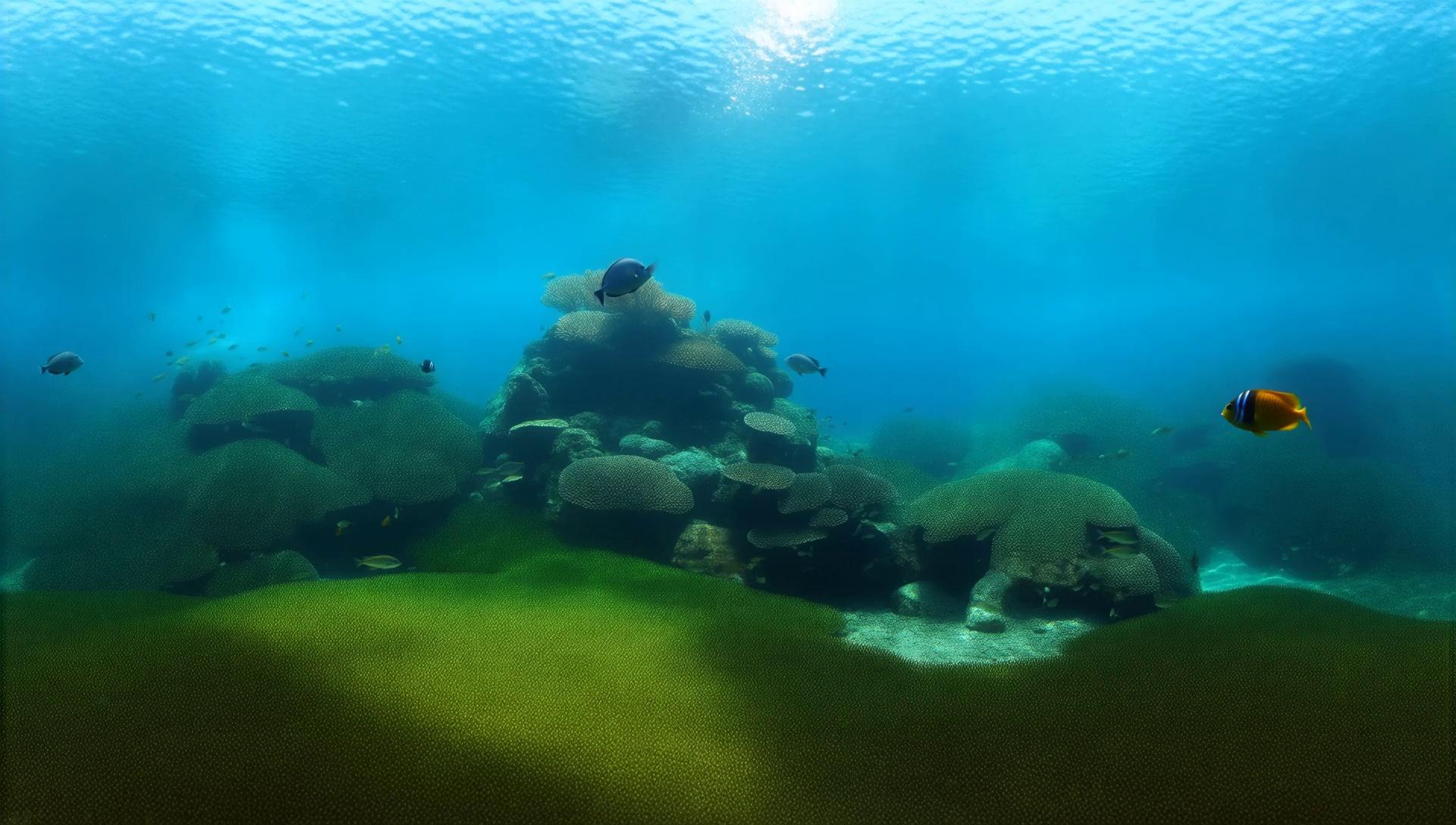} & 
  \includegraphics[width=\linewidth]{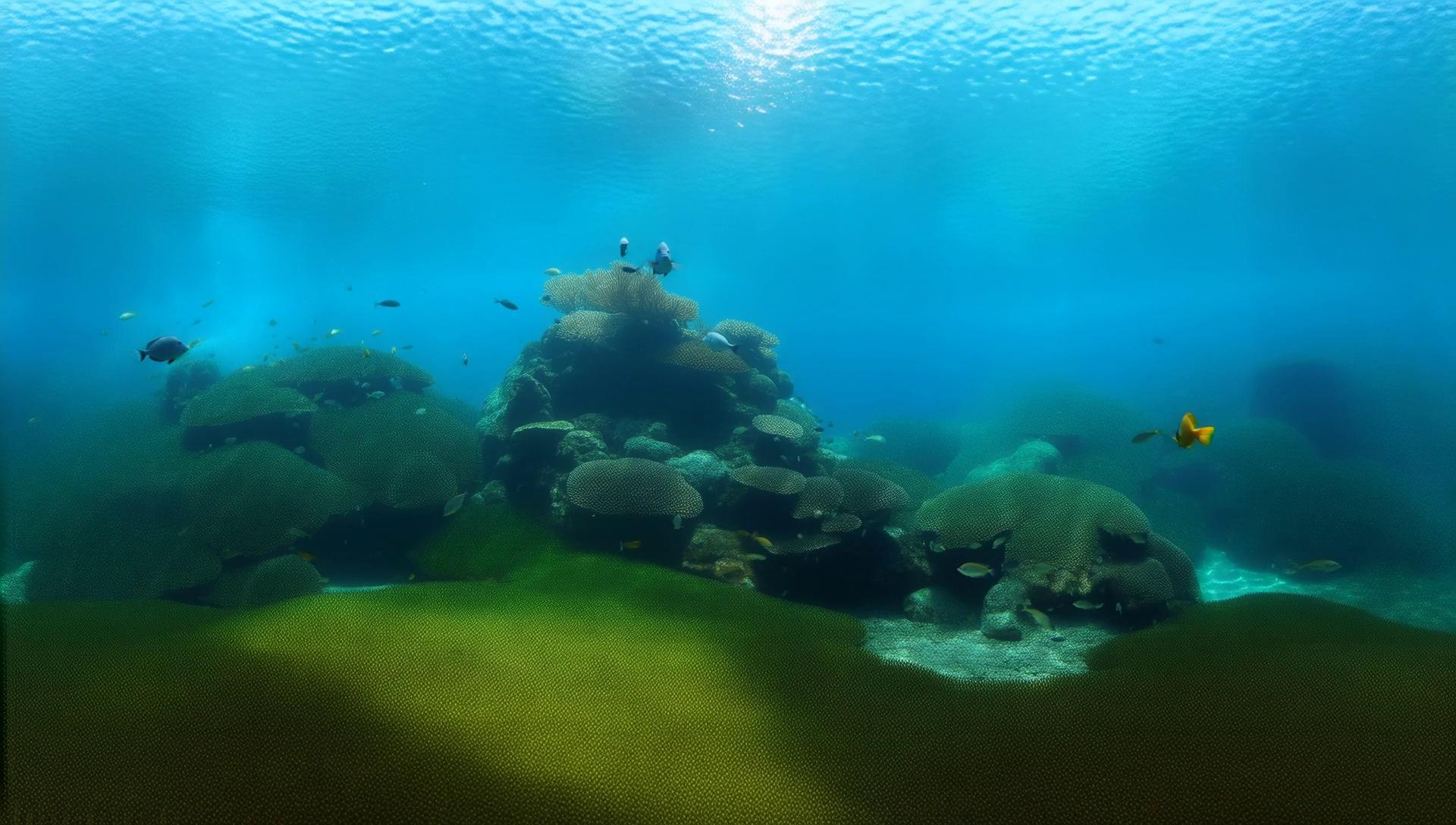} & 
  \includegraphics[width=\linewidth]{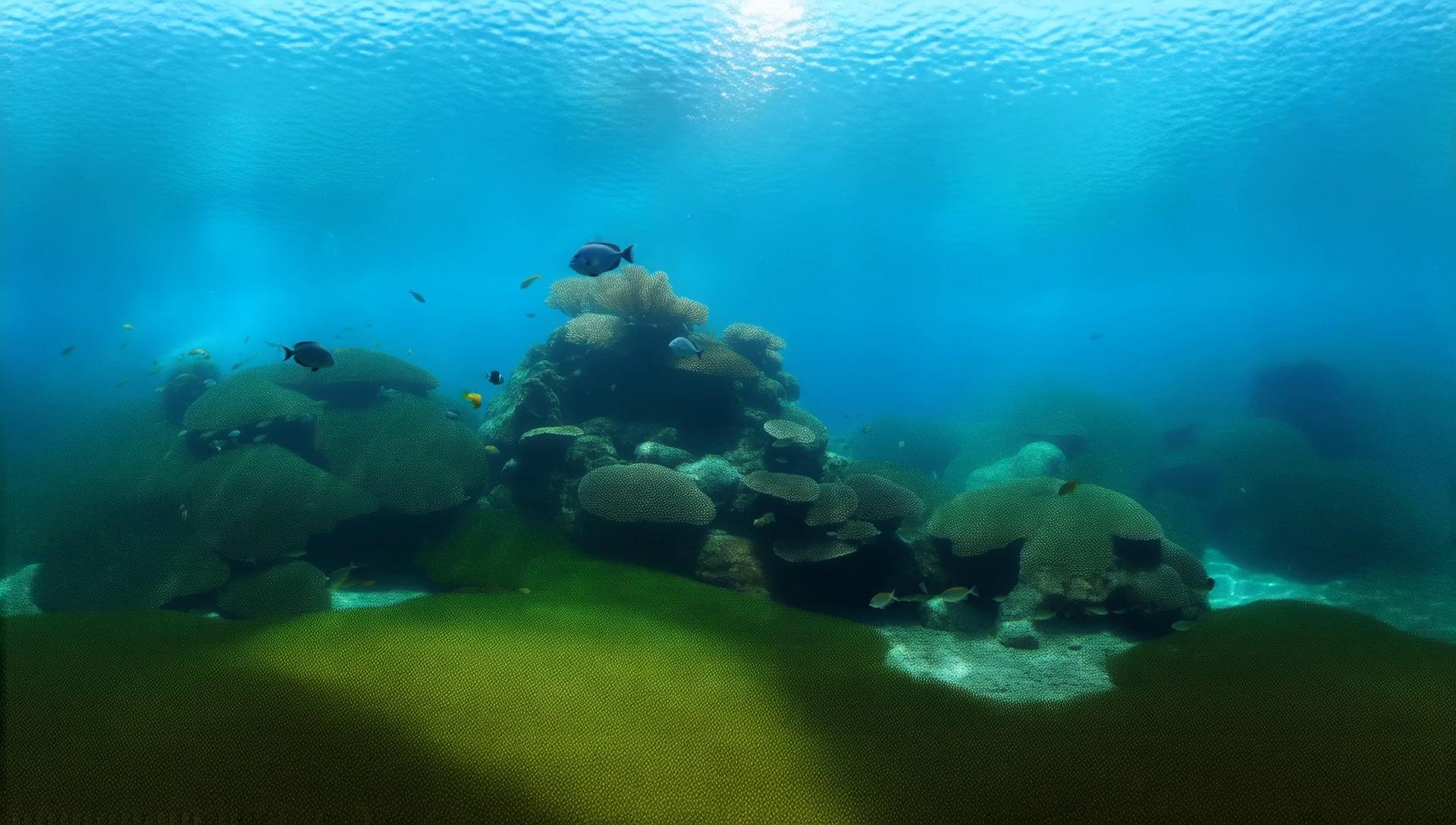} & 
  \includegraphics[width=\linewidth]{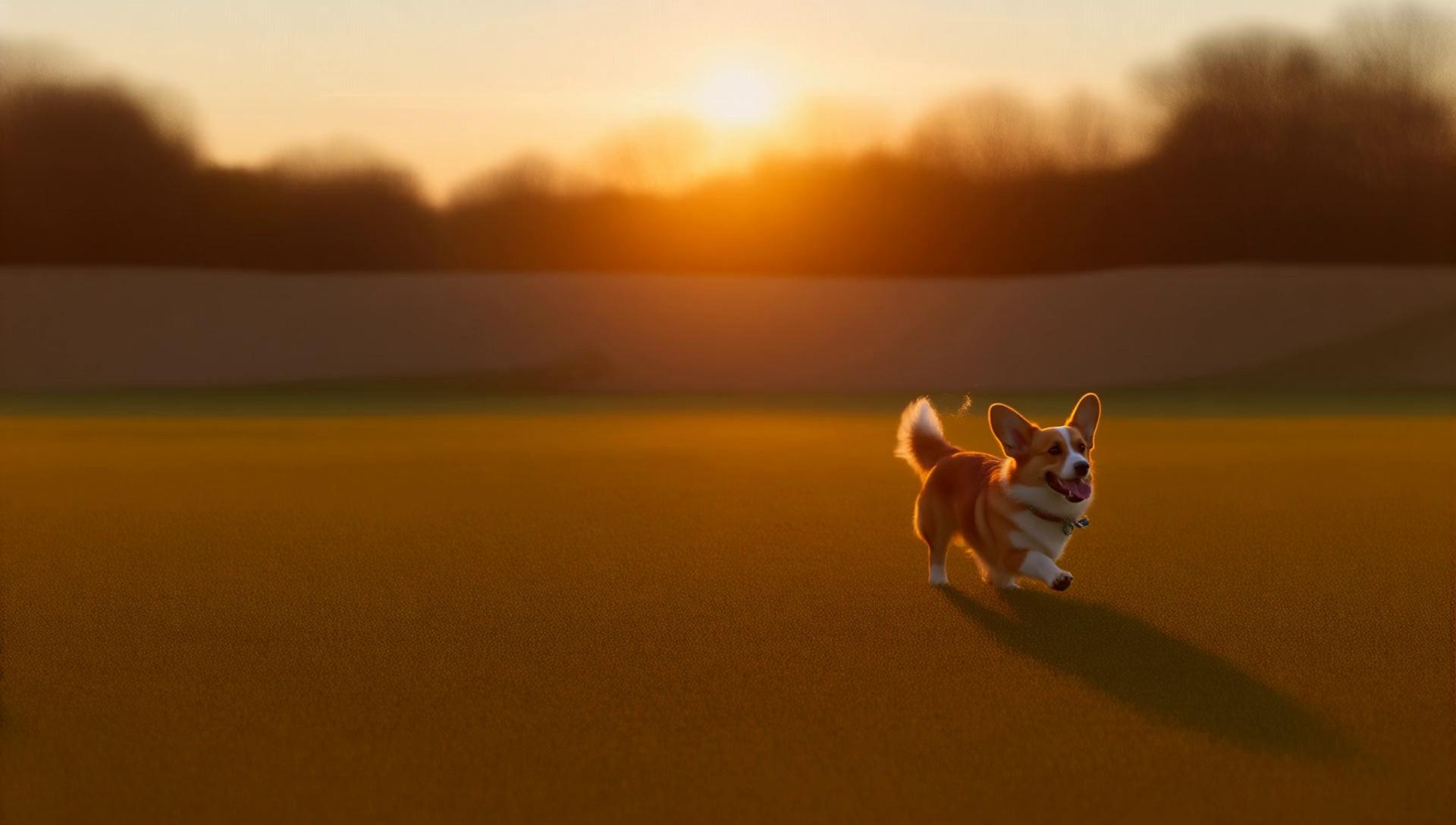} & 
  \includegraphics[width=\linewidth]{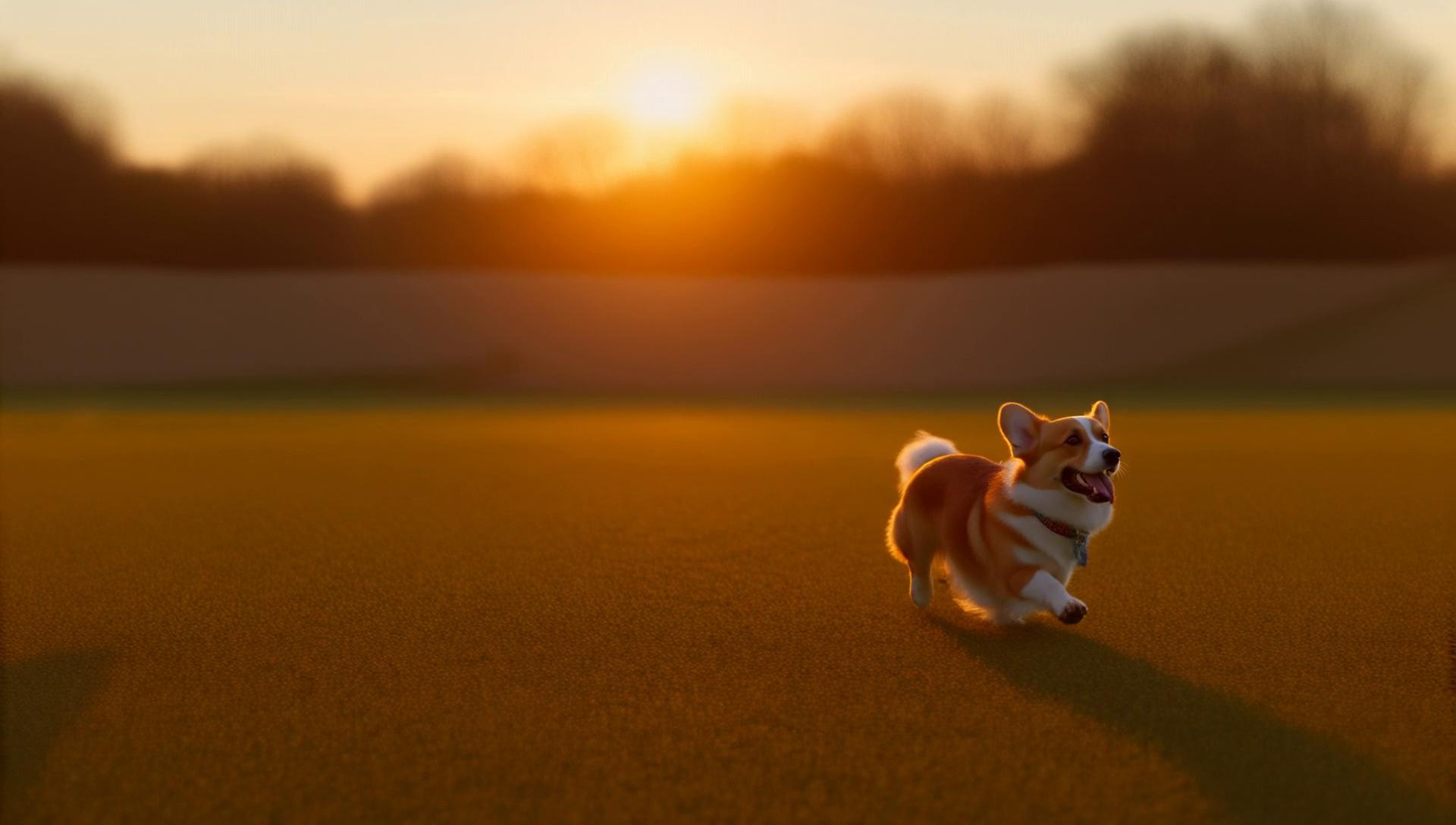} & 
  \includegraphics[width=\linewidth]{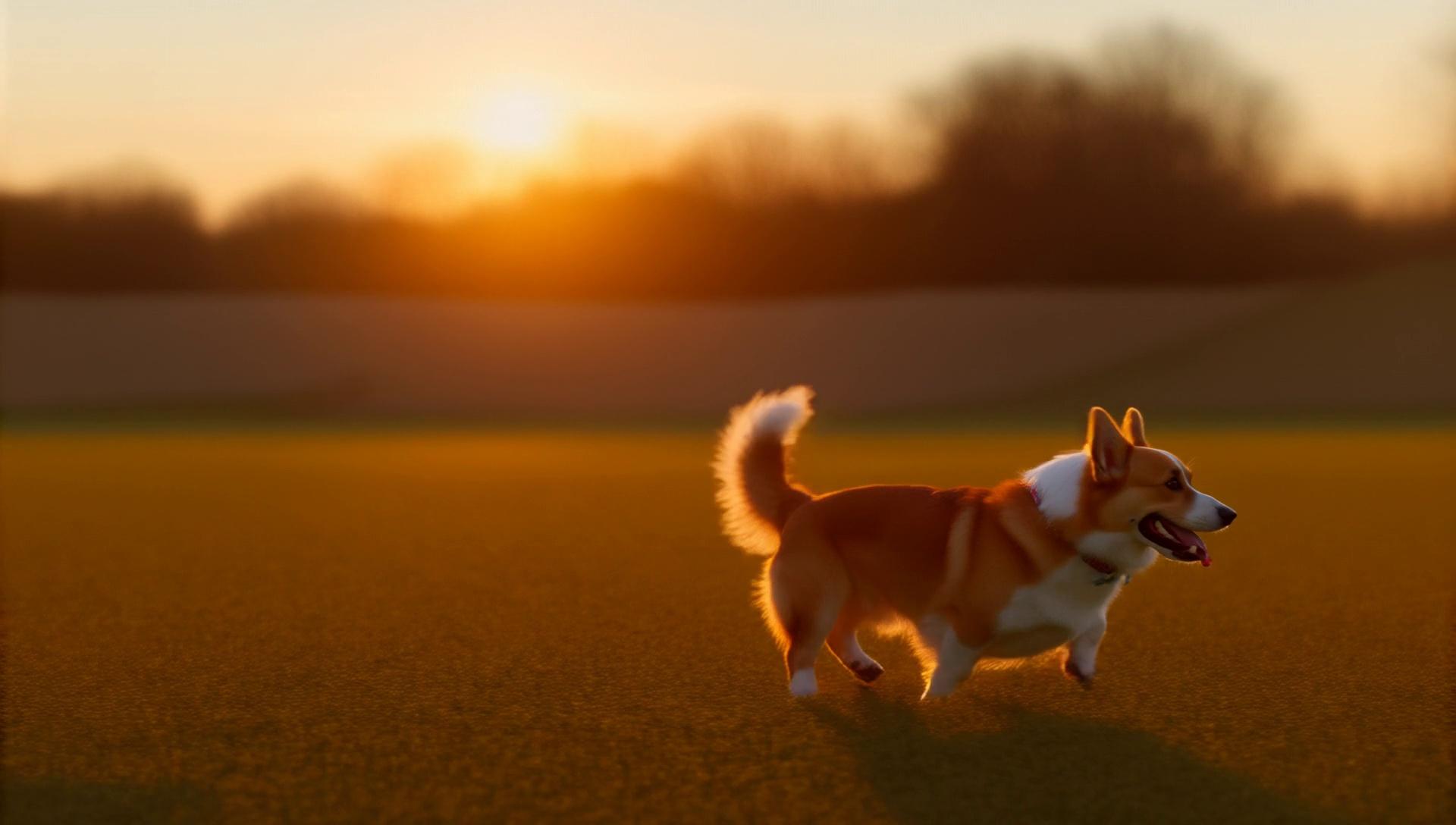} \\
  Self Forcing & 
  \includegraphics[width=\linewidth]{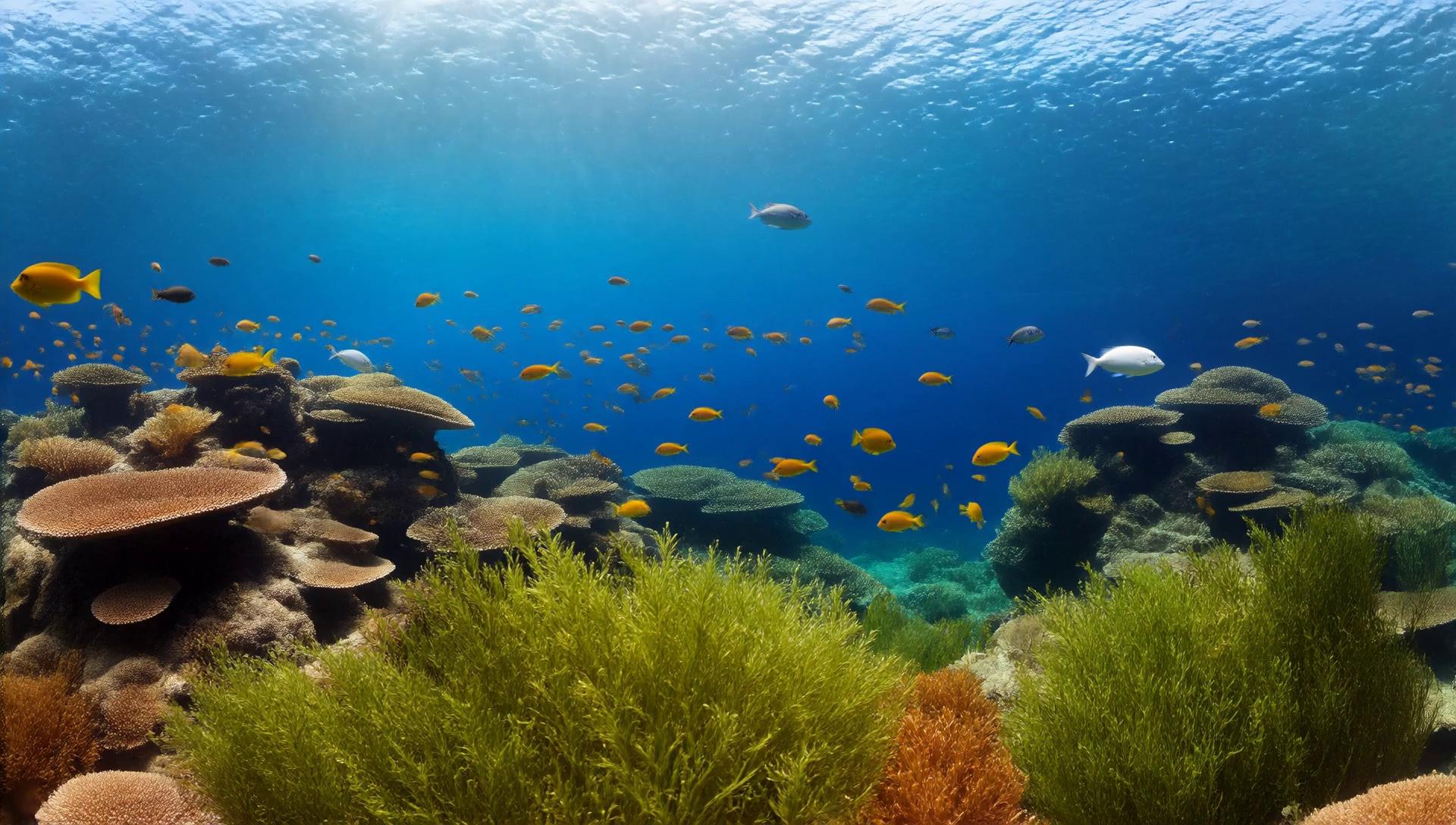} & 
  \includegraphics[width=\linewidth]{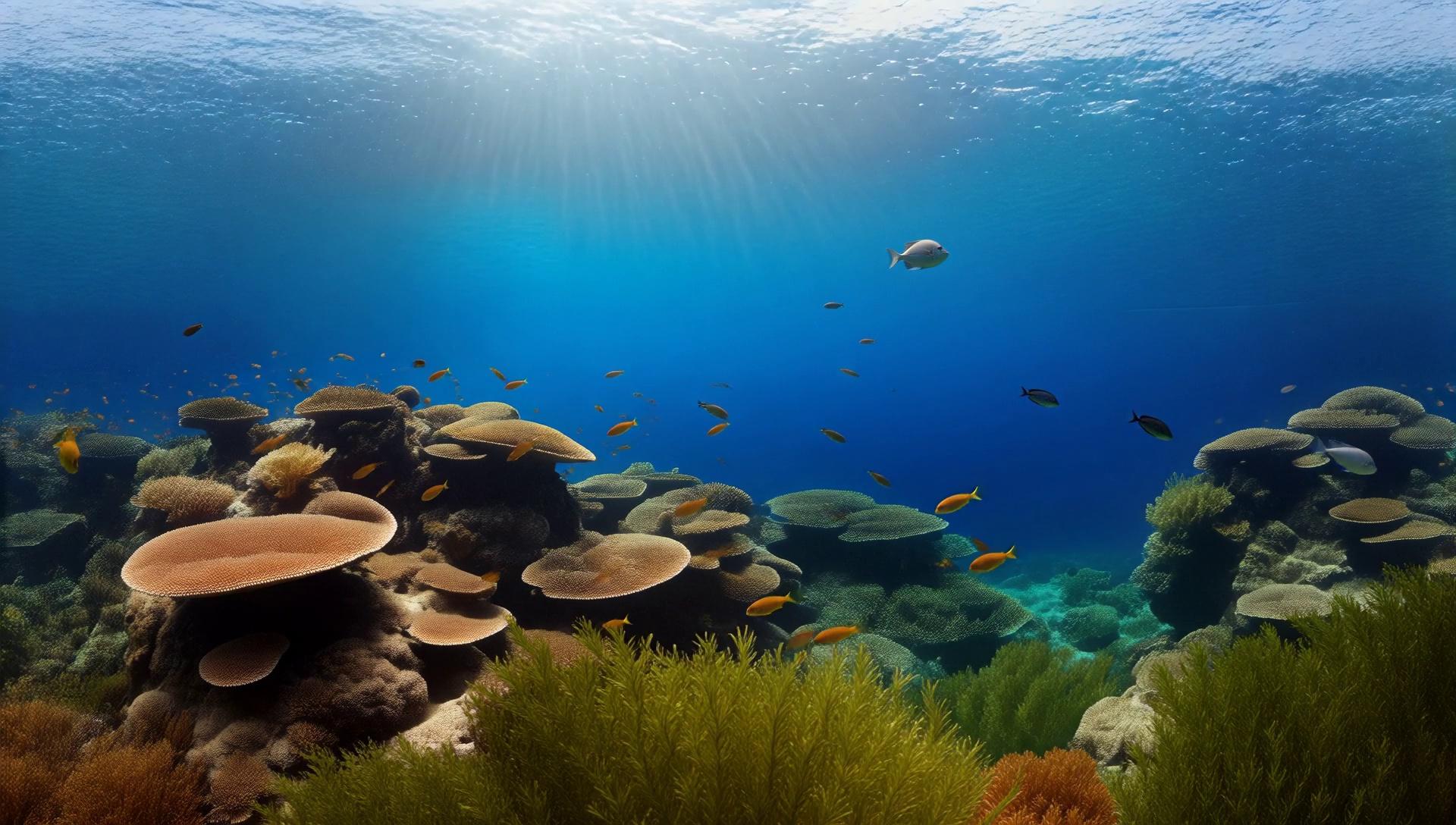} & 
  \includegraphics[width=\linewidth]{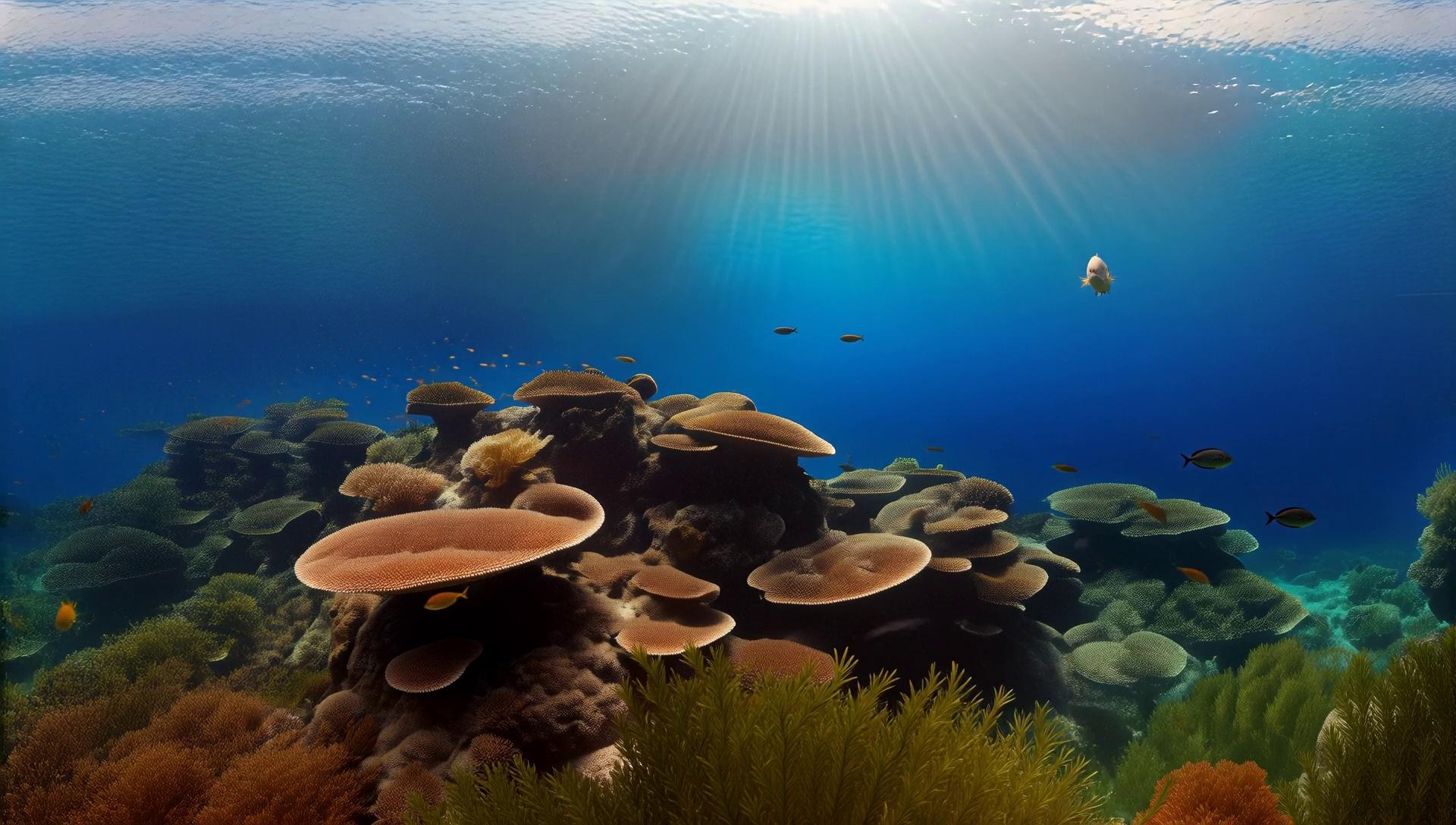} & 
  \includegraphics[width=\linewidth]{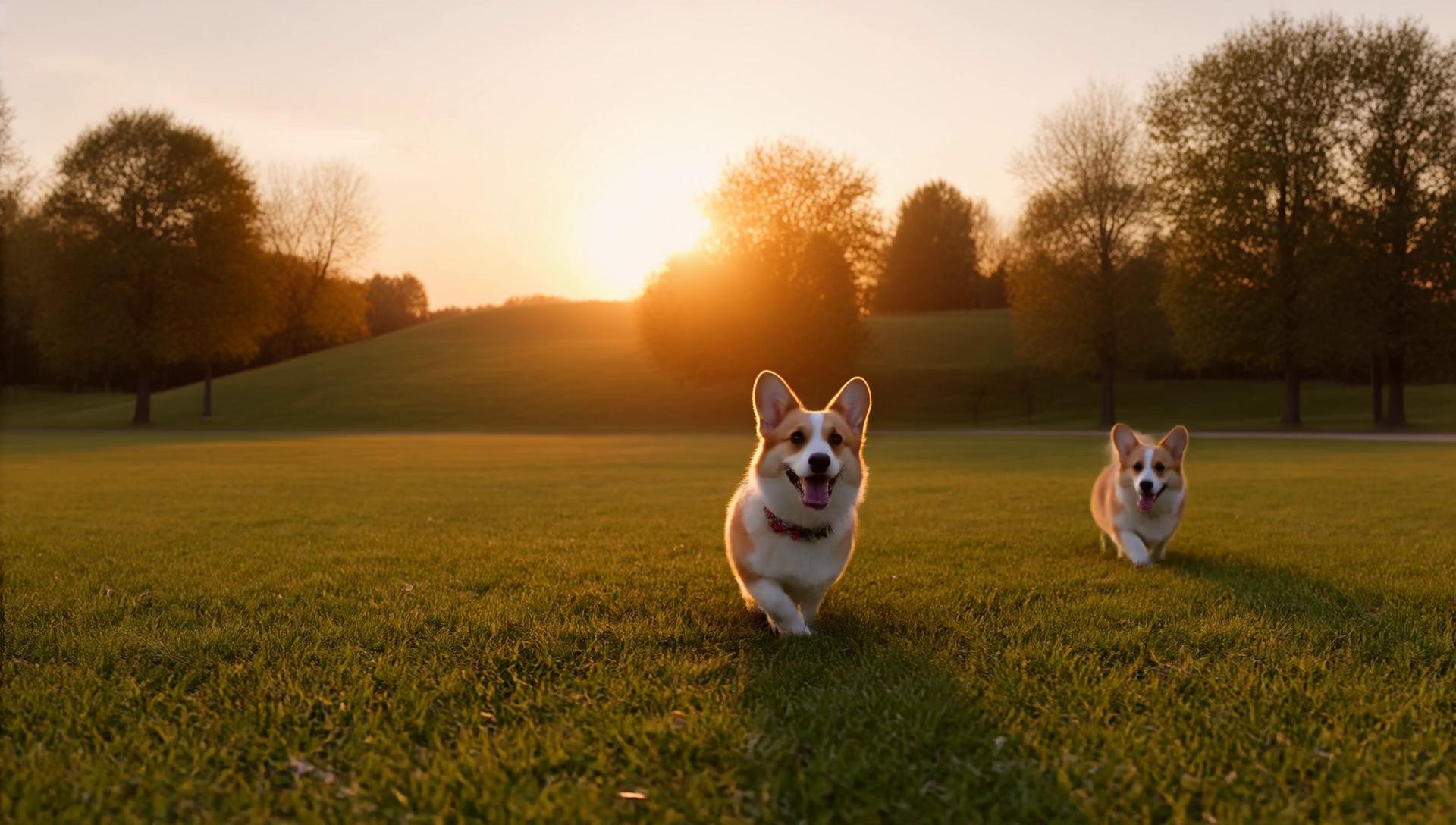} & 
  \includegraphics[width=\linewidth]{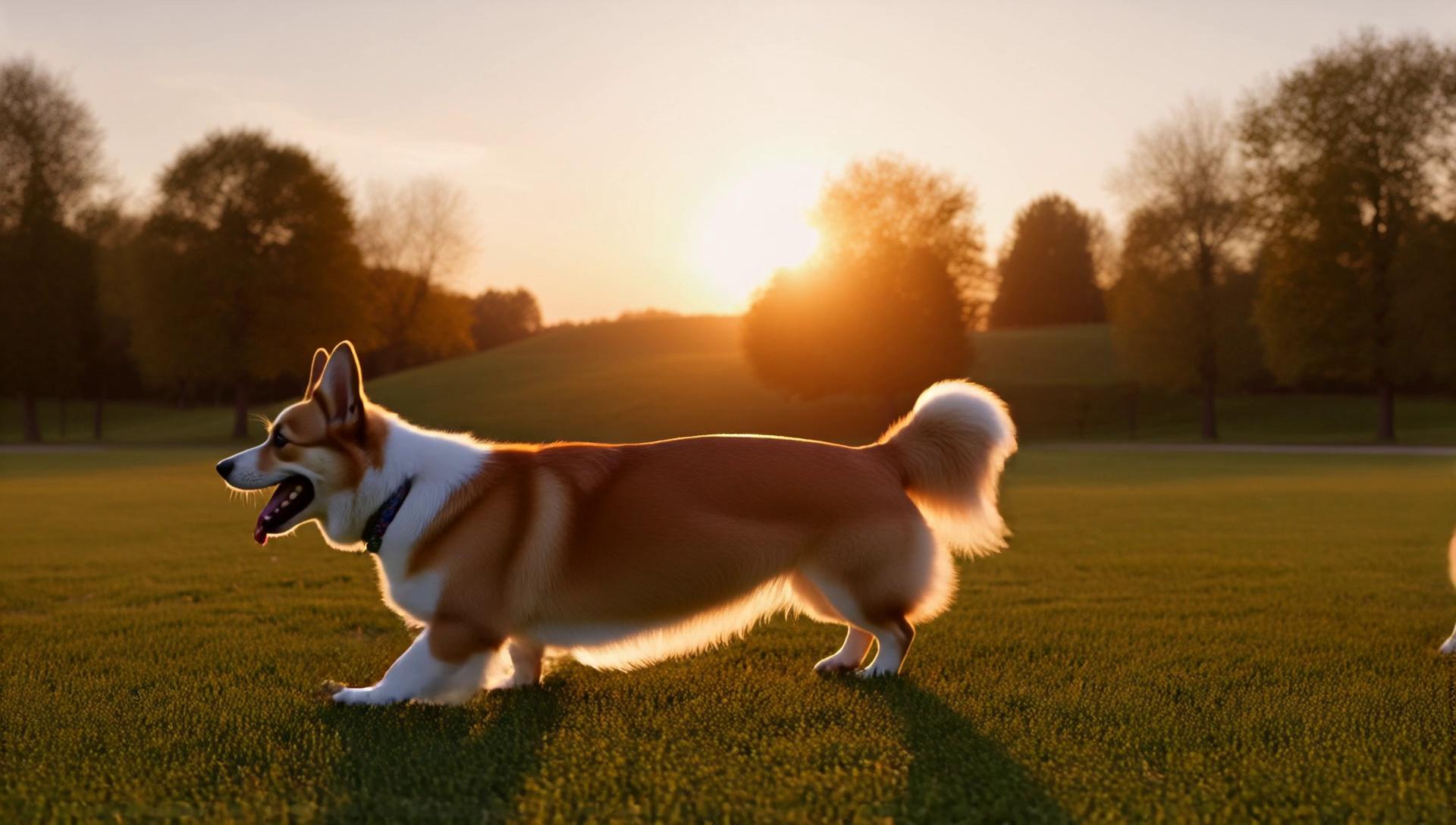} & 
  \includegraphics[width=\linewidth]{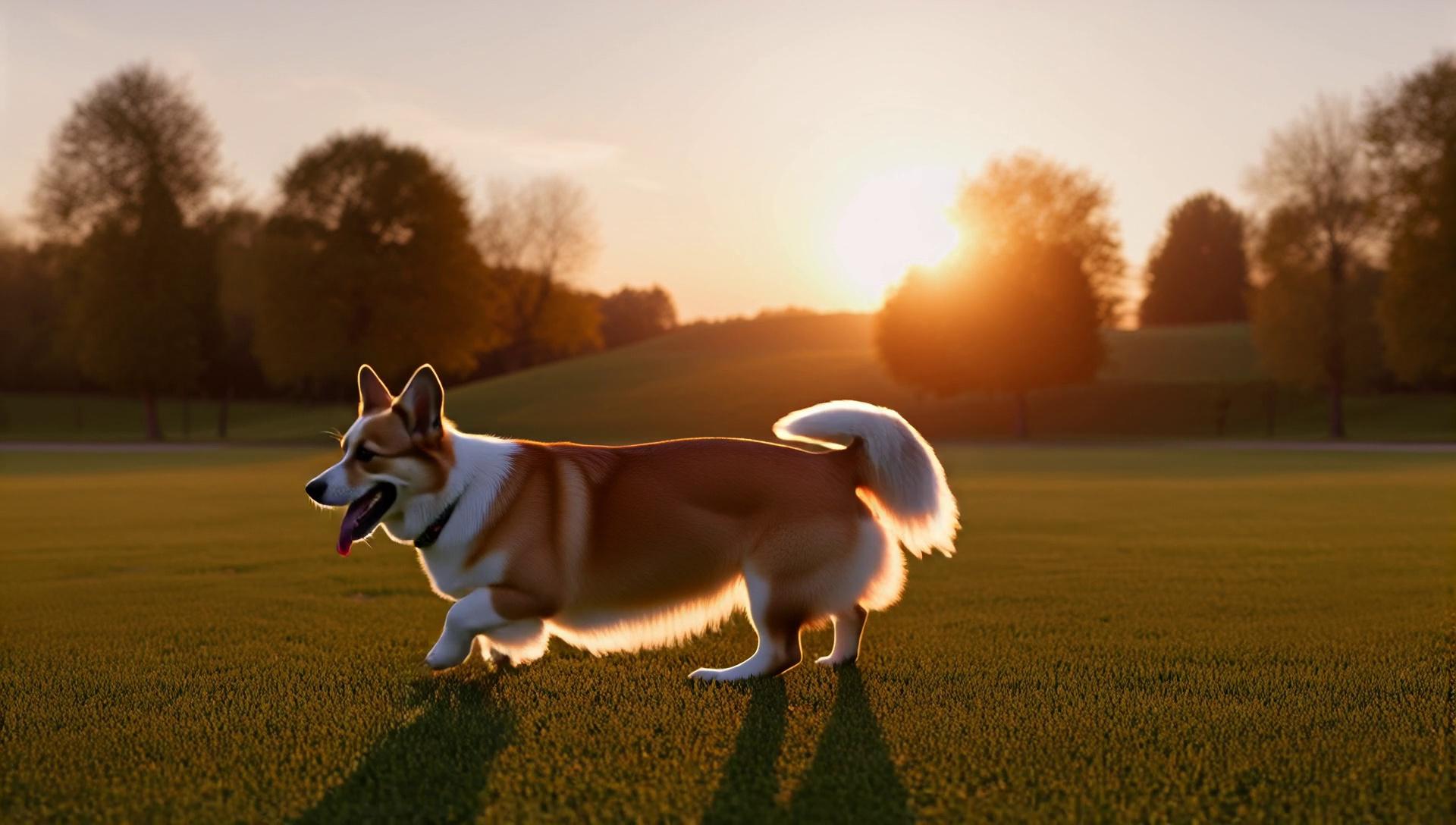} \\
  LTX & 
  \includegraphics[width=\linewidth]{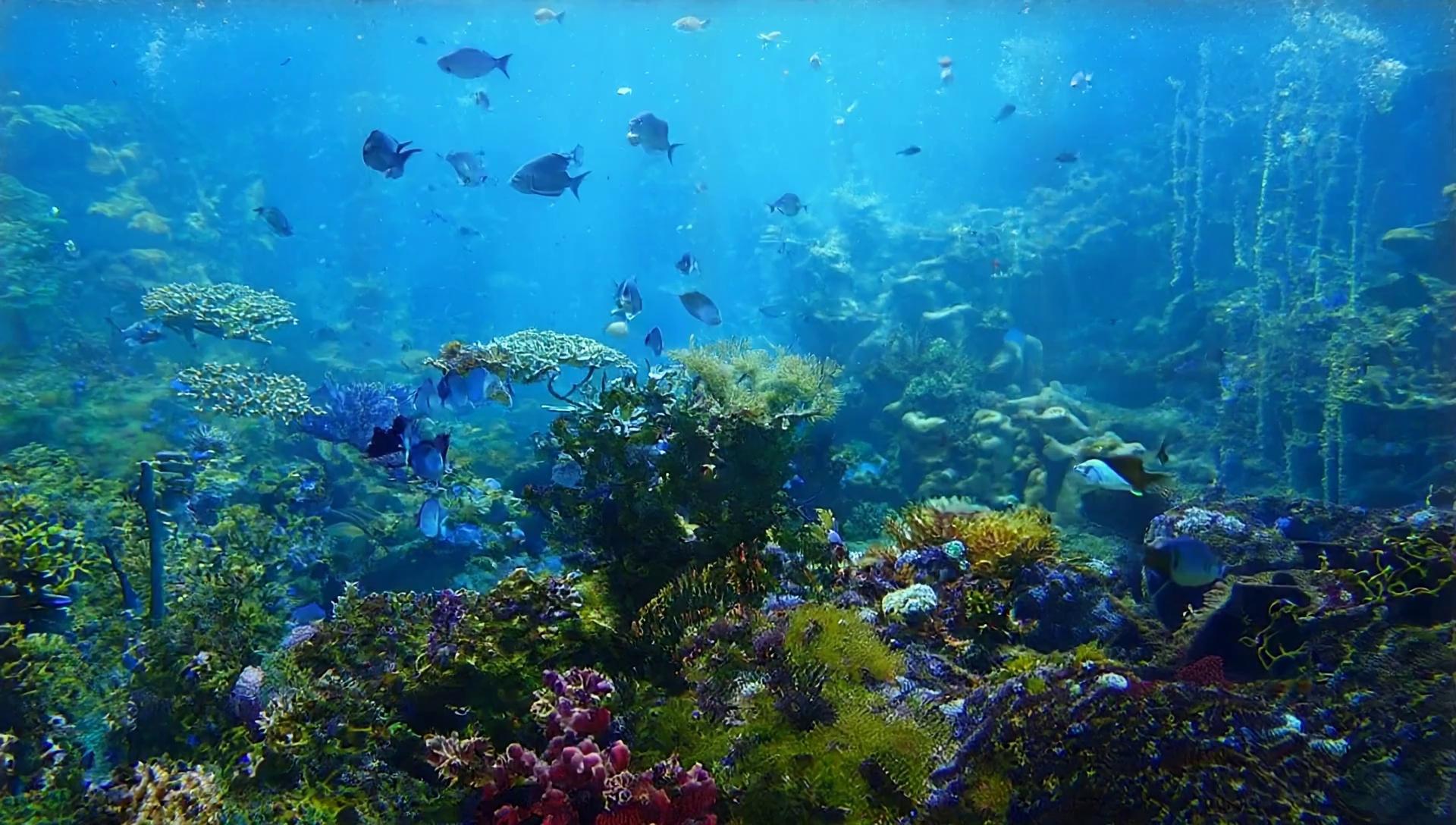} & 
  \includegraphics[width=\linewidth]{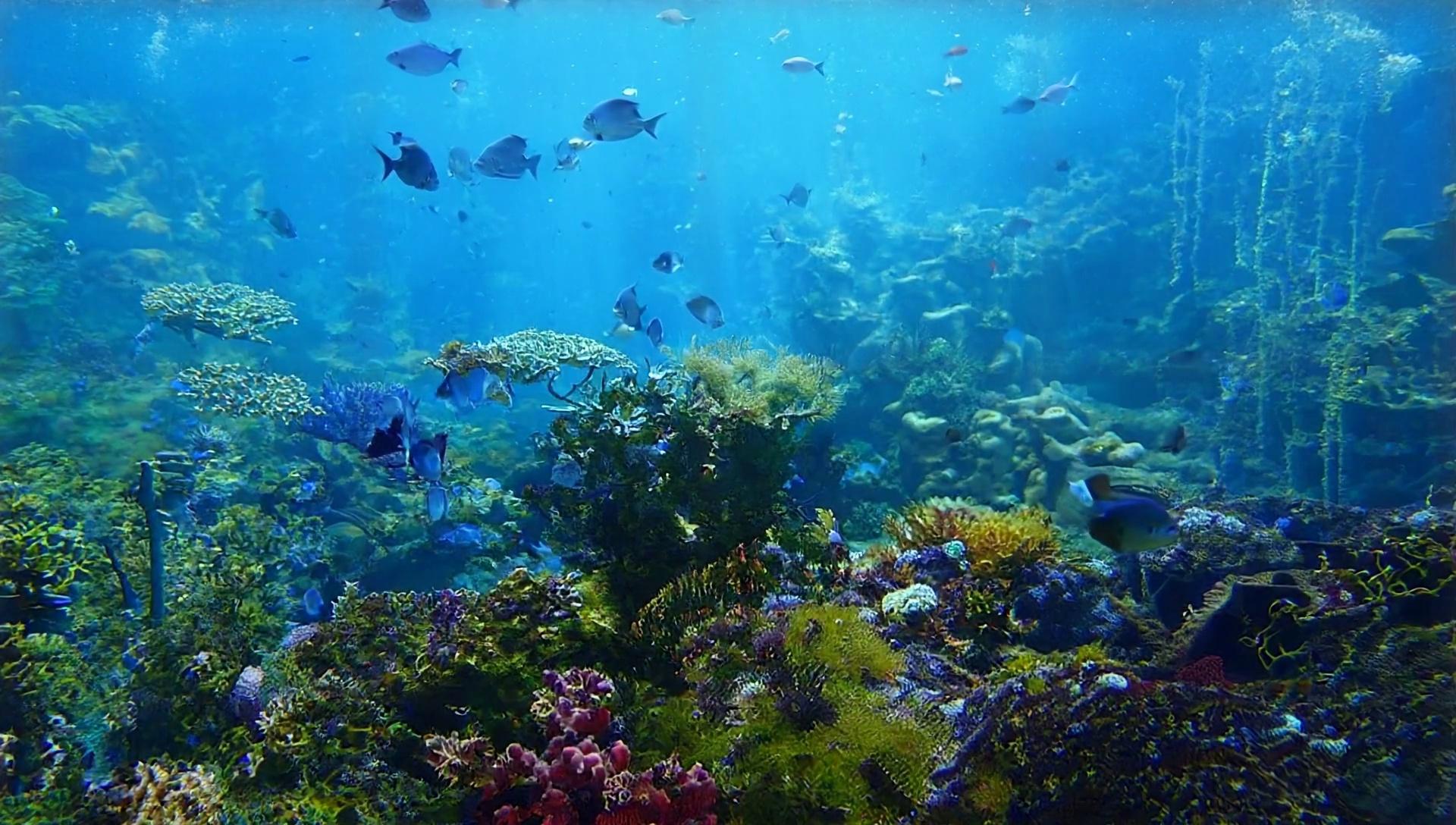} & 
  \includegraphics[width=\linewidth]{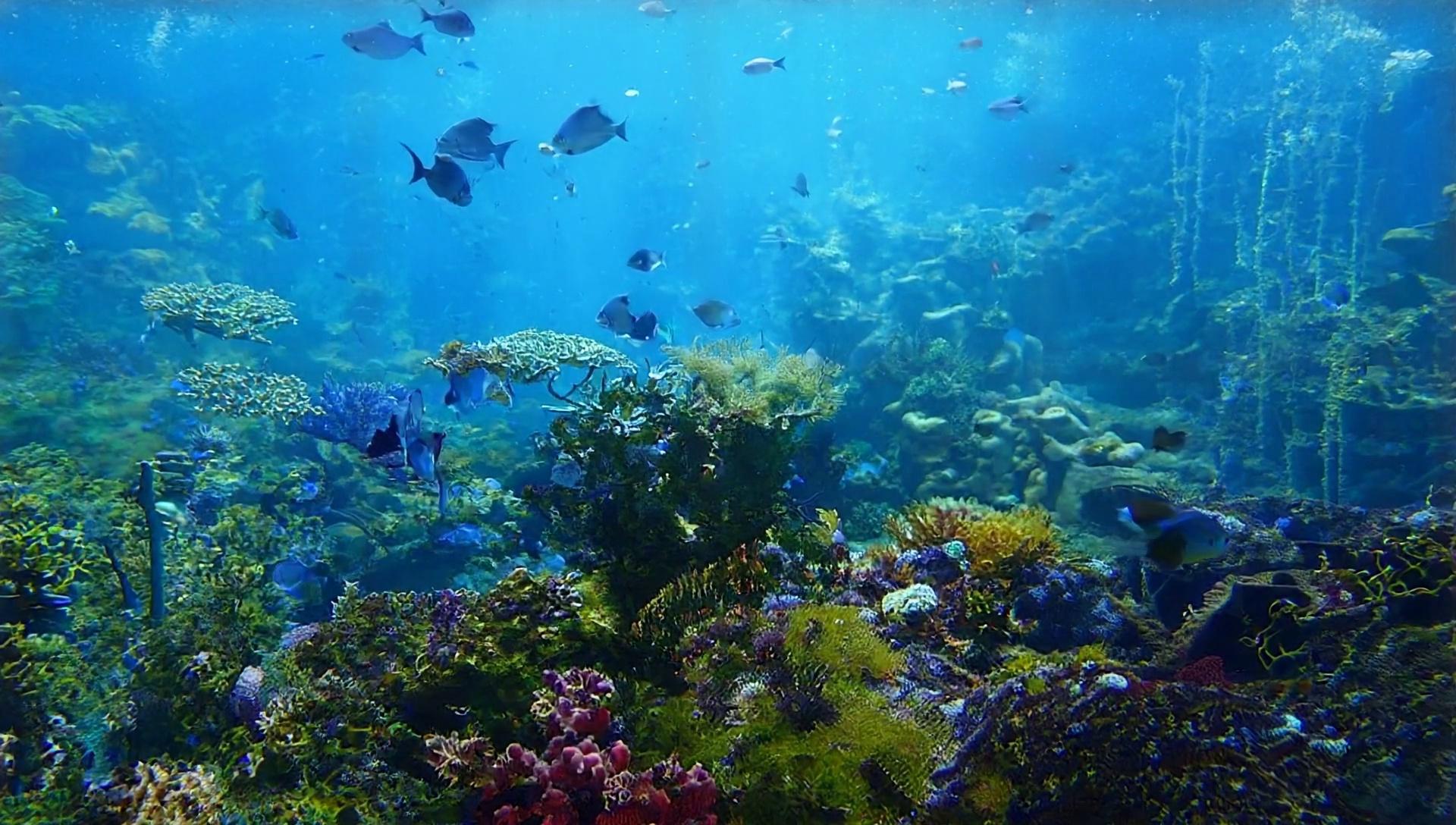} & 
  \includegraphics[width=\linewidth]{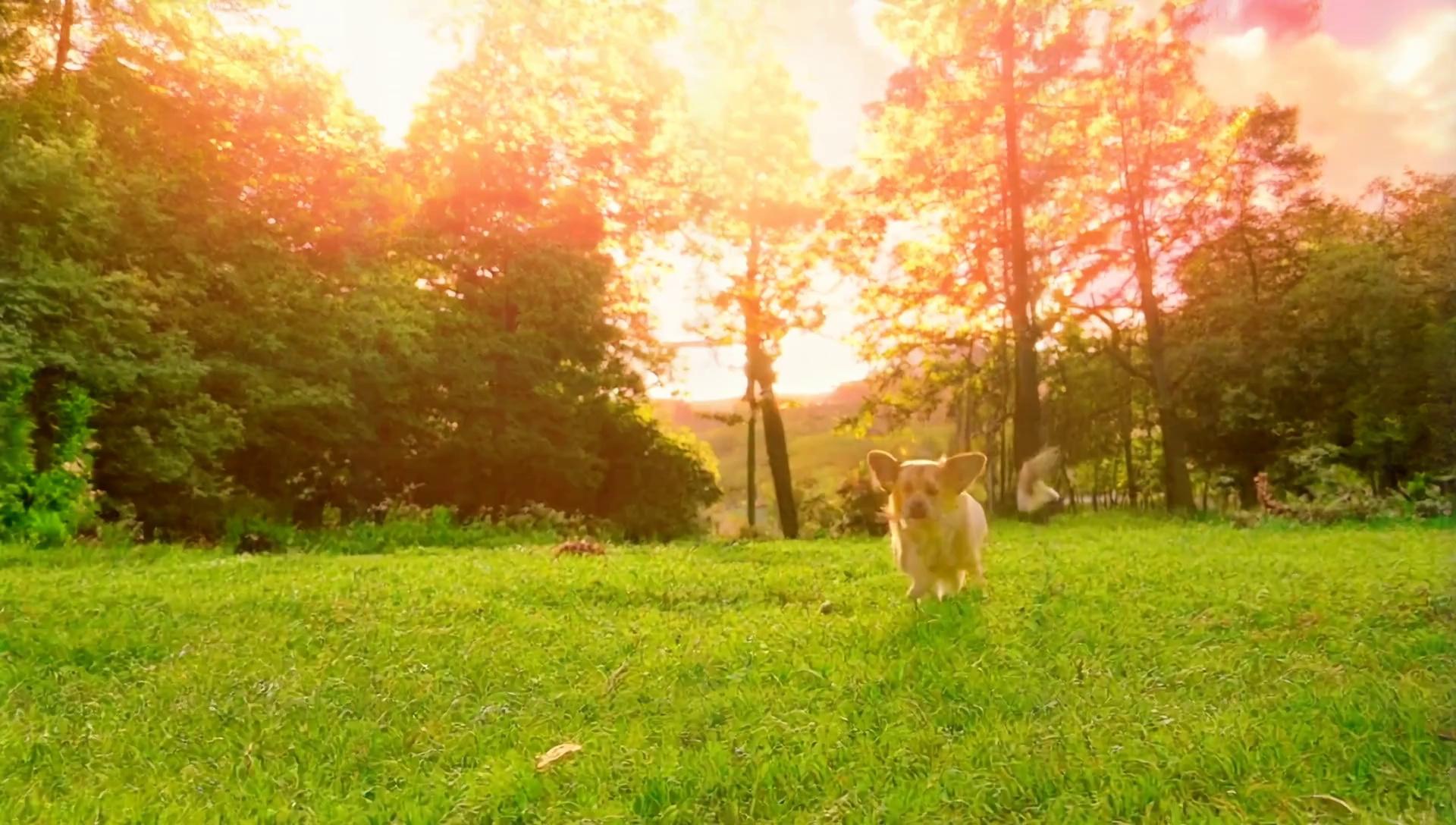} & 
  \includegraphics[width=\linewidth]{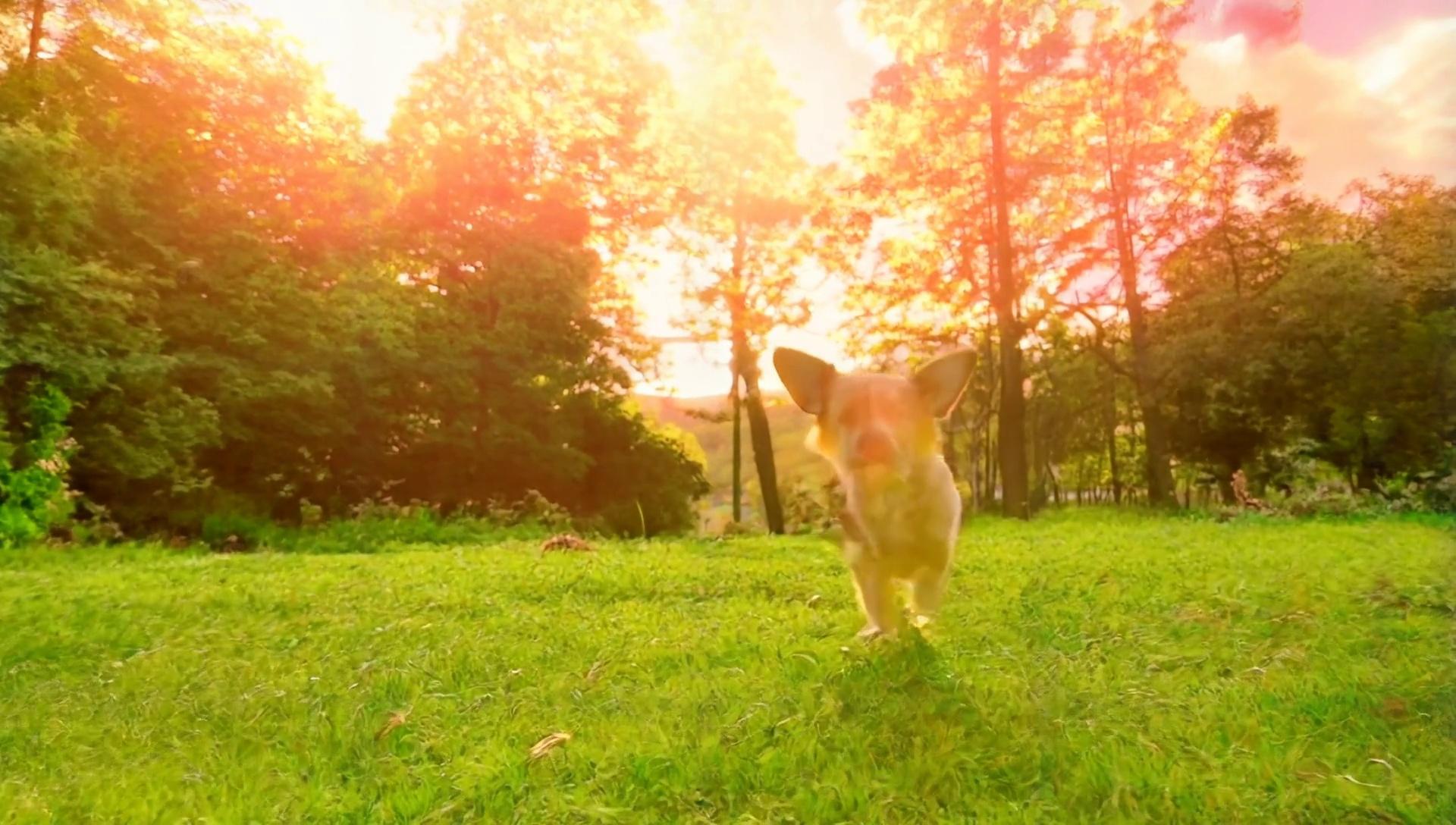} & 
  \includegraphics[width=\linewidth]{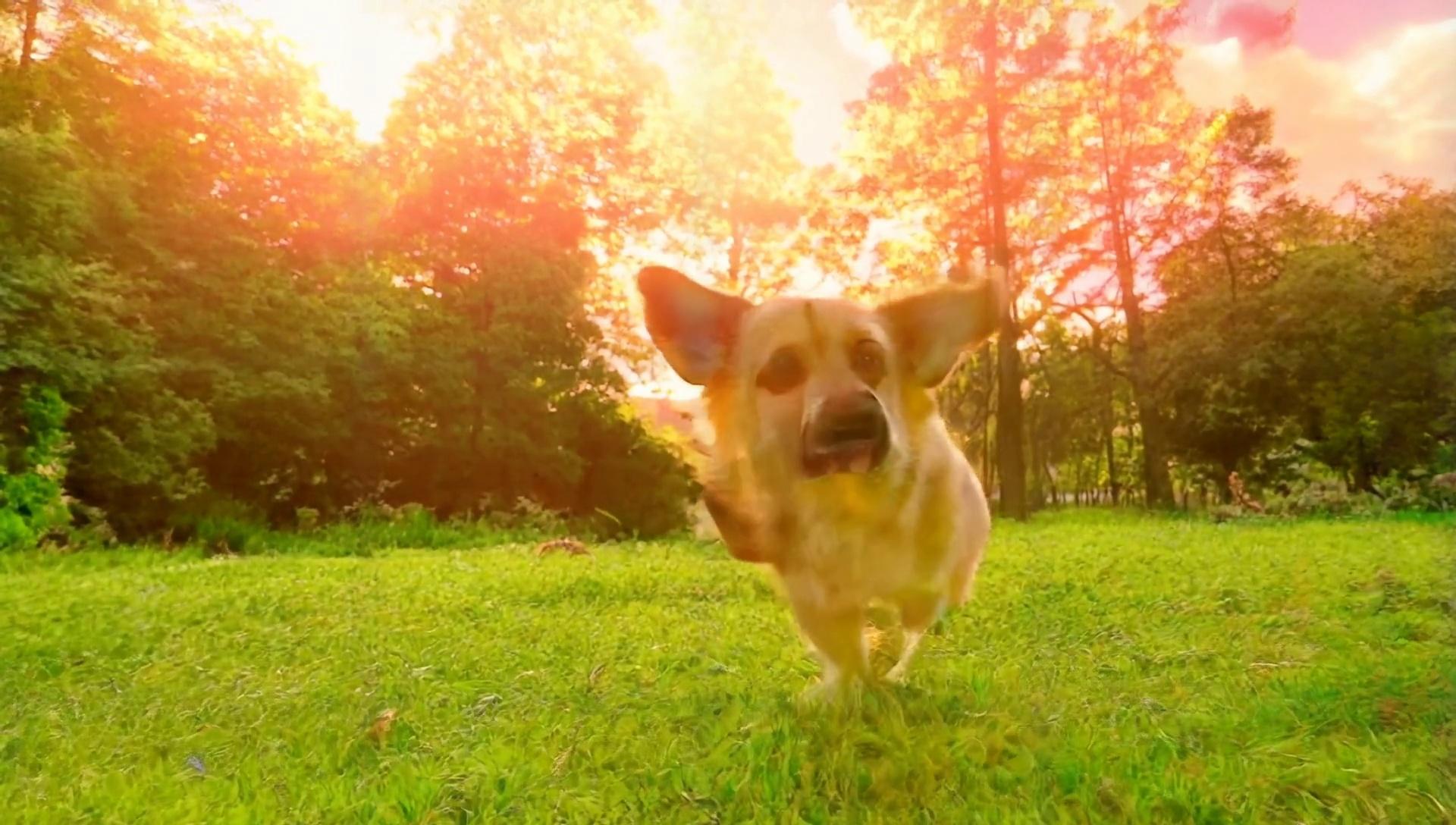} \\  
  Flash
  Video & 
  \includegraphics[width=\linewidth]{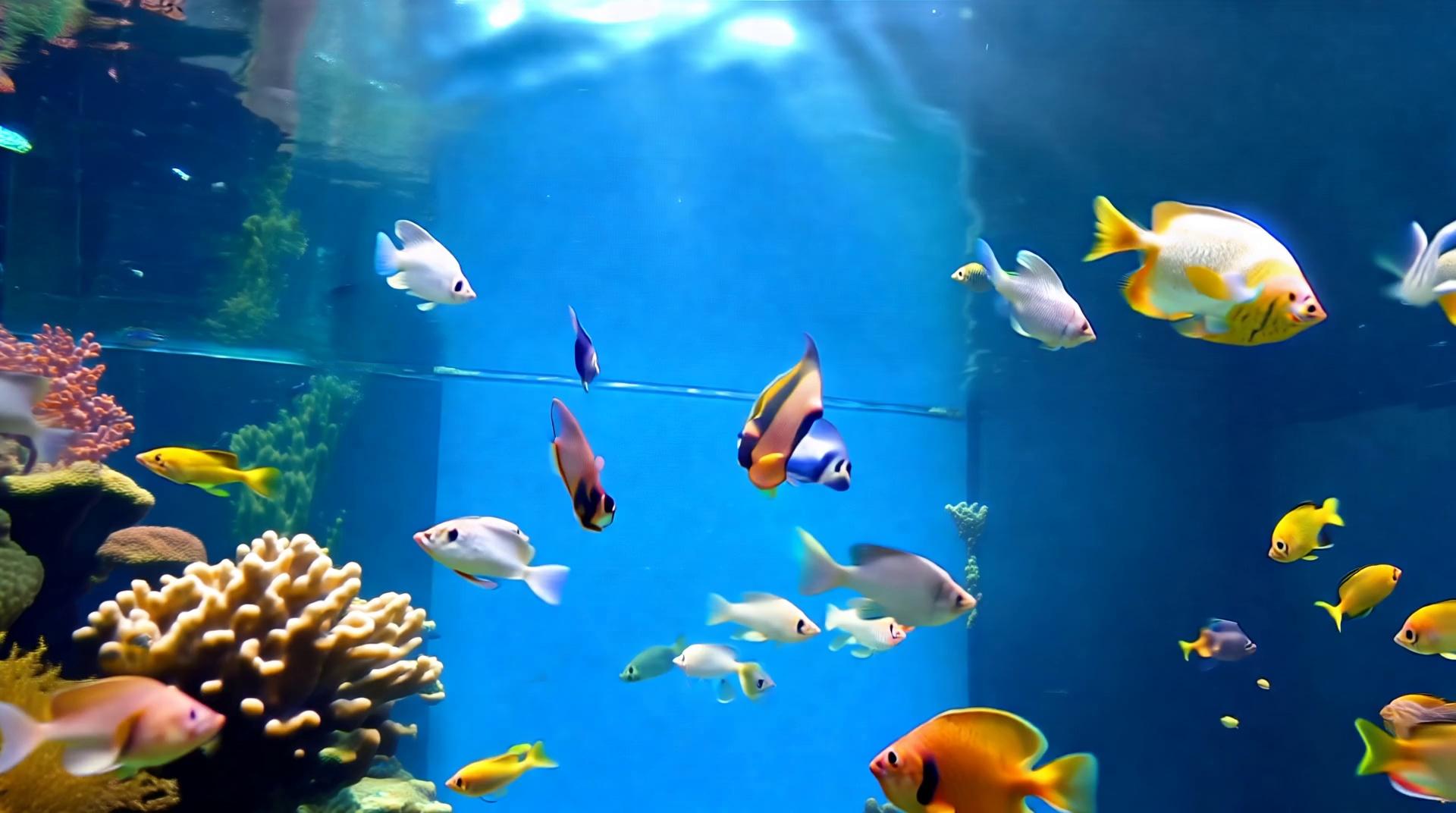} & 
  \includegraphics[width=\linewidth]{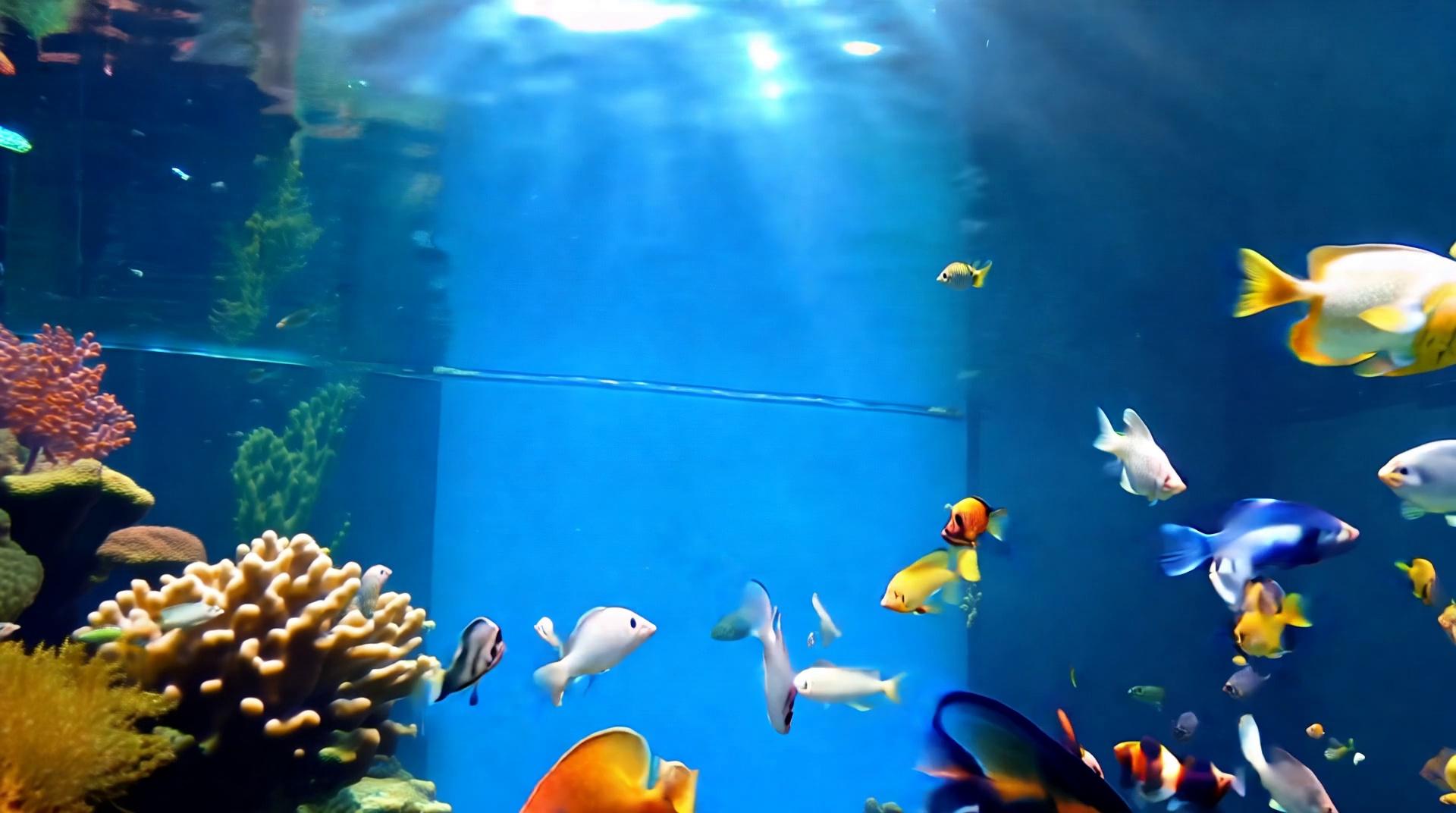} & 
  \includegraphics[width=\linewidth]{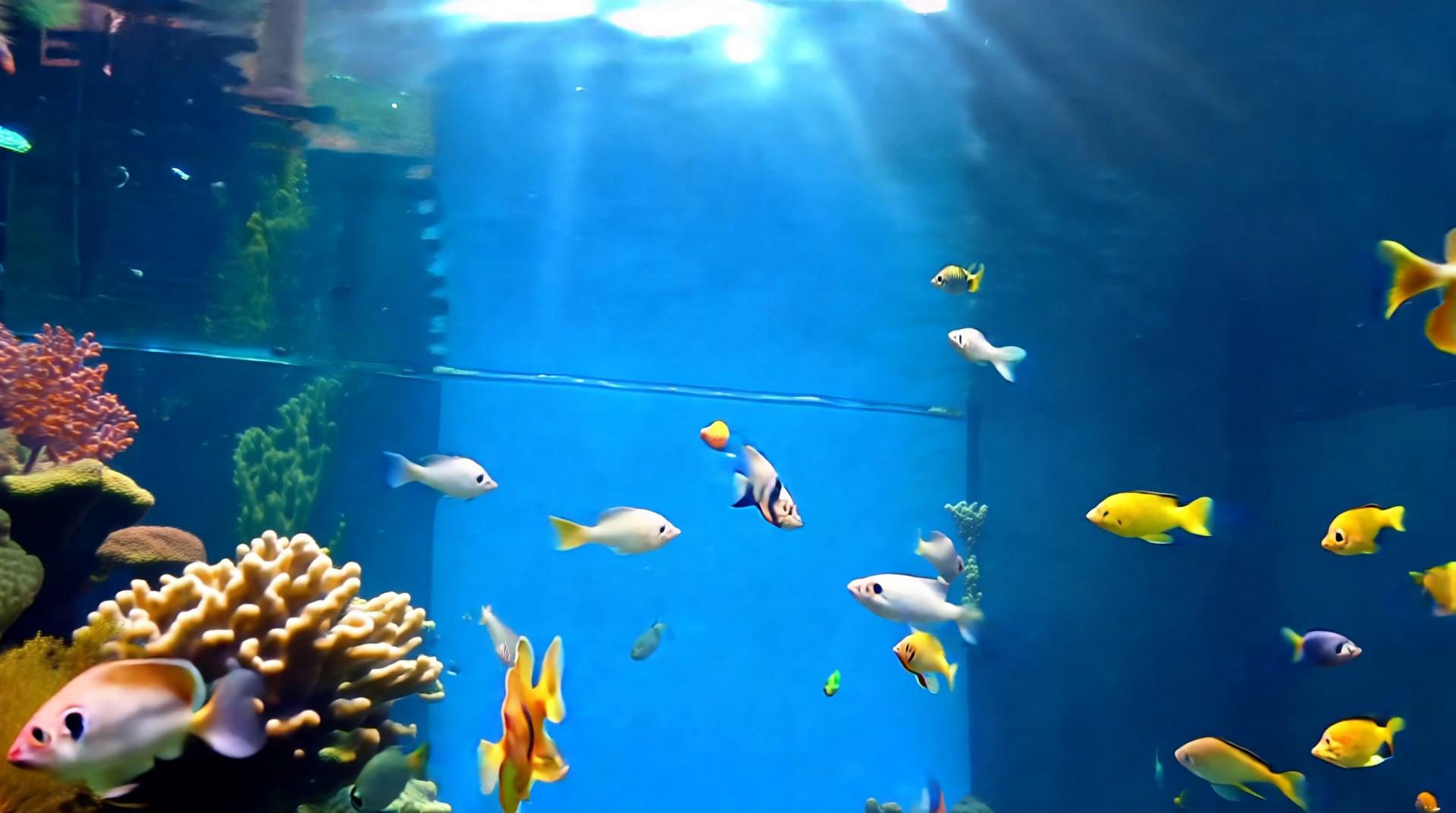} & 
  \includegraphics[width=\linewidth]{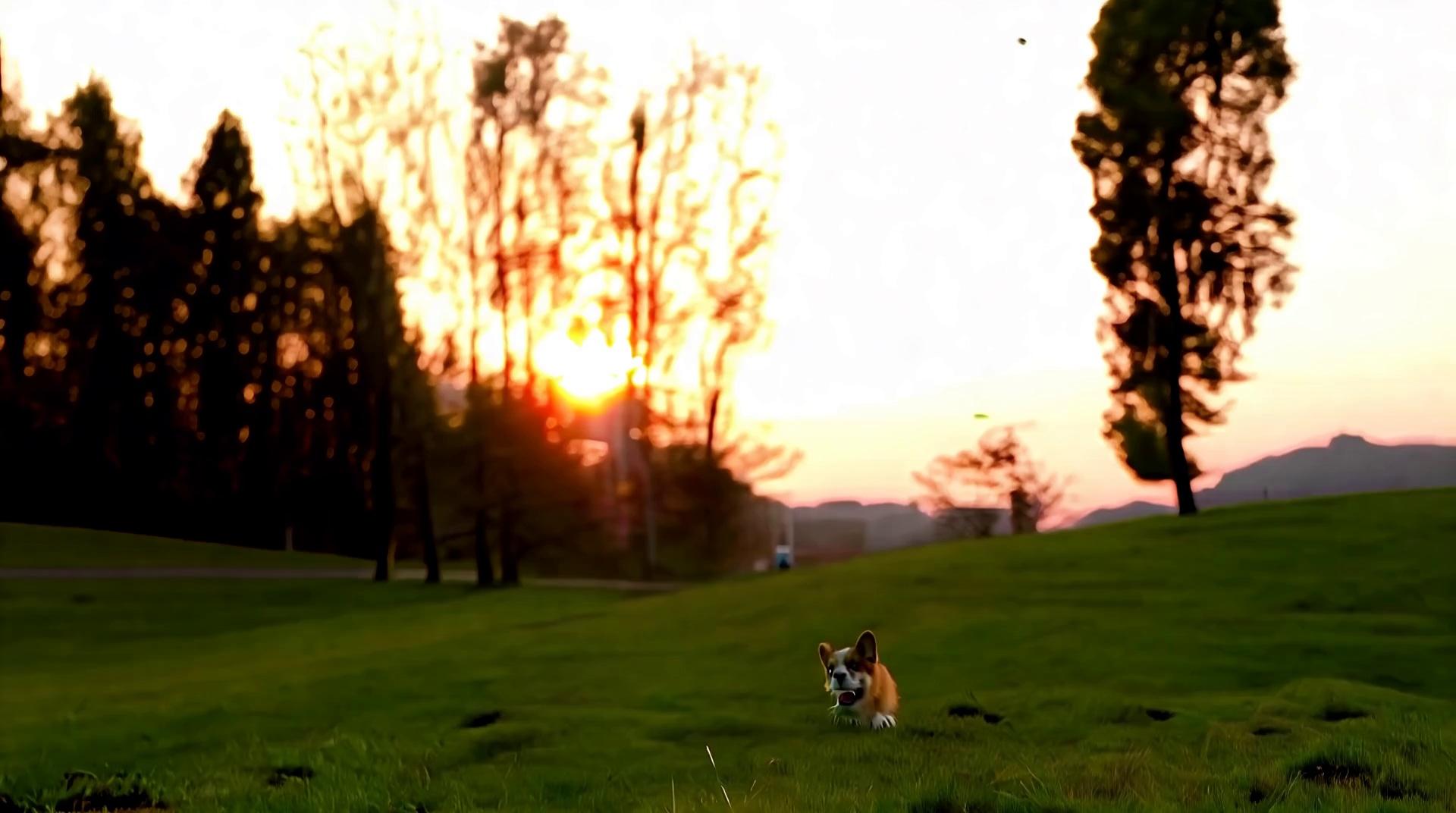} & 
  \includegraphics[width=\linewidth]{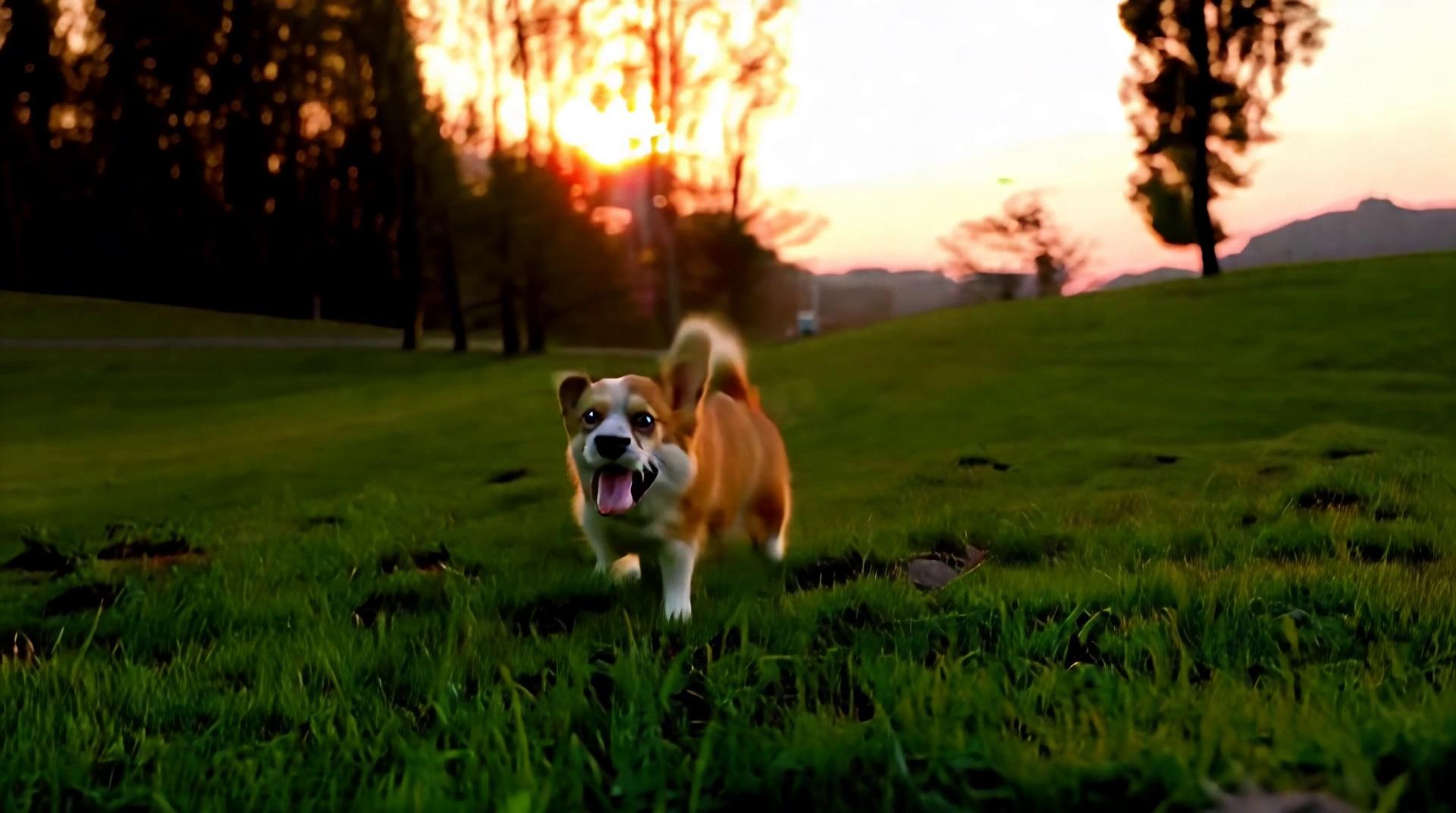} & 
  \includegraphics[width=\linewidth]{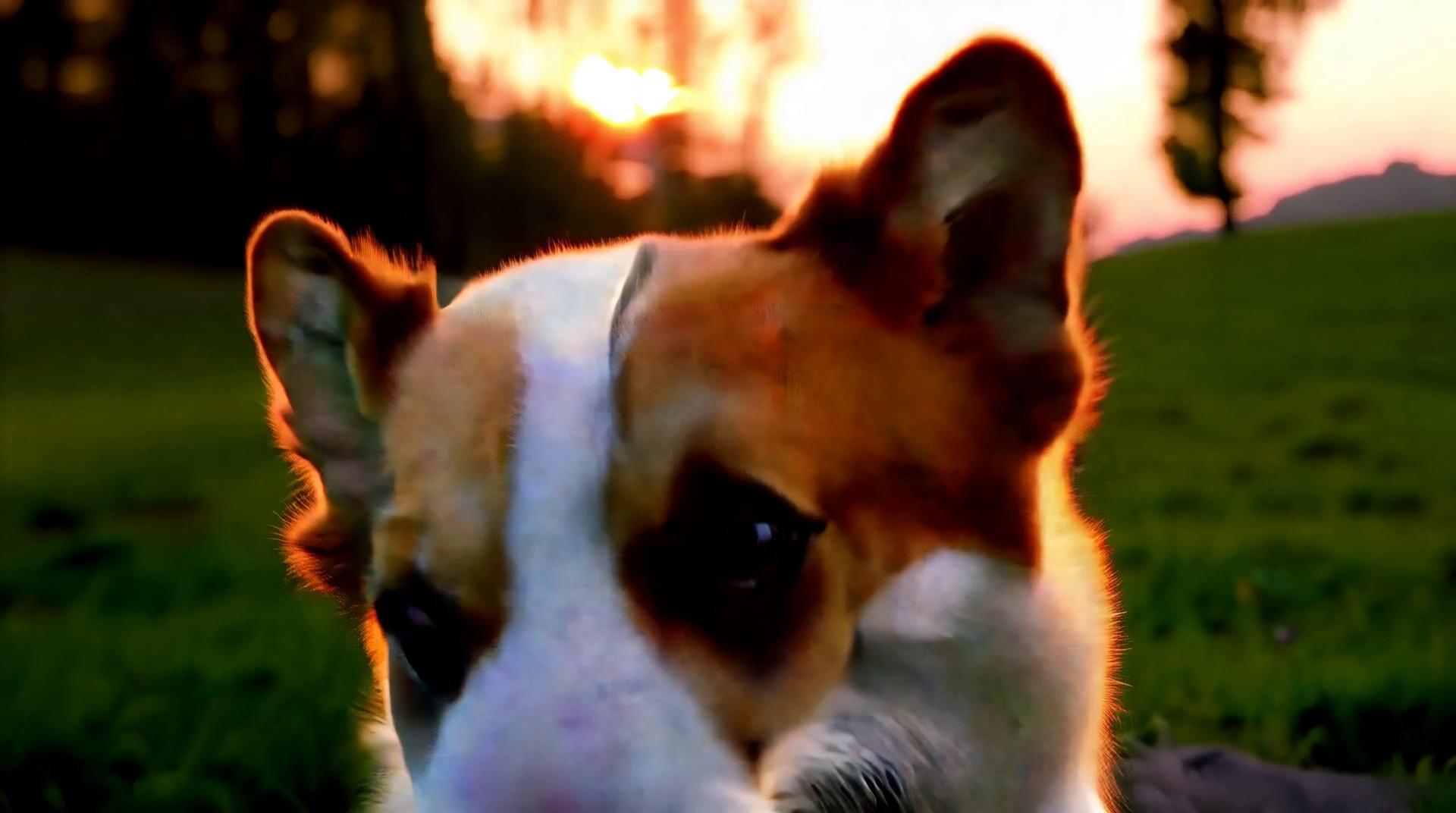} \\ 
  \textbf{HiStream}
  \textbf{(Ours)} & 
  \includegraphics[width=\linewidth]{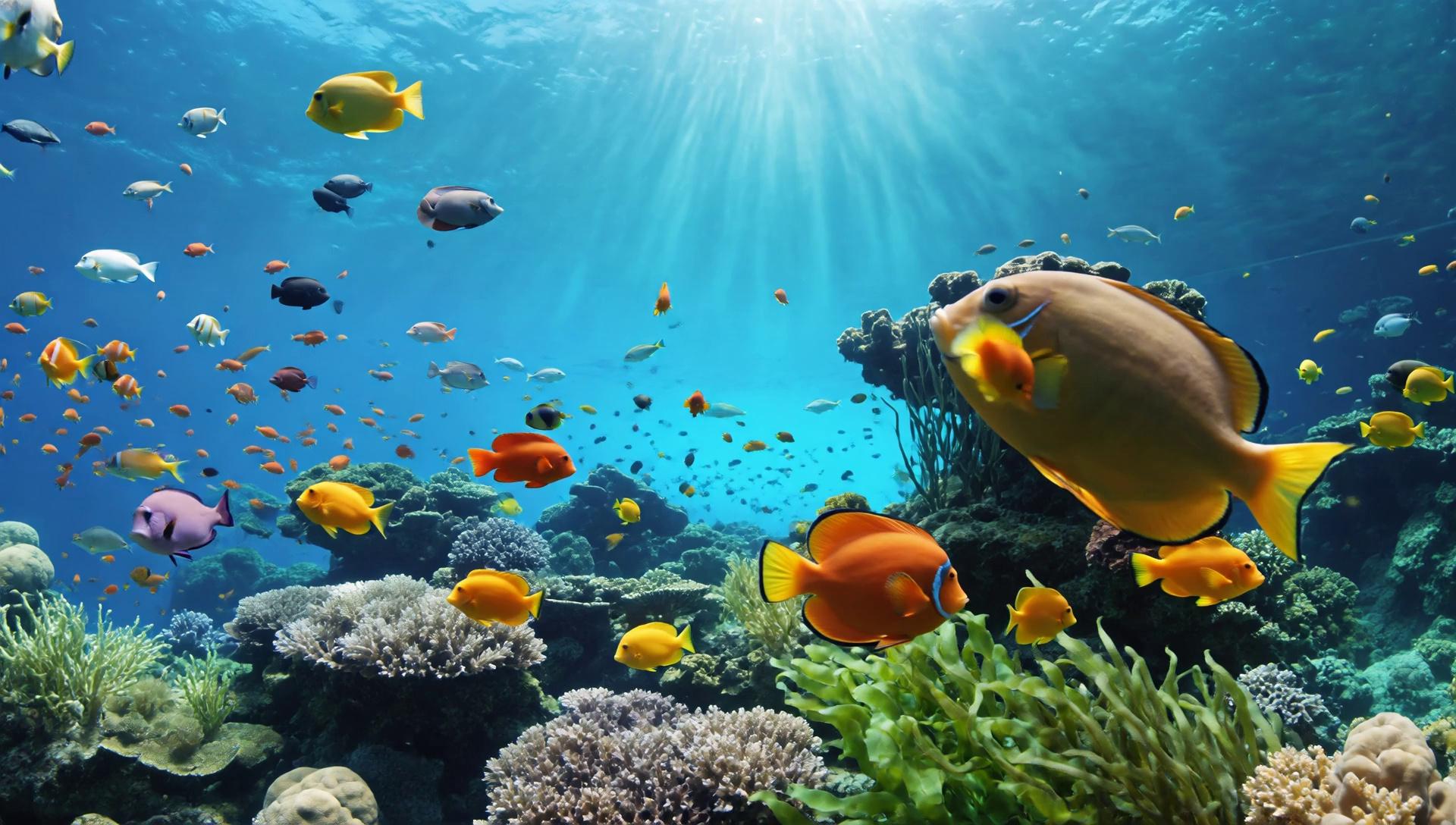} & 
  \includegraphics[width=\linewidth]{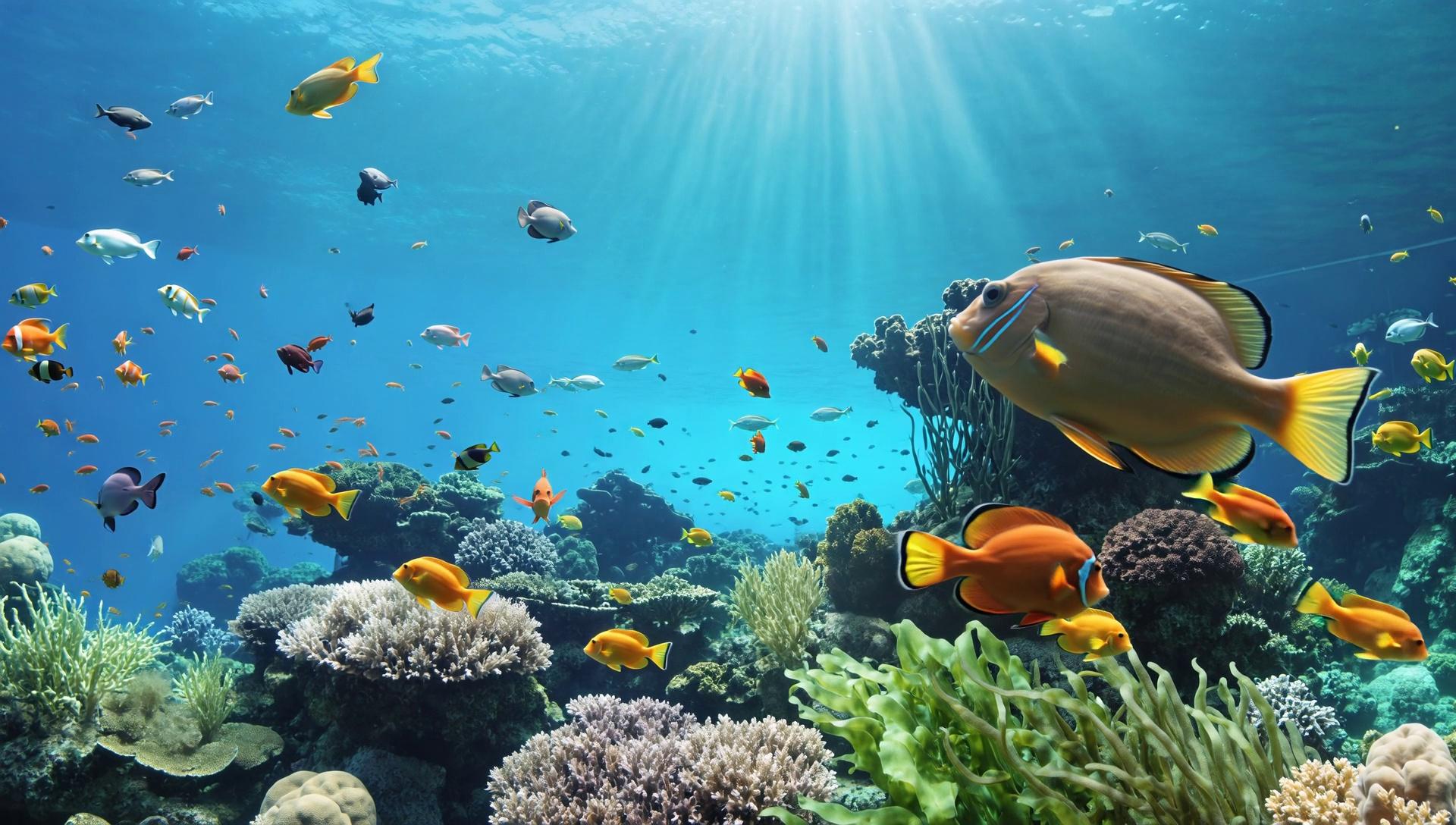} & 
  \includegraphics[width=\linewidth]{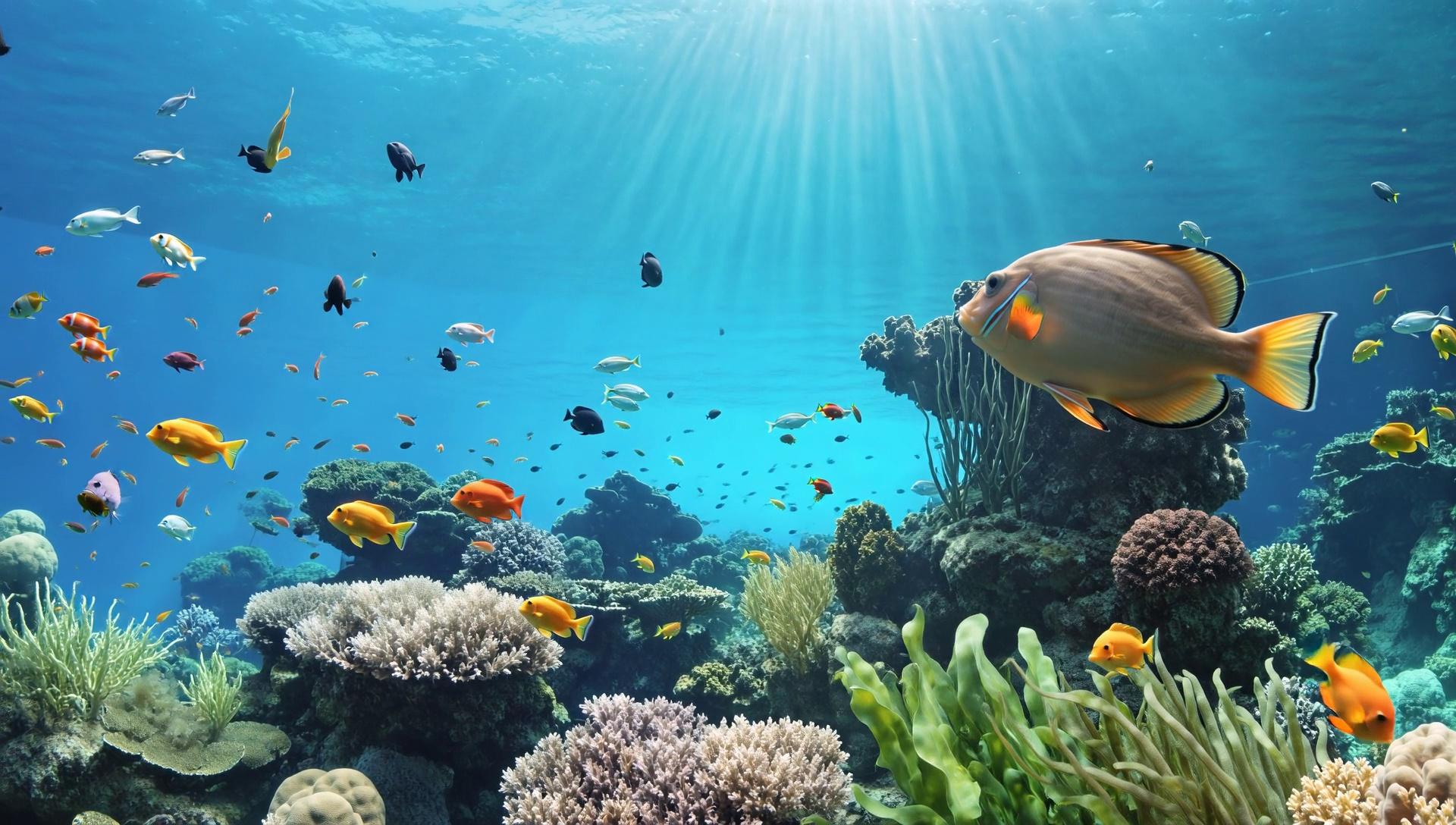} & 
  \includegraphics[width=\linewidth]{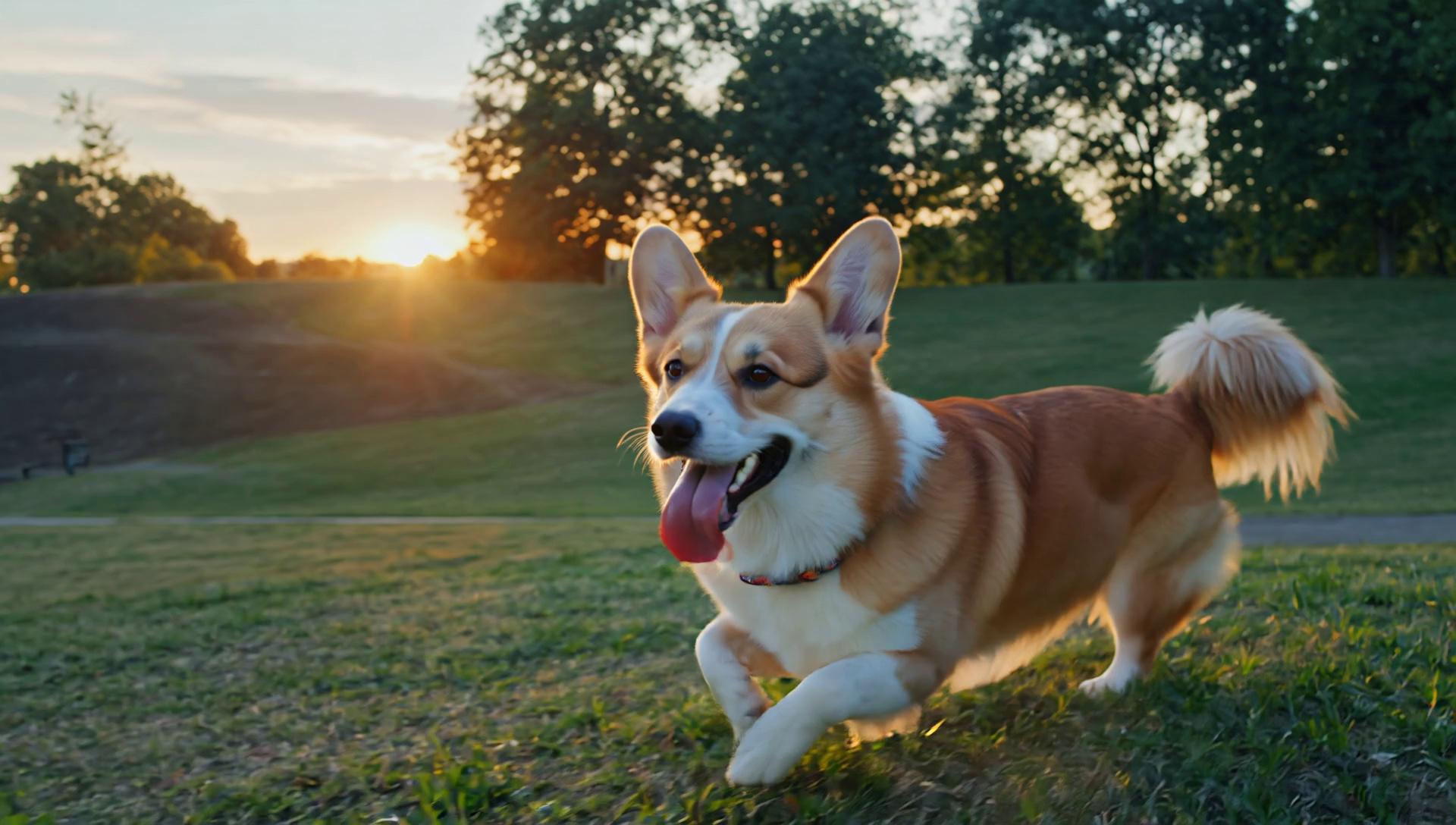} & 
  \includegraphics[width=\linewidth]{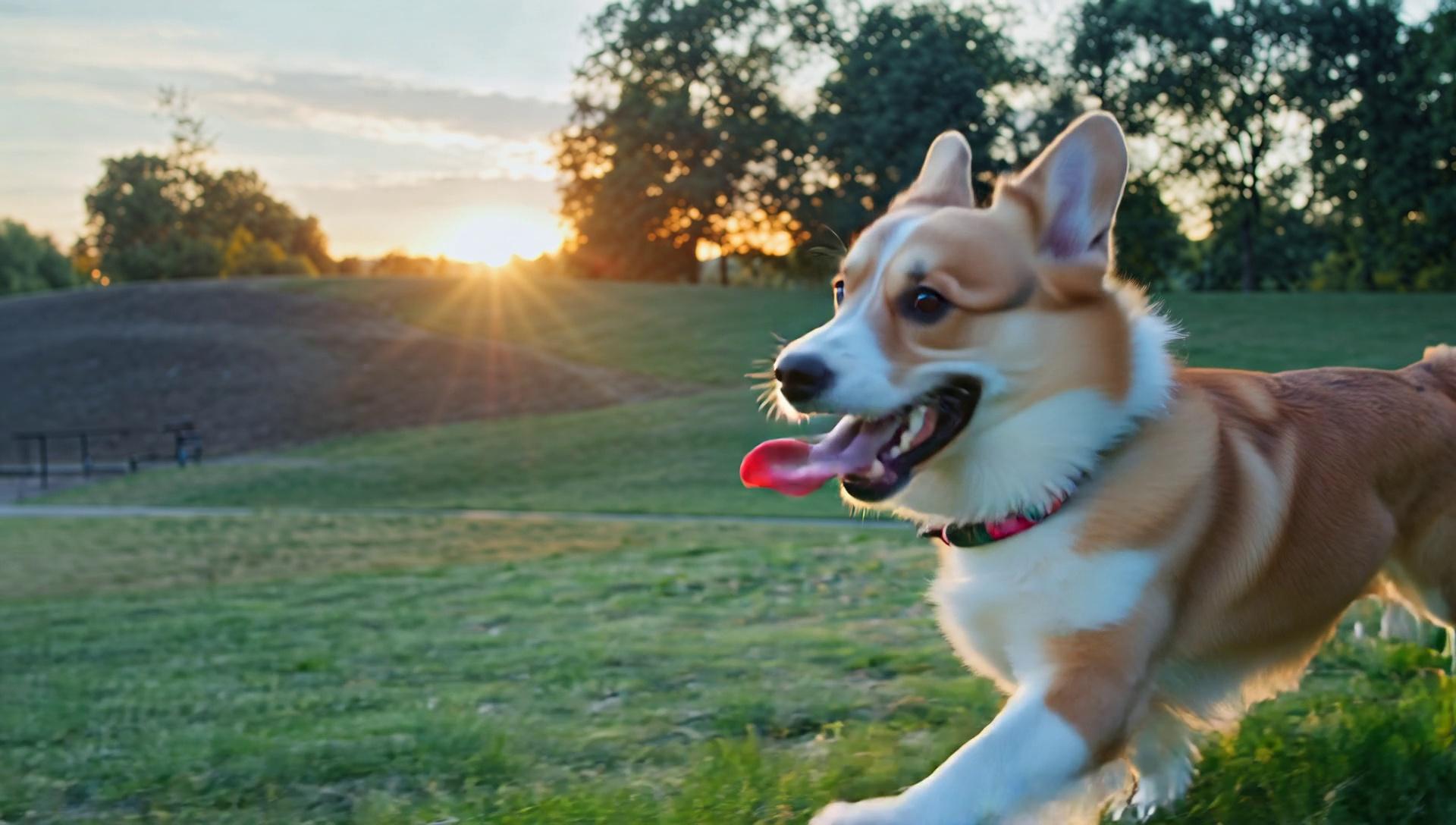} & 
  \includegraphics[width=\linewidth]{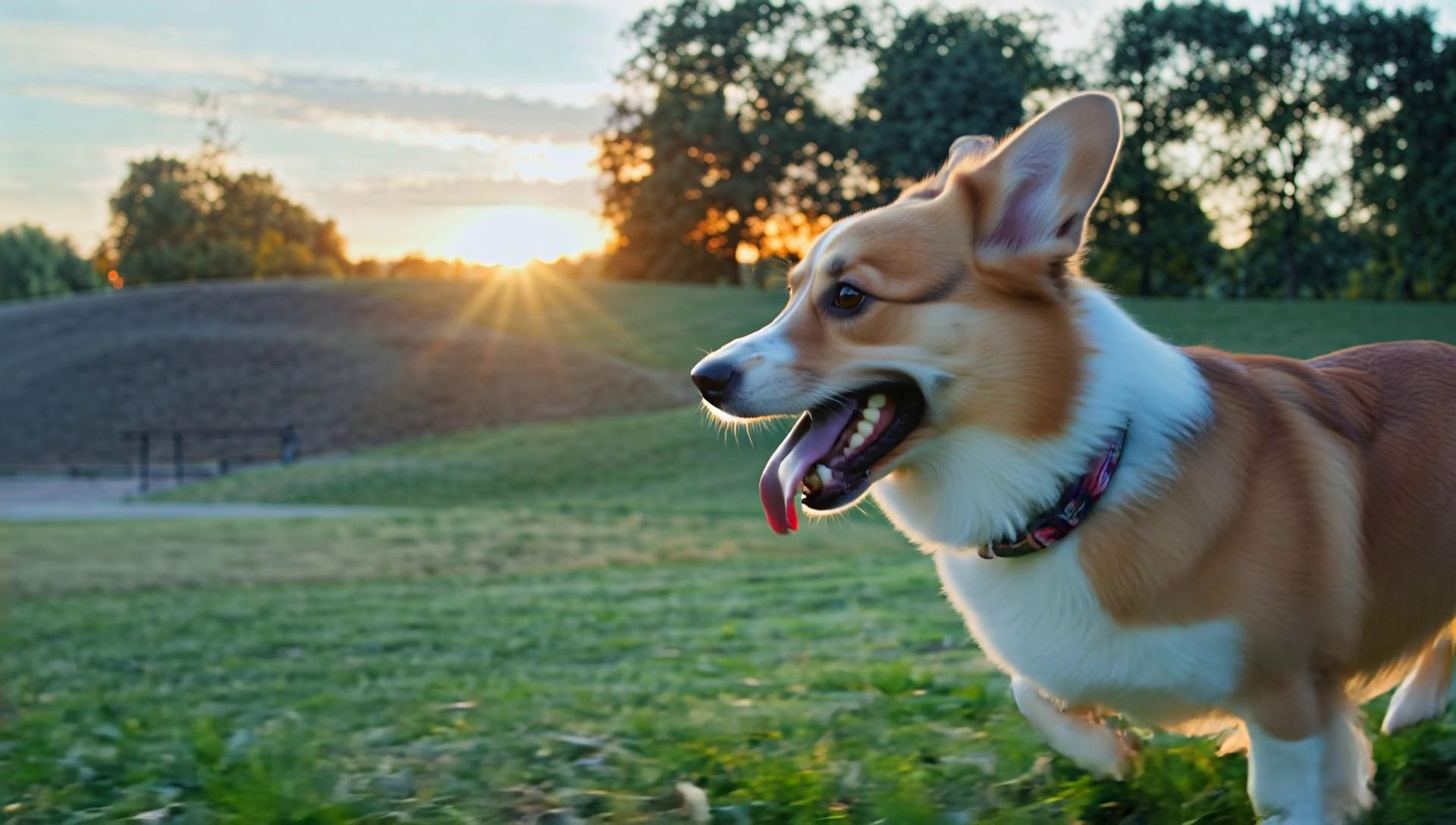} \\
  \noalign{\vspace{0.5ex}}
  \hdashline
  \noalign{\vspace{0.5ex}}
  Wan2.1 & 
  \includegraphics[width=\linewidth]{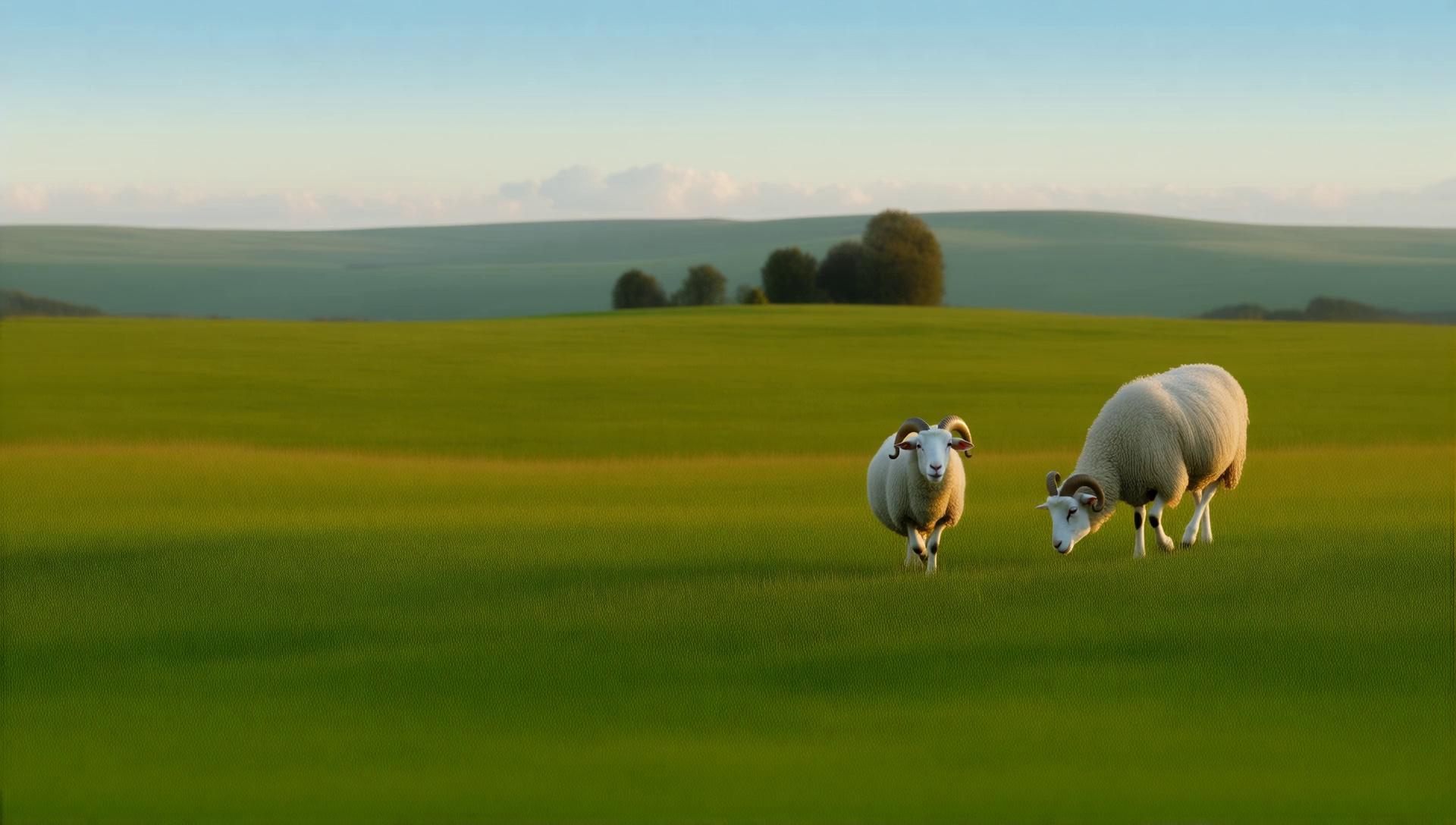} & 
  \includegraphics[width=\linewidth]{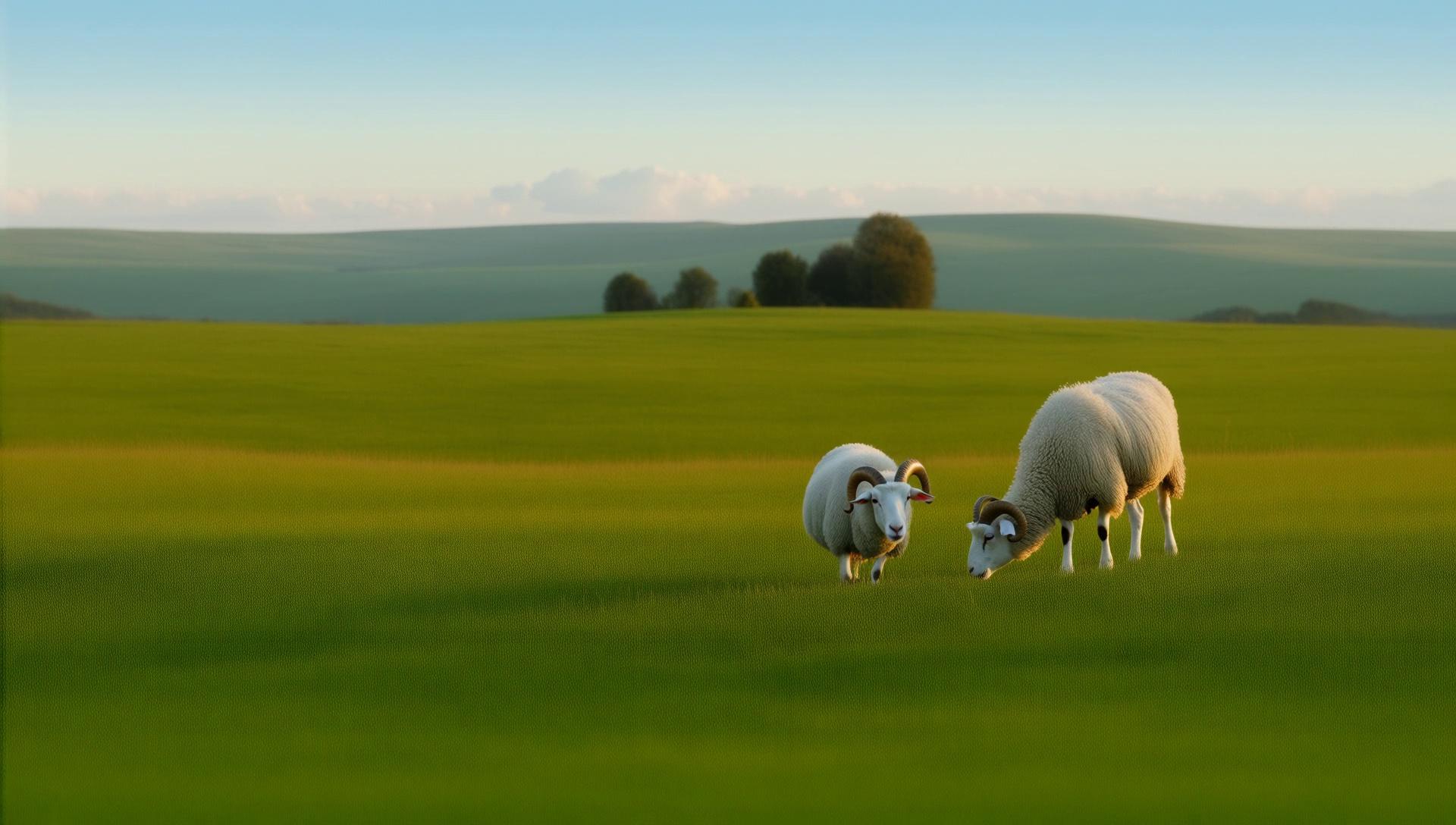} & 
  \includegraphics[width=\linewidth]{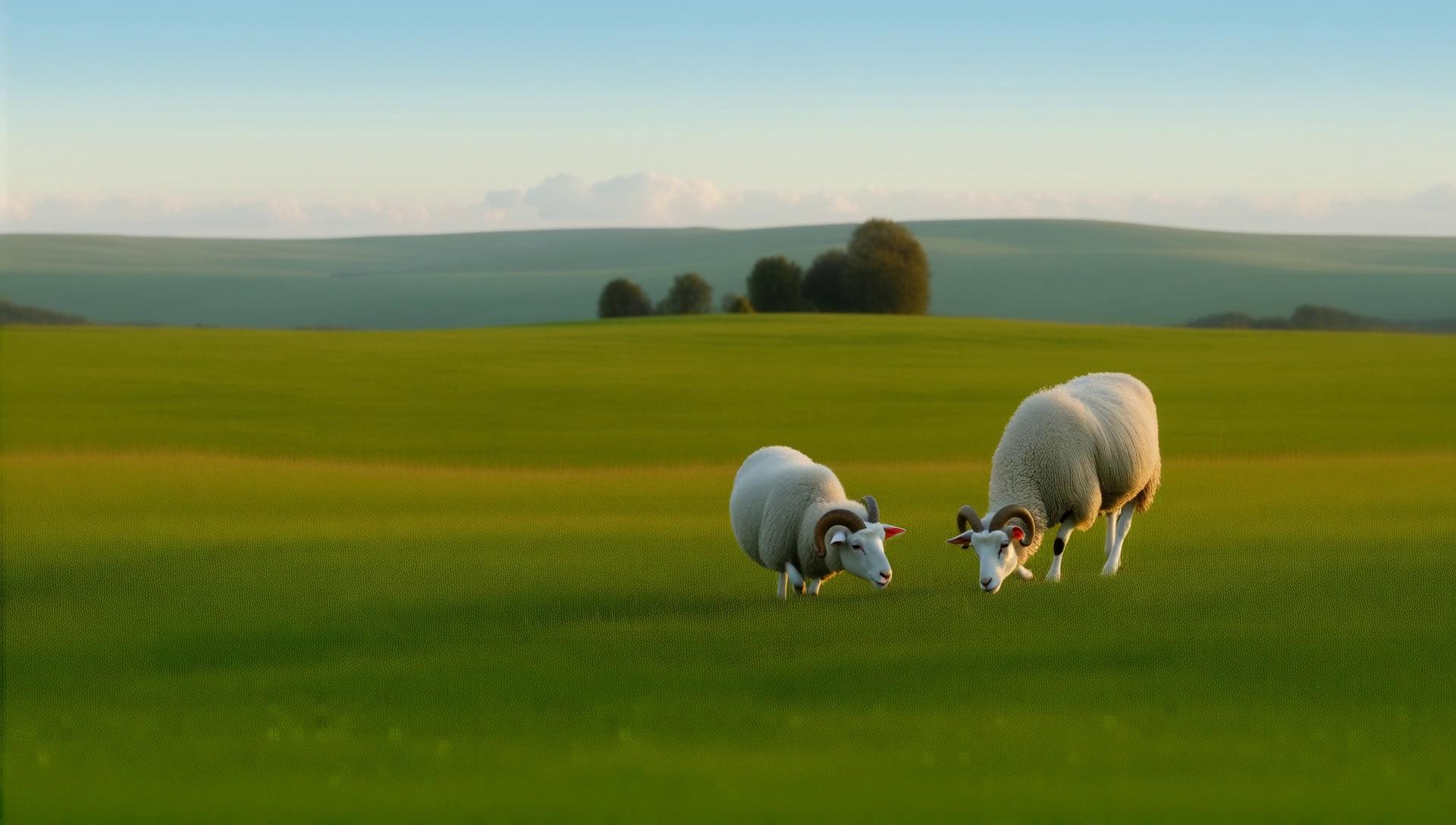} & 
  \includegraphics[width=\linewidth]{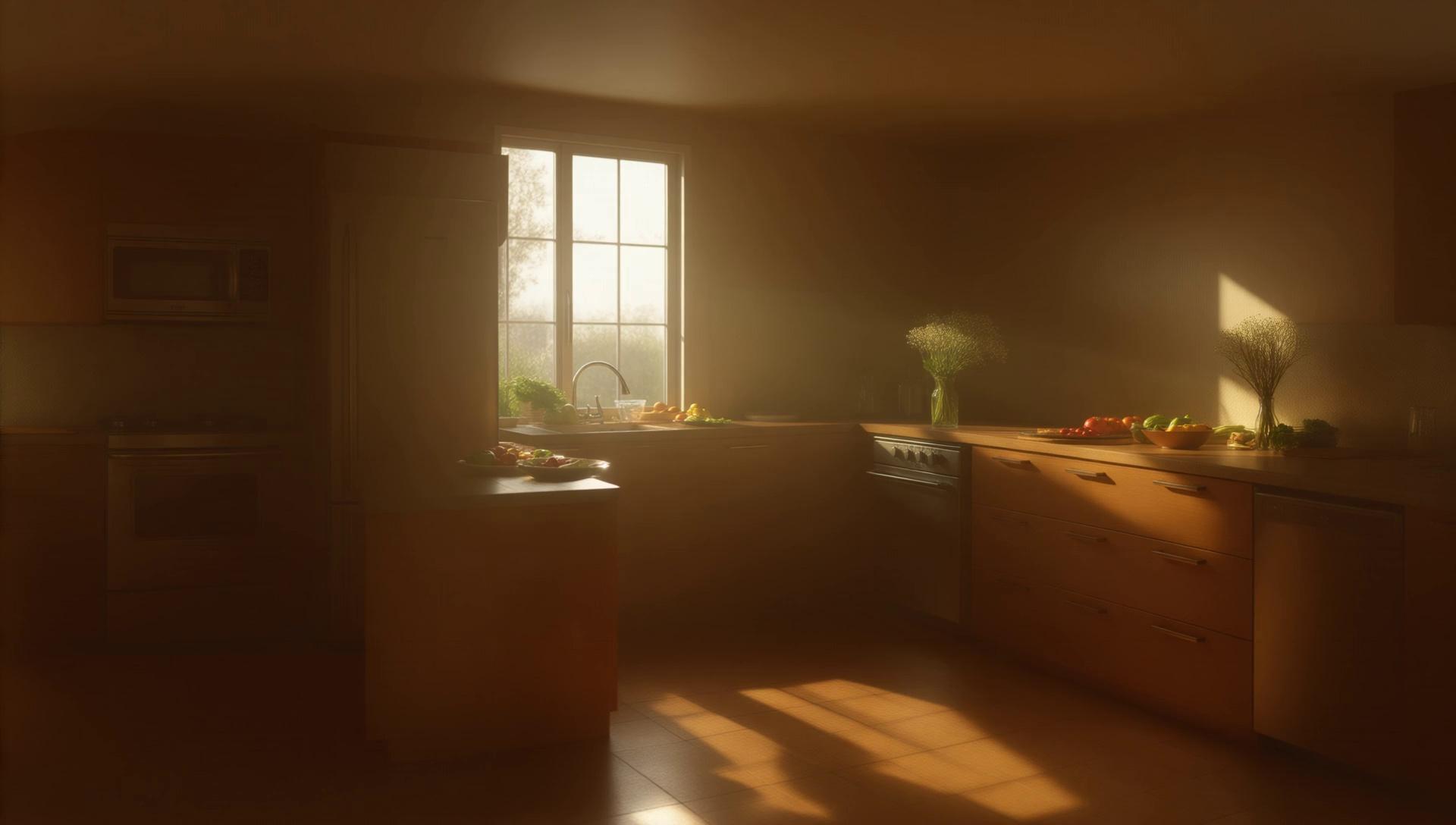} & 
  \includegraphics[width=\linewidth]{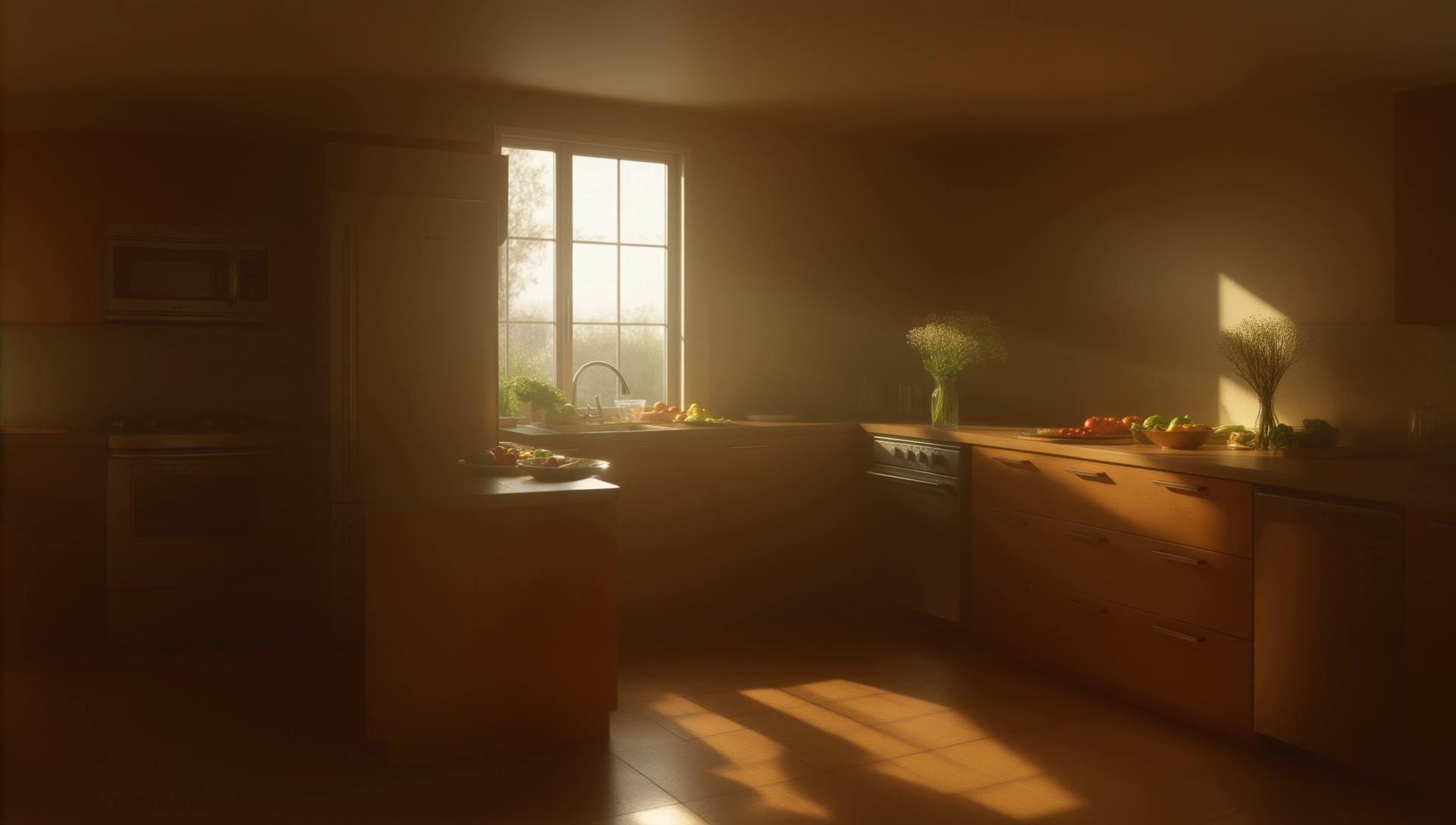} & 
  \includegraphics[width=\linewidth]{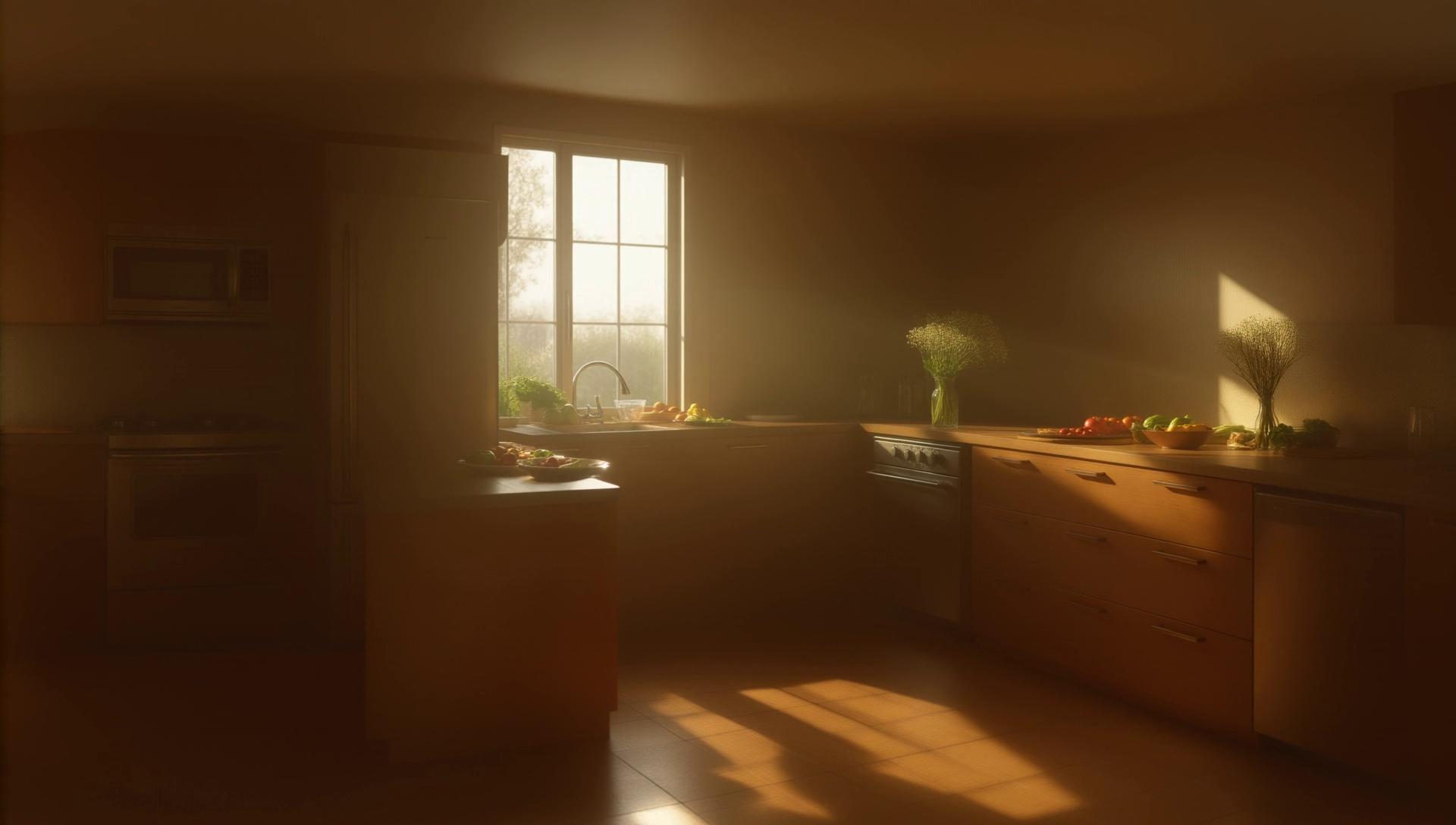} \\
  Self Forcing & 
  \includegraphics[width=\linewidth]{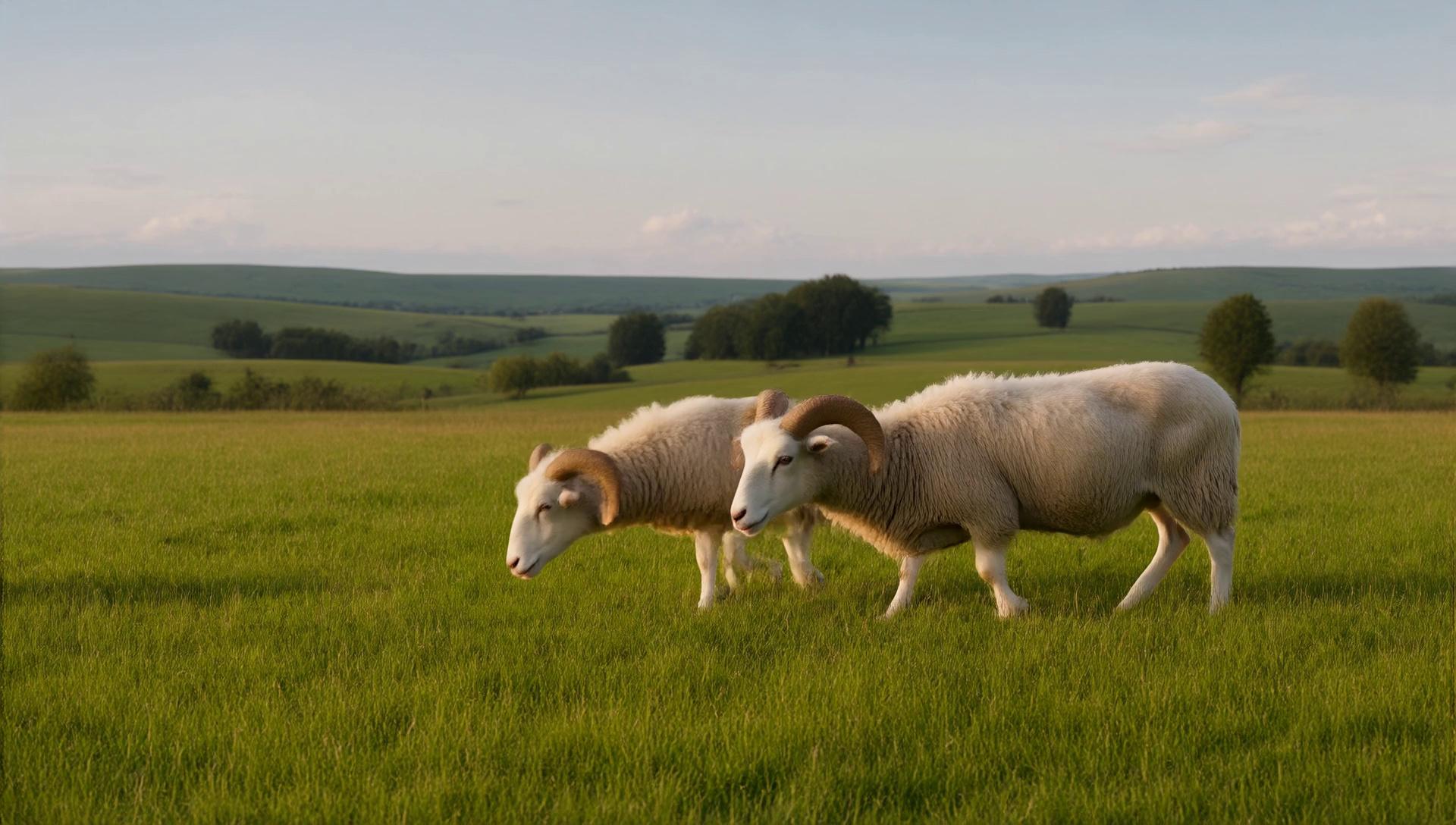} & 
  \includegraphics[width=\linewidth]{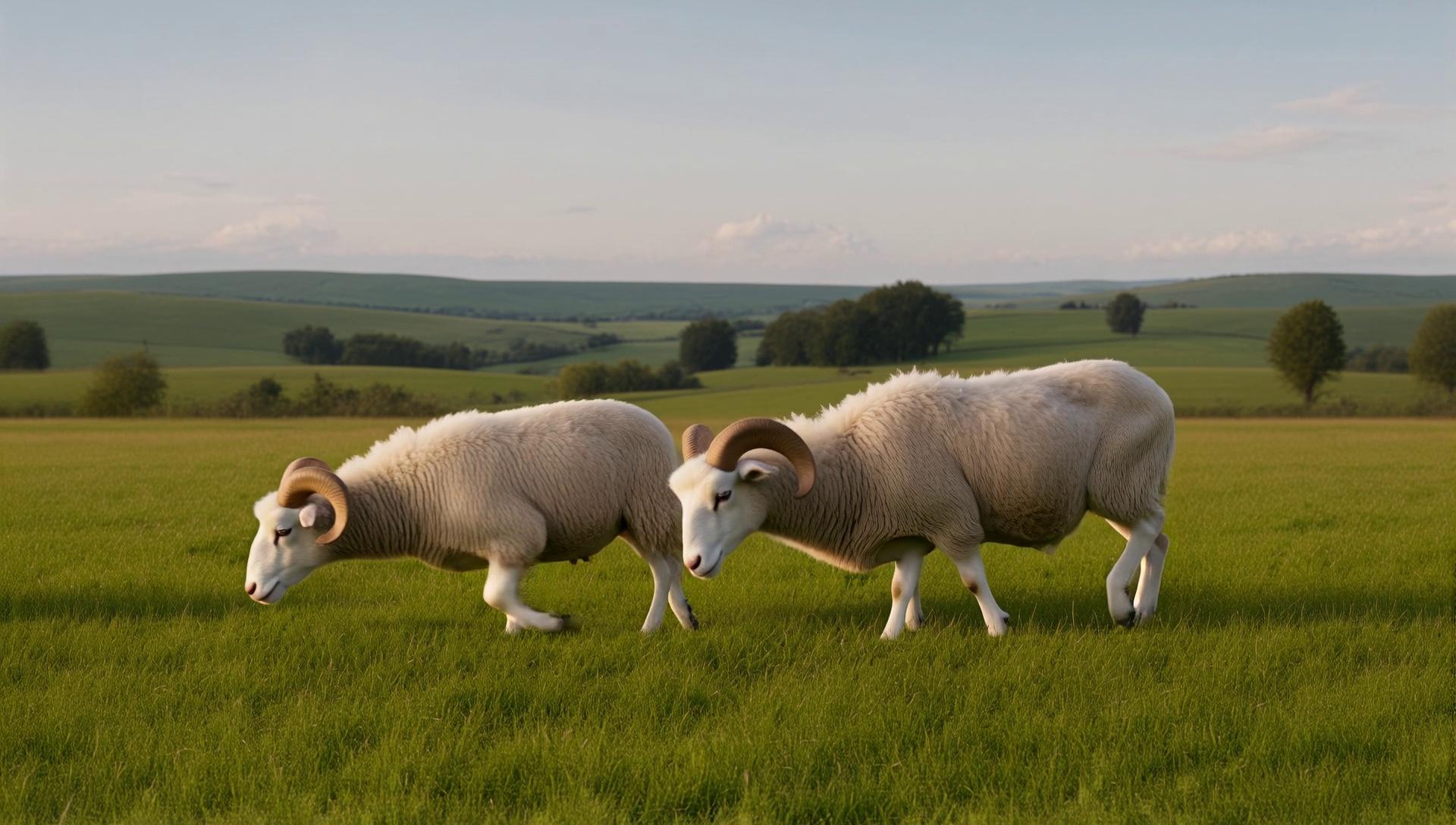} & 
  \includegraphics[width=\linewidth]{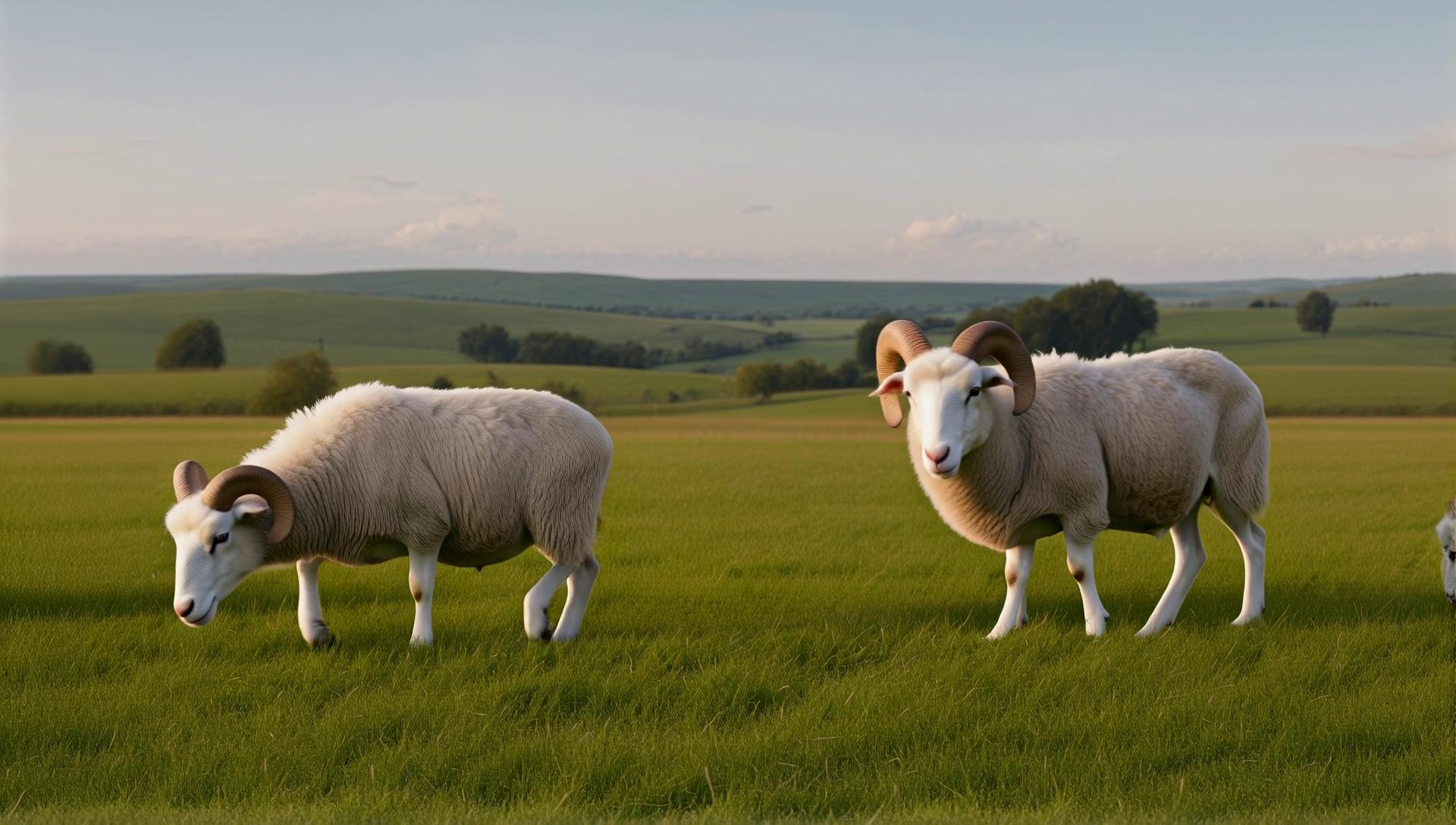} & 
  \includegraphics[width=\linewidth]{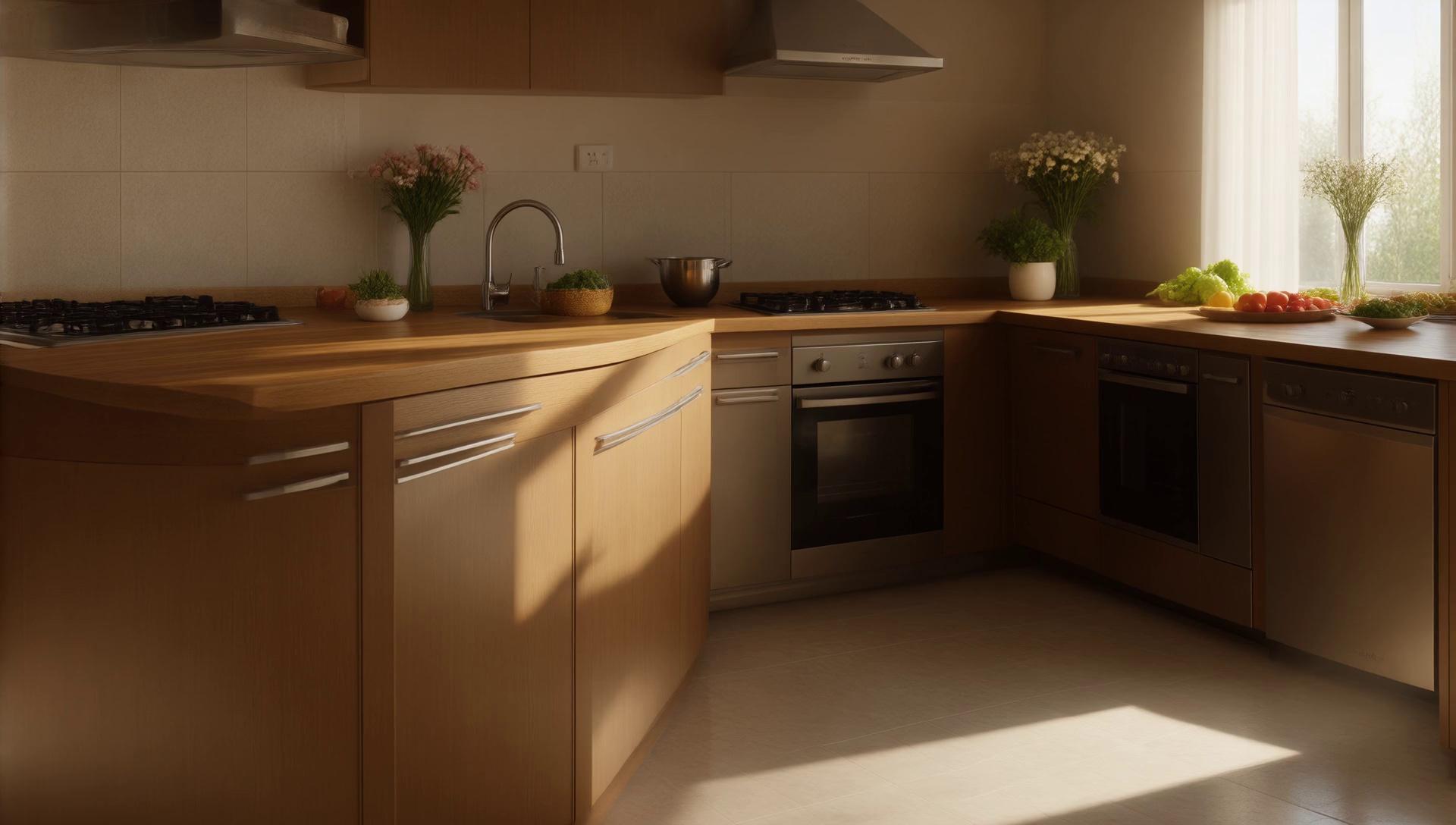} & 
  \includegraphics[width=\linewidth]{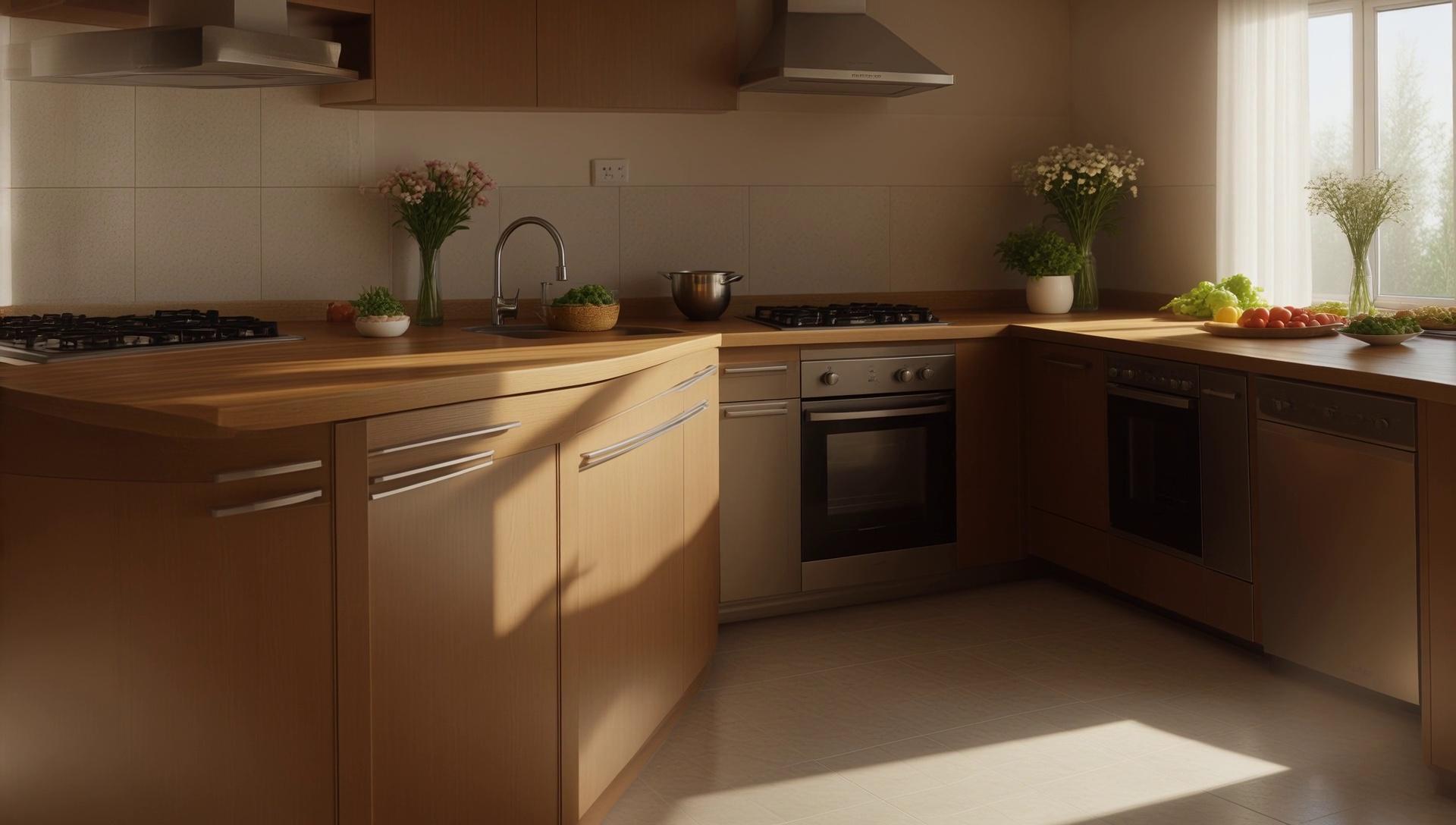} & 
  \includegraphics[width=\linewidth]{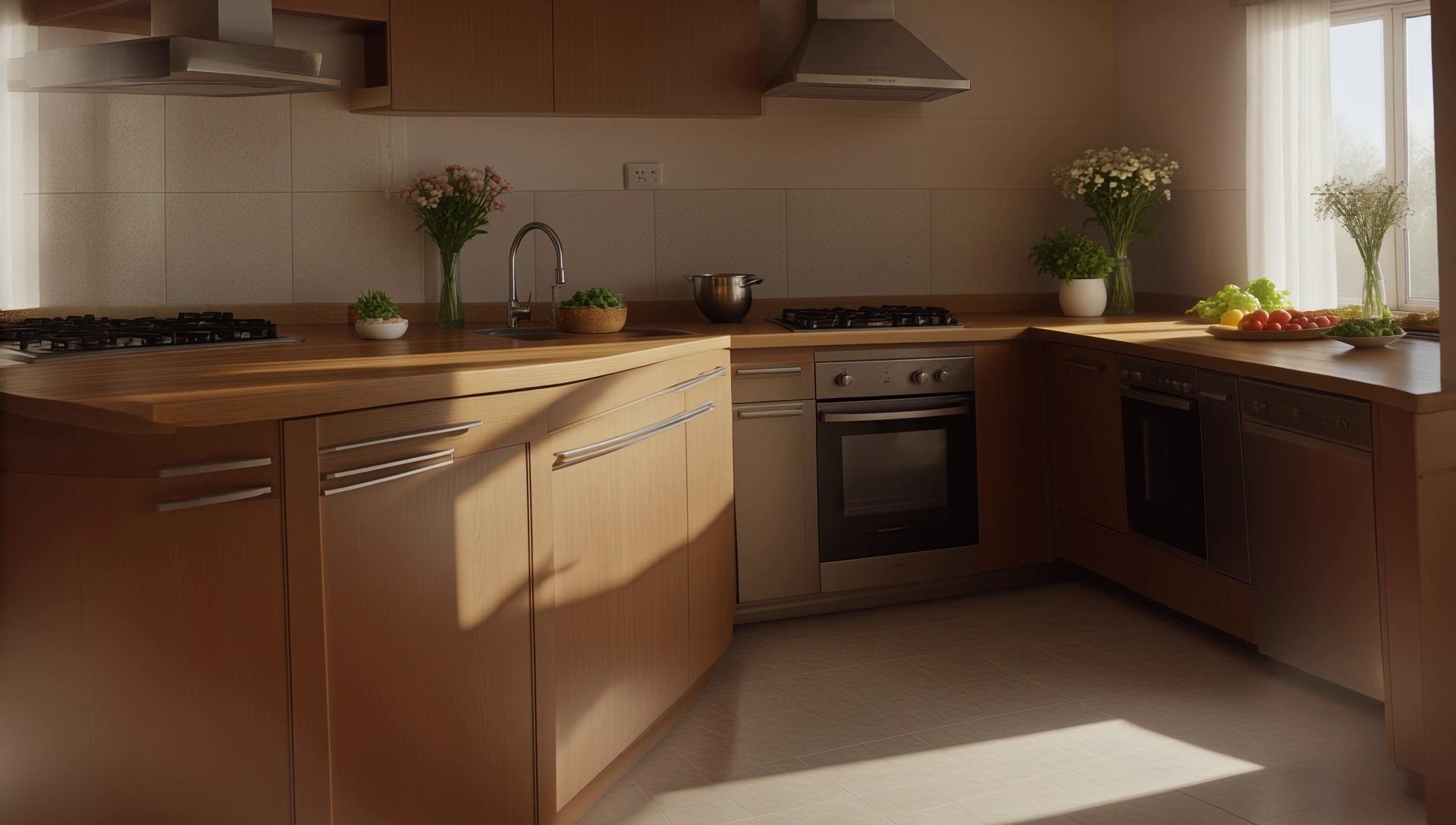} \\
  LTX & 
  \includegraphics[width=\linewidth]{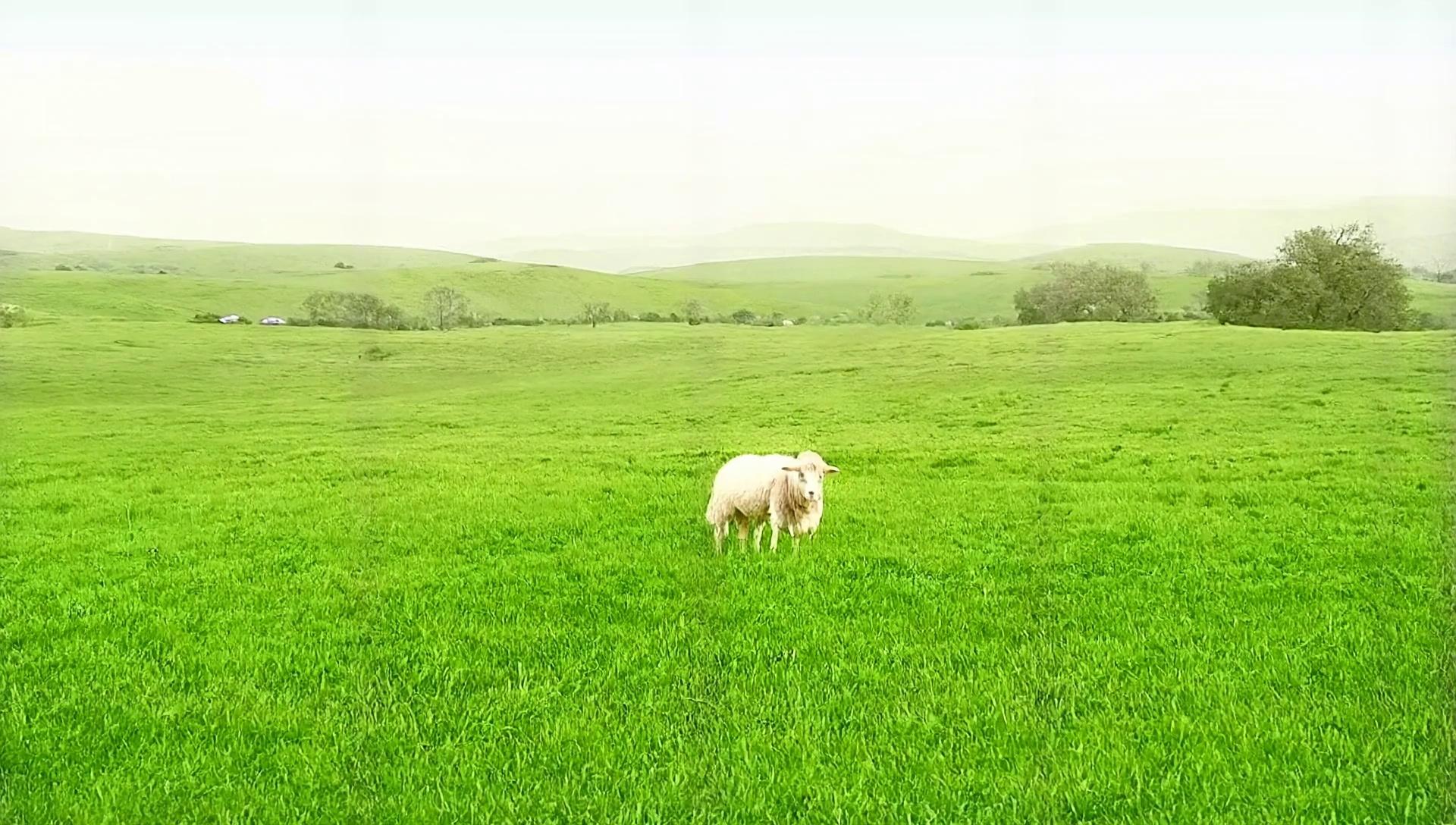} & 
  \includegraphics[width=\linewidth]{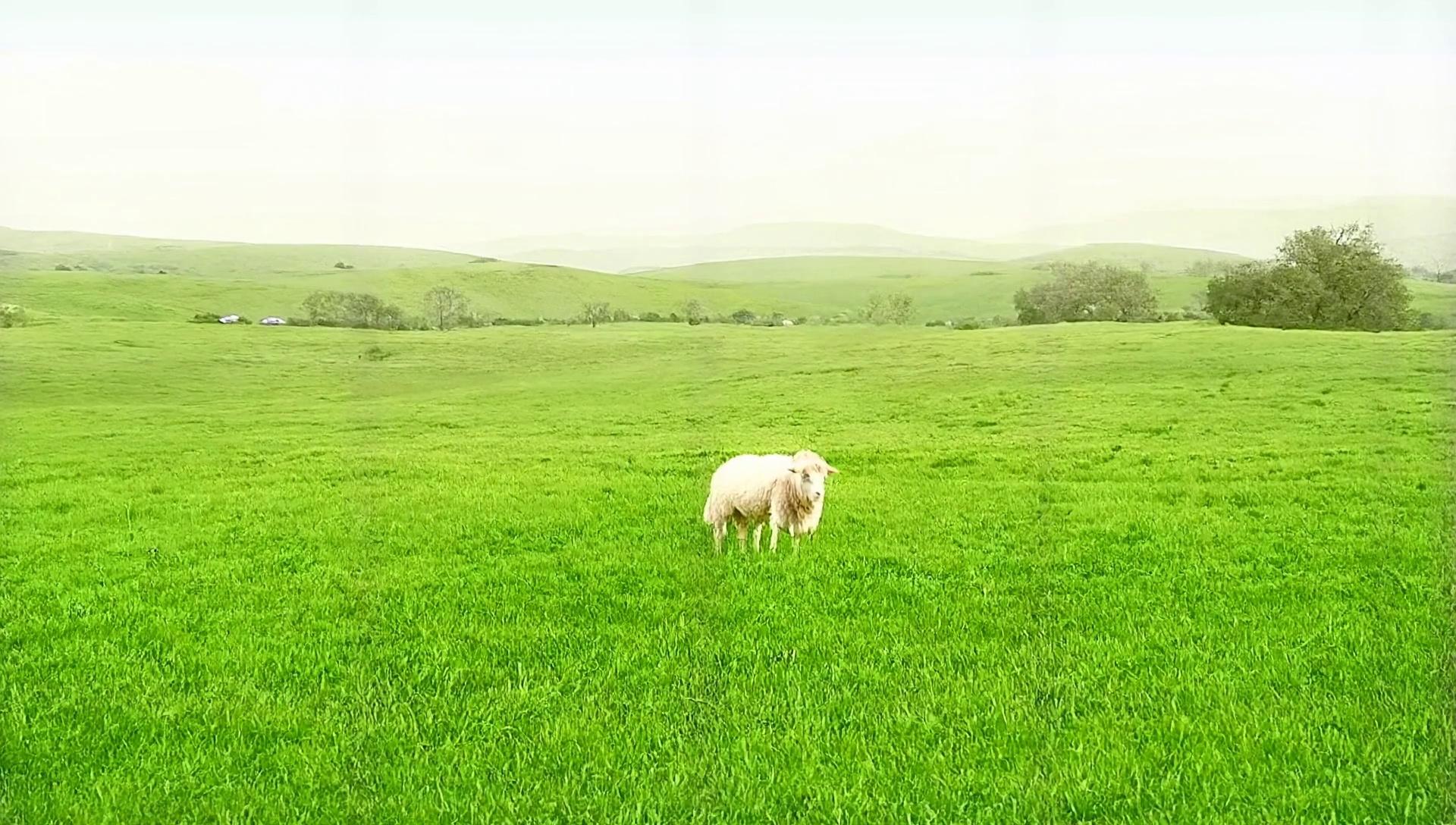} & 
  \includegraphics[width=\linewidth]{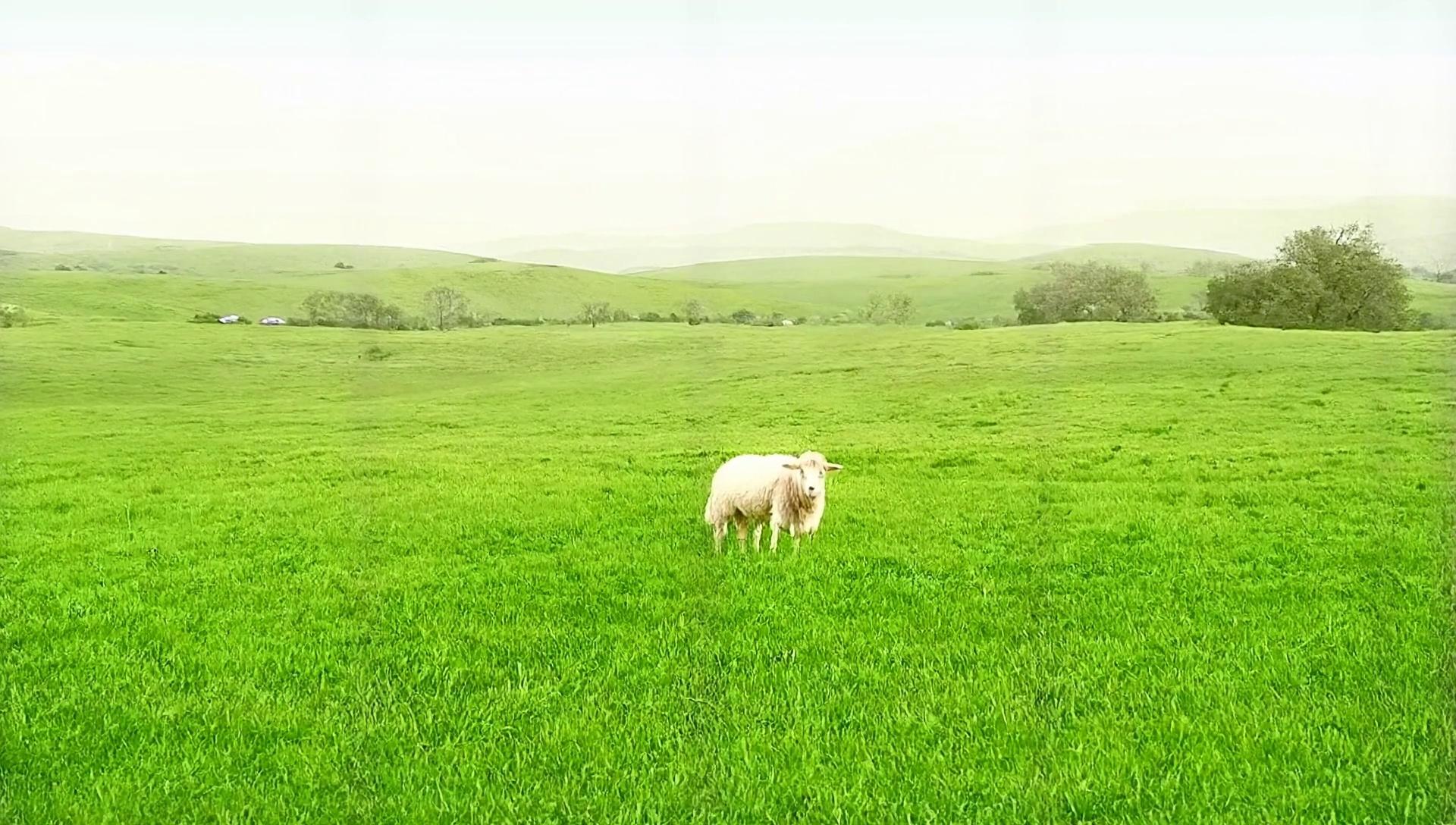} & 
  \includegraphics[width=\linewidth]{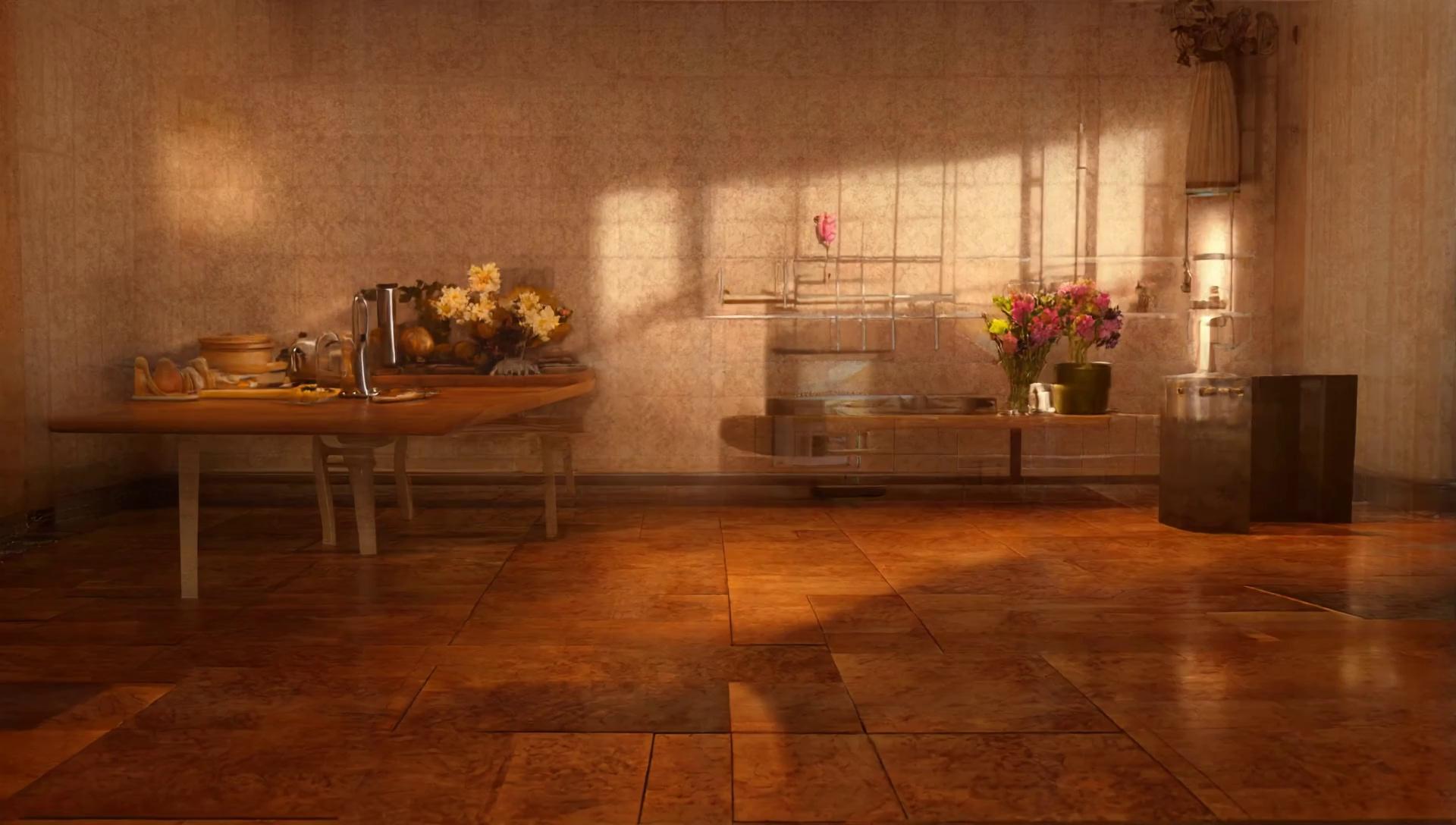} & 
  \includegraphics[width=\linewidth]{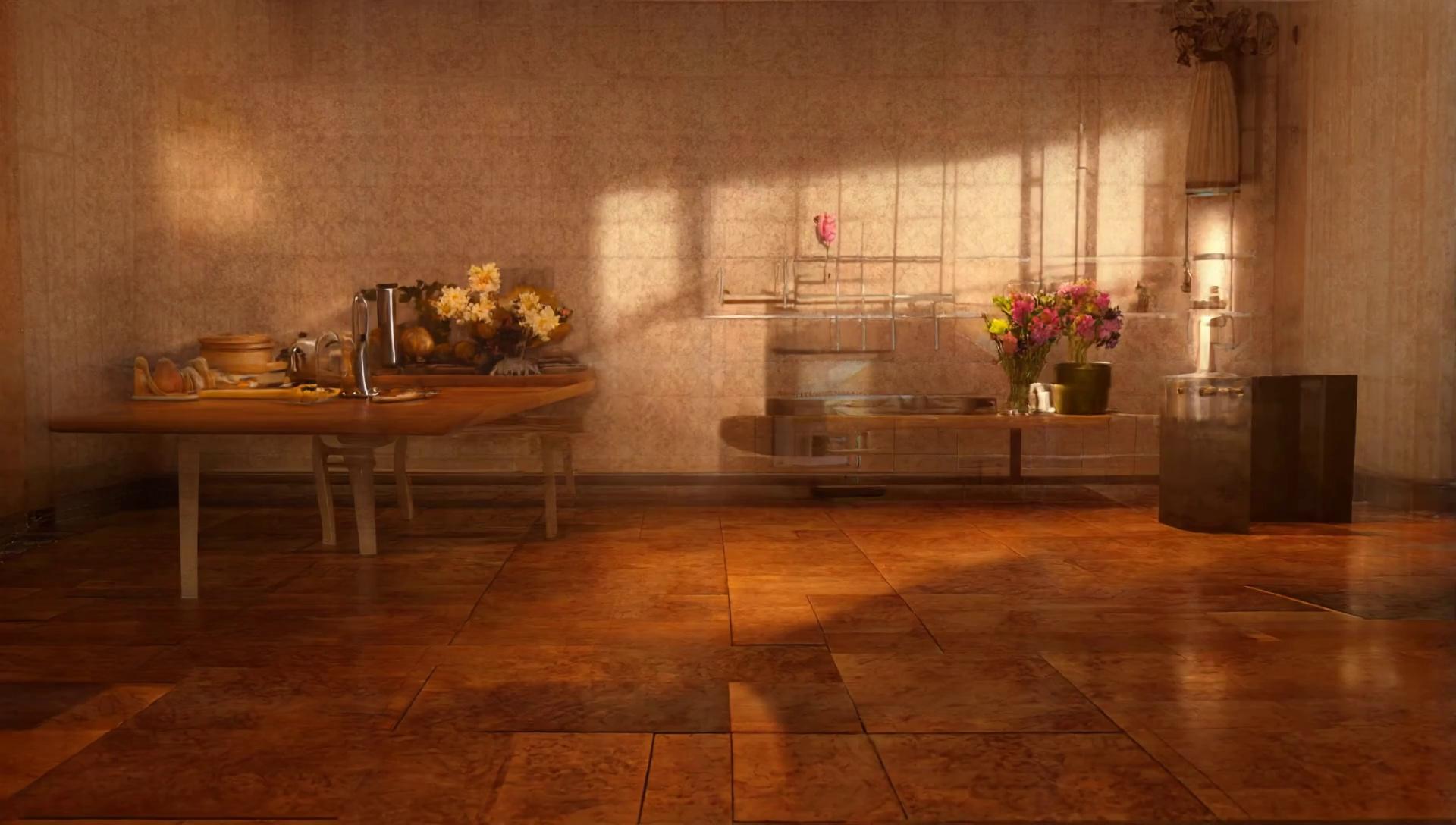} & 
  \includegraphics[width=\linewidth]{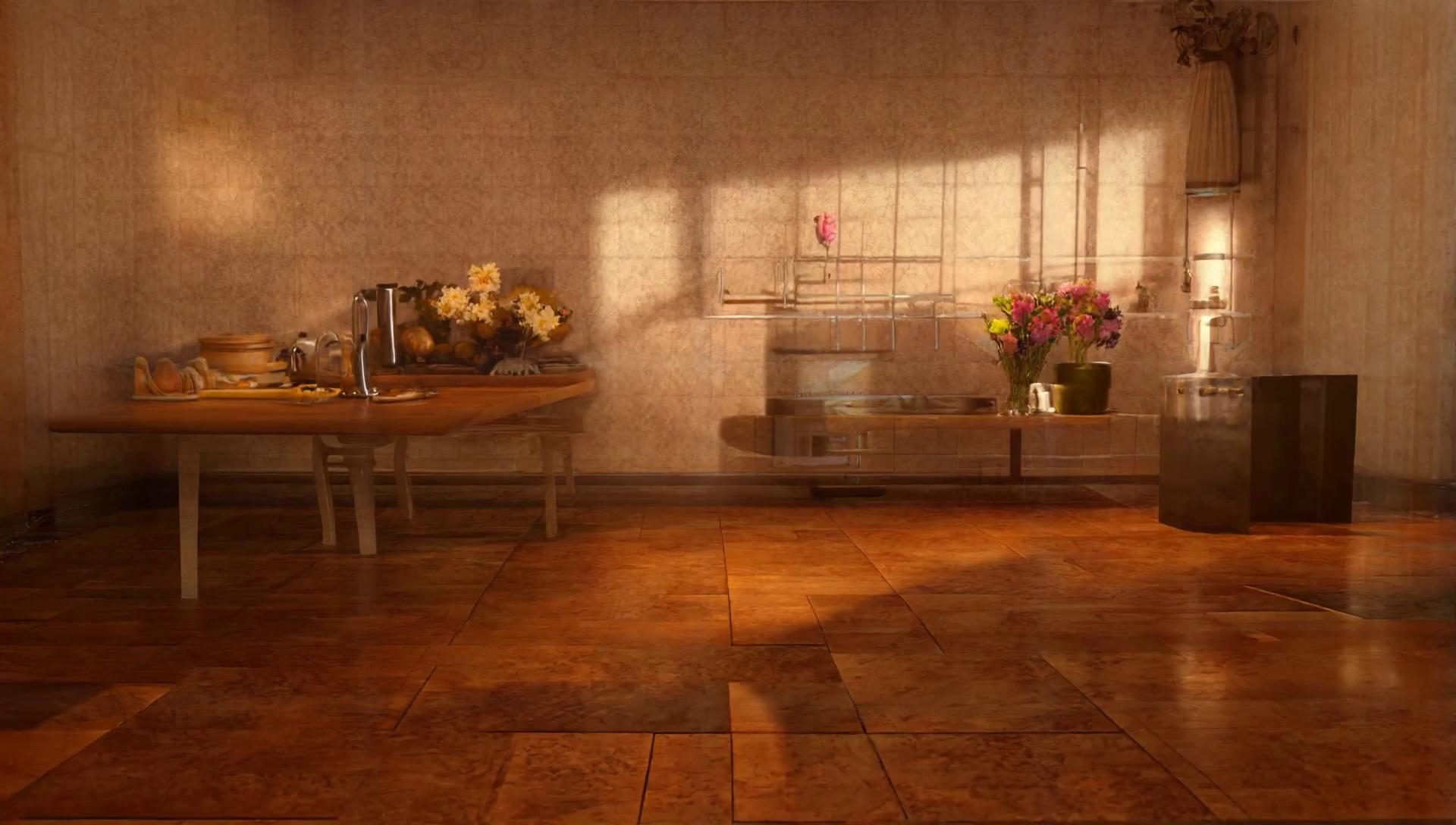} \\  
  Flash
  Video & 
  \includegraphics[width=\linewidth]{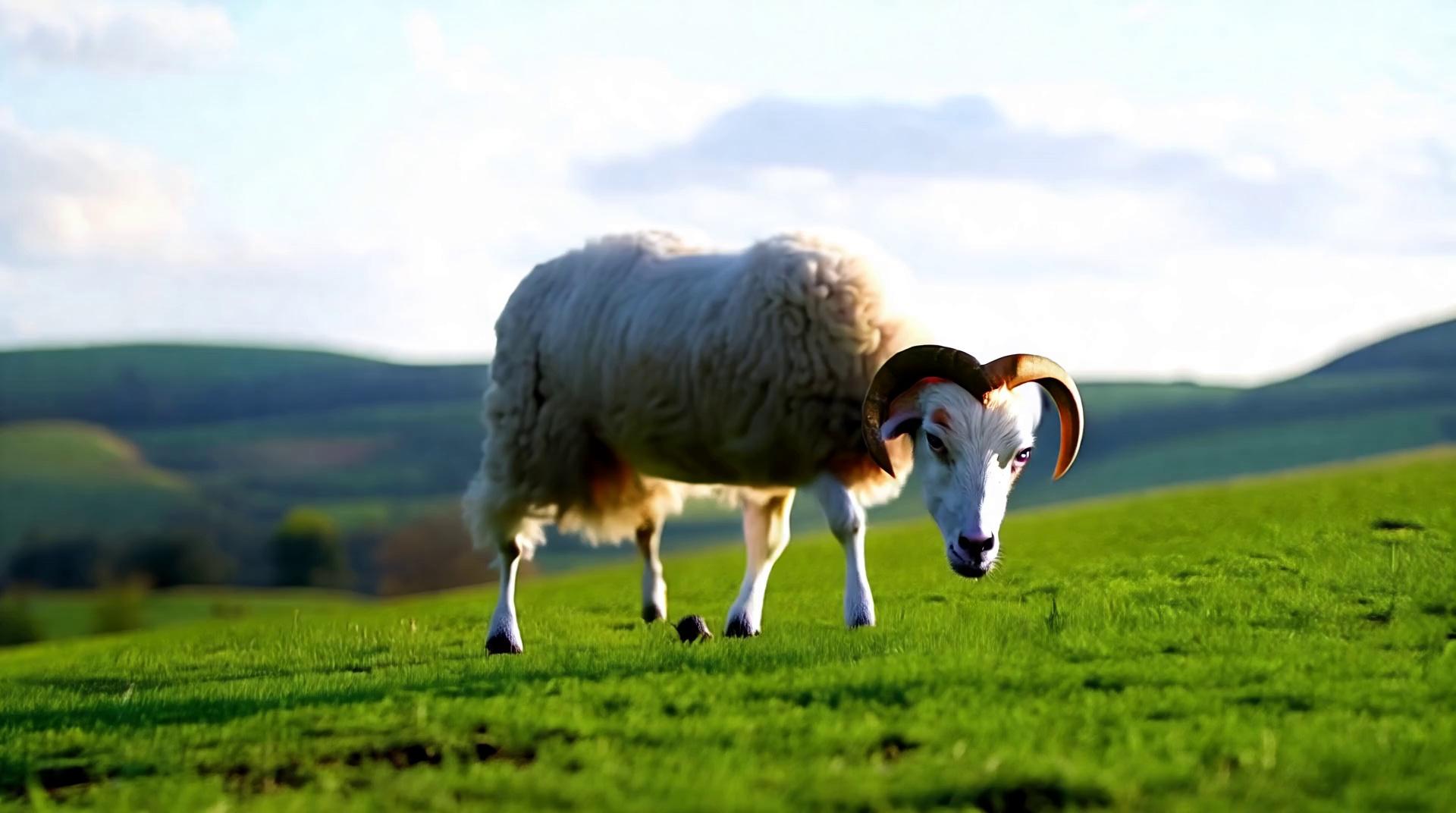} & 
  \includegraphics[width=\linewidth]{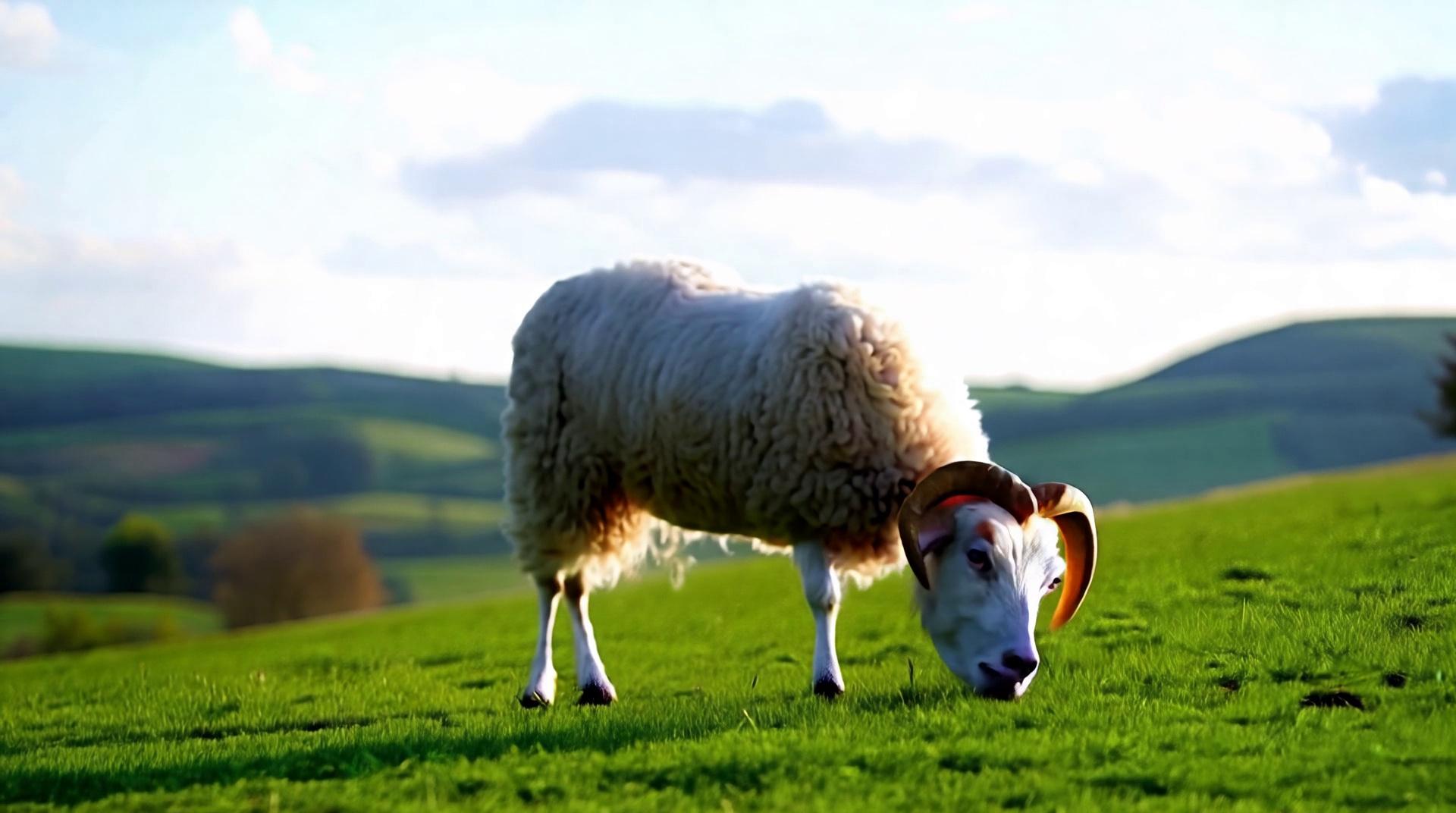} & 
  \includegraphics[width=\linewidth]{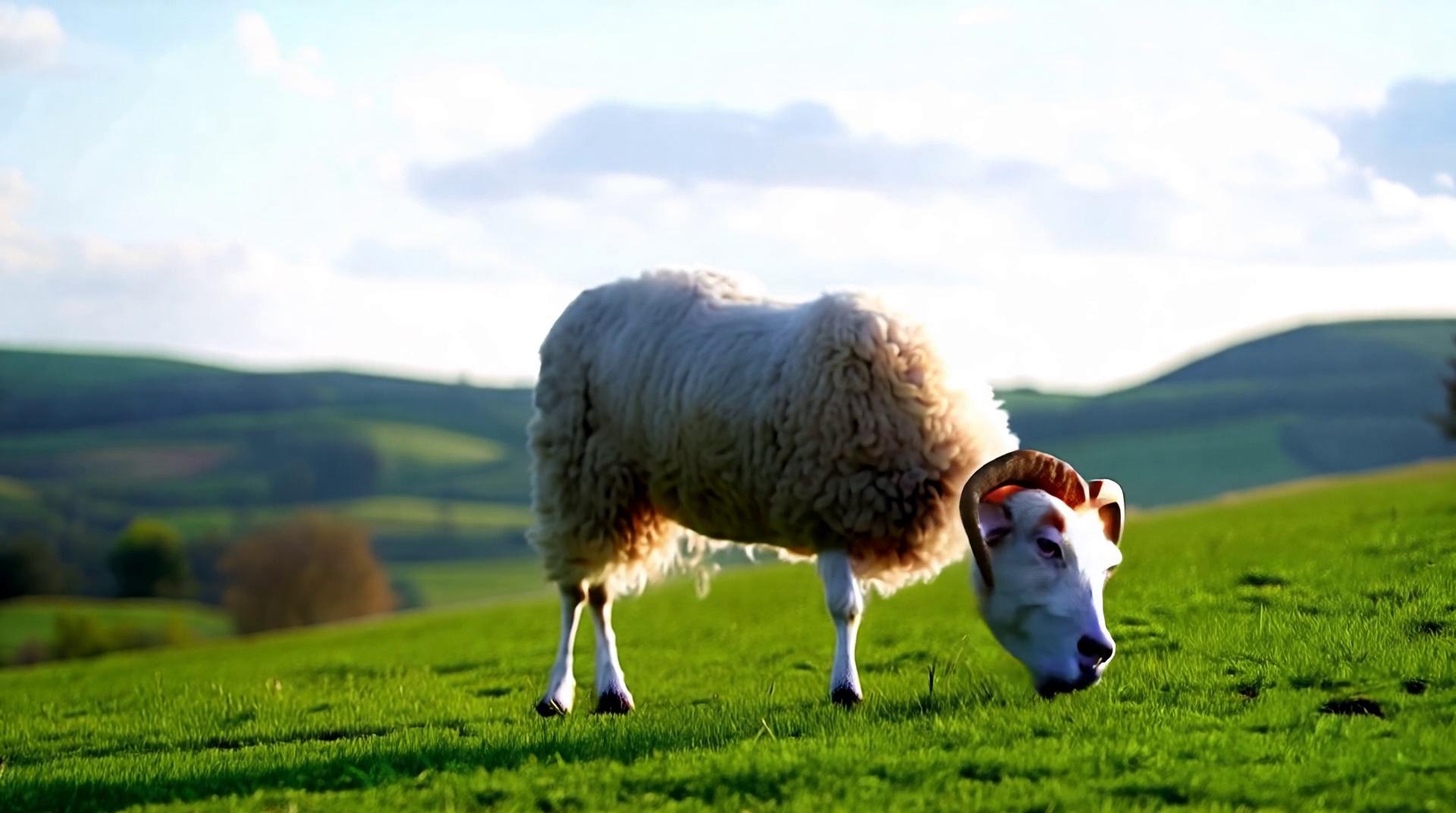} & 
  \includegraphics[width=\linewidth]{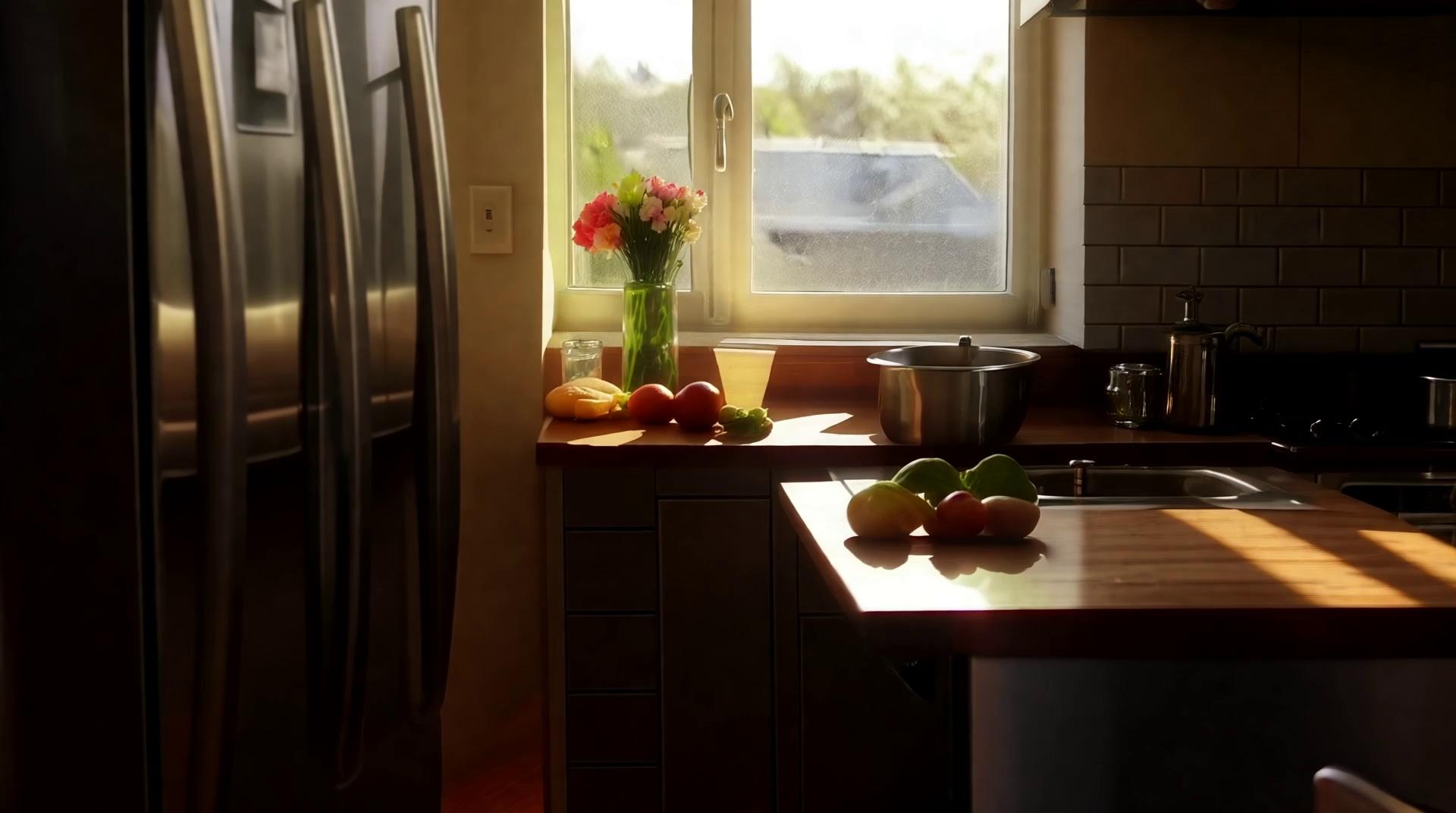} & 
  \includegraphics[width=\linewidth]{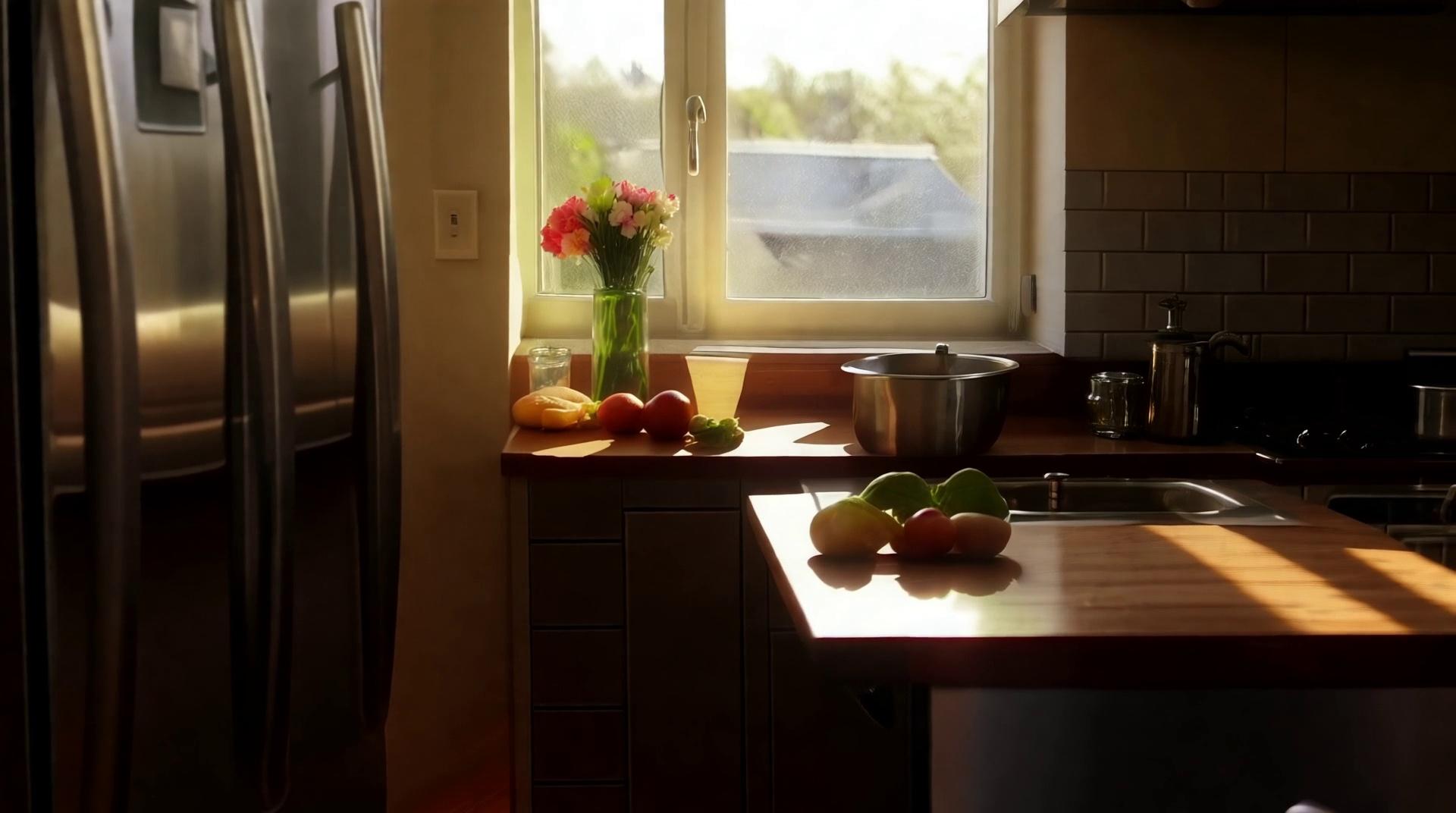} & 
  \includegraphics[width=\linewidth]{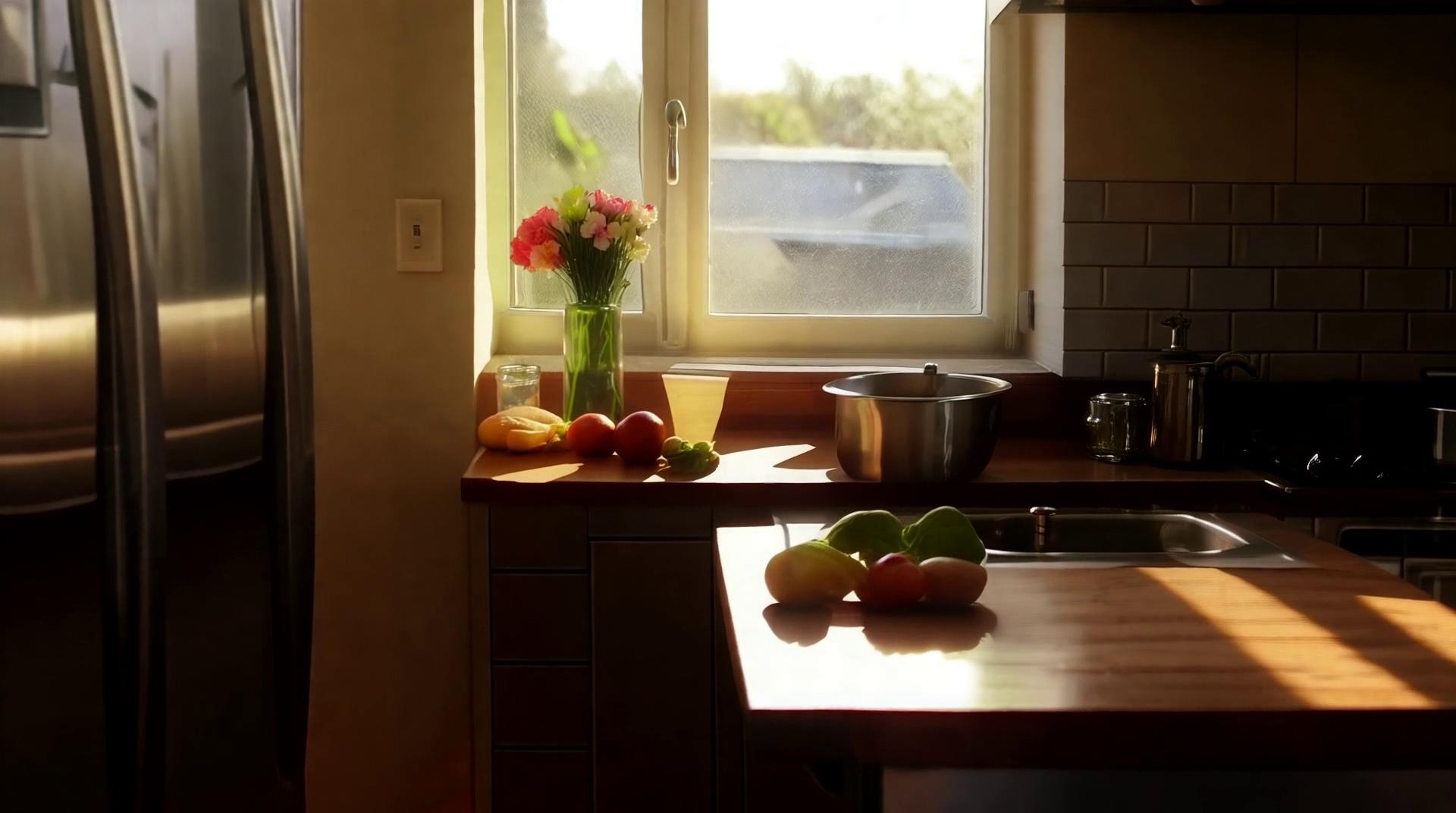} \\  
  \textbf{HiStream}
  \textbf{(Ours)}& 
  \includegraphics[width=\linewidth]{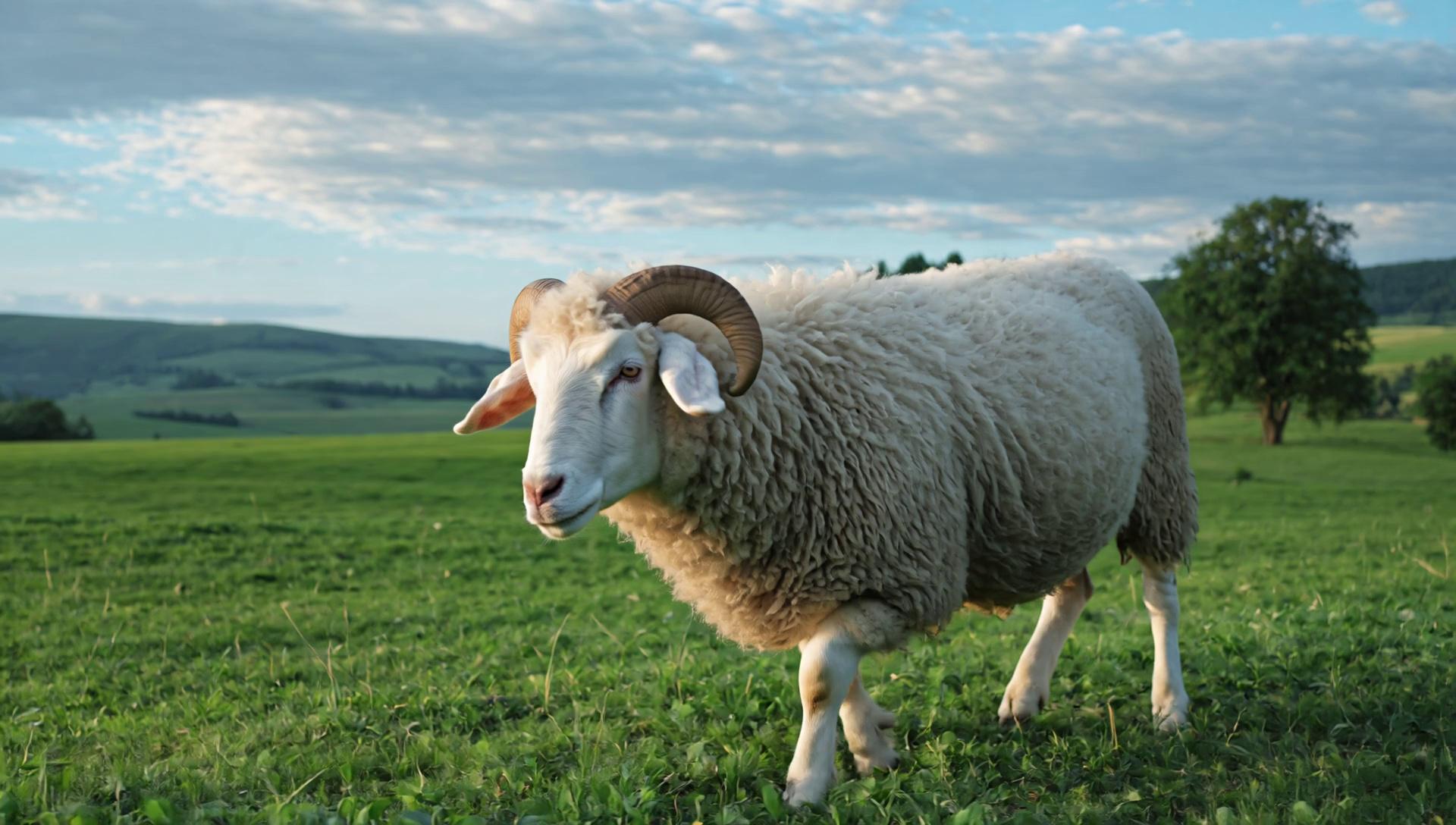} & 
  \includegraphics[width=\linewidth]{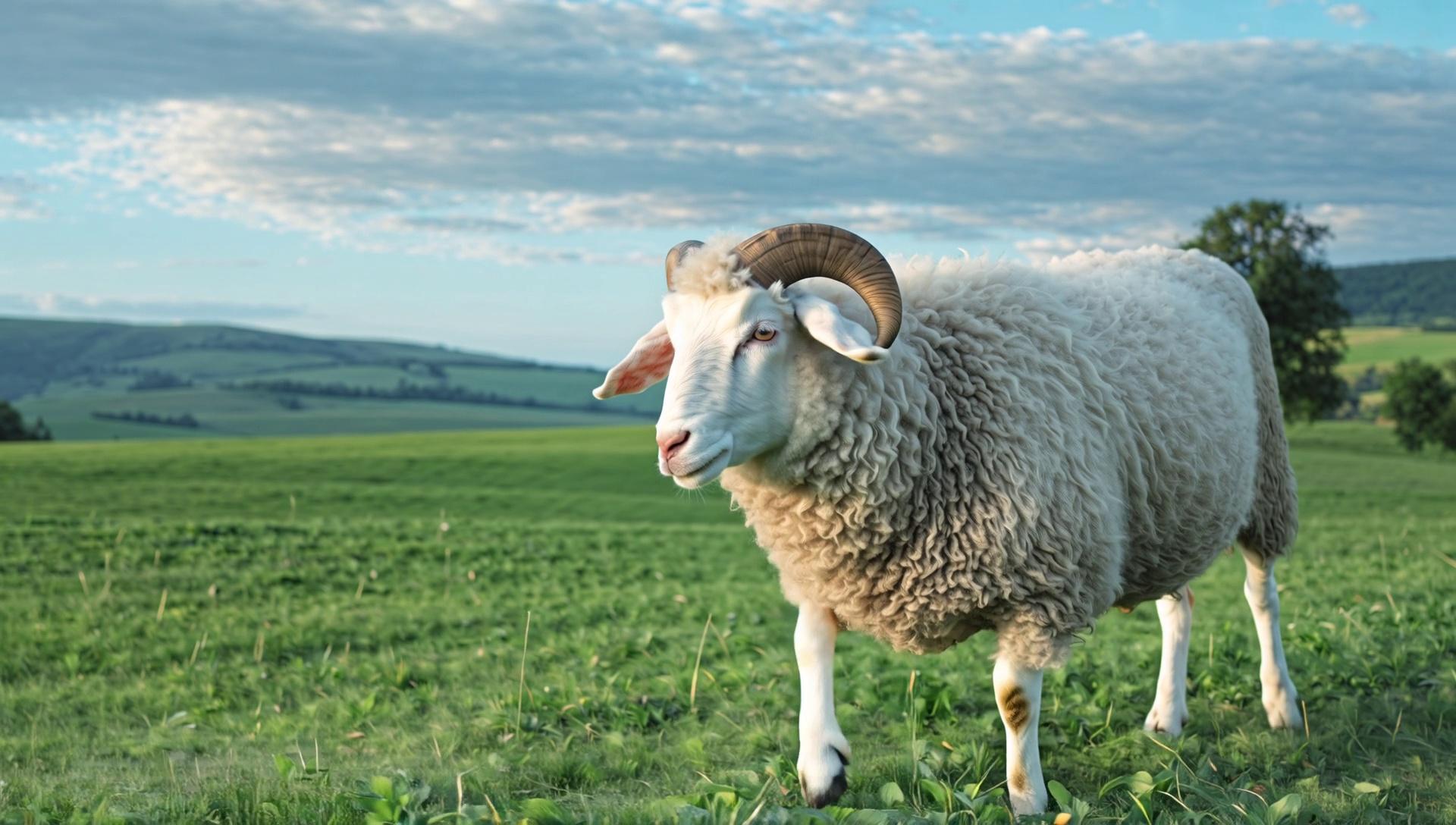} & 
  \includegraphics[width=\linewidth]{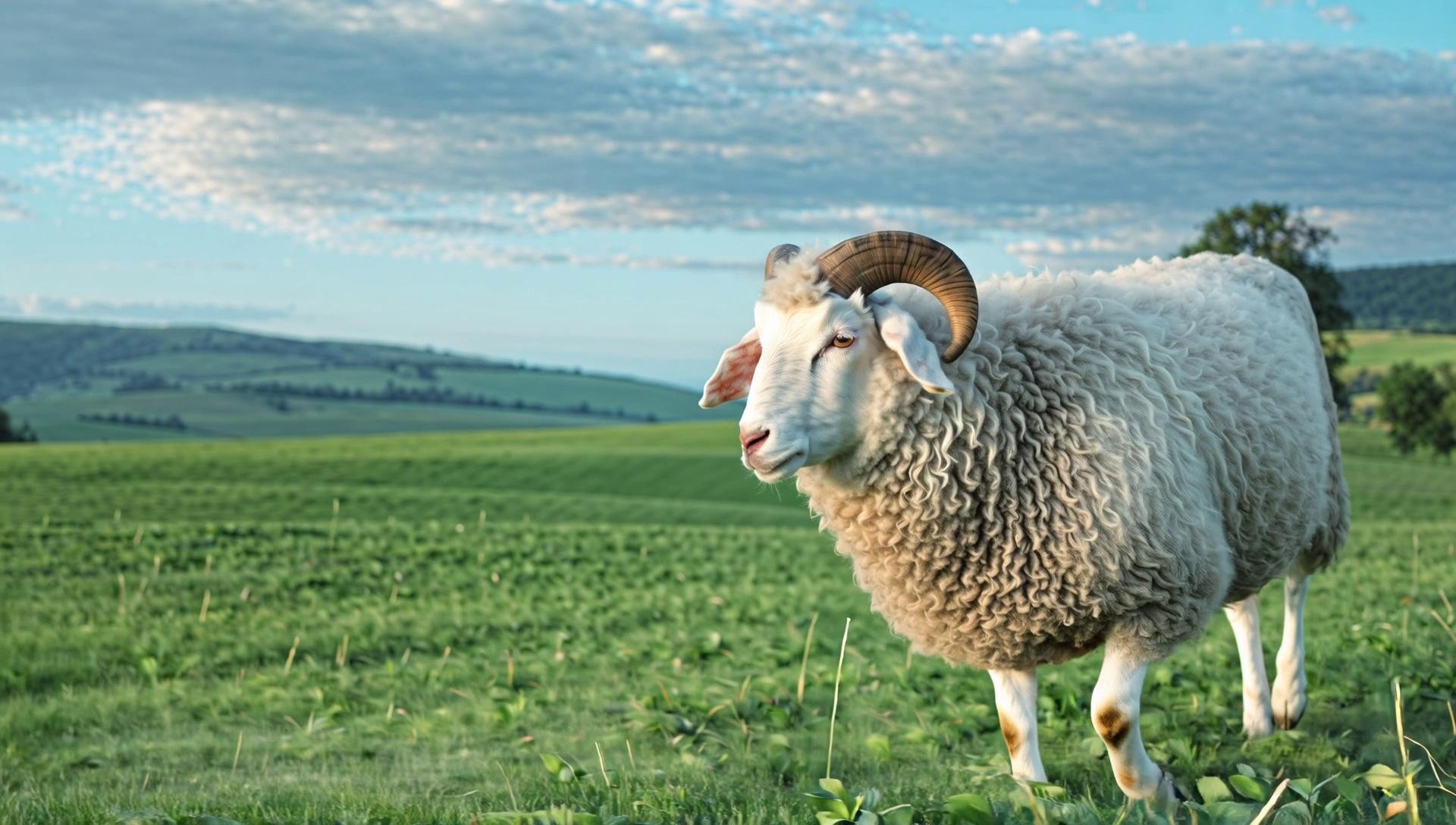} & 
  \includegraphics[width=\linewidth]{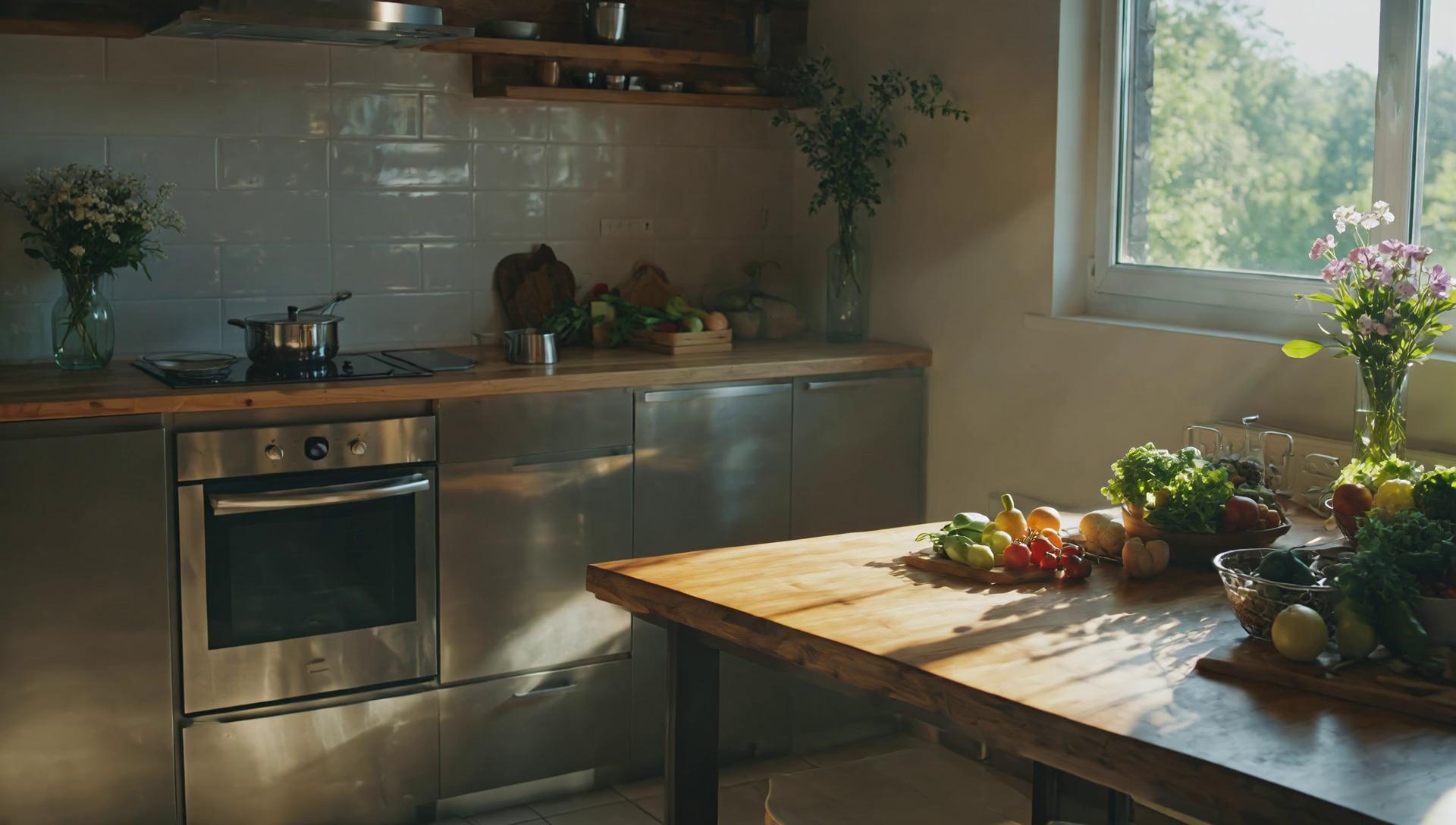} & 
  \includegraphics[width=\linewidth]{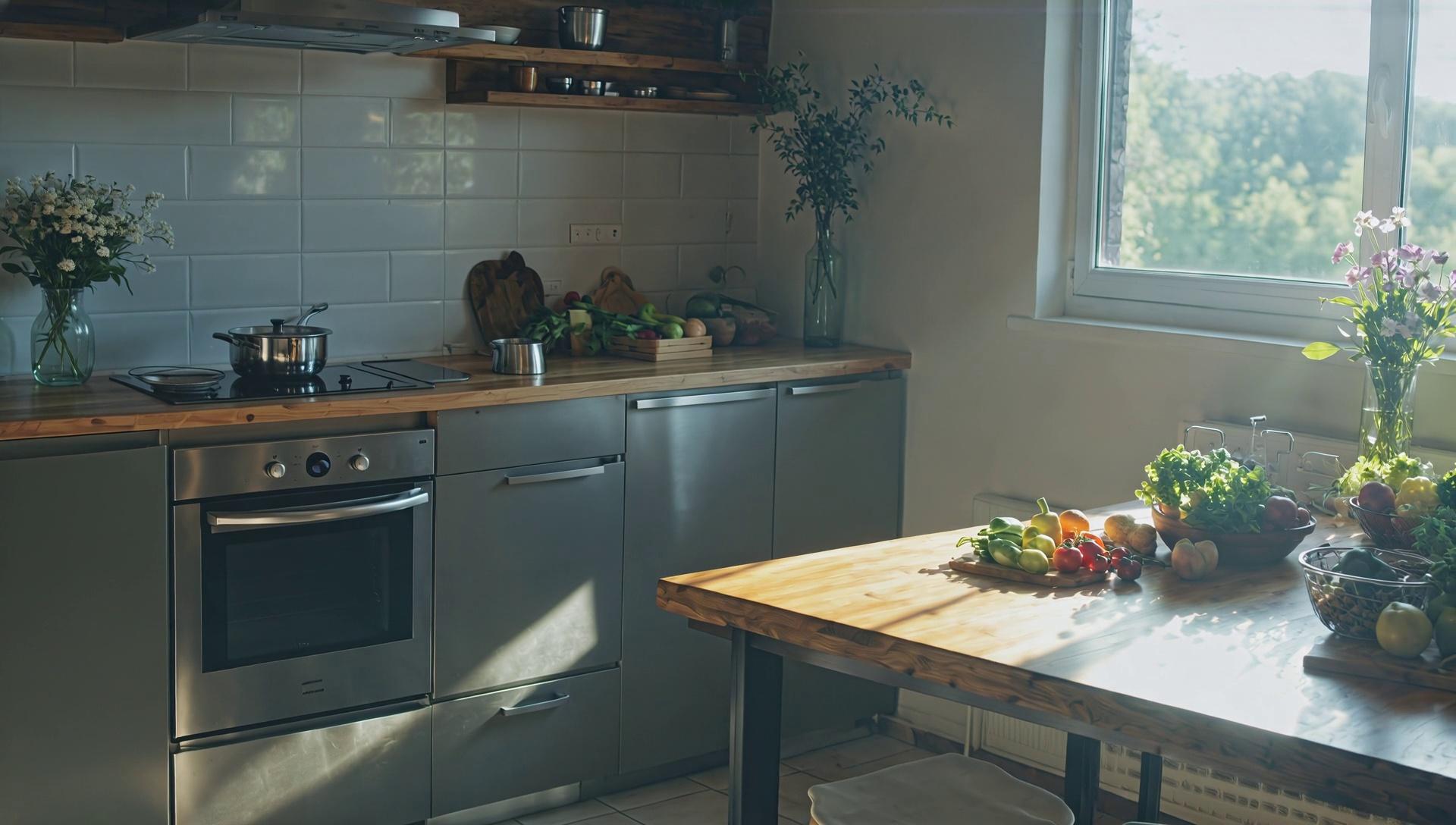} & 
  \includegraphics[width=\linewidth]{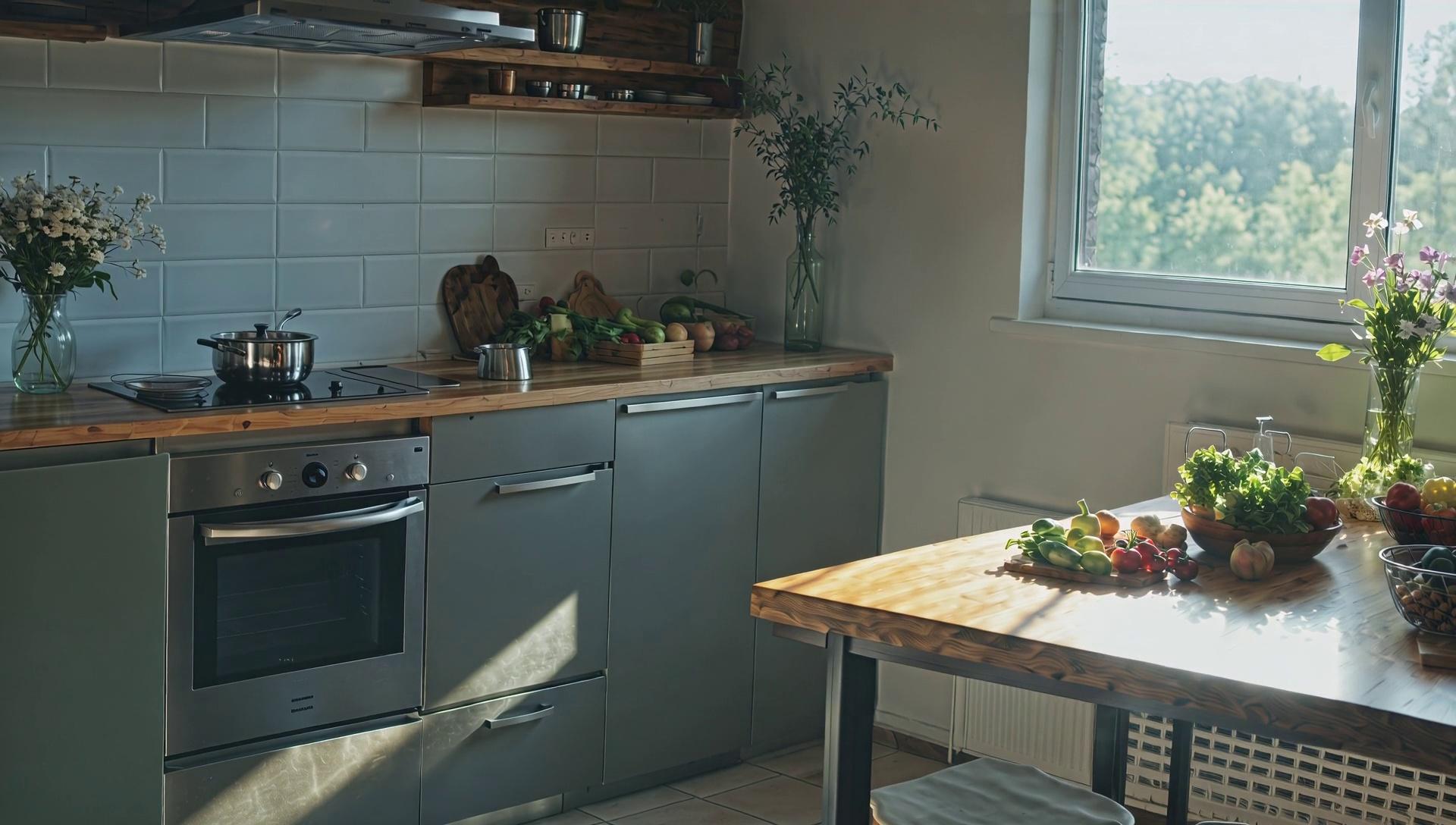} \\
\end{tabular}
\caption{\textbf{Qualitative comparisons.} We visualize high-resolution videos generated by HiStream (Ours) against those by Wan2.1~\citep{wan2025}, Self Forcing~\citep{huang2025self}, LTX~\citep{HaCohen2024LTXVideo}, and FlashVideo~\citep{zhang2025flashvideo}. The videos generated by HiStream exhibit the highest visual fidelity and the cleanest texture, free from spurious patterns or visible artifacts. Best viewed \textbf{ZOOMED-IN}.}
\vspace{-1.0em}
\label{fig:baseline}
\end{figure*}

\begin{figure*}[t!]
\centering
\setlength{\tabcolsep}{0.1em}  
\renewcommand{\arraystretch}{0.2}
 \begin{tabular}{C{0.09\linewidth} C{0.145\linewidth} C{0.145\linewidth} C{0.145\linewidth} @{\hspace{0.5em}} C{0.145\linewidth} C{0.145\linewidth} C{0.145\linewidth}}
  w/o HD Tech & 
  \includegraphics[width=\linewidth]{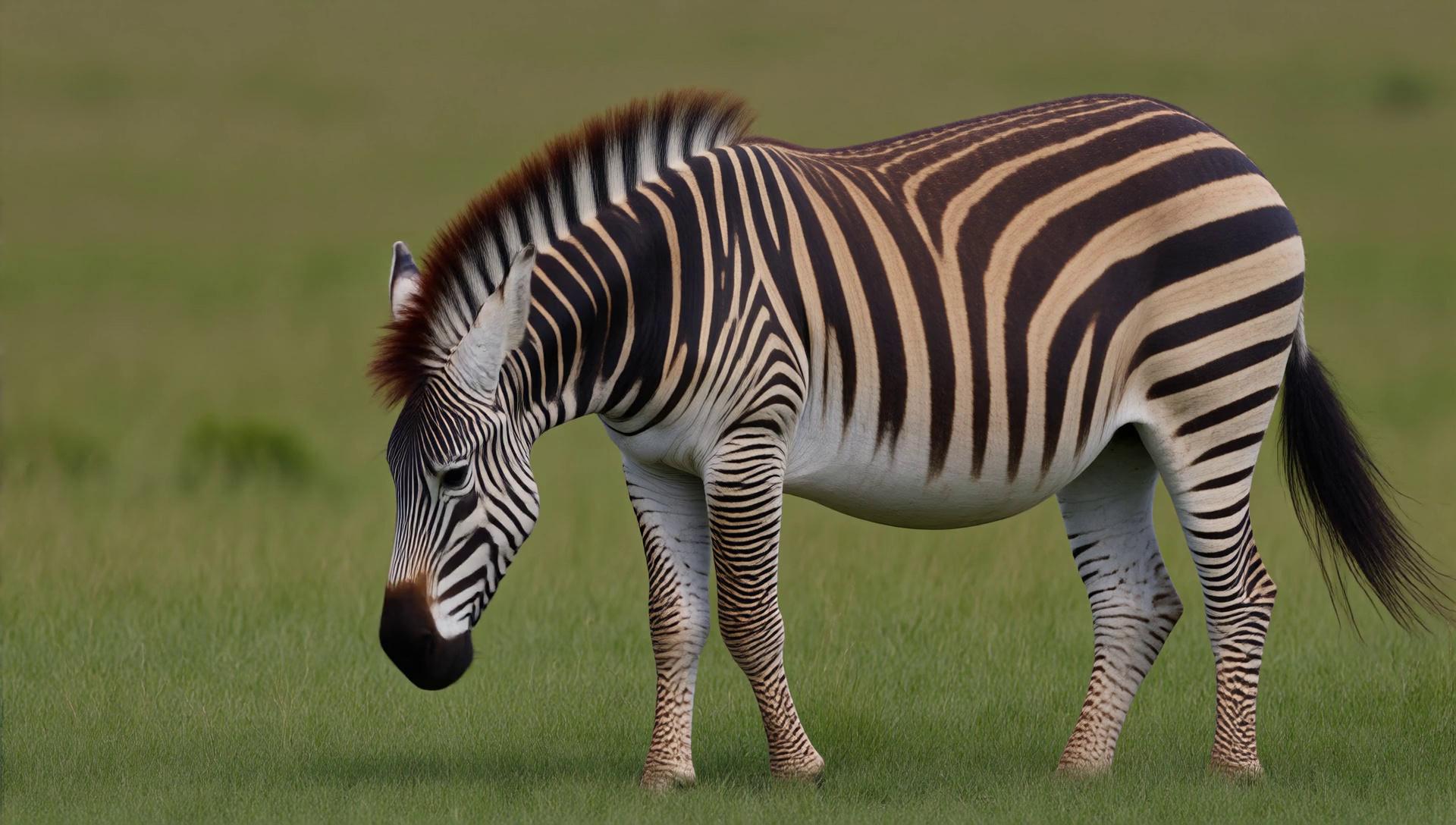} & 
  \includegraphics[width=\linewidth]{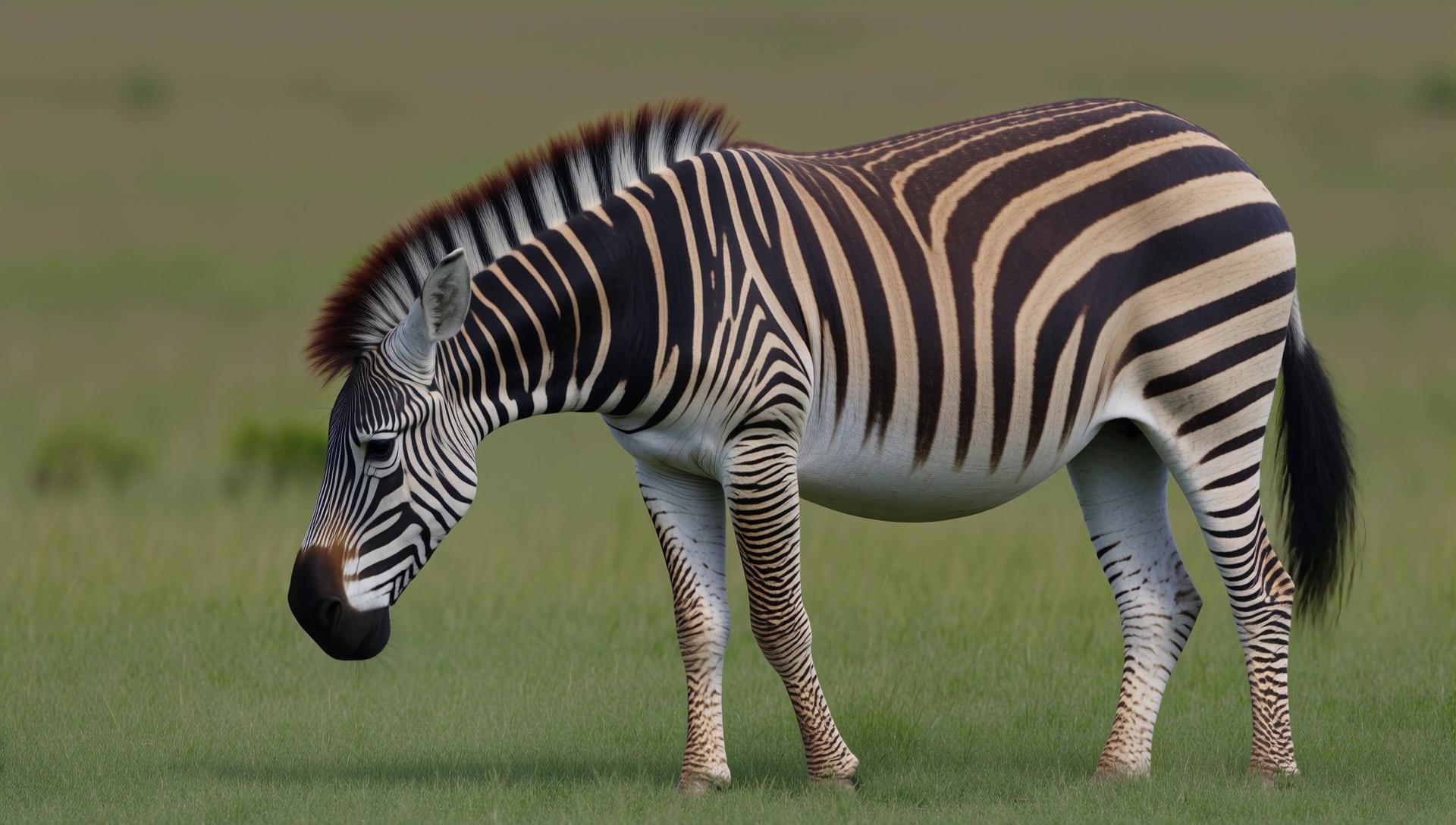} & 
  \includegraphics[width=\linewidth]{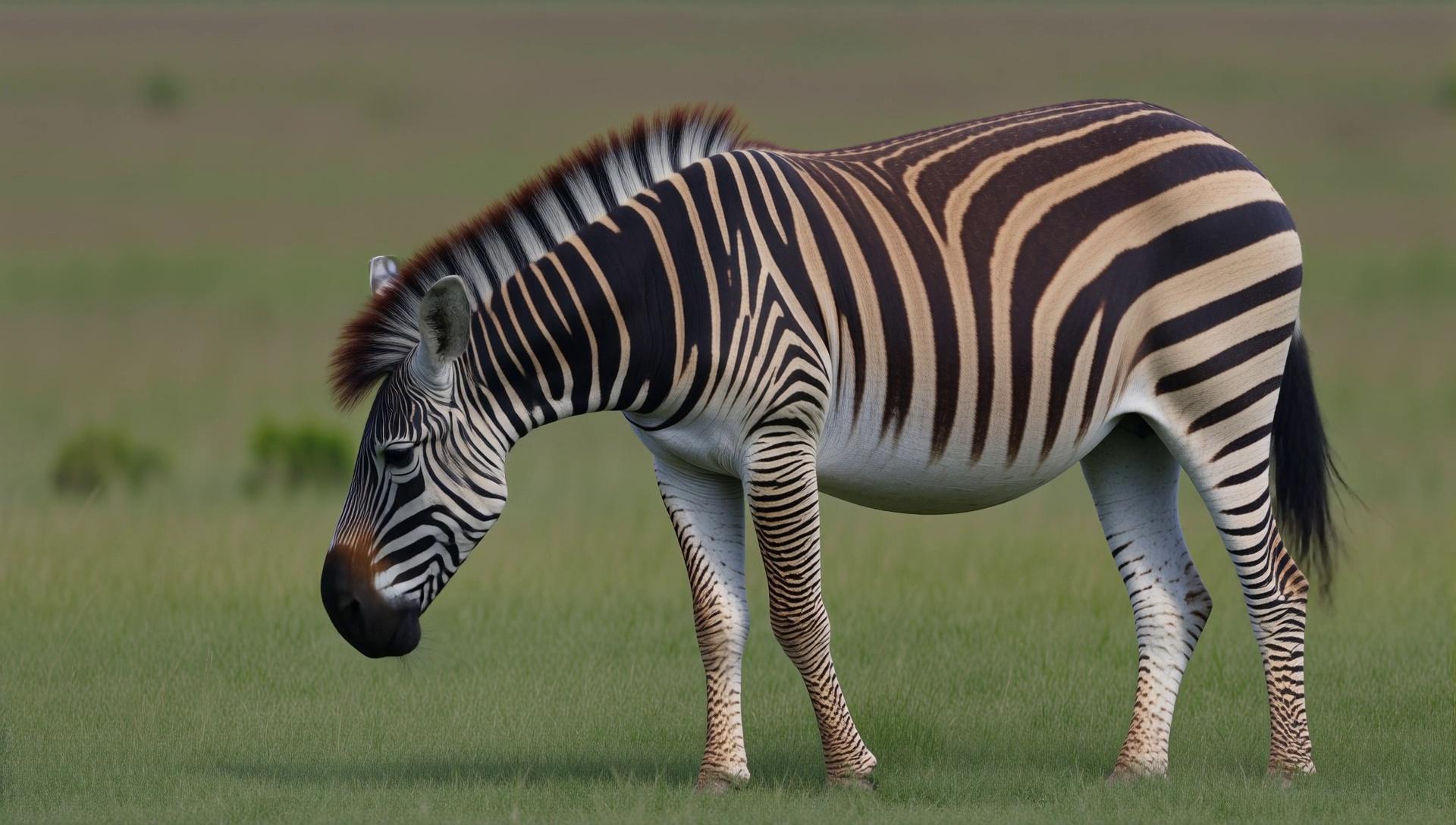} & 
  \includegraphics[width=\linewidth]{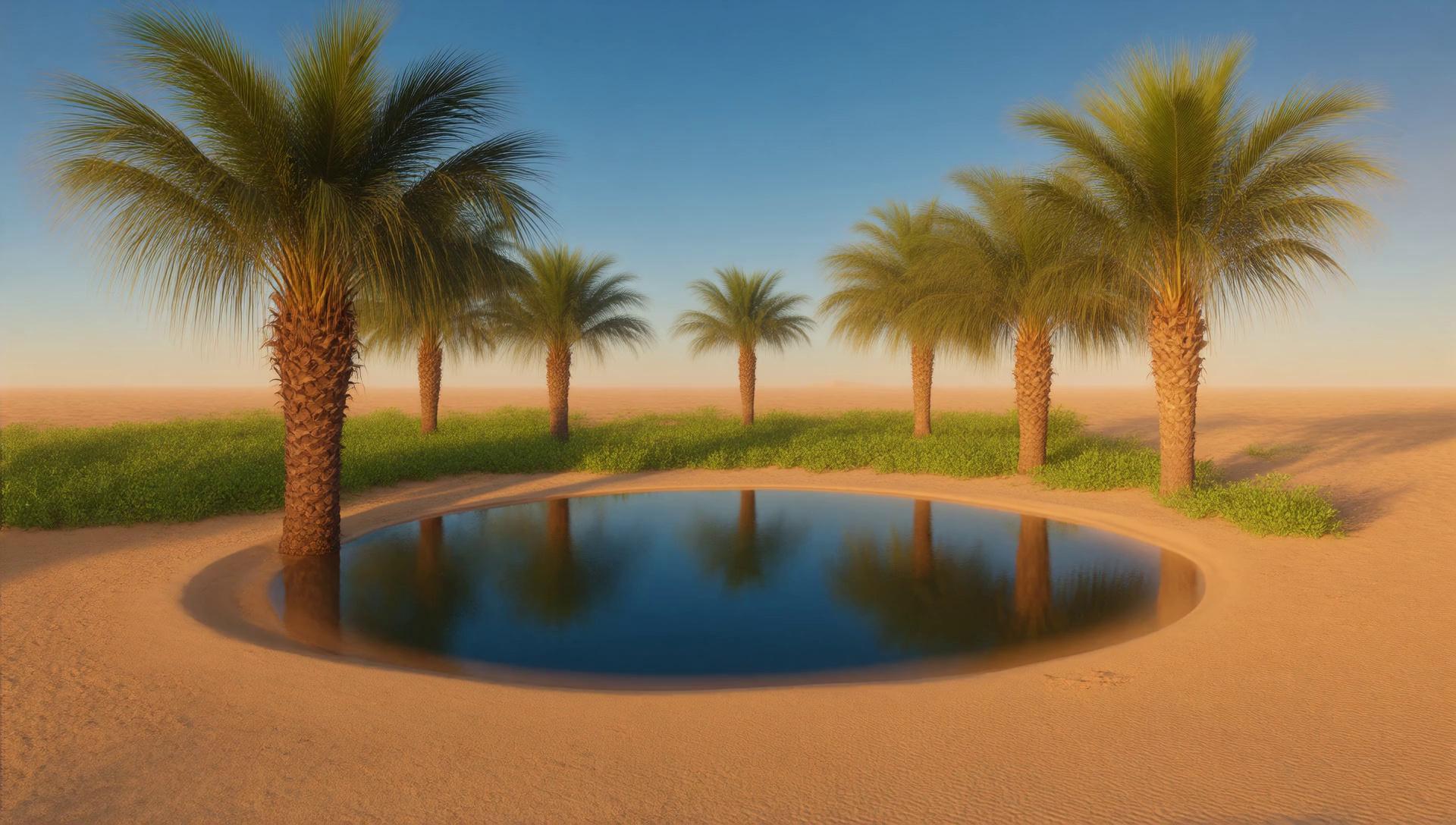} & 
  \includegraphics[width=\linewidth]{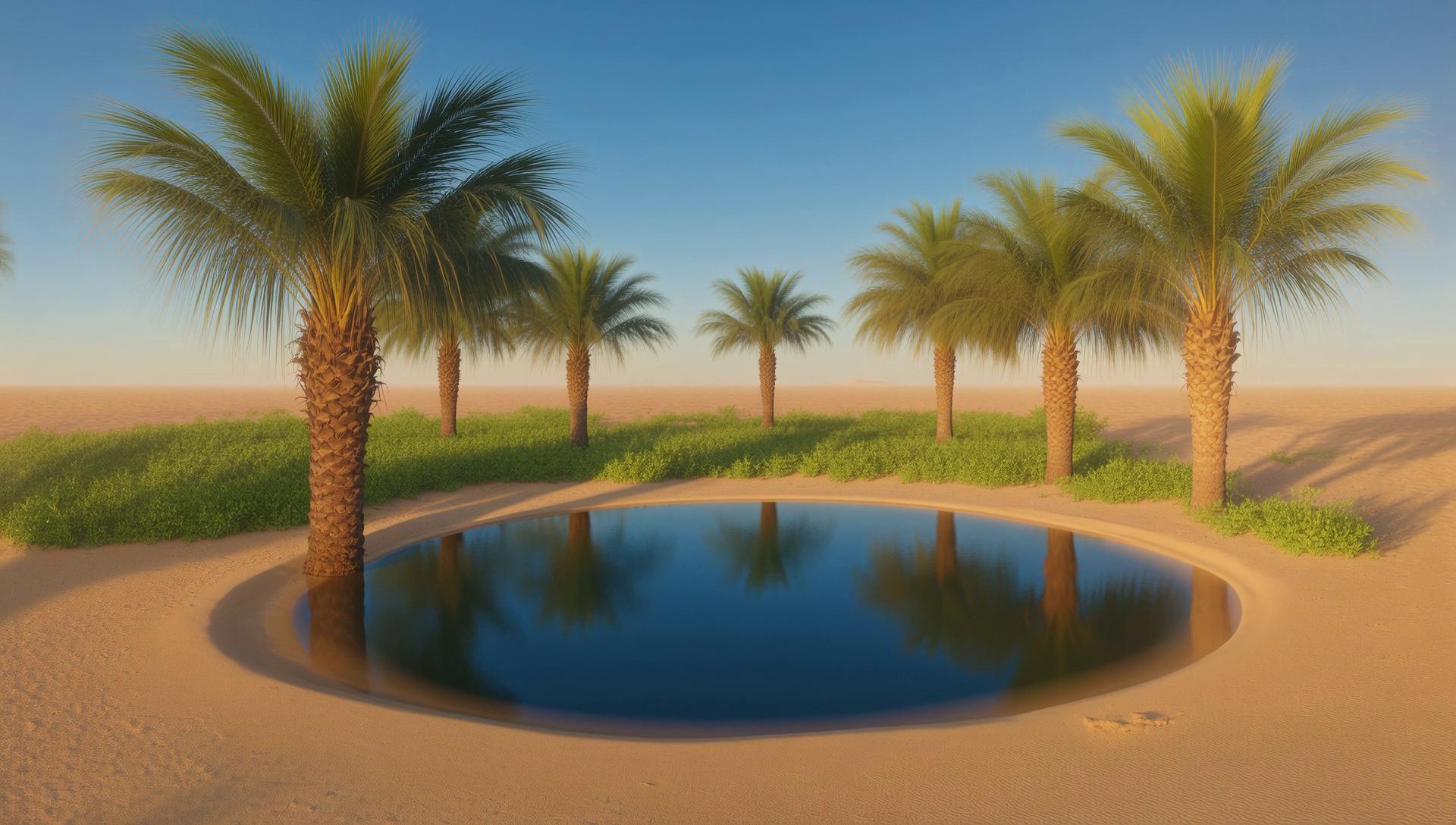} & 
  \includegraphics[width=\linewidth]{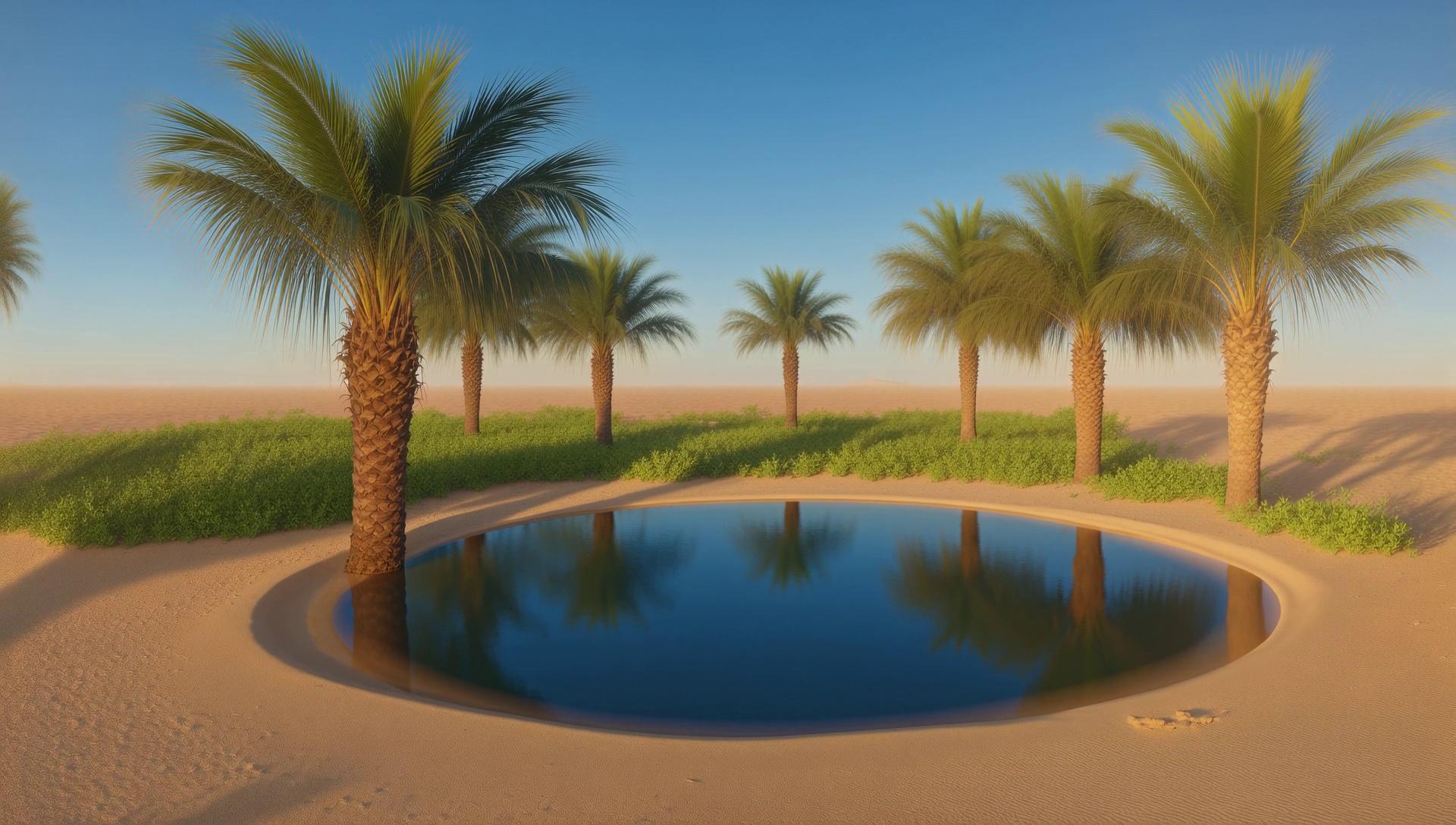} \\
  w/o Tuning & 
  \includegraphics[width=\linewidth]{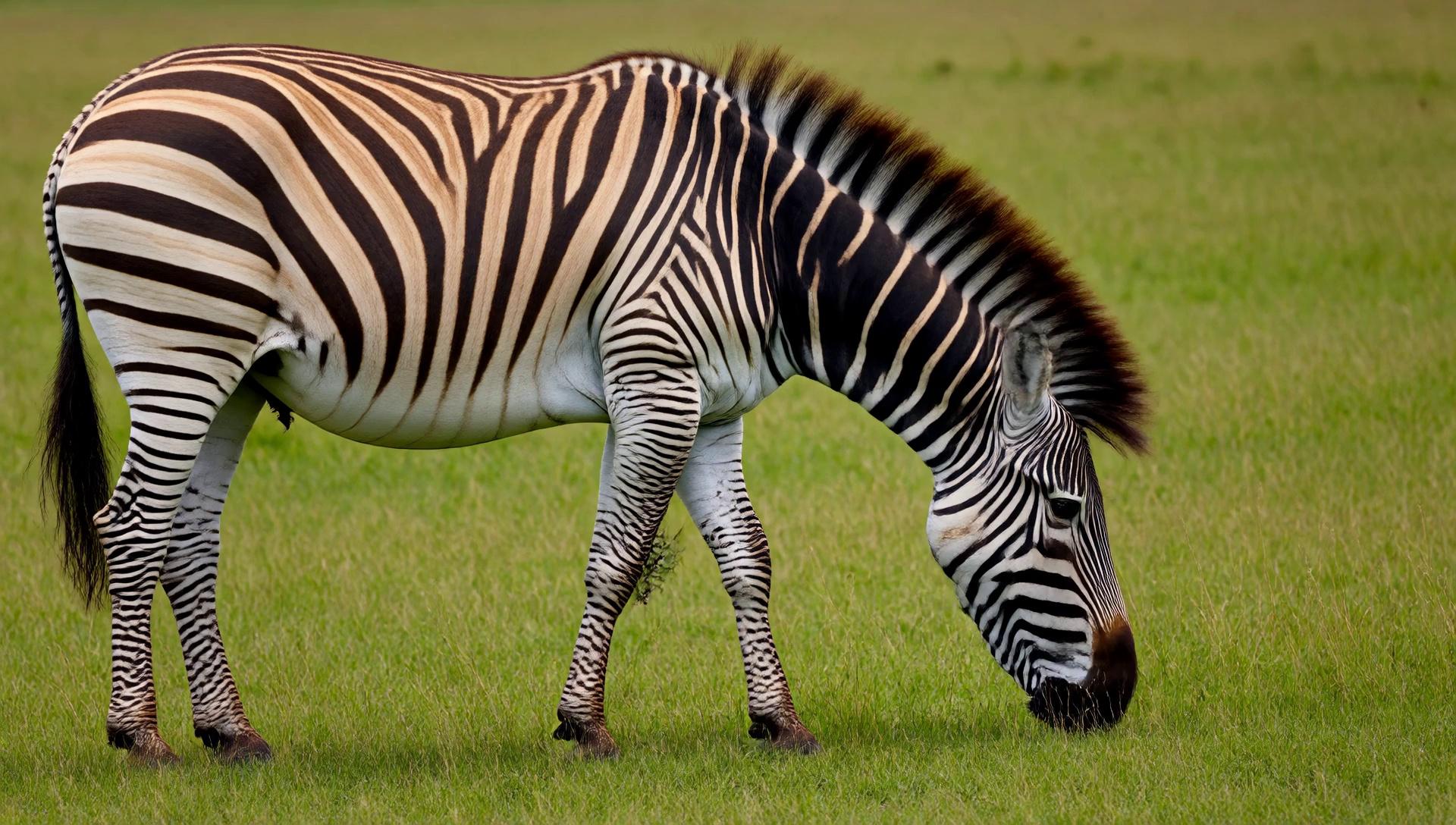} & 
  \includegraphics[width=\linewidth]{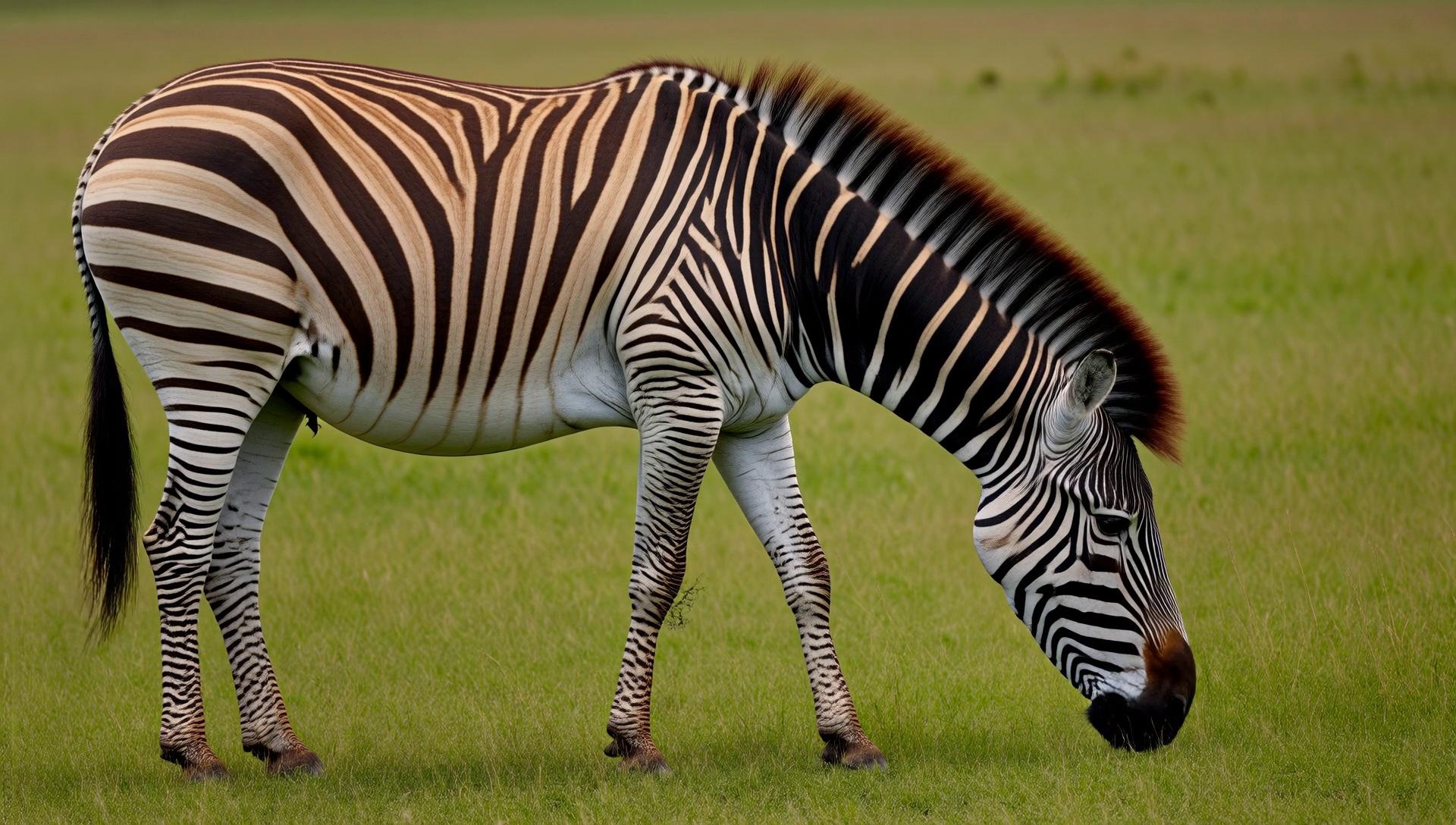} & 
  \includegraphics[width=\linewidth]{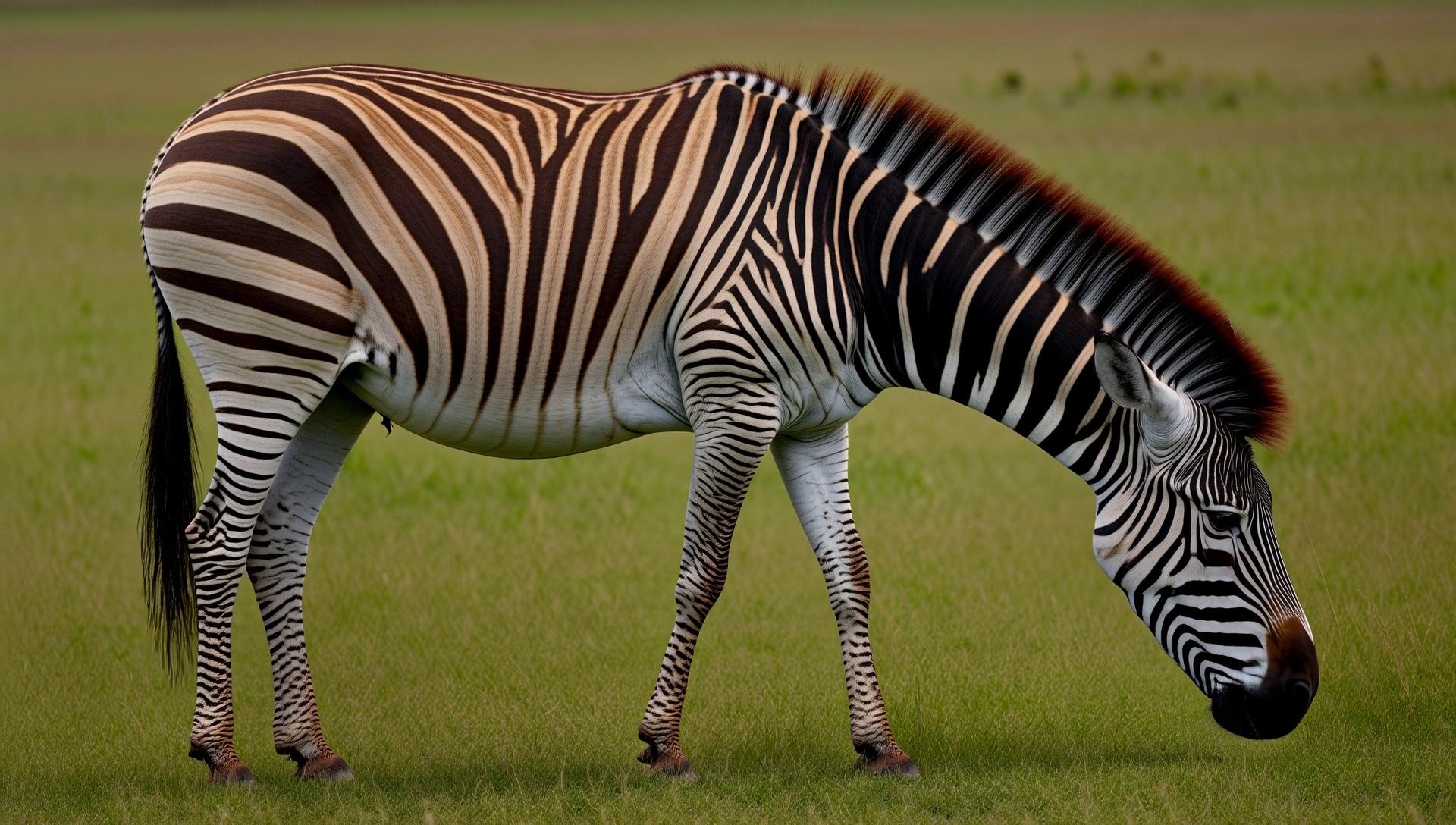} & 
  \includegraphics[width=\linewidth]{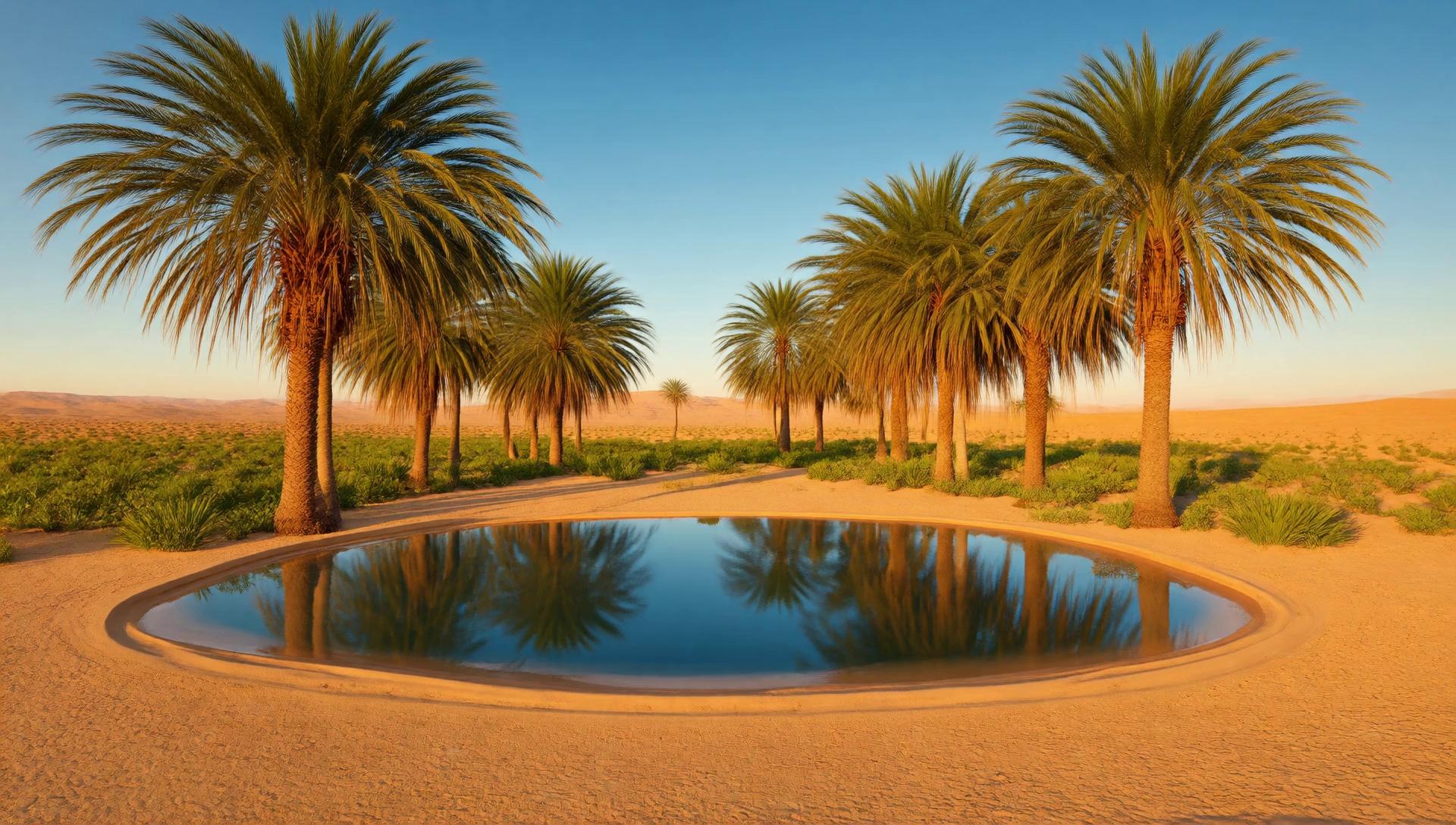} & 
  \includegraphics[width=\linewidth]{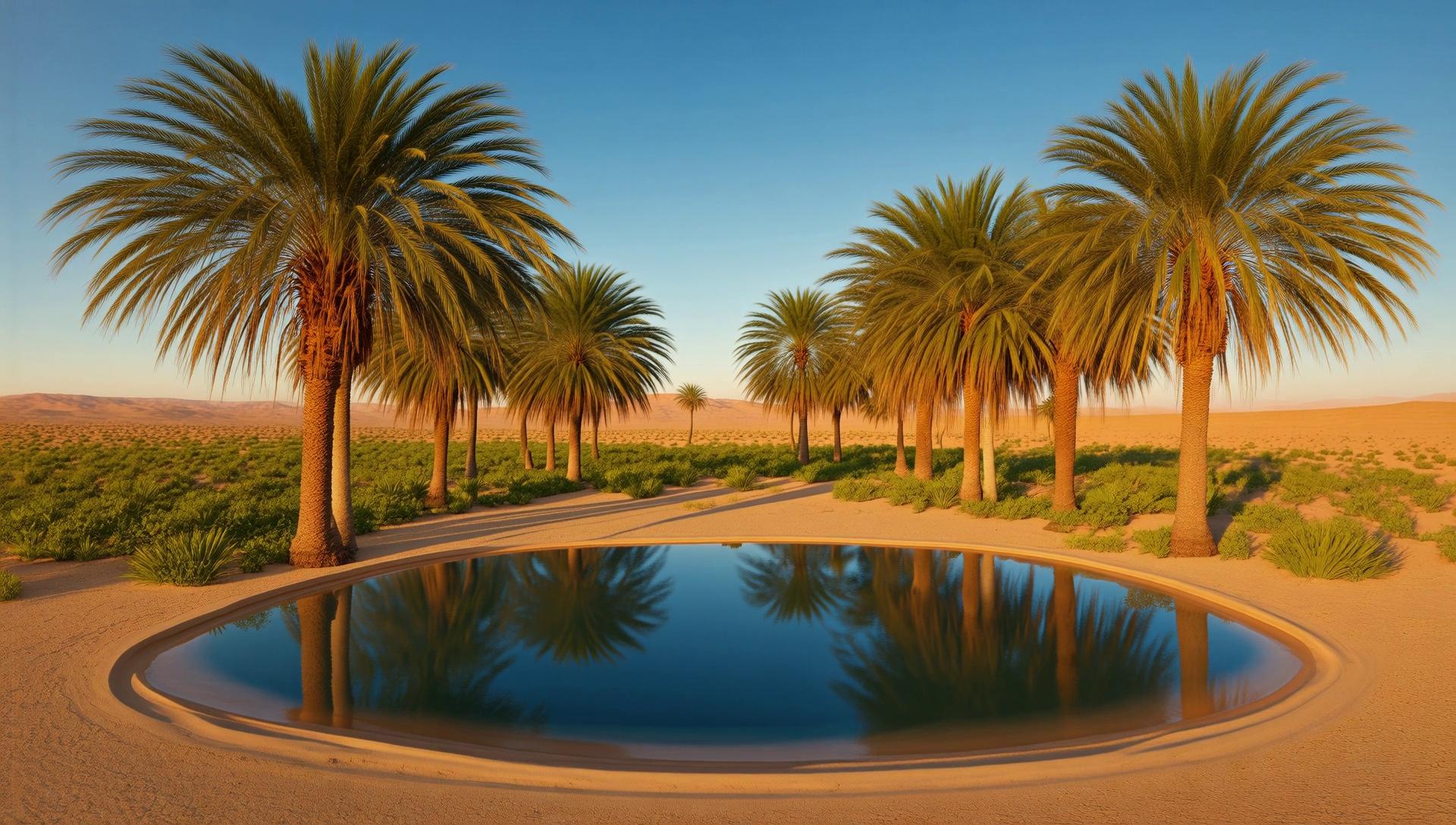} & 
  \includegraphics[width=\linewidth]{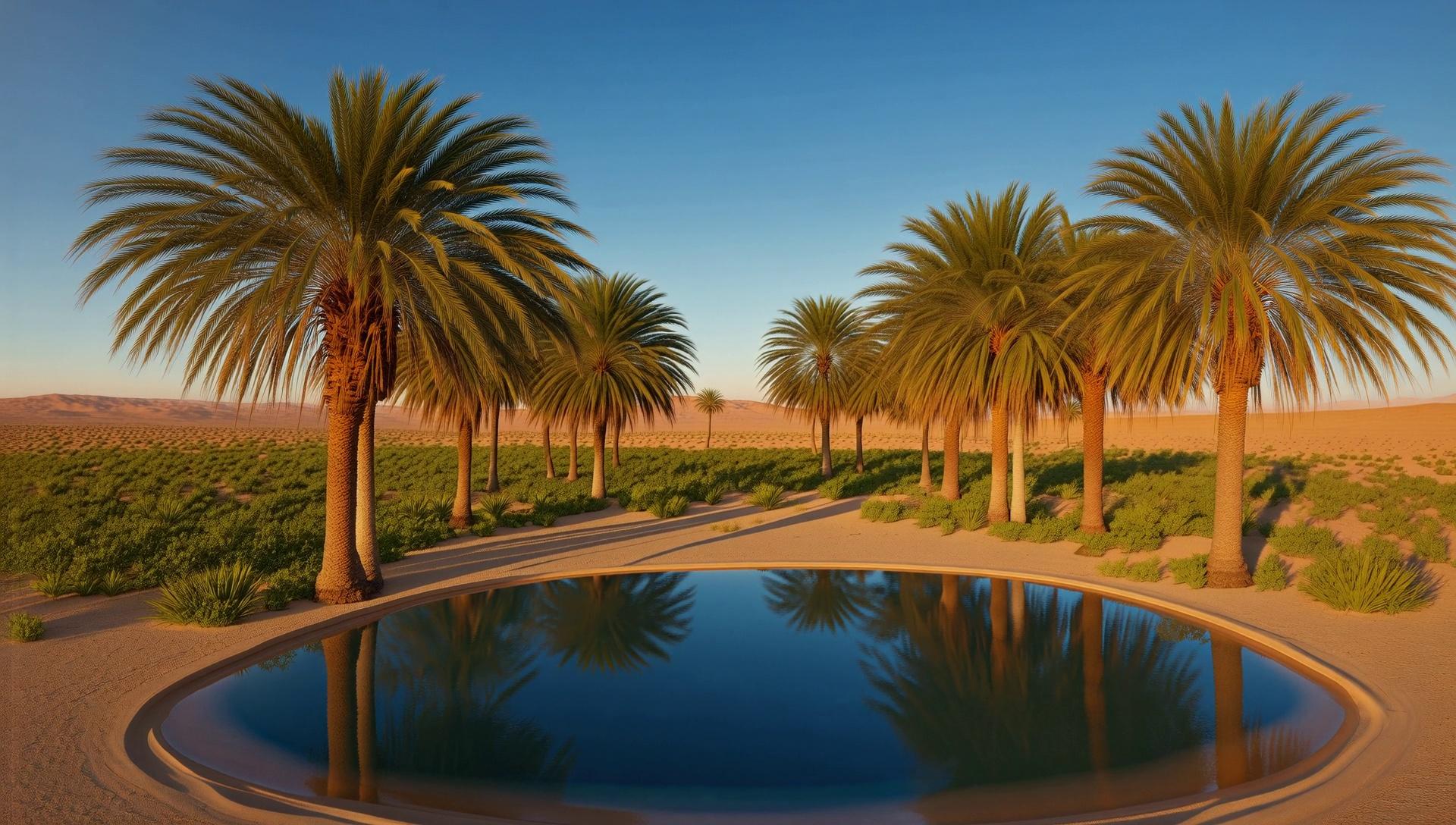} \\
  w/o DRC & 
  \includegraphics[width=\linewidth]{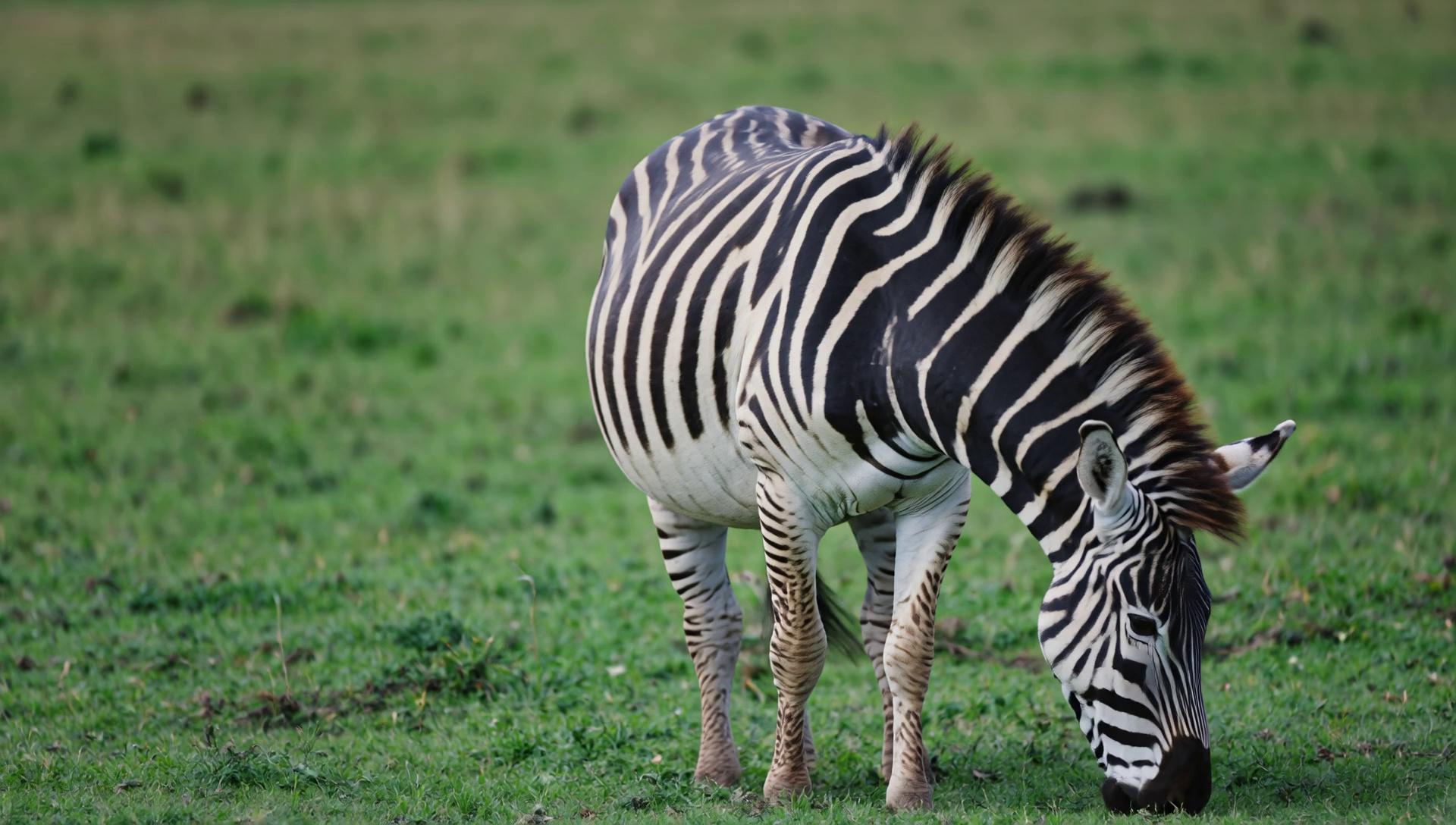} & 
  \includegraphics[width=\linewidth]{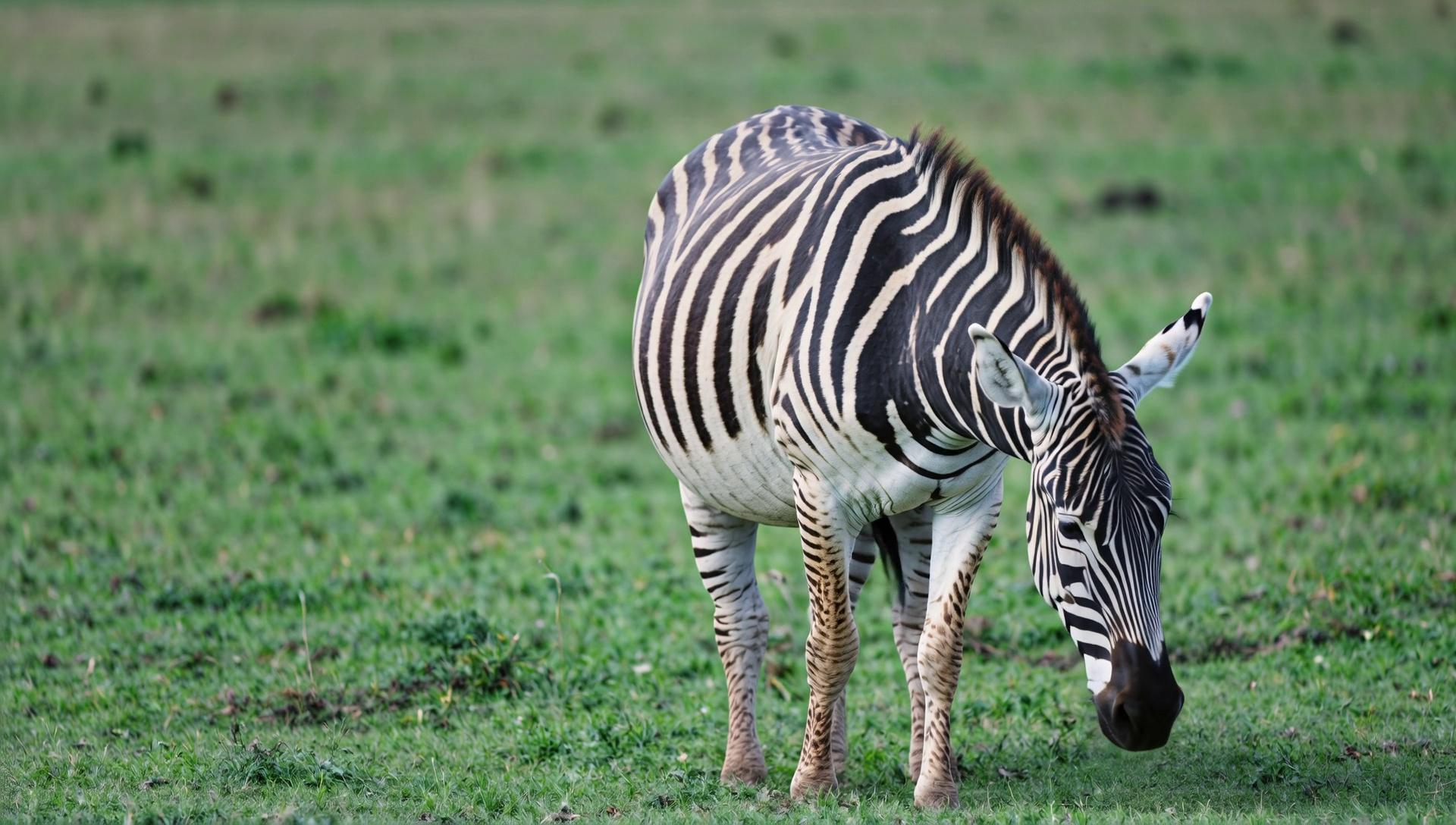} & 
  \includegraphics[width=\linewidth]{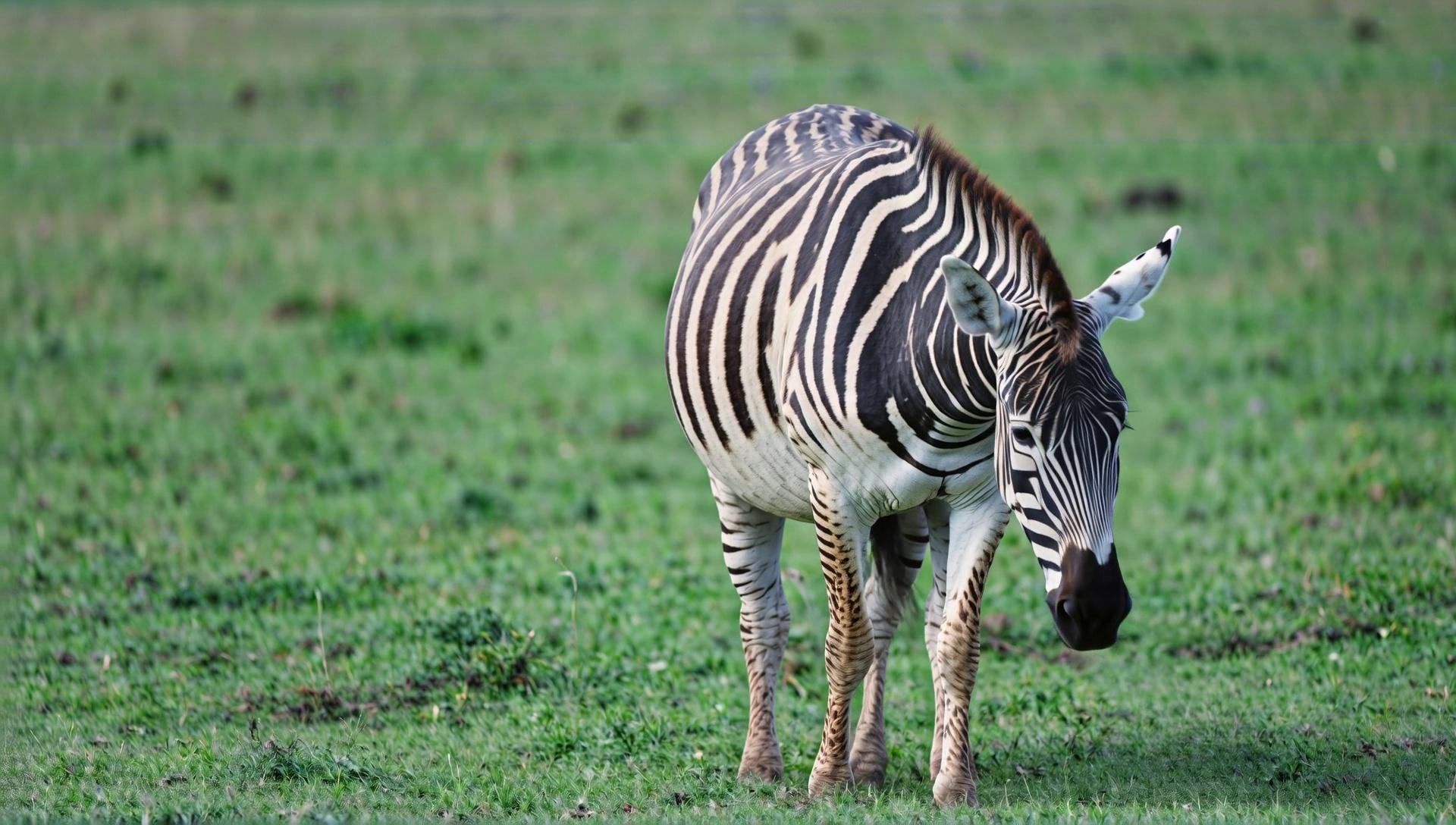} & 
  \includegraphics[width=\linewidth]{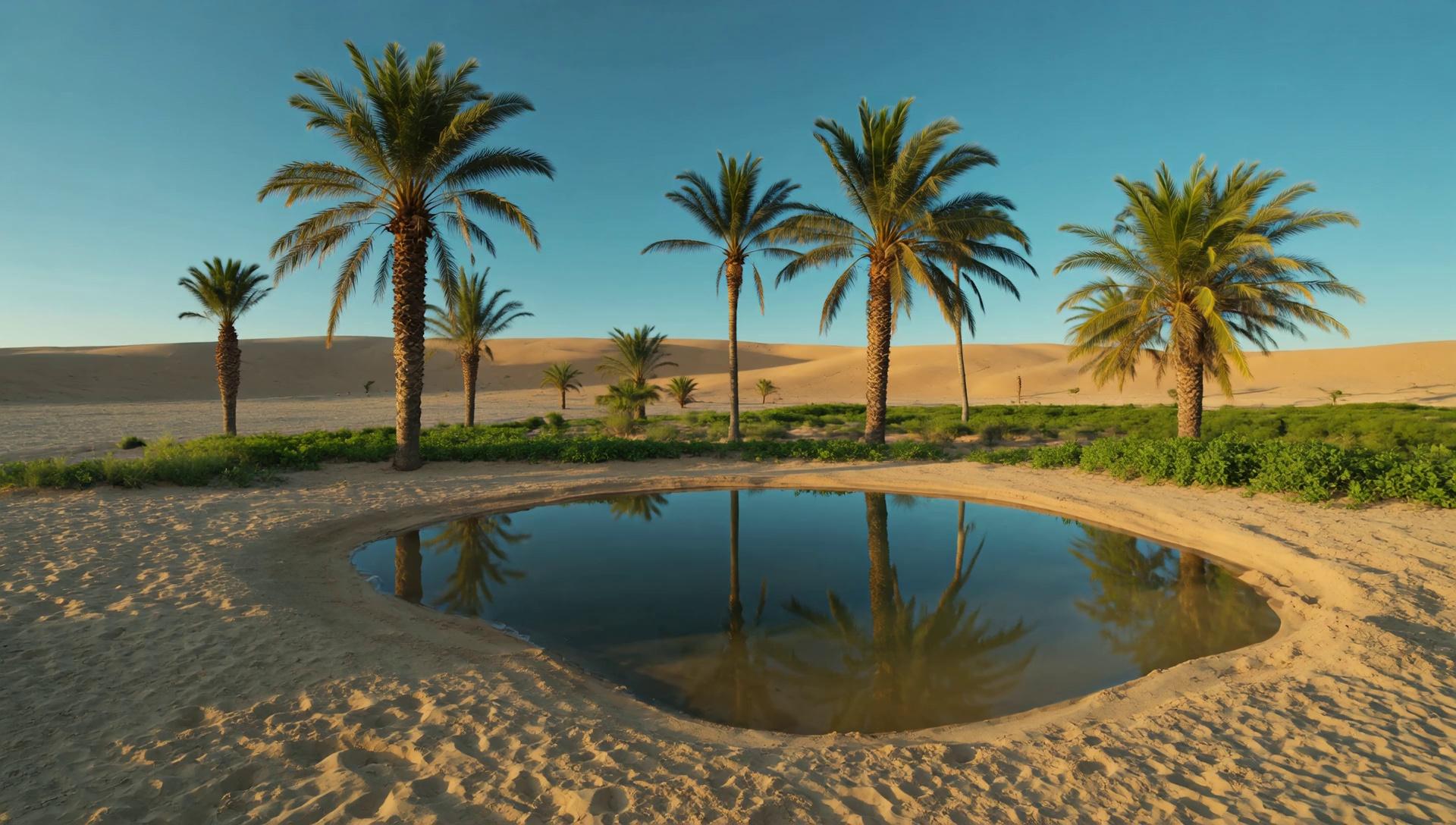} & 
  \includegraphics[width=\linewidth]{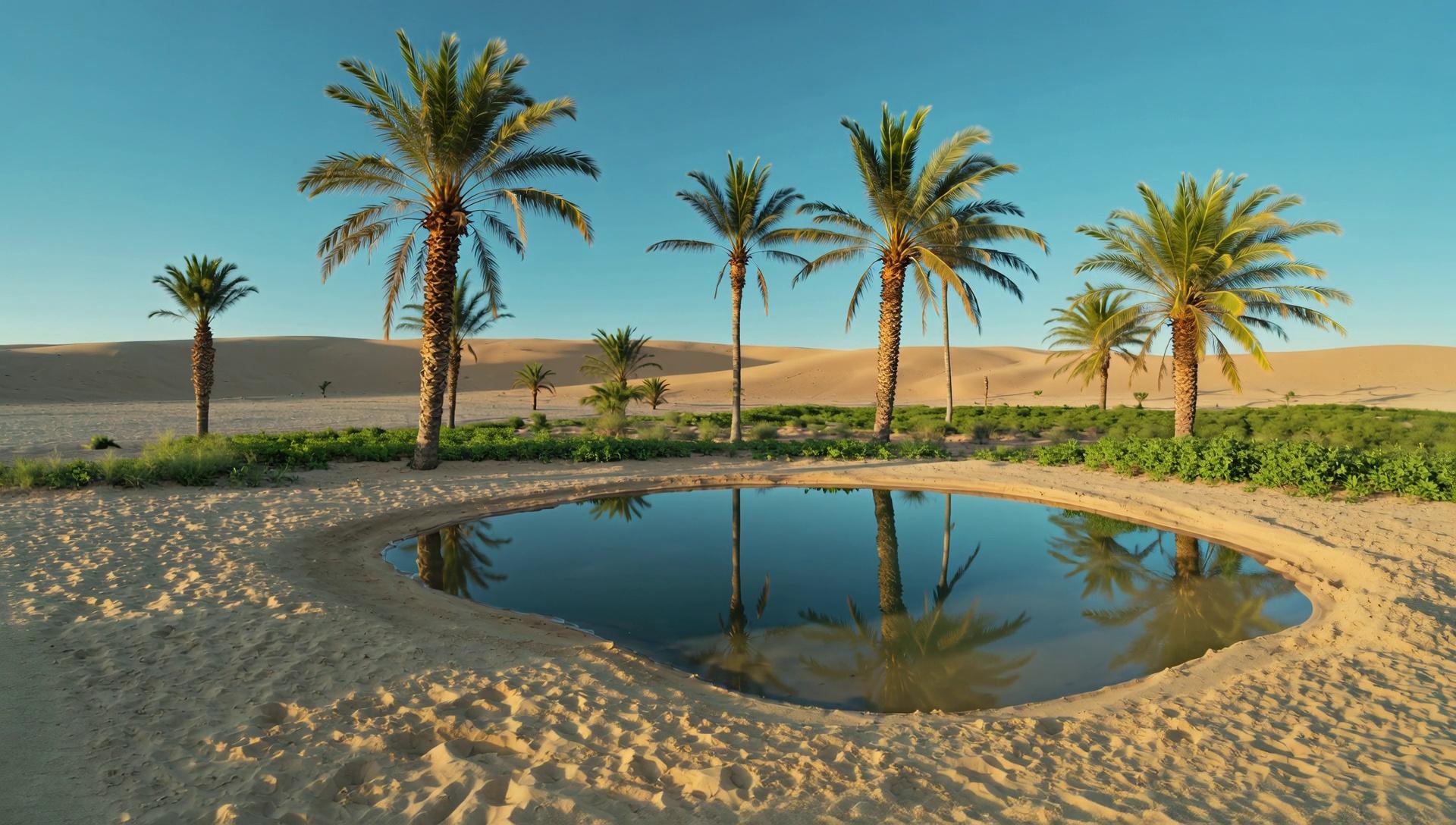} & 
  \includegraphics[width=\linewidth]{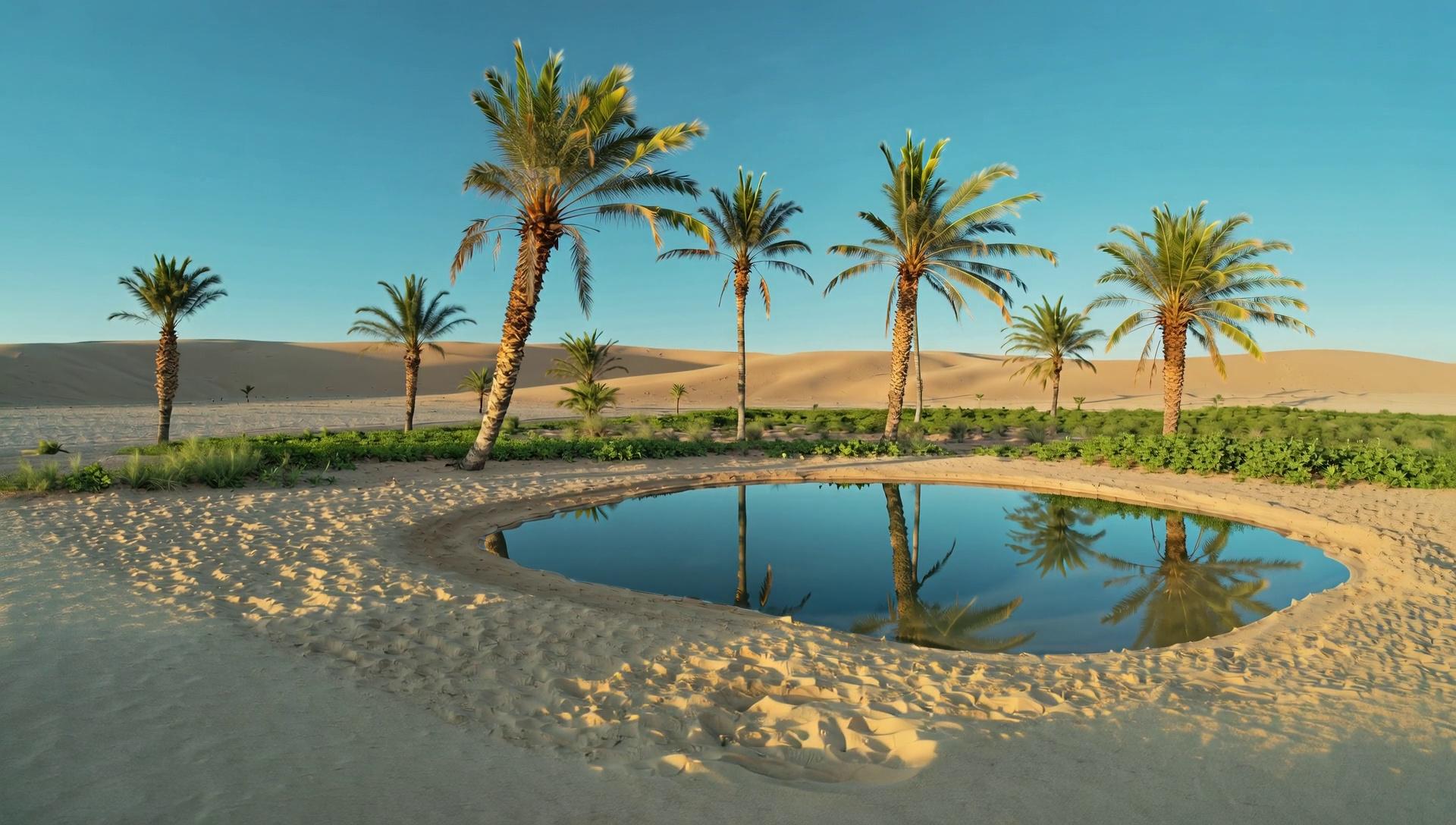} \\  
  w/o AGSW & 
  \includegraphics[width=\linewidth]{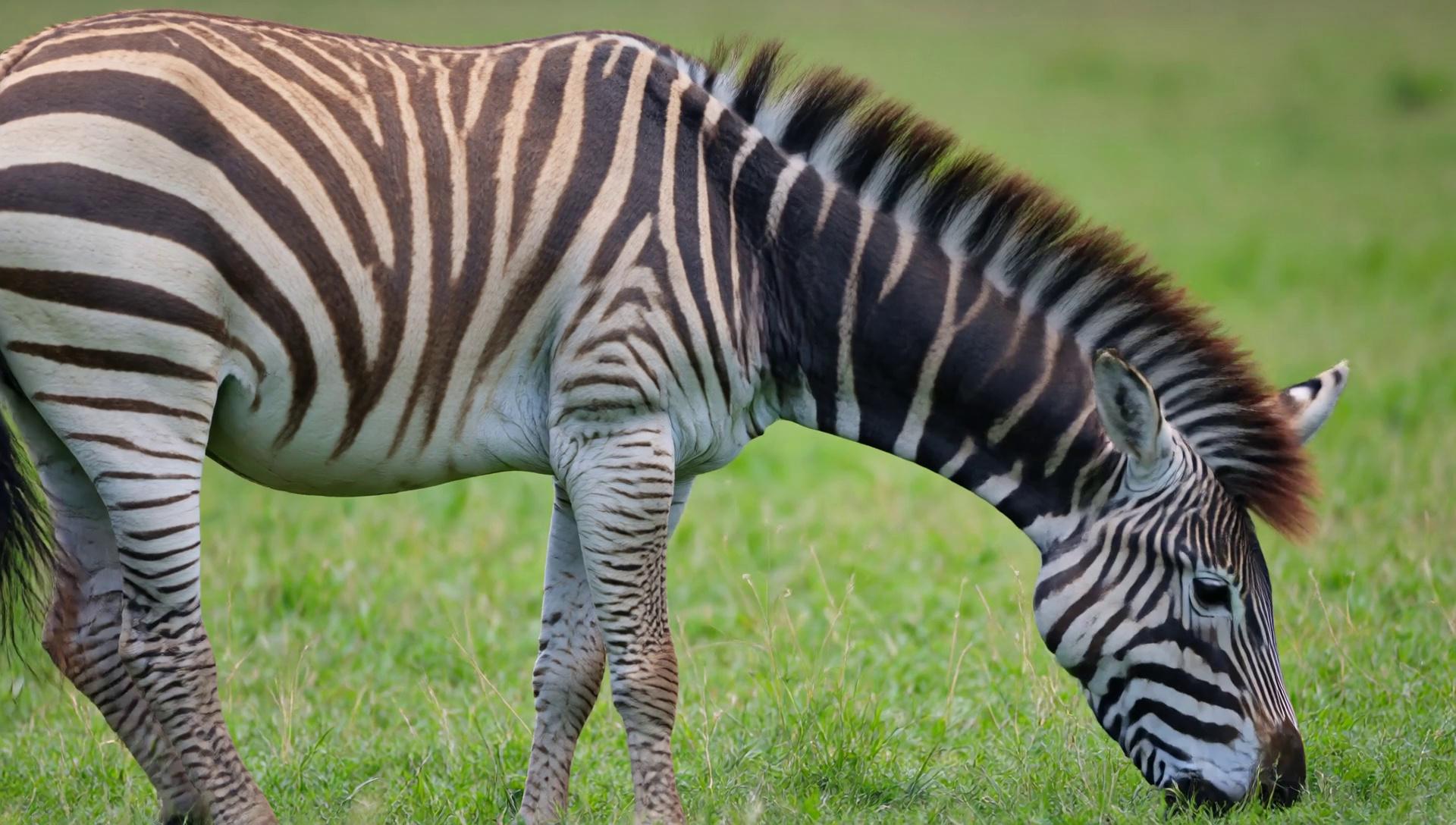} & 
  \includegraphics[width=\linewidth]{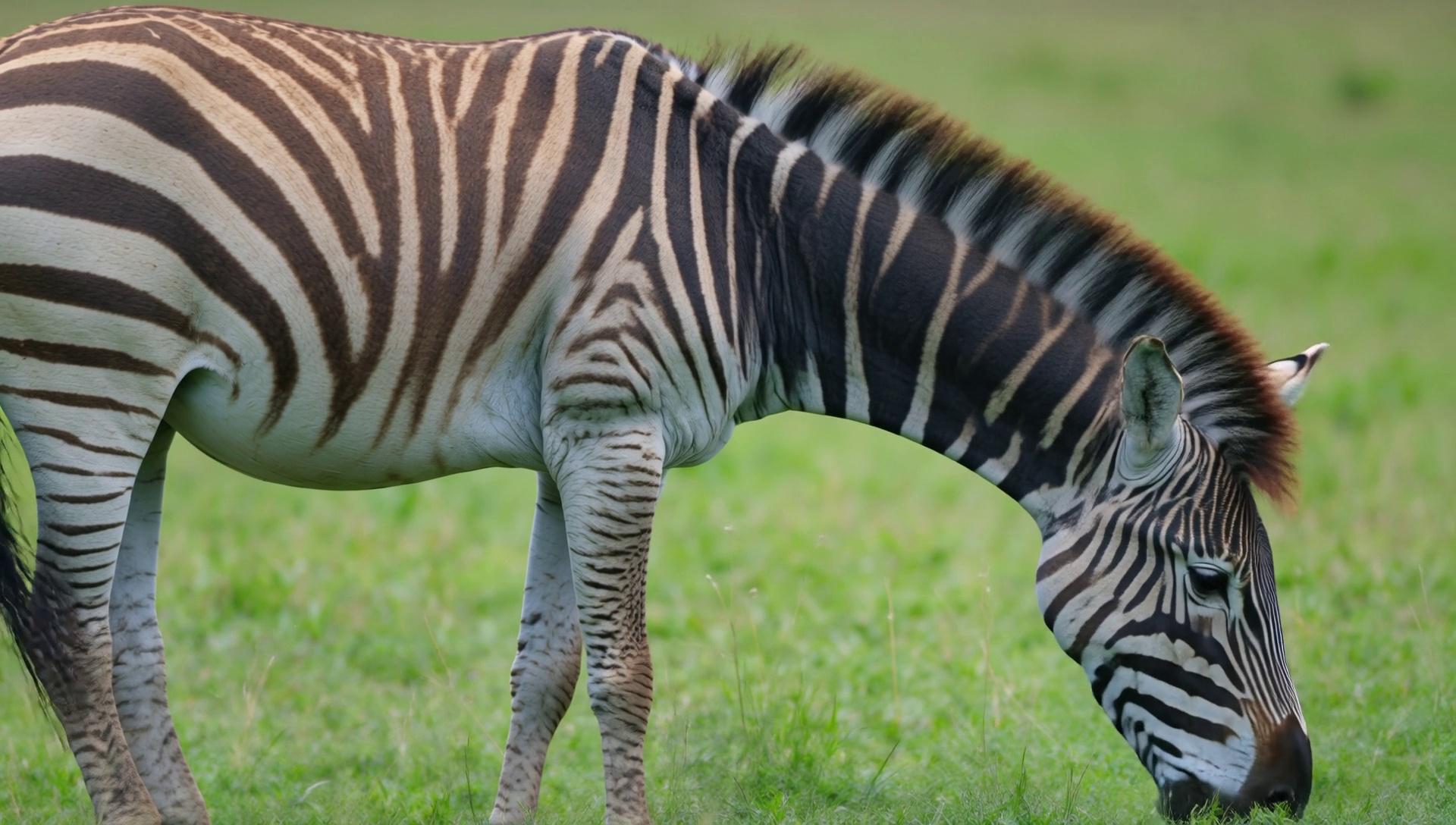} & 
  \includegraphics[width=\linewidth]{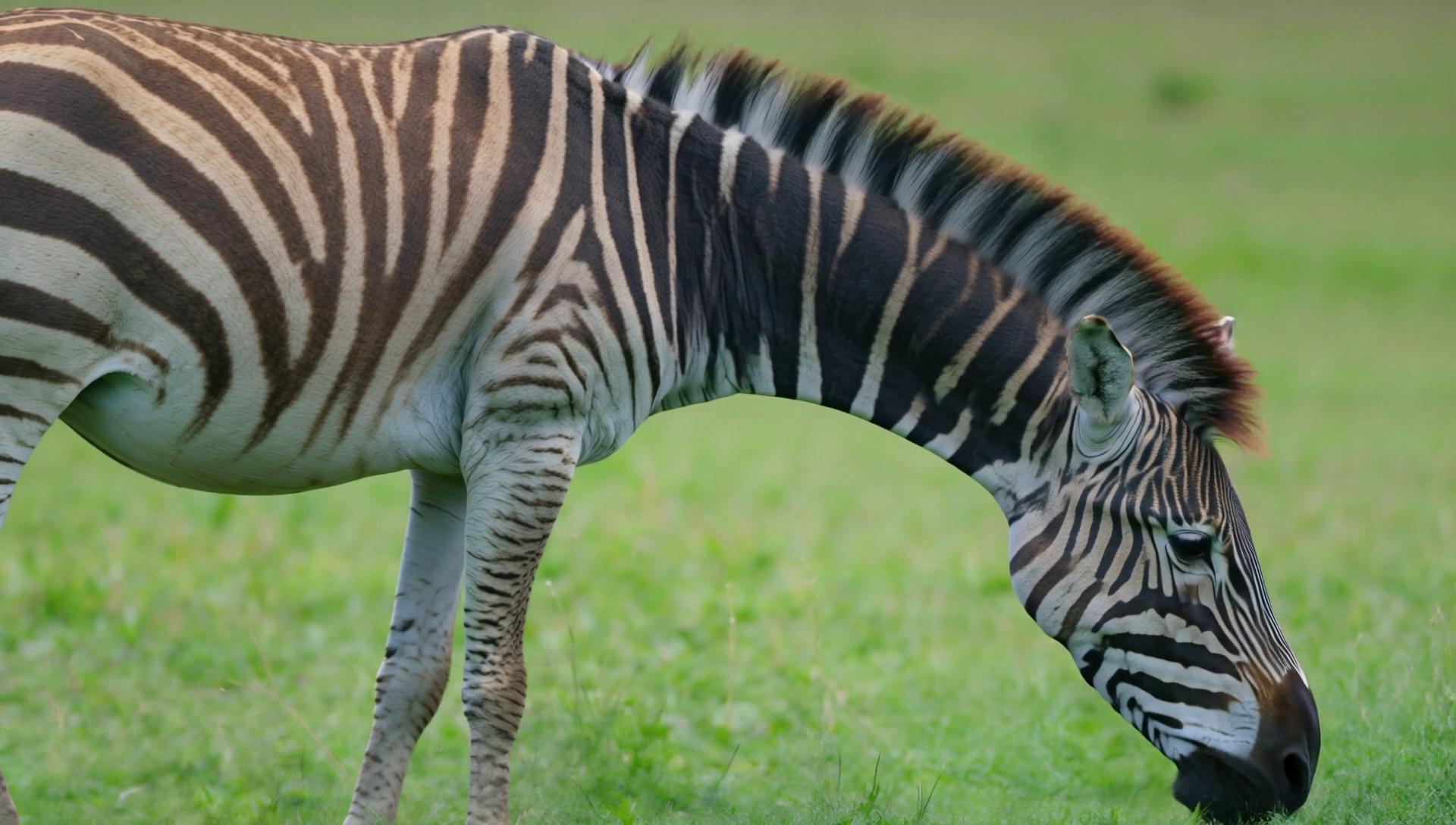} & 
  \includegraphics[width=\linewidth]{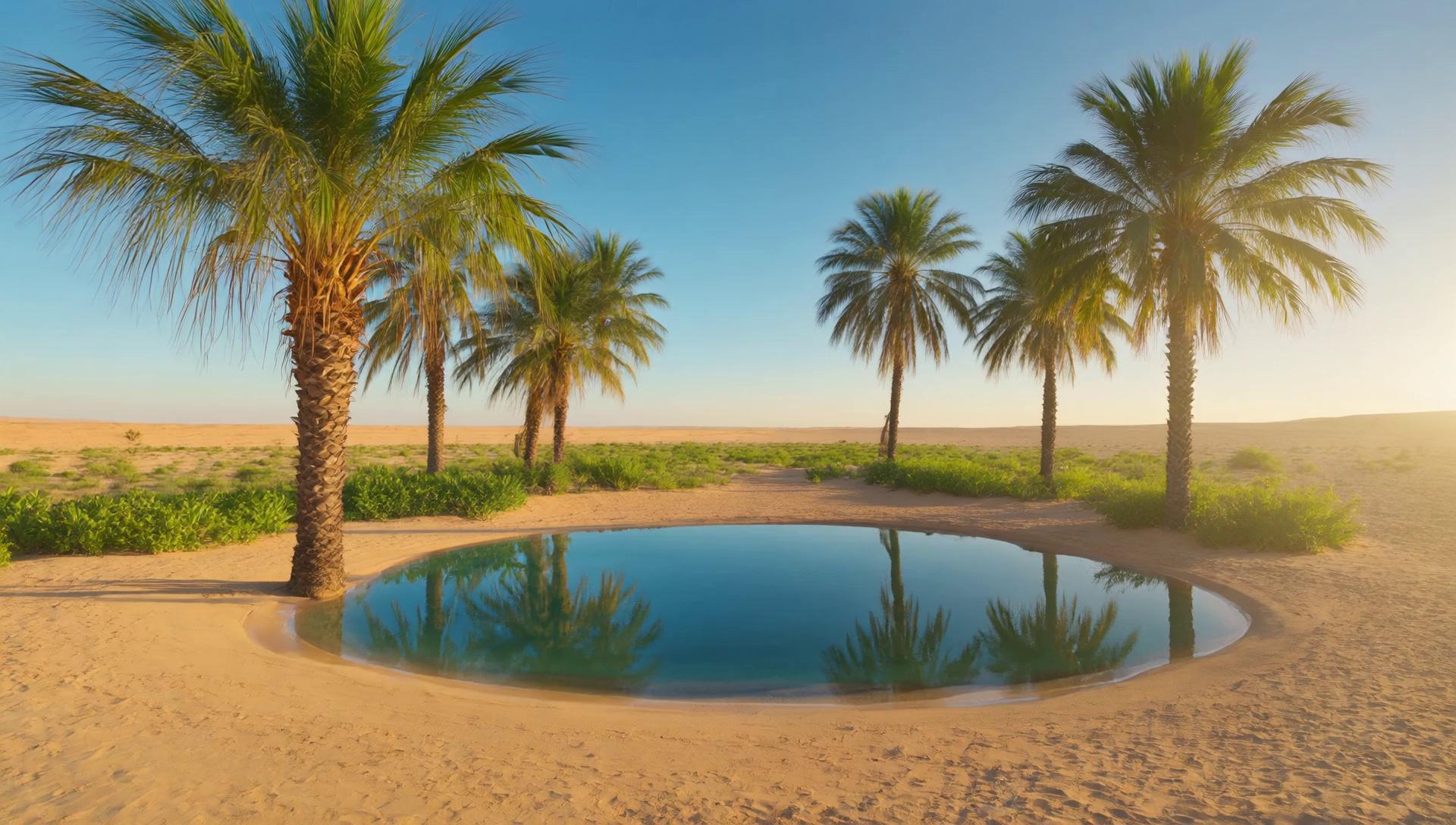} & 
  \includegraphics[width=\linewidth]{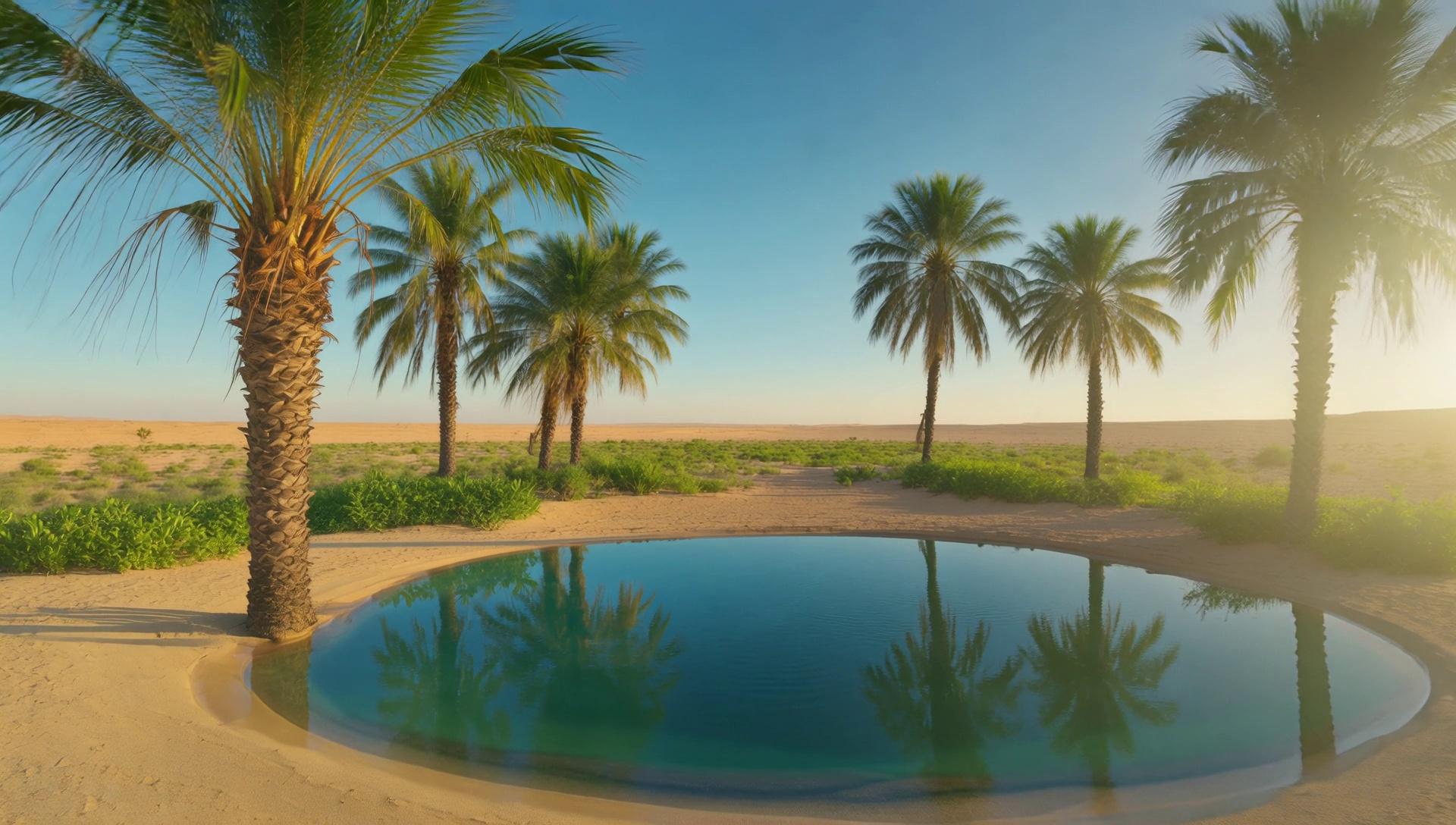} & 
  \includegraphics[width=\linewidth]{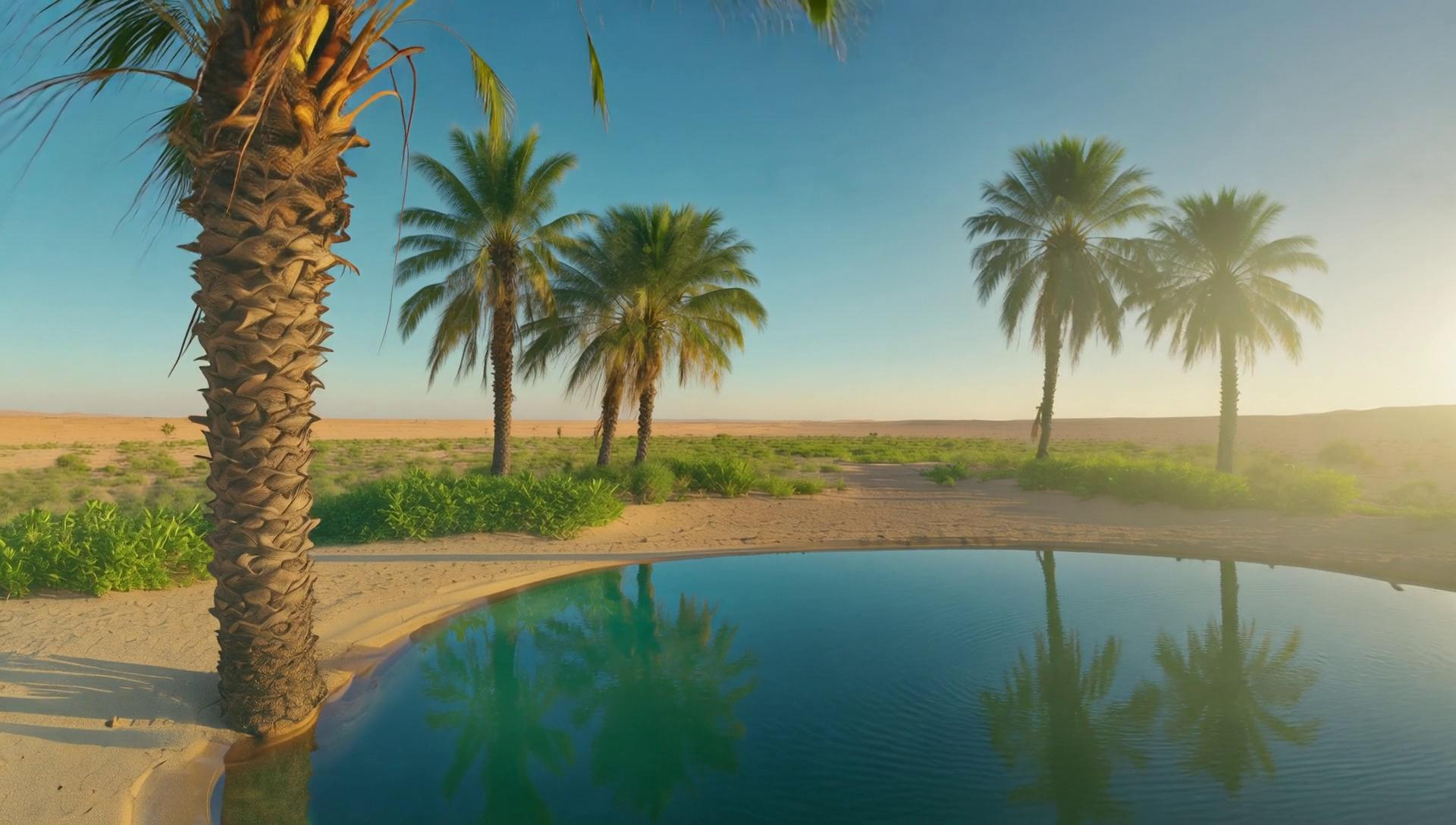} \\ 
  \textbf{HiStream}
  \textbf{(Ours)} & 
  \includegraphics[width=\linewidth]{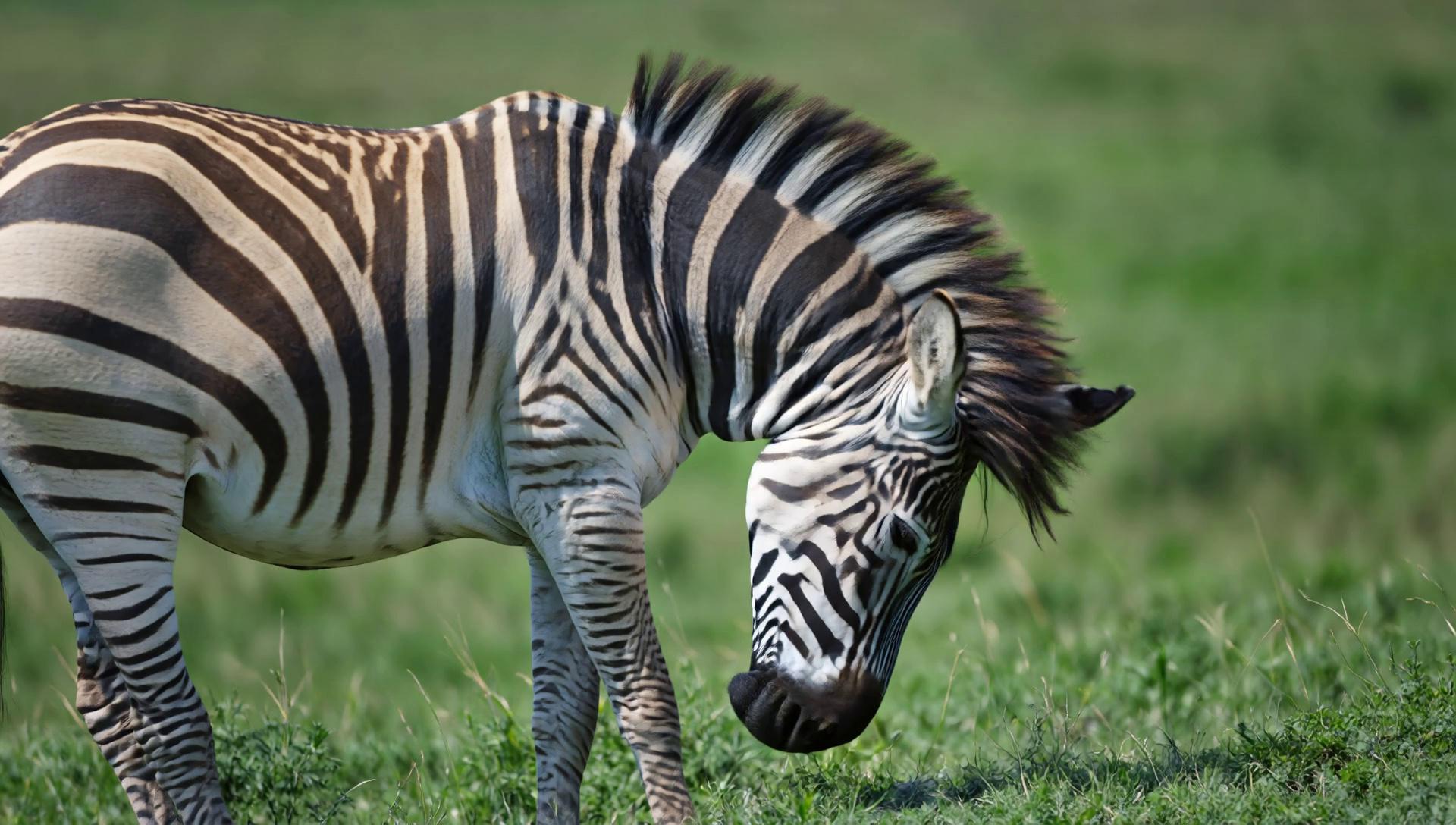} & 
  \includegraphics[width=\linewidth]{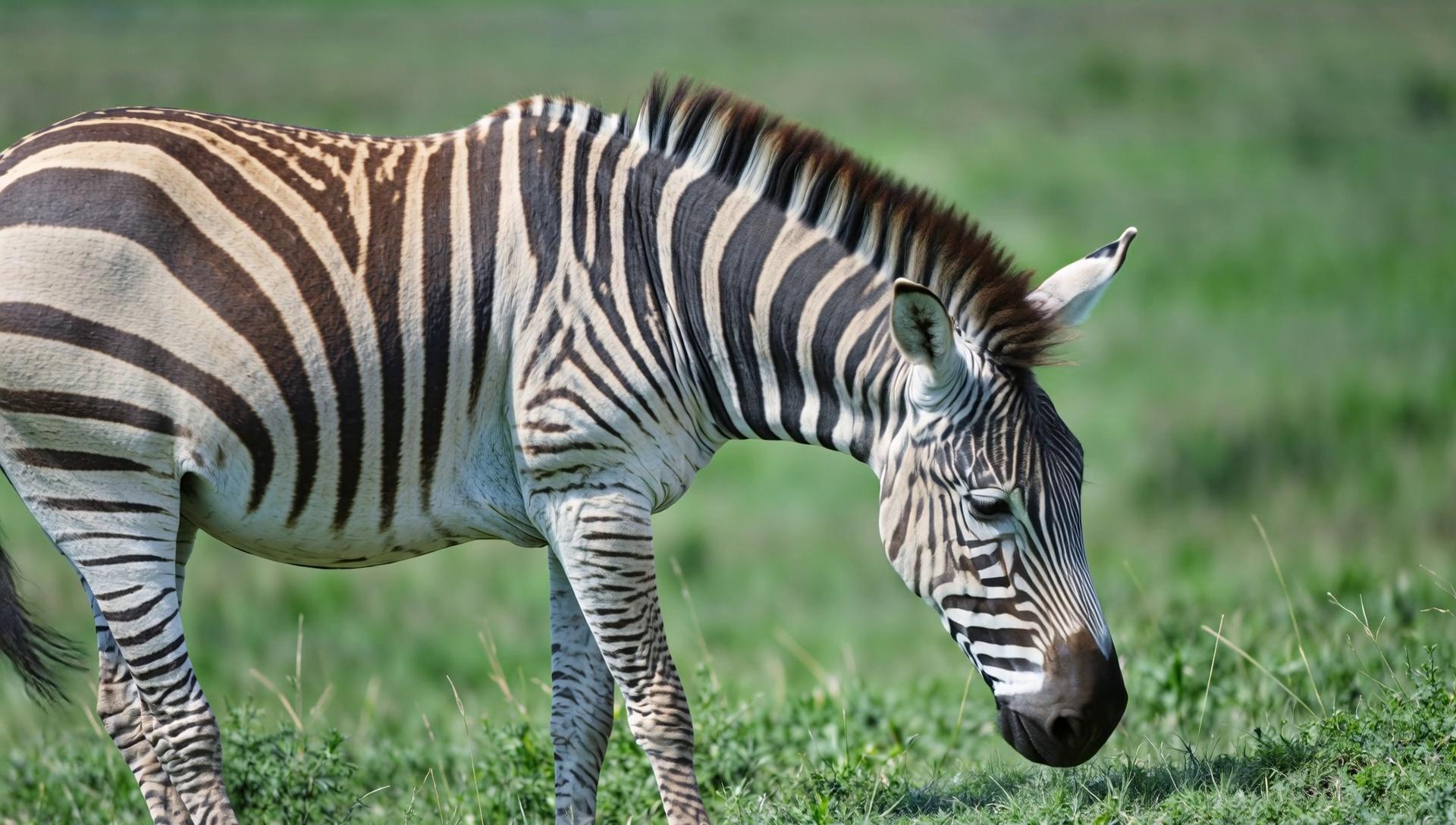} & 
  \includegraphics[width=\linewidth]{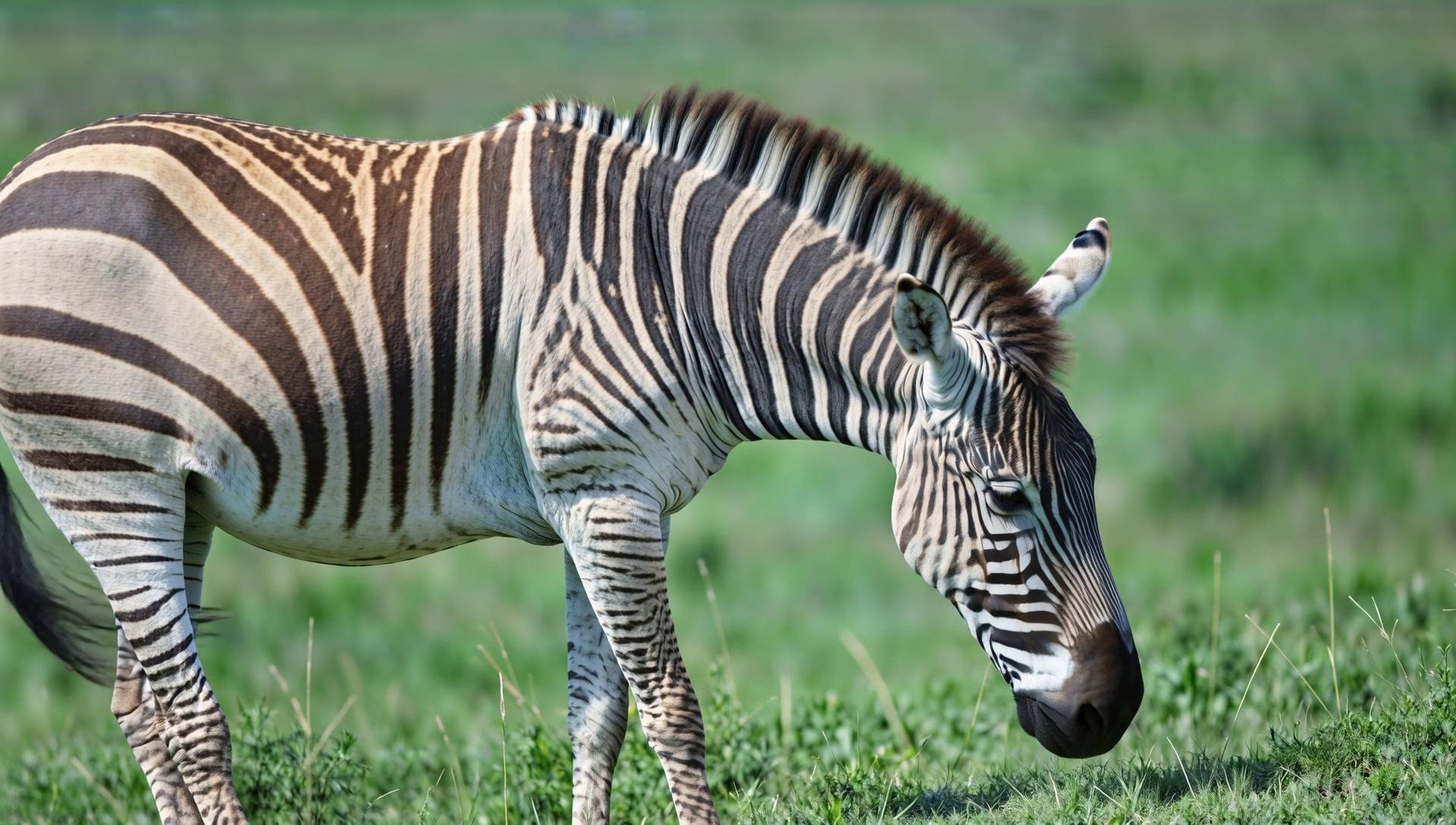} & 
  \includegraphics[width=\linewidth]{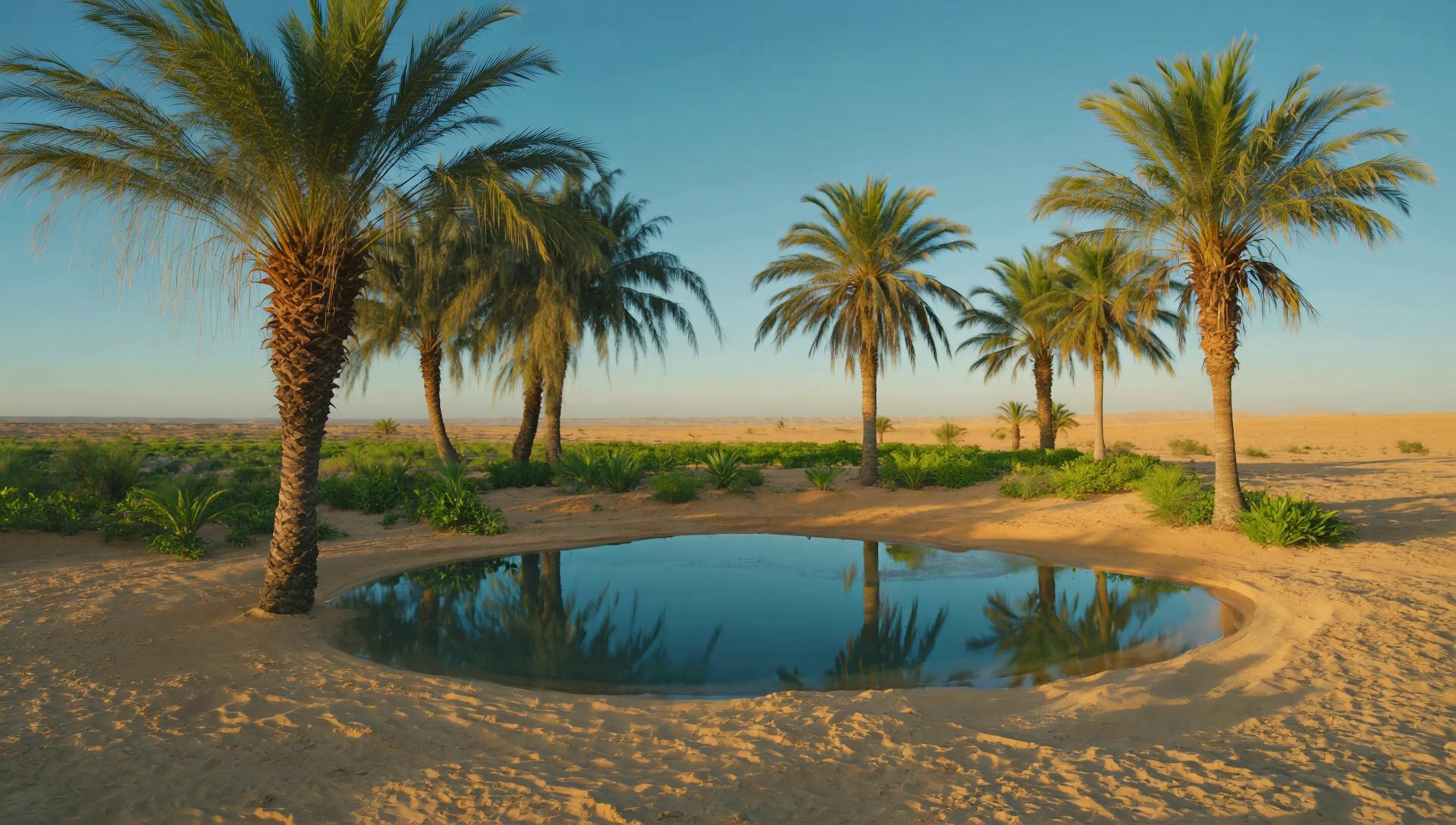} & 
  \includegraphics[width=\linewidth]{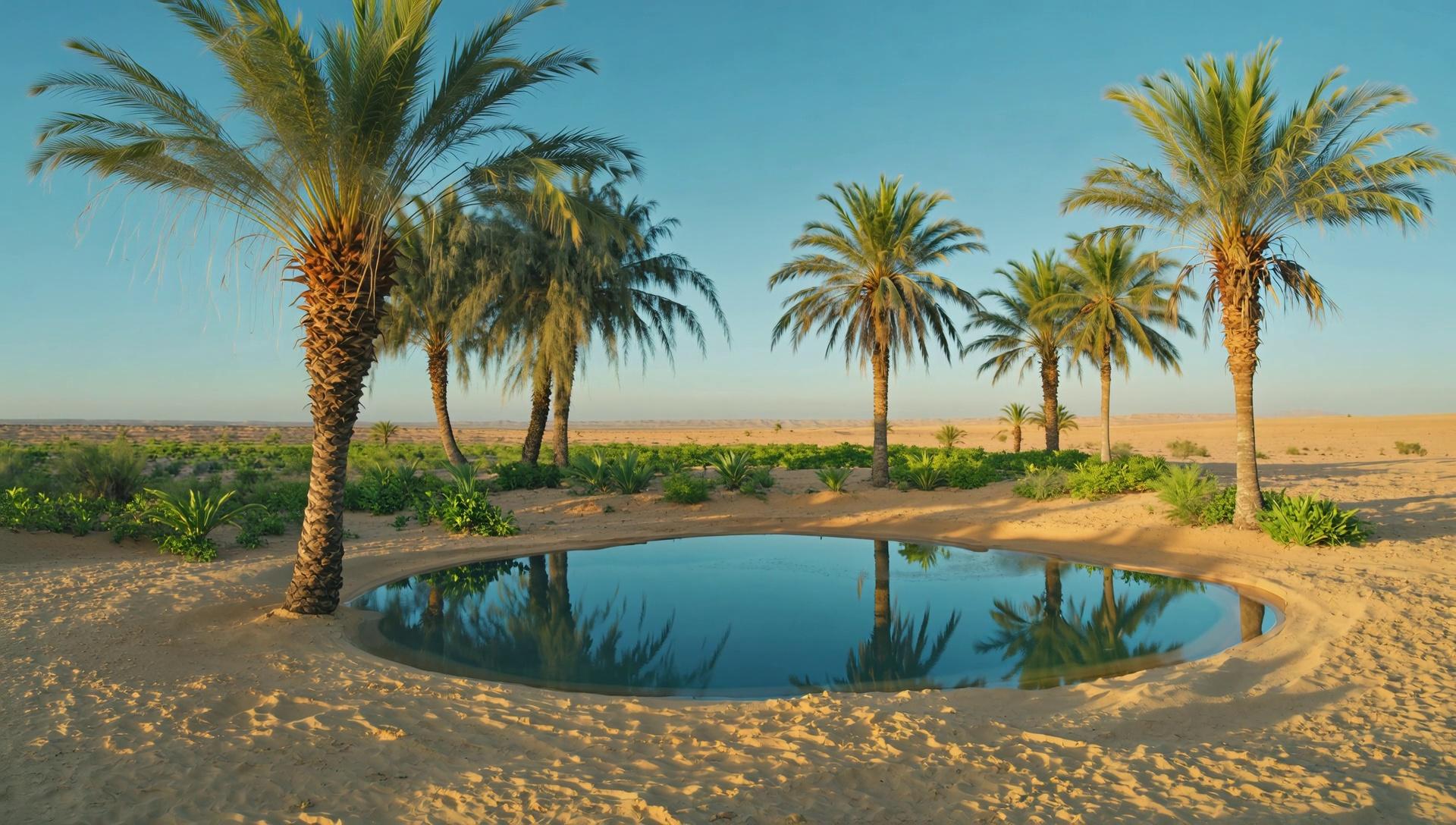} & 
  \includegraphics[width=\linewidth]{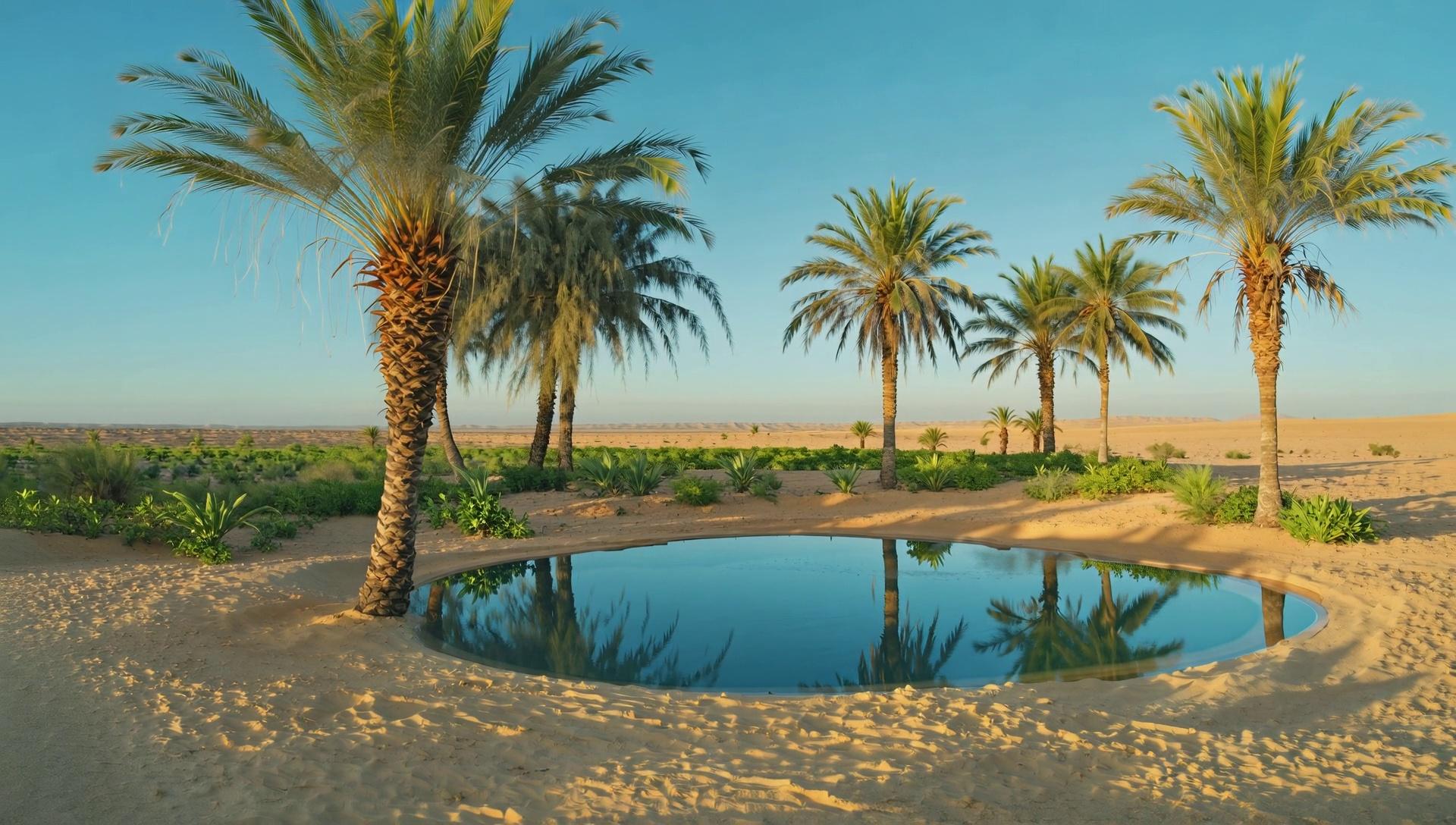} \\
\end{tabular}
\vspace{-0.5em}
\caption{\textbf{Qualitative ablations.} We perform controlled comparisons of HiStream with alternative variants. Best viewed \textbf{ZOOMED-IN}.}
\vspace{-1.0em}
\label{fig:abl}
\end{figure*}

\paragraph{Implementation details.} We design HiStream on top of the Wan2.1-T2V-1.3B model~\citep{wan2025}, whose original resolution is $832 \times 480$. Following the Self Forcing paradigm~\citep{huang2025self}, we leverage the pre-trained Wan2.1-T2V-14B model as a teacher to guide our student model's training. Notably, HiStream requires no additional real video during this process. All training prompts are obtained from VidProM~\citep{wang2024vidprom} as we follow Self Forcing~\citep{huang2025self}. We tune the model at $960 \times 544$ resolution (memory limitation), while generating content at $1920 \times 1088$ during inference. HiStream uses $4$-step diffusion and implements chunk-wise autoregressive variants, generating chunks of $M=3$ latent frames.

\paragraph{Evaluation metrics.} We perform extensive evaluations using VBench~\citep{huang2024vbench} and a user preference study, covering both visual fidelity and semantic alignment. We use augmented prompts from VBench as all baselines require. We also rigorously examine the runtime efficiency of our approach to validate its applicability.

\subsection{Comparison with Existing Baselines}

% \noindent\textbf{Comparison with Existing Baselines.}
%
We compare our model with relevant open-source efficient/high-resolution video generation models of similar scale. Our comparisons include three diffusion models: Wan2.1-1.3B~\citep{wan2025} (our base model structure), Self Forcing~\citep{huang2025self} (our initialization weights), LTX~\citep{HaCohen2024LTXVideo} (known for efficiency), and FlashVideo~\citep{zhang2025flashvideo} (known for high-resolution generation).
Table~\ref{tab:baseline} exhibits that our method achieves the best Quality Score and Total Score on VBench~\citep{huang2024vbench} in the high-resolution generation. While scores in high-resolution generation are generally lower than those at low resolutions, our method nonetheless achieves a score that is highly competitive, approaching the best result recorded in the low-resolution setting. 
In terms of Per-Frame Denoising Latency, our proposed HiStream is the most efficient, achieving a minimal latency of only $\mathbf{0.48\,s}$. This performance represents an impressive $\mathbf{76.2\times}$ acceleration compared to the baseline Wan2.1-T2V-1.3B (36.56\,s). Crucially, HiStream also significantly outperforms the existing efficient method Self Forcing (1.18\,s), demonstrating a near $\mathbf{2.5\times}$ speedup.
As shown in Figure~\ref{fig:baseline}, the videos generated by HiStream exhibit the highest visual fidelity and the cleanest texture, free from spurious patterns or visible artifacts.

\subsection{Ablation Studies}

We perform controlled comparisons of HiStream with alternative variants. We evaluate: (1) NTK-Rope + attention scaling (HD Tech), (2) Dual-Resolution Caching (DRC), (3) Anchor-Guided Sliding Window (AGSW), (4) Tuning, and (5) Asymmetric Denoising (HiStream+).  
As illustrated by the quantitative results in Table~\ref{tab:abl} and the qualitative analysis in Figure~\ref{fig:abl} (and Appendix), we draw the following conclusions regarding the contribution of each component:

\noindent\textbf{Foundation of HD Inference (HD Tech):} The implementation of NTK-RoPE and attention scaling is the fundamental prerequisite for stable high-resolution inference. Experiments show that \emph{w/o HD Tech}, the model either fails to generate coherent video content at the target resolution or exhibits severe stability issues, confirming that these techniques form the necessary quality basis.

\noindent\textbf{Dual-Resolution Caching (DRC):} The DRC mechanism provides a dual benefit. Ablating this technique (\emph{w/o DRC}) results in a latency of 0.70\,s. Incorporating DRC not only improves efficiency significantly (0.70\,s $\rightarrow$ 0.48\,s), but also leads to superior compositional quality. This is attributed to the overall structural planning of the generated video being supervised and executed at the trained low-resolution, resulting in a more coherent spatial layout.

\noindent\textbf{Anchor-Guided Sliding Window (AGSW):} The AGSW primarily serves as an efficiency booster. Removing the AGSW mechanism (\emph{w/o AGSW}) increases the latency to 0.78\,s. Crucially, AGSW accelerates the process to 0.48\,s with only negligible impact on the final video quality. This confirms that AGSW successfully exploits temporal redundancy without compromising video coherence.

\noindent\textbf{Necessity of Comprehensive Fine-Tuning:} The introduction of the full HiStream framework (encompassing HD Tech, DRC, and AGSW) creates a significant gap between the model's original training process and the modified inference paradigm. Our comprehensive fine-tuning is designed precisely to align the model's weights with this new protocol. As demonstrated in Figure~\ref{fig:abl}, the absence of this tuning (\emph{w/o Tuning}) results in a substantial degradation of video quality, confirming that bridging this training-inference gap is critical for achieving optimal high-fidelity synthesis.

\noindent\textbf{Ablation on Denoising Steps (HiStream+).}
We further investigate the impact of the denoising step count, comparing our 4-step main setting (0.48\,s) with two accelerated 2-step variants. 
Interestingly, the quantitative results in Table~\ref{tab:abl} show that a \emph{uniform 2-step} process (applying 2 steps to all chunks) and our \textbf{HiStream+} variant (4 steps for the first chunk, 2 for the rest) achieve nearly identical \texttt{vbench} scores. However, the qualitative results in Appendix reveal a critical difference. Compared to our main HiStream setting, the \emph{uniform 2-step} model suffers from severe blur and artifacts in the initial chunk. These initial errors then propagate, degrading the quality of all subsequent chunks. In contrast, HiStream+ avoids this issue by dedicating 4 steps to the initial chunk, establishing a robust, high-quality cache. As a result, HiStream+ exhibits far less visual degradation. This confirms that our asymmetric strategy is the superior approach, achieving a significant latency reduction (0.48\,s $\rightarrow$ 0.34\,s) with only a minimal sacrifice in visual fidelity. When benchmarked on a single H100 GPU, the per-frame denoising latency further drops to 0.21\,s (4.8\,FPS). This result suggests that real-time 1080p video generation is now within reach, requiring only further hardware and engineering optimizations.

In summary, our ablations validate the synergistic design of HiStream: \textbf{HD Tech} first establishes the stable high-resolution foundation. Building on this, \textbf{DRC} enhances both compositional quality and efficiency, while \textbf{AGSW} provides significant acceleration. Critically, our comprehensive \textbf{Tuning} then aligns the model to this modified inference protocol, bridging the training-inference gap to ensure all components work cohesively. We also demonstrate that this framework can be pushed to even higher speeds with HiStream+, where an \textbf{Asymmetric Denoising} strategy achieves a latency of 0.34\,s by intelligently managing the quality-efficiency trade-off, vastly surpassing a naive 2-step approach in visual quality.

\begin{table}[t!]
\centering
\small
\caption{\textbf{Quantitative ablations.} HiStream achieves the optimal overall video quality, while HiStream+ offers a compelling trade-off between video fidelity and generation efficiency.}
\vspace{-2mm}
\label{tab:abl}
\scalebox{0.92}{\begin{tabular}{lcccc}
\toprule
Method  & \makecell{Quality \\ Score $\uparrow$} & \makecell{Semantic \\ Score $\uparrow$} & \makecell{Total \\ Score $\uparrow$} & \makecell{Per-Frame \\ \textbf{Denoising} Latency (s) $\downarrow$} \\ 
\midrule
\rowcolor{lightgray}
\multicolumn{5}{l}{\textit{Stable Setting}} \\
w/o HD Tech &  84.58  &  72.99    &  82.26    &  0.48       \\ 
w/o Tuning &  84.63   &  79.98    &  83.70    &  0.48     \\ 
w/o DRC &   84.46   &  \textbf{81.23}    &  83.81    &  0.70     \\ 
w/o AGSW &  \textbf{85.15}   &  80.28    &  \underline{84.17}    &  0.78     \\ 
HiStream (Ours) &   \underline{85.00}    & \underline{80.97}     &  \textbf{84.20} & 0.48  \\ 
\midrule
\rowcolor{lightgray}
\multicolumn{5}{l}{\textit{Faster Setting}} \\
Naive Two Steps &   84.68    & 80.30     &  83.80 & \textbf{0.32}  \\ 
HiStream+ (Ours) &   84.69    & 80.25     &  83.80 & \underline{0.34}  \\ 
\bottomrule
\end{tabular}}
\vspace{-6mm}
\end{table}
\section{Conclusion}
\label{sec:conclusion}

In summary, HiStream makes high-resolution video generation practical and scalable by combining Dual-Resolution Caching with an Anchor-Guided Sliding Window. Our main framework drastically reduces redundant computation while preserving temporal consistency, achieving state-of-the-art visual quality with significantly improved inference efficiency. Furthermore, our HiStream+ variant leverages Asymmetric Denoising to push this efficiency to its limit (achieving a $107.5\times$ acceleration over the baseline), offering a powerful and flexible trade-off for ultra-fast generation. This work provides a robust foundation for accessible, high-fidelity cinematic content generation.

However, key limitations still remain. Although our framework optimized the denoising pipeline, the VAE decoder has emerged as the new primary bottleneck; decoding 81 frames at 1080p still requires $16.45\,\text{s}$ on an A100 and $9.40\,\text{s}$ on an H100. Furthermore, high memory costs during distillation constrained our experiments to a 1.3B student model, also not supervised on full 1080p data. This likely contributes to remaining flaws, such as a lack of physical realism and object interpenetration. Addressing these VAE and training-memory bottlenecks is the critical next step.

% \clearpage
% \newpage
\bibliographystyle{assets/plainnat}
\bibliography{main}

\clearpage
\newpage
\beginappendix

\setcounter{page}{1}

\noindent\textbf{Overview.} 
In the supplementary material, we introduce more implementation details and additional experimental results.

\section{More Implementation Details}

\subsection{Training details}

We utilize the DMD2~\citep{yin2024improved} distillation framework for our training pipeline. This setup employs three distinct models based on the Wan2.1~\citep{wan2025} architecture: a frozen 14B model serves as the real diffusion model (\textit{teacher}), an updatable 1.3B model acts as the final generator (\textit{student}), and a second updatable 1.3B model functions as the fake diffusion model to assist gradient calculation. The fake diffusion model is updated using the standard diffusion loss. Crucially, the final generator is updated by the distribution matching gradient, which is calculated from the divergence between the real and fake diffusion models. Further technical details regarding the training process are available in the original DMD2 paper.

To adapt the models for high-resolution generation, we set the timestep shift to 5 and employ NTK-RoPE with a scaling factor of $2\times$ throughout the training process. Training clips have a spatial resolution of $960 \times 544$ and a temporal length of 81 frames. After VAE compression, this translates to 21 latent frames, which are processed in 7 chunks of 3 latent frames each. The entire procedure was efficiently performed on 64 $\times$ H100 GPUs and achieved convergence in approximately 12 hours.

\subsection{Inference details}

Our inference pipeline utilizes only the final 1.3B generator model. The process is conducted at a high resolution of $1920 \times 1088$ (twice the spatial resolution used during training), maintaining 81 frames (21 latent frames) which are generated chunk-by-chunk in 7 total chunks (3 latent frames per chunk). 

To manage this significant resolution jump while maintaining quality, we apply the \textbf{HD Tech}: the timestep shift parameter is increased from 5 to 7, and the NTK-RoPE factor is increased from $2\times$ (used in training) to $5\times$. We also set the attention scaling factor to 2 exclusively for the initial chunk, allowing the model to focus on generating strong, detailed high-resolution content for the anchor frames. For all subsequent chunks, attention scaling is disabled (factor set to 1). This is highly effective because the high-resolution details generated in Chunk 1 are efficiently transferred and propagated to subsequent chunks, preventing the generation of redundant artifacts. For alignment with established baselines, all 1080p benchmark experiments reported in the main paper were uniformly conducted on a single A100 GPU.

\section{Additional Experimental Results}

\subsection{Comparison with Super-Resolution}

\begin{figure}[t!]
\centering
\setlength{\tabcolsep}{0.1em}  
\renewcommand{\arraystretch}{0.2}
 \begin{tabular}{C{0.18\linewidth} C{0.26\linewidth} C{0.26\linewidth} C{0.26\linewidth}}
  FlashVSR & 
  \includegraphics[width=\linewidth]{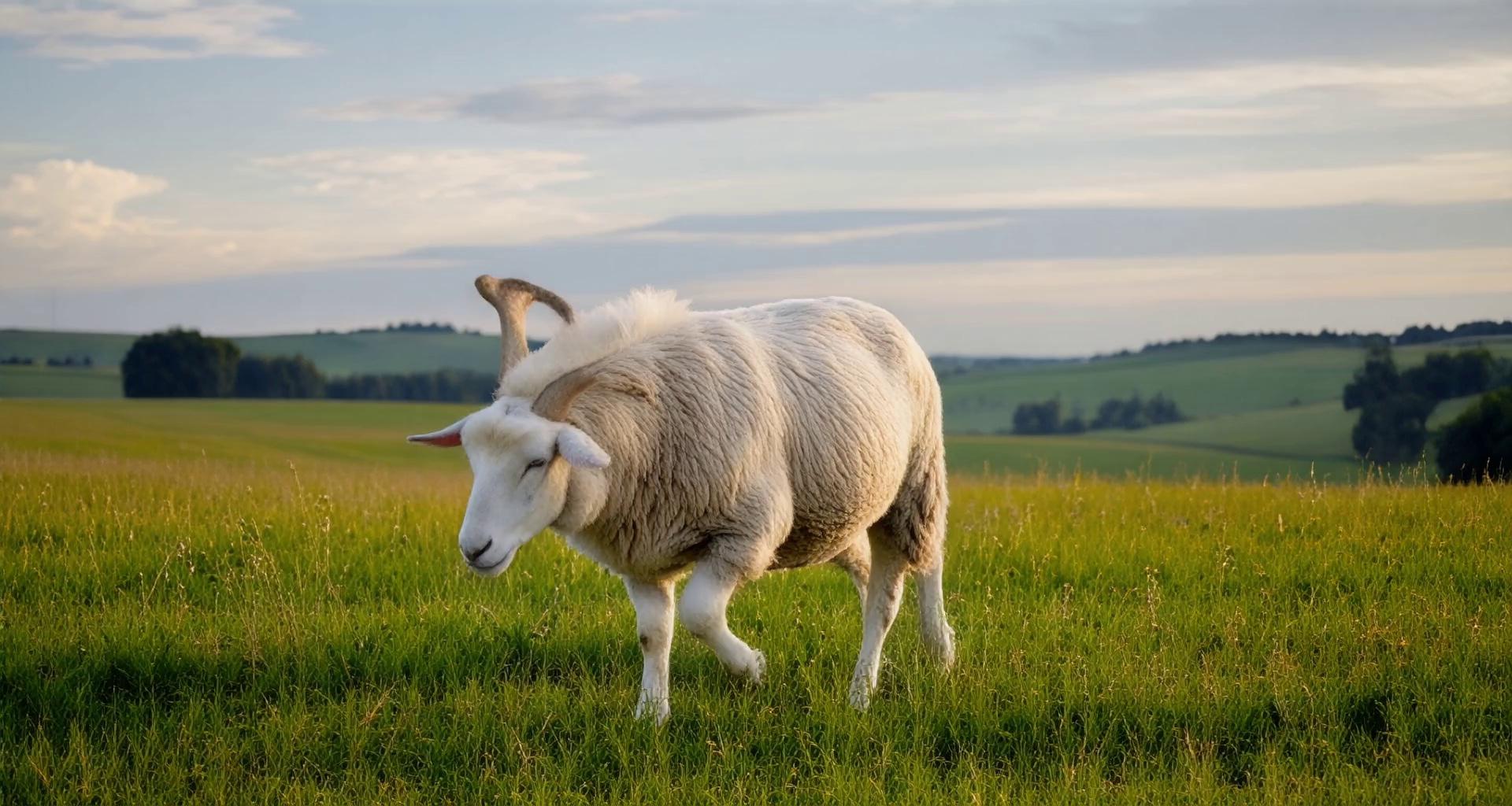} & 
  \includegraphics[width=\linewidth]{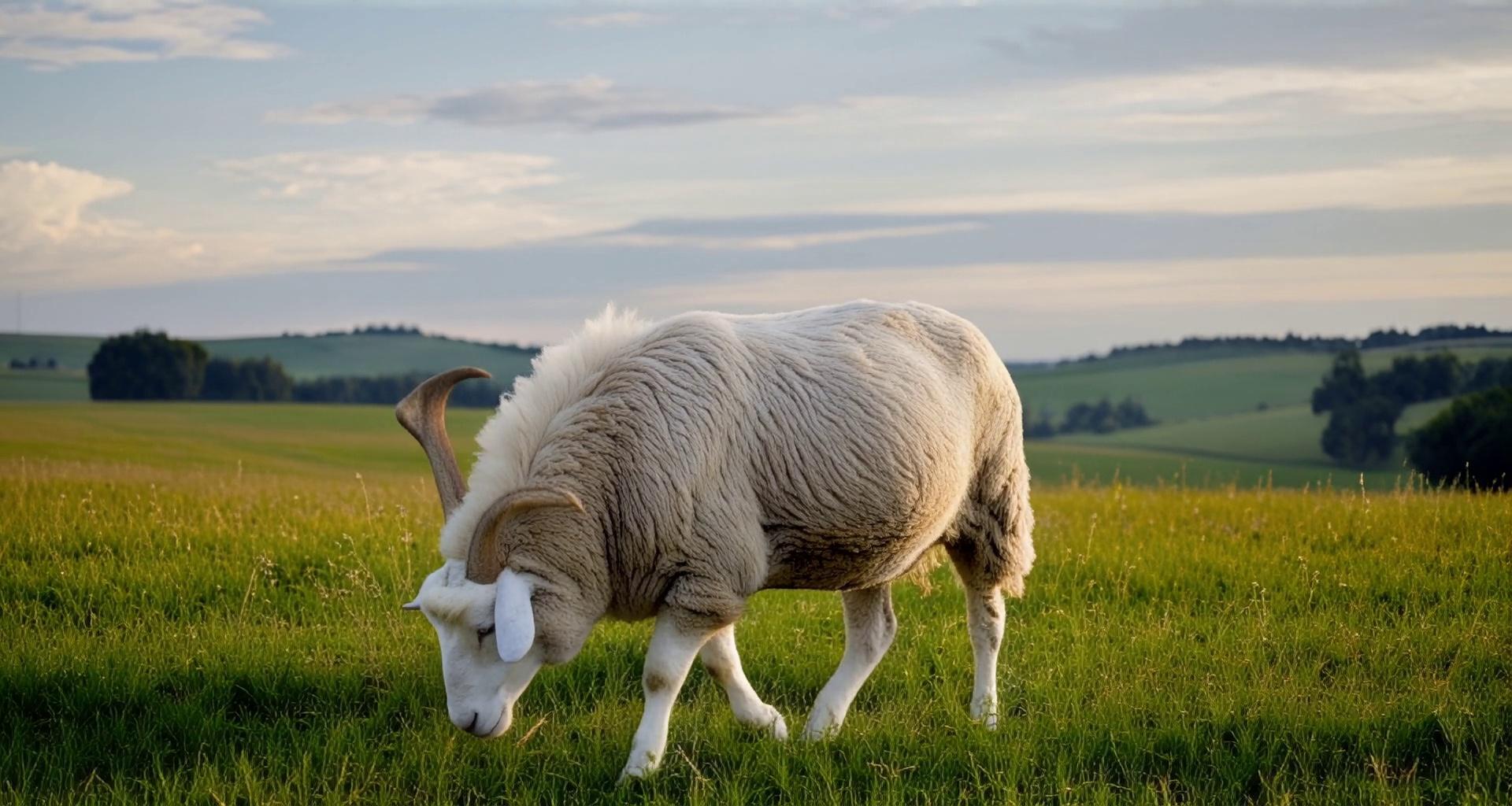} & 
  \includegraphics[width=\linewidth]{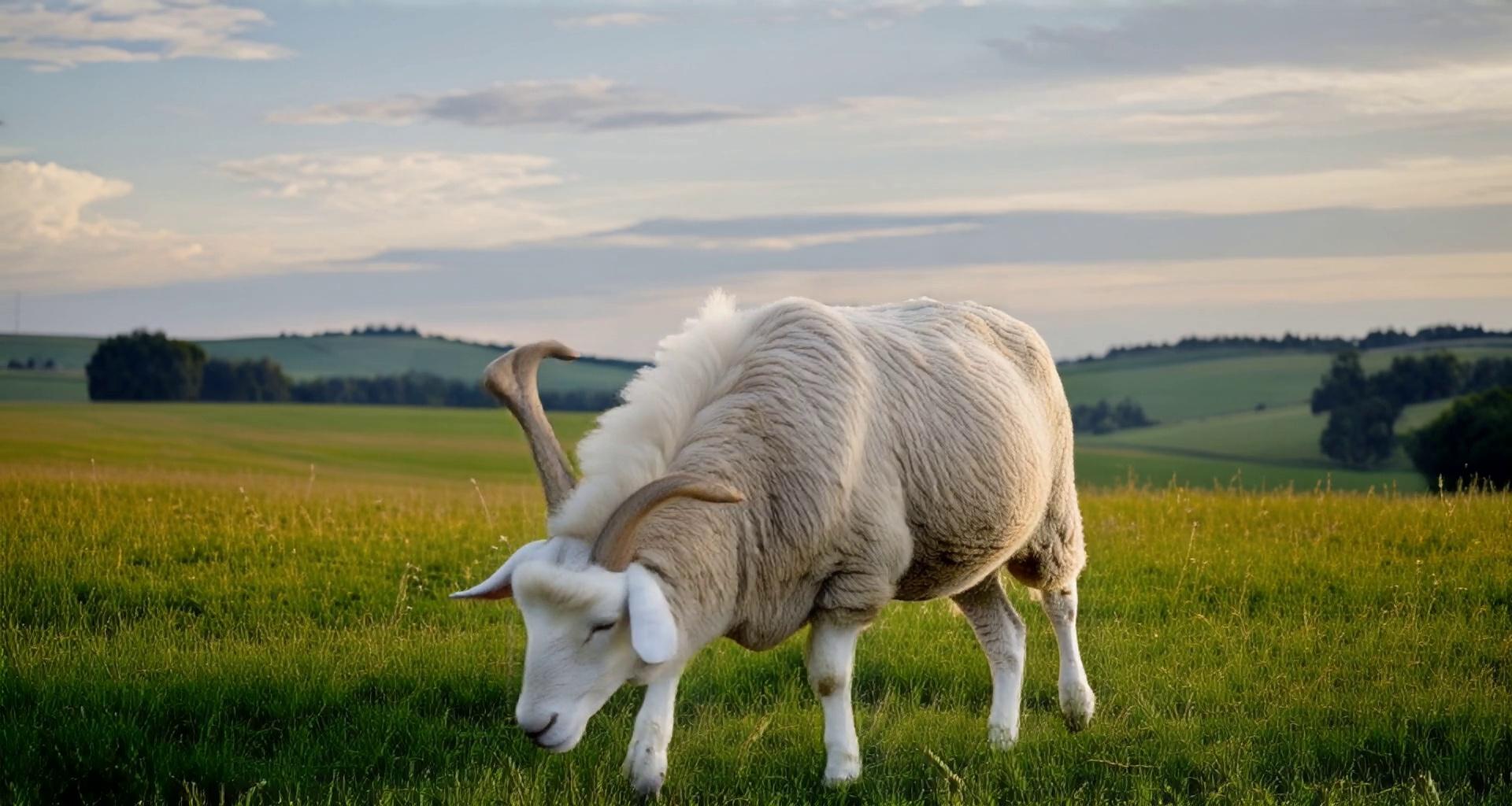}   \\ 
  \textbf{HiStream} 
  \textbf{(Ours)} & 
  \includegraphics[width=\linewidth]{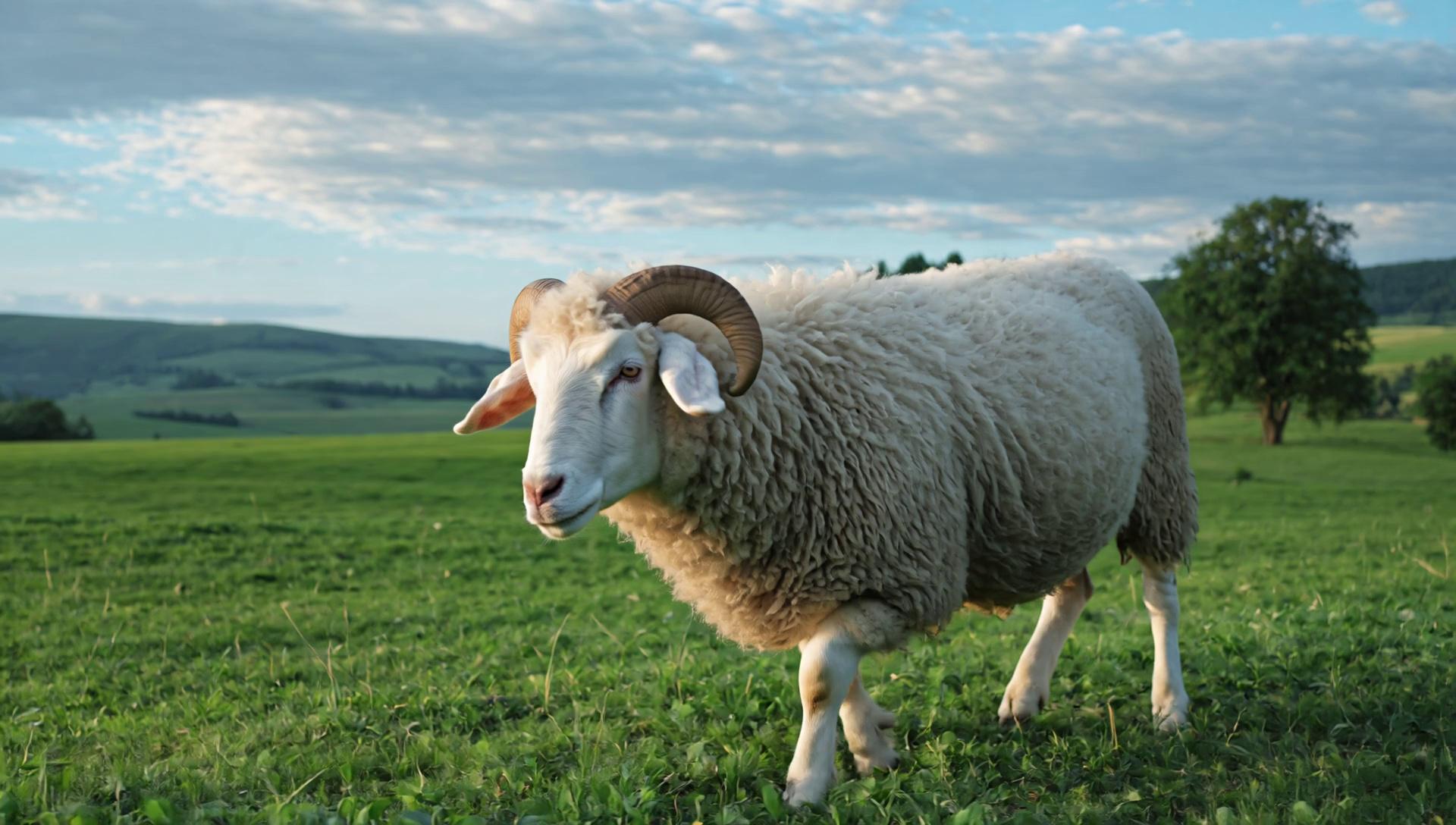} & 
  \includegraphics[width=\linewidth]{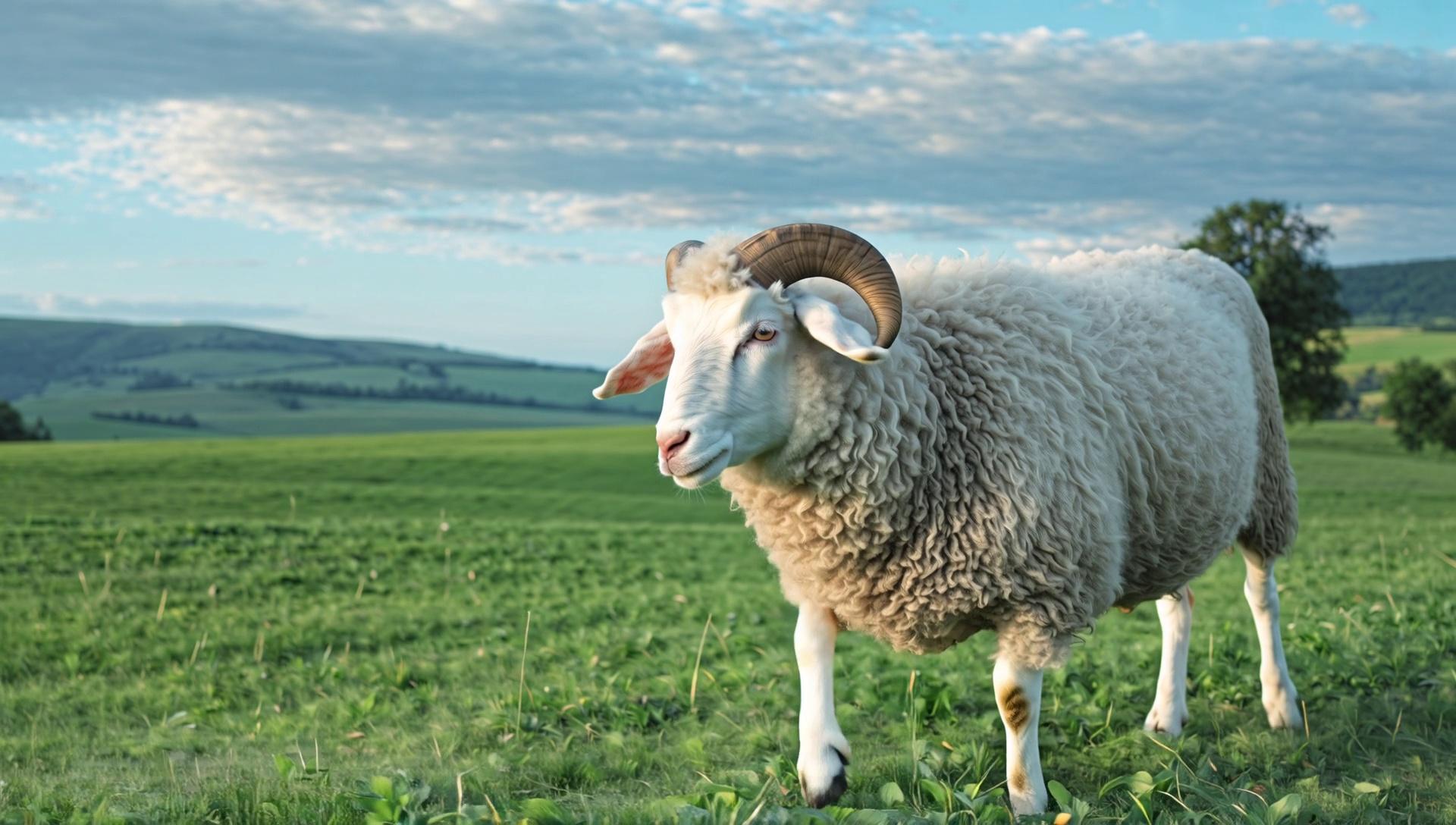} & 
  \includegraphics[width=\linewidth]{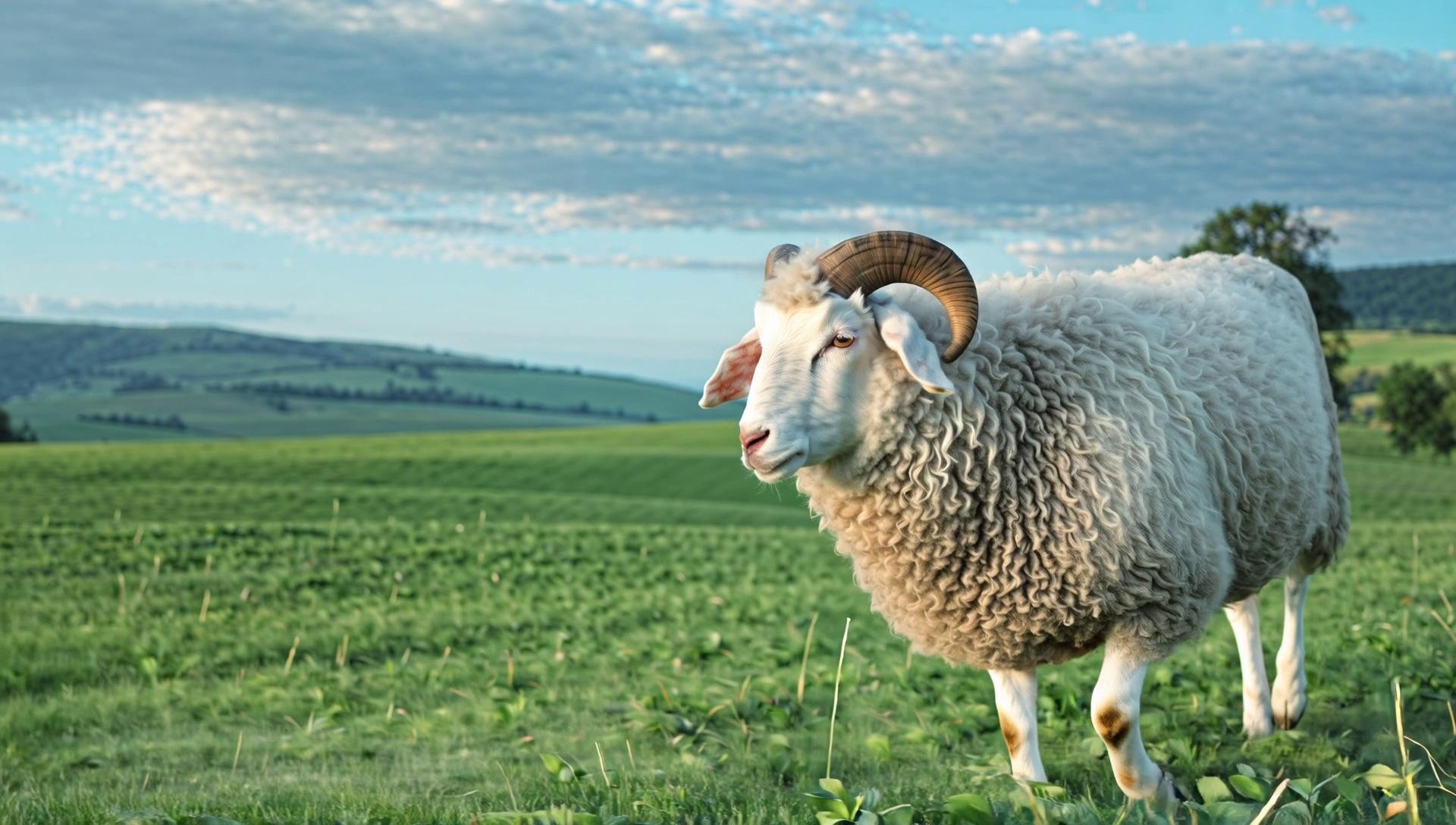}   \\
\end{tabular}
\caption{\textbf{Qualitative comparisons with super-resolution.} Our native high-resolution synthesis captures fine textures with greater accuracy than two-stage super-resolution pipelines, which often miss or hallucinate details. Best viewed \textbf{ZOOMED-IN}.}
\vspace{-1.0em}
\label{fig:sr}
\end{figure}

In contrast to conventional super-resolution (SR) methodologies, our higher-resolution generation approach is designed to exploit the latent capabilities of the pre-trained model. Therefore, the resultant performance is derived directly from the underlying base model, obviating the need for a separate, dedicated SR model. We evaluate our method against a standard super-resolution post-processing configuration: namely, applying FlashVSR~\citep{zhuang2025flashvsr} to the Self Forcing~\citep{huang2025self} output. As shown in Table~\ref{tab:sr}, HiStream achieves a better Quality Score and Total Score. In addition, Figure~\ref{fig:sr} shows that our native high-resolution synthesis provides better detail accuracy, capturing fine textures that are often missed or incorrectly generated by two-stage pipelines that depend on super-resolution.

\begin{table}[h!]
\centering
\small
\caption{\textbf{Quantitative comparisons with super-resolution.} Compared to super-resolution post-processing setting Self Forcing + FlashVSR, HiStream achieves a higher Quality Score and Total Score.}
% \vspace{-2mm}
\label{tab:sr}
\scalebox{0.91}{\begin{tabular}{lcccc}
\toprule
Method  & \makecell{Quality \\ Score $\uparrow$} & \makecell{Semantic \\ Score $\uparrow$} & \makecell{Total \\ Score $\uparrow$} \\ 
\midrule
Self Forcing~\citep{huang2025self} + FlashVSR~\citep{zhuang2025flashvsr} &  84.71   &  \textbf{81.04}    &  83.98      \\ 
HiStream (Ours) &   \textbf{85.00}    & \underline{80.97}     &  \textbf{84.20} \\
\bottomrule
\end{tabular}}
% \vspace{-2mm}
\end{table}

\begin{figure*}[t!]
\centering
\setlength{\tabcolsep}{0.1em}  
\renewcommand{\arraystretch}{0.2}
 \begin{tabular}{C{0.09\linewidth} C{0.145\linewidth} C{0.145\linewidth} C{0.145\linewidth} @{\hspace{0.5em}} C{0.145\linewidth} C{0.145\linewidth} C{0.145\linewidth}}
  Naive Two Steps & 
  \includegraphics[width=\linewidth]{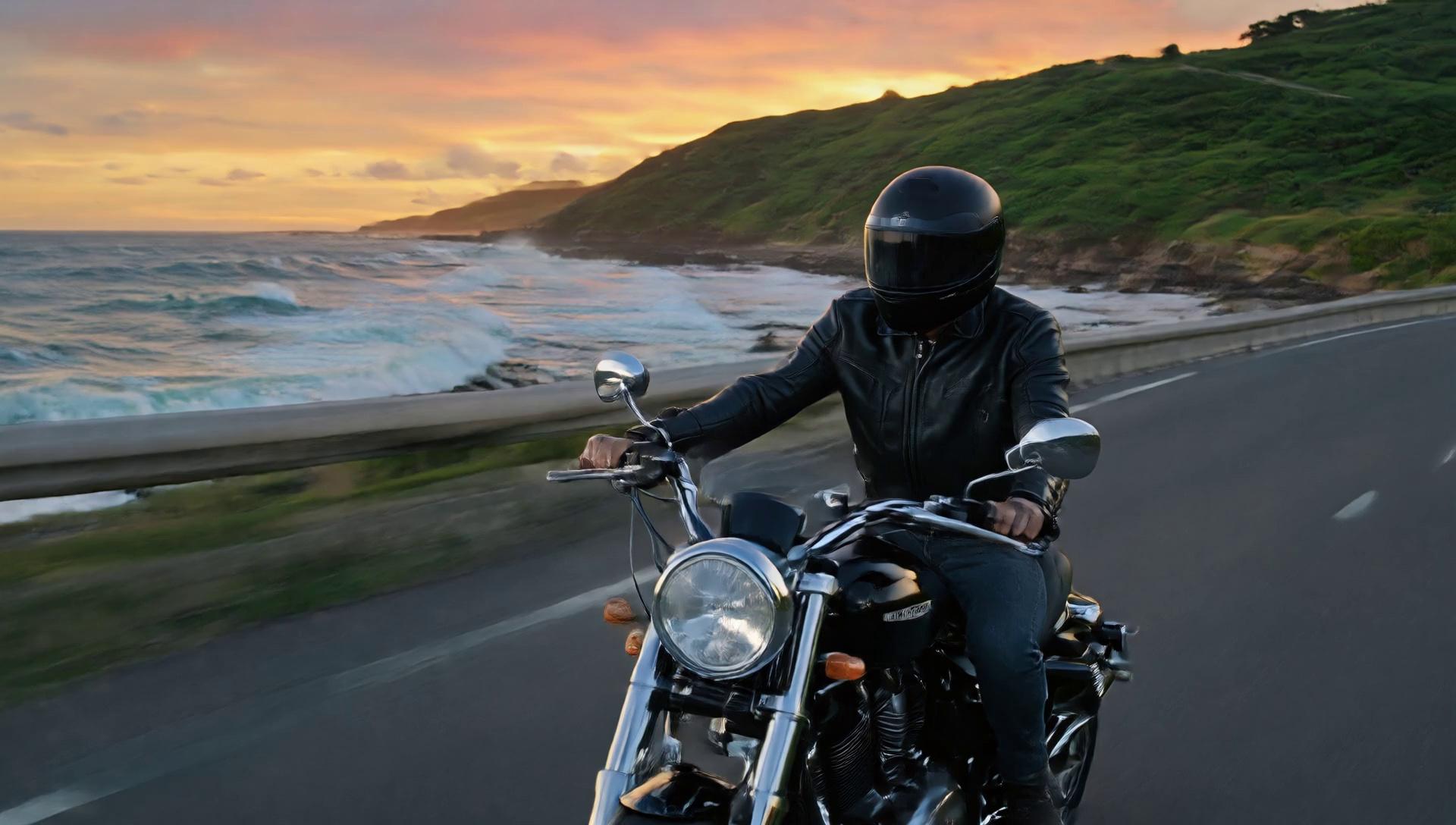} & 
  \includegraphics[width=\linewidth]{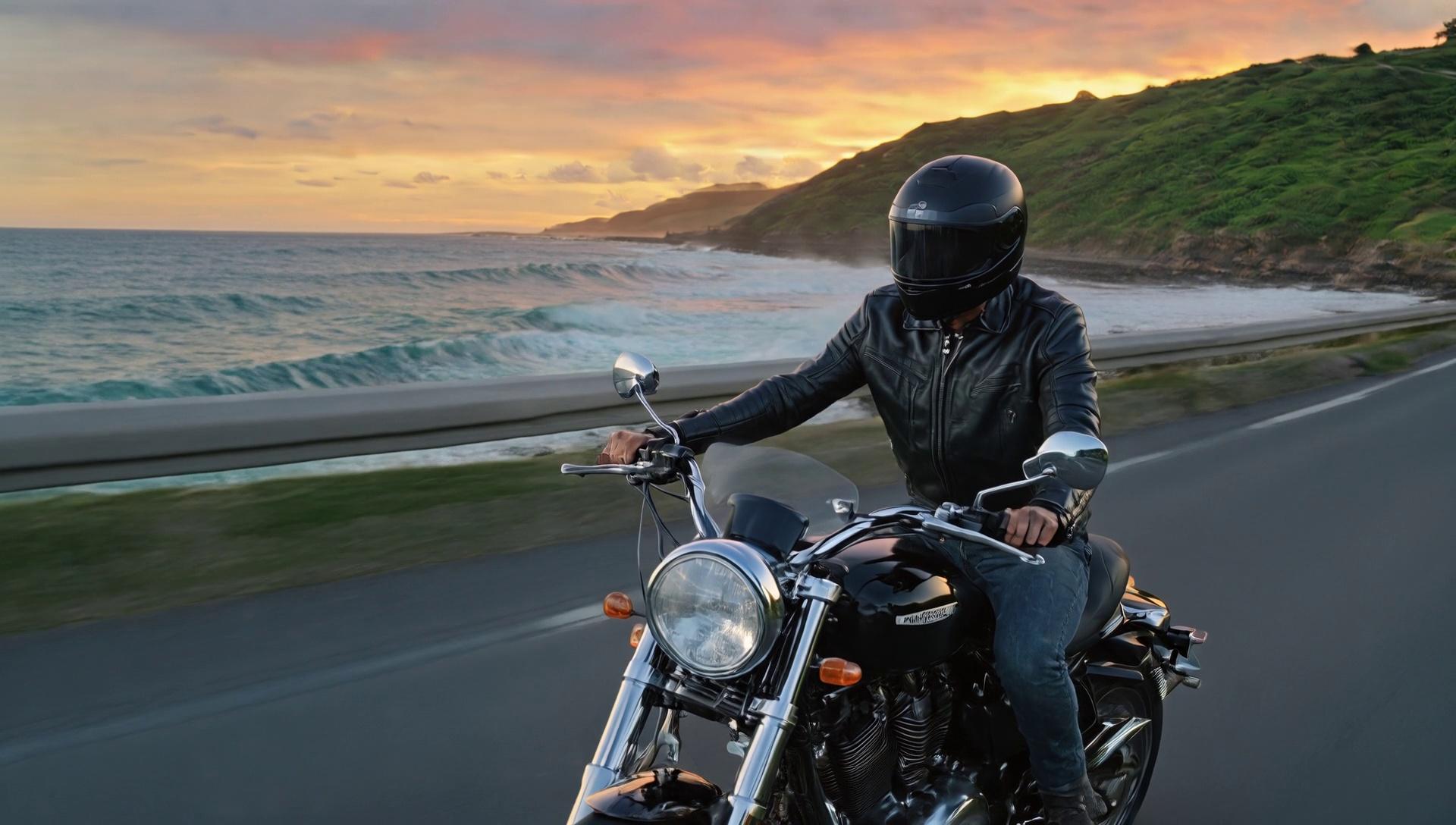} & 
  \includegraphics[width=\linewidth]{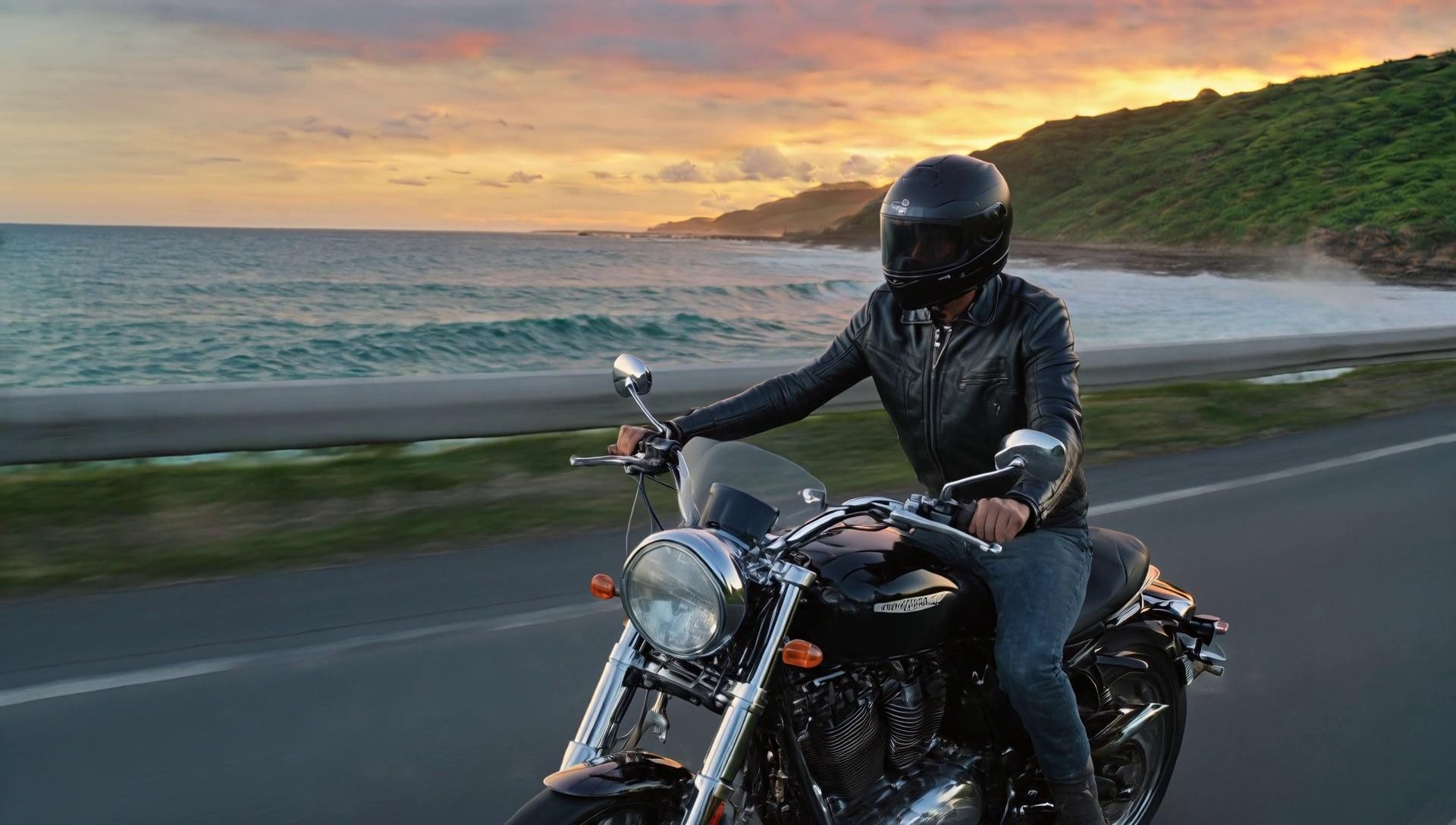} & 
  \includegraphics[width=\linewidth]{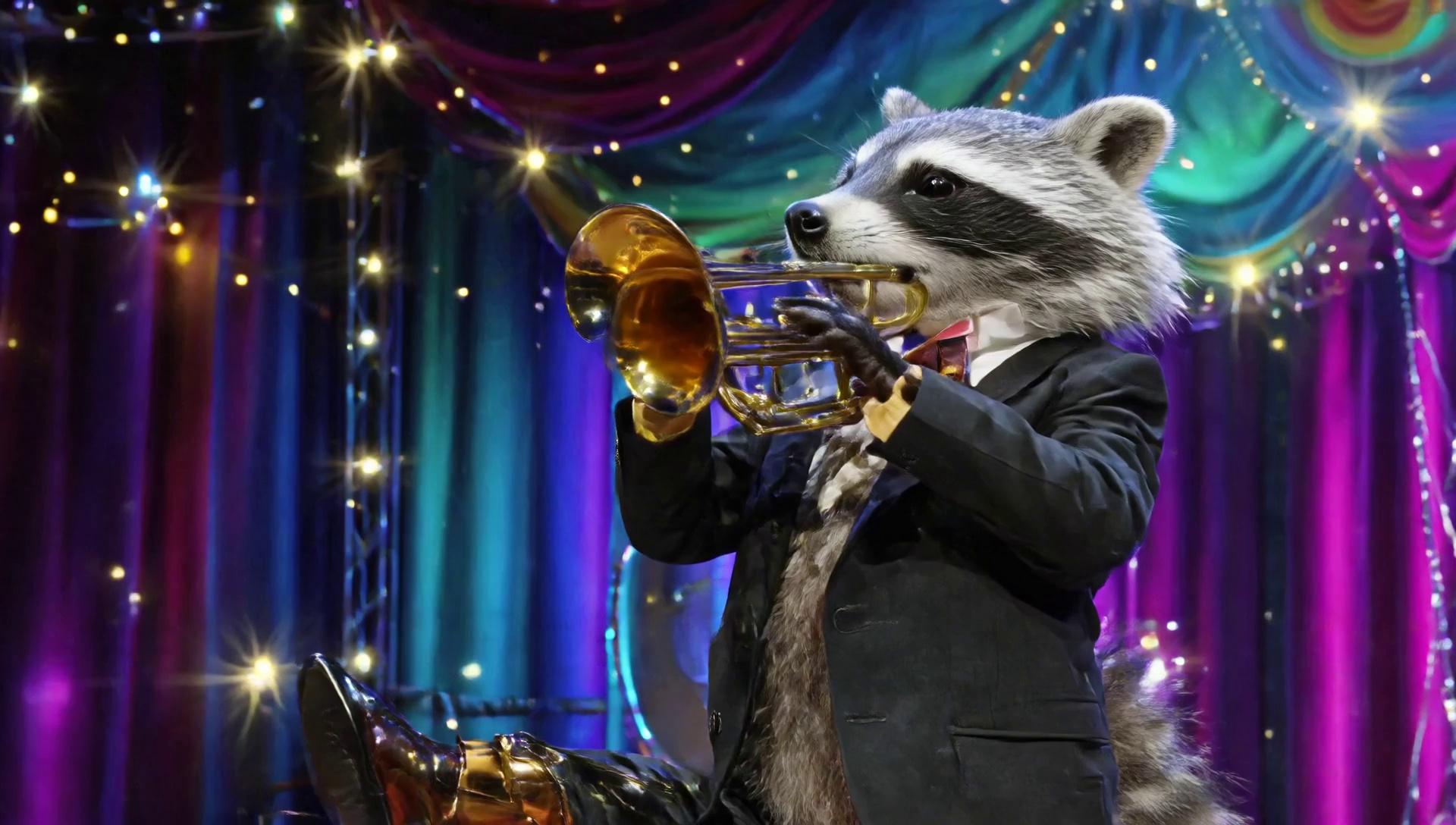} & 
  \includegraphics[width=\linewidth]{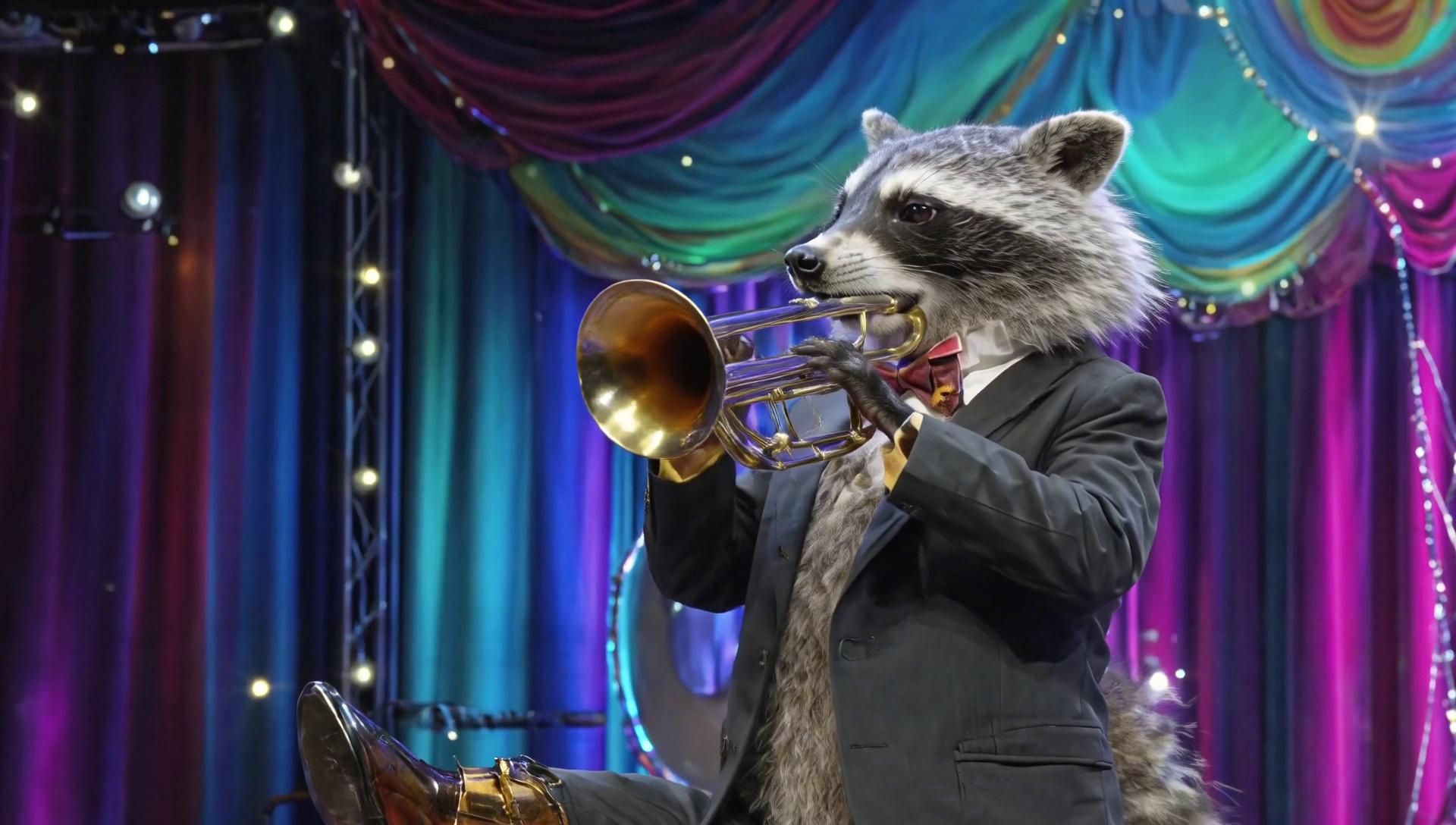} & 
  \includegraphics[width=\linewidth]{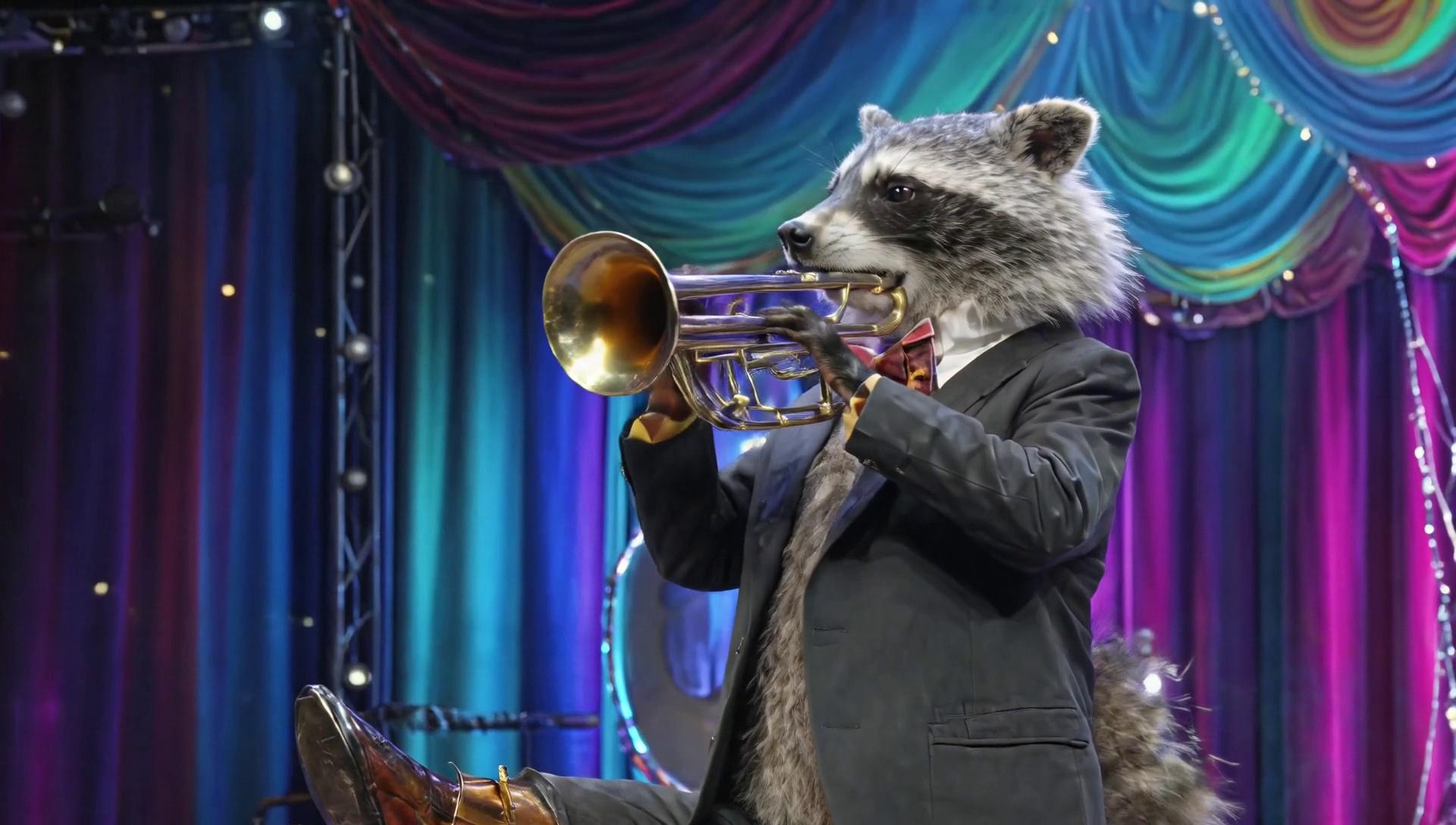} \\
  HiStream+ & 
  \includegraphics[width=\linewidth]{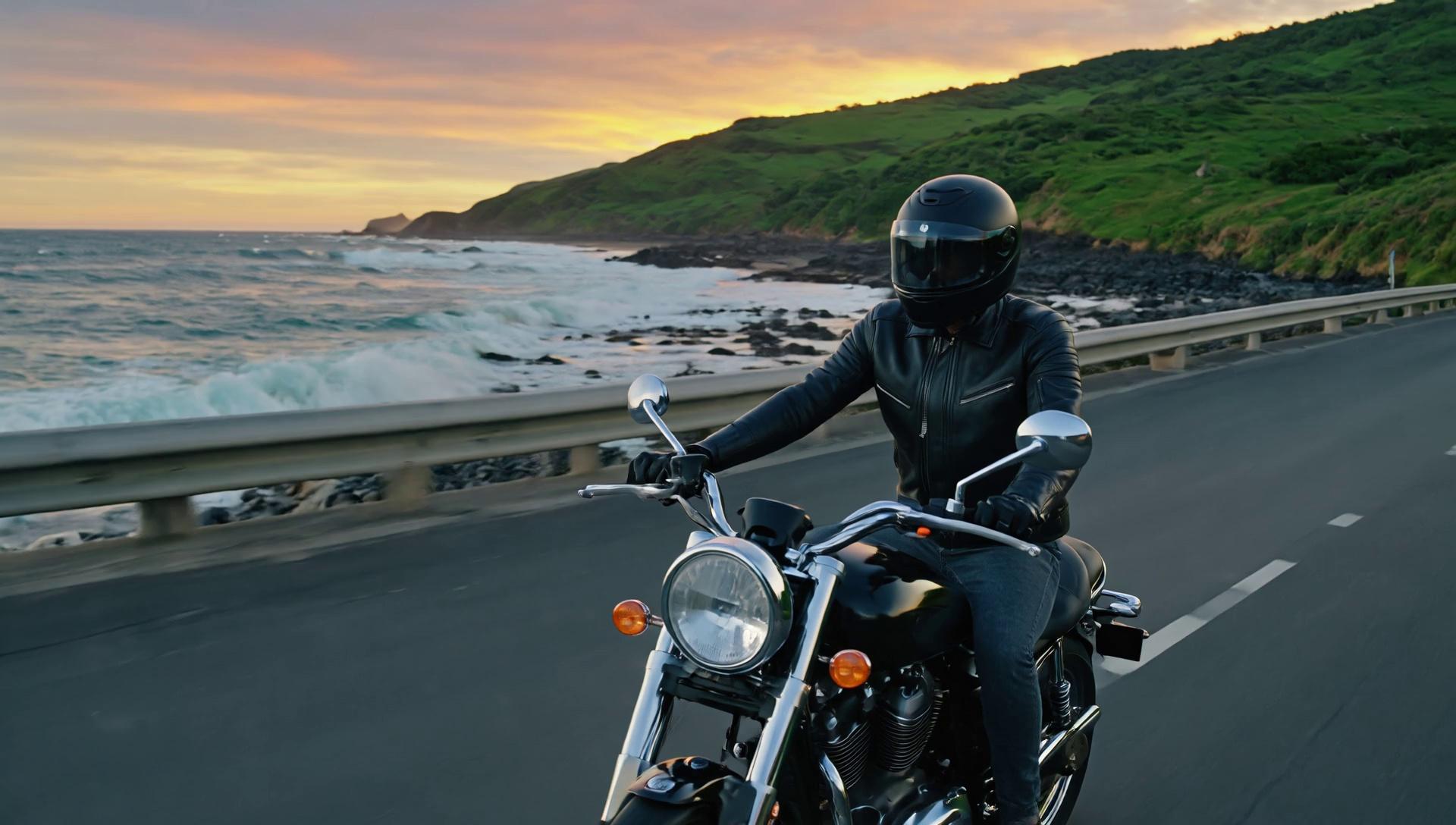} & 
  \includegraphics[width=\linewidth]{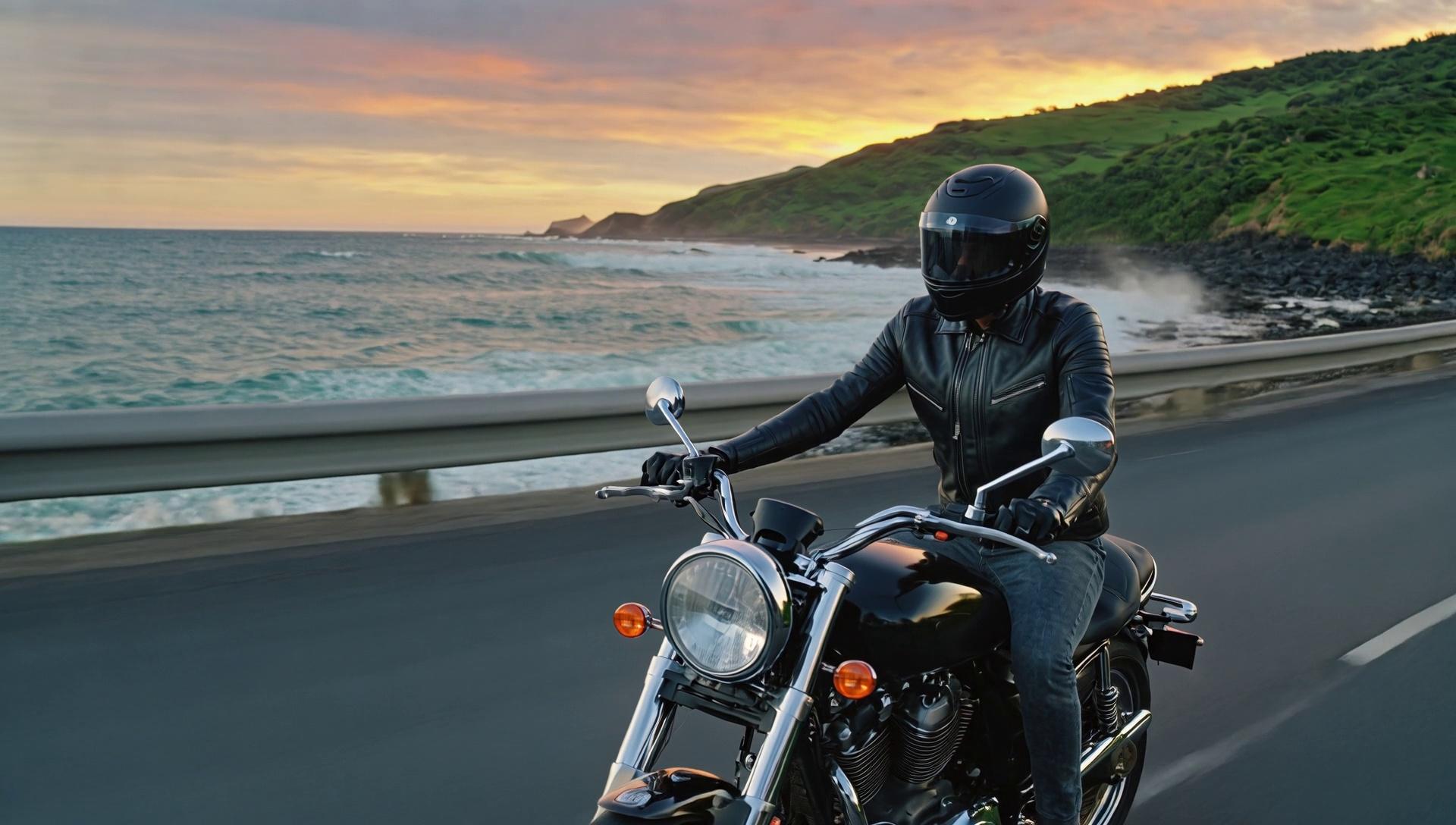} & 
  \includegraphics[width=\linewidth]{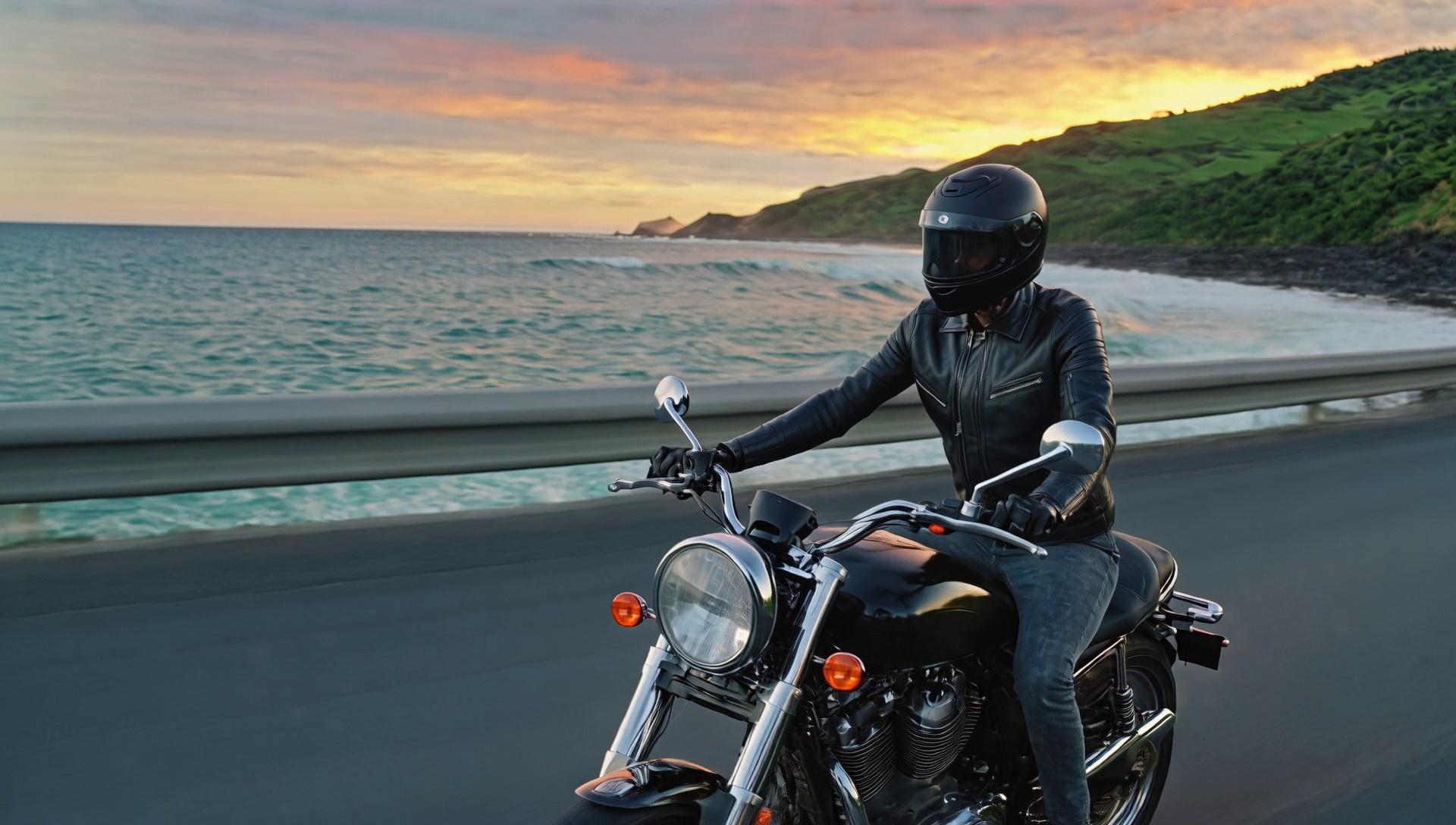} & 
  \includegraphics[width=\linewidth]{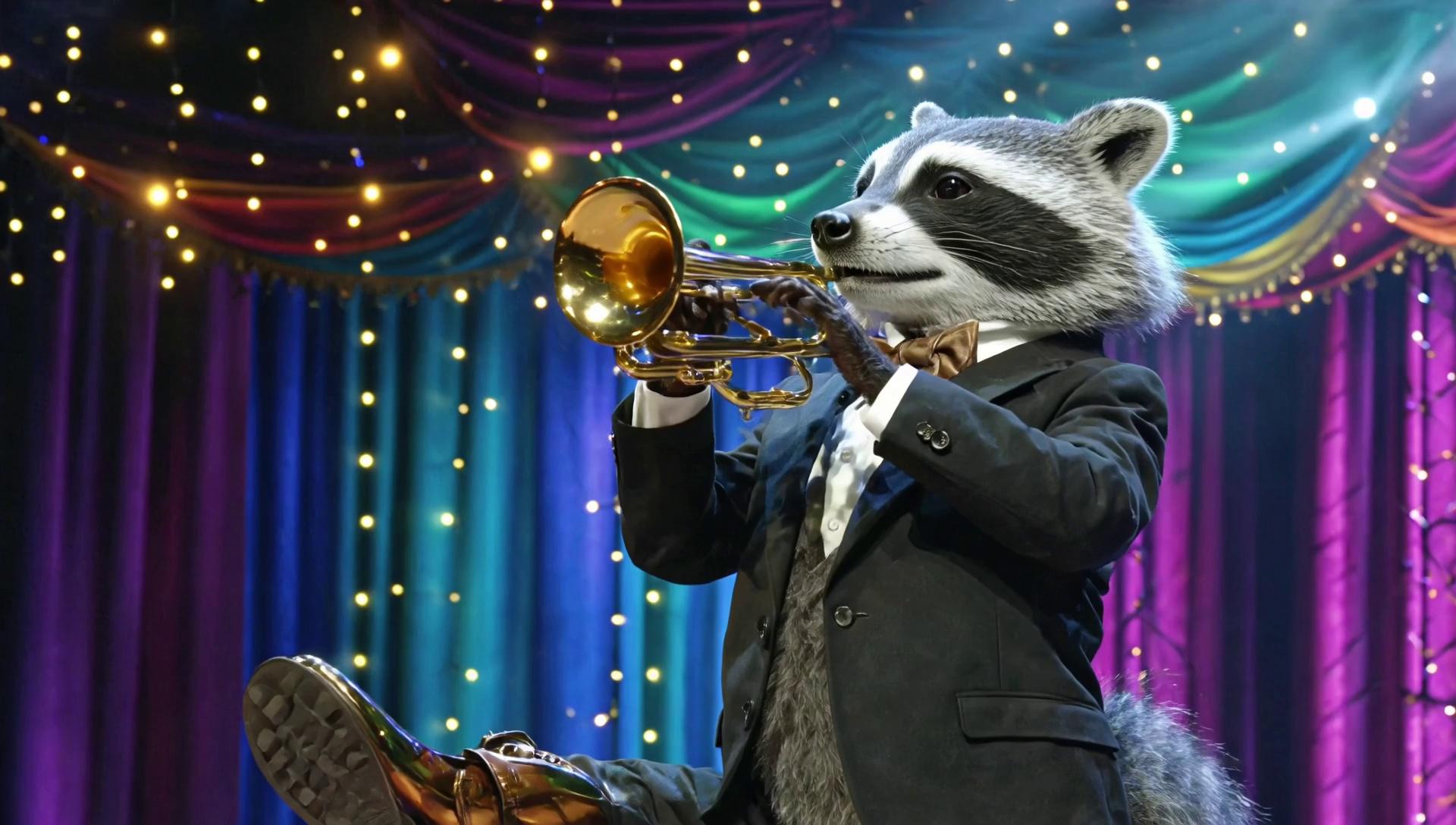} & 
  \includegraphics[width=\linewidth]{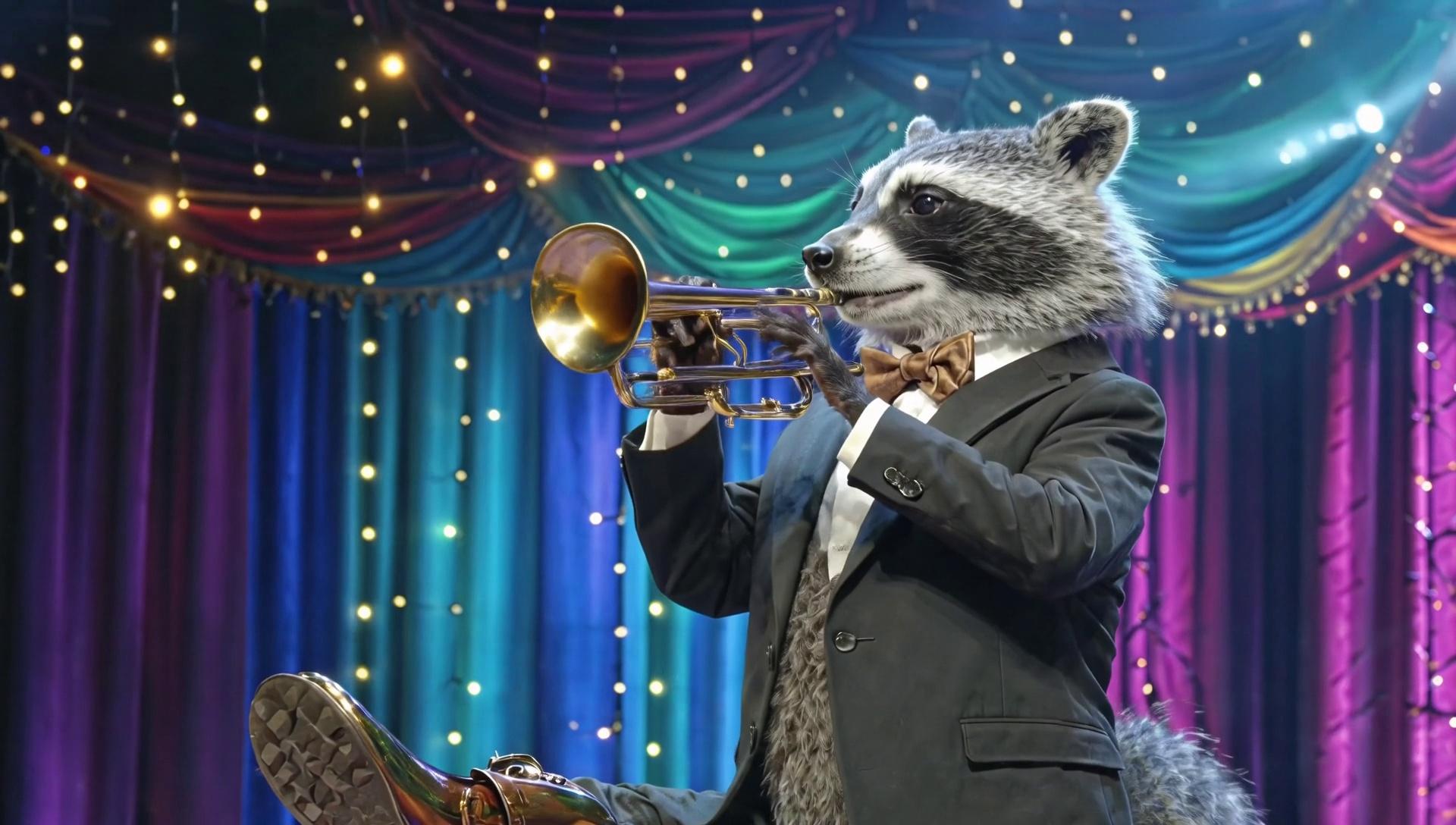} & 
  \includegraphics[width=\linewidth]{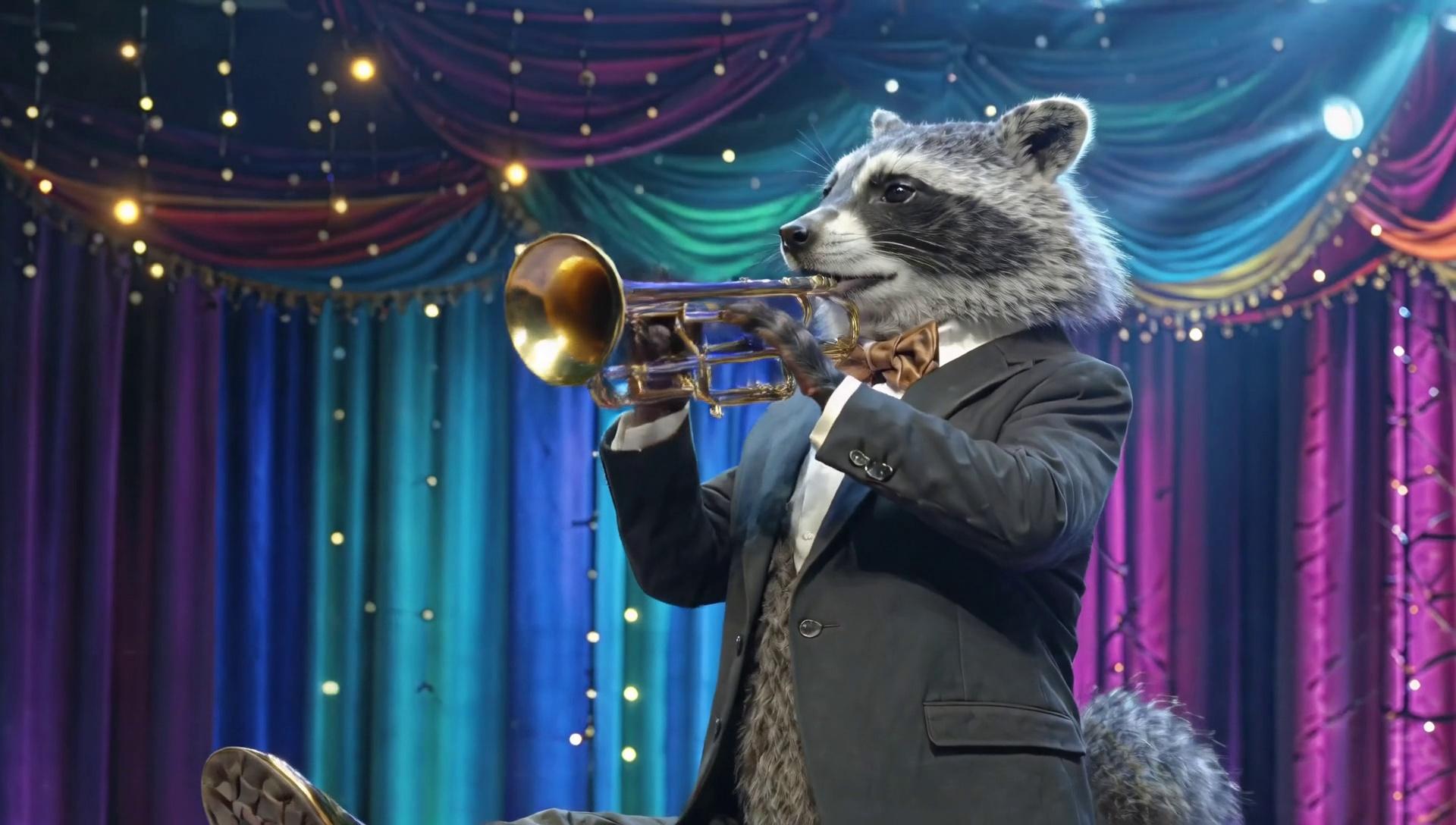} \\
  HiStream & 
  \includegraphics[width=\linewidth]{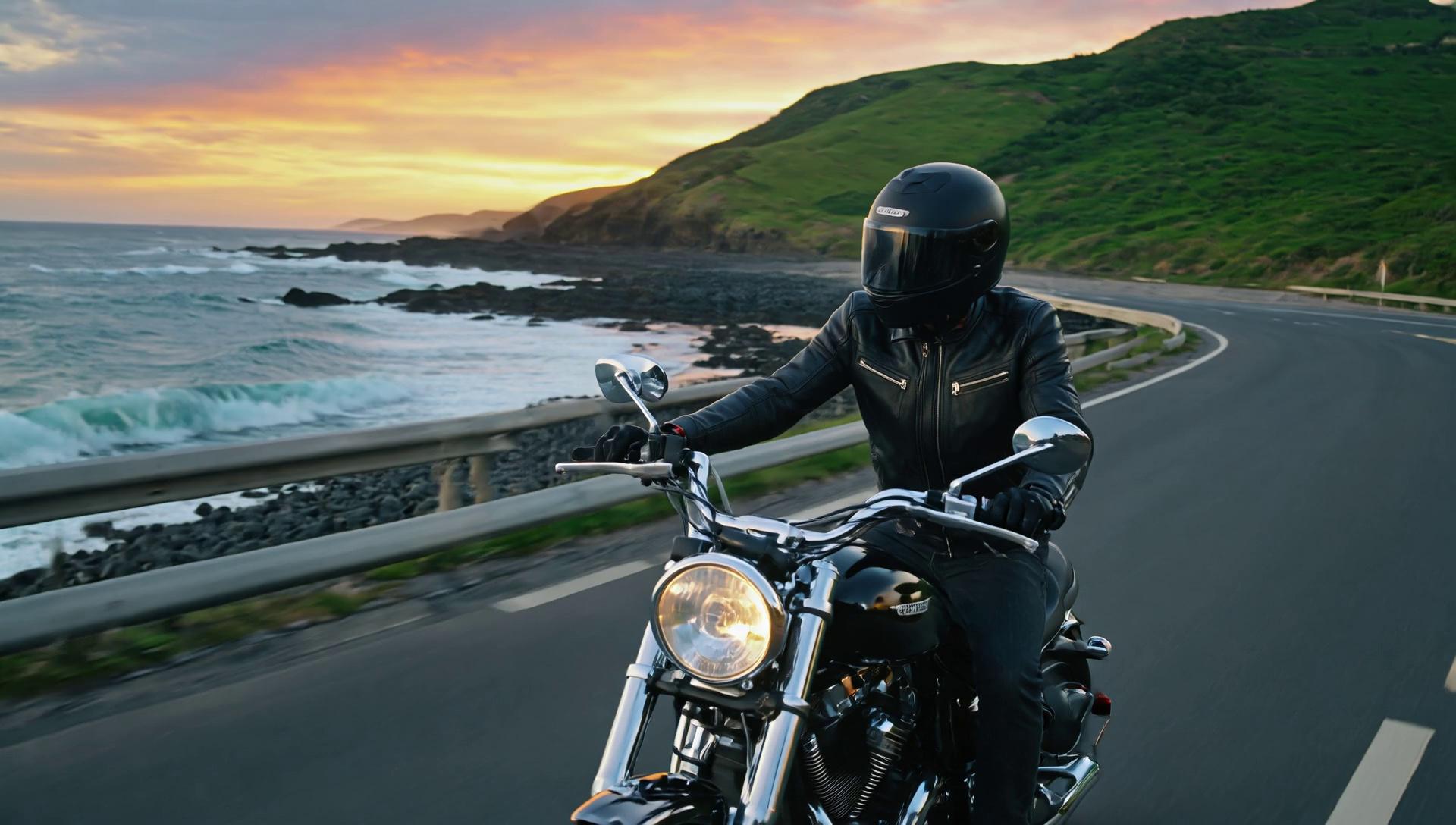} & 
  \includegraphics[width=\linewidth]{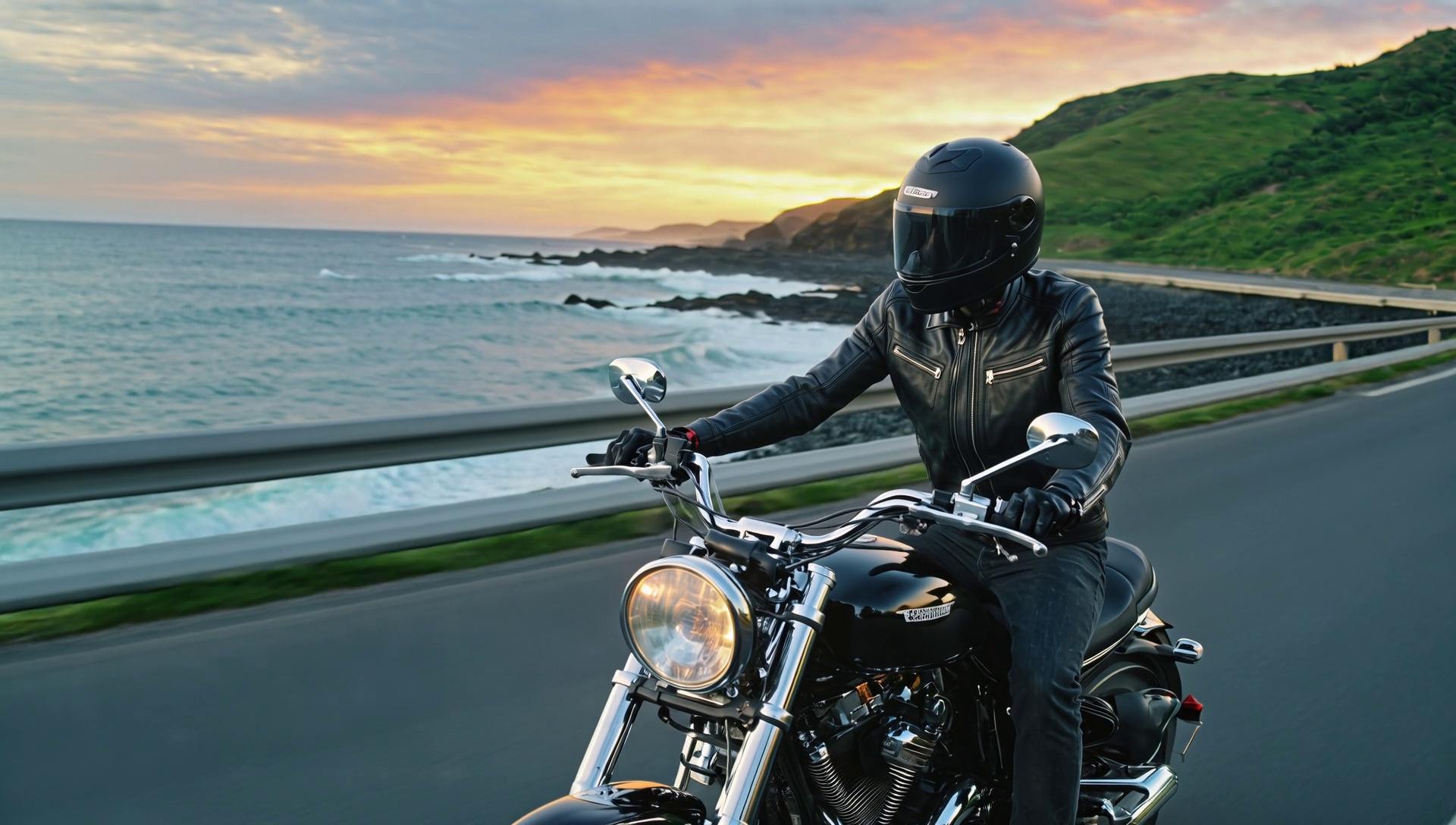} & 
  \includegraphics[width=\linewidth]{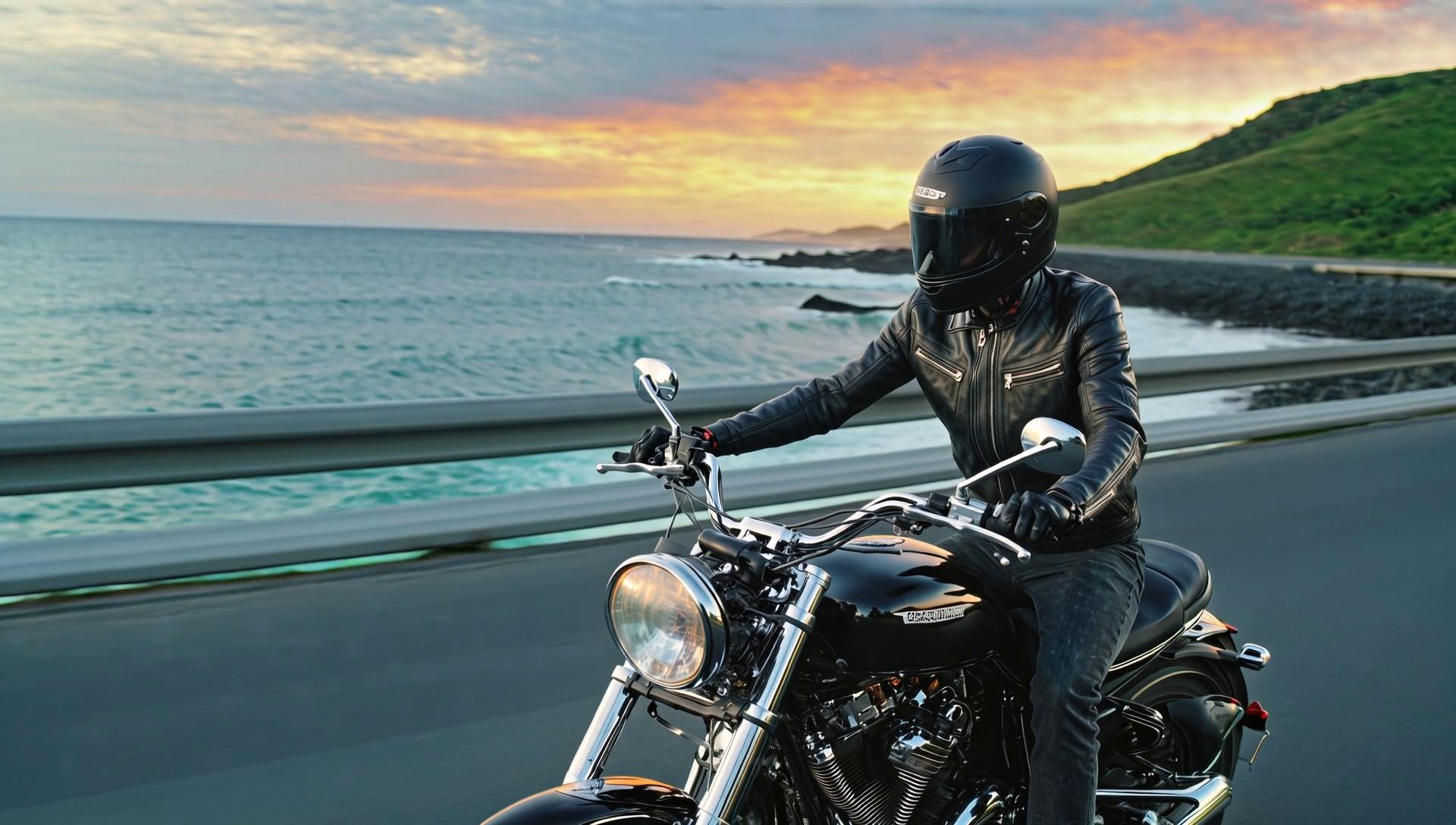} & 
  \includegraphics[width=\linewidth]{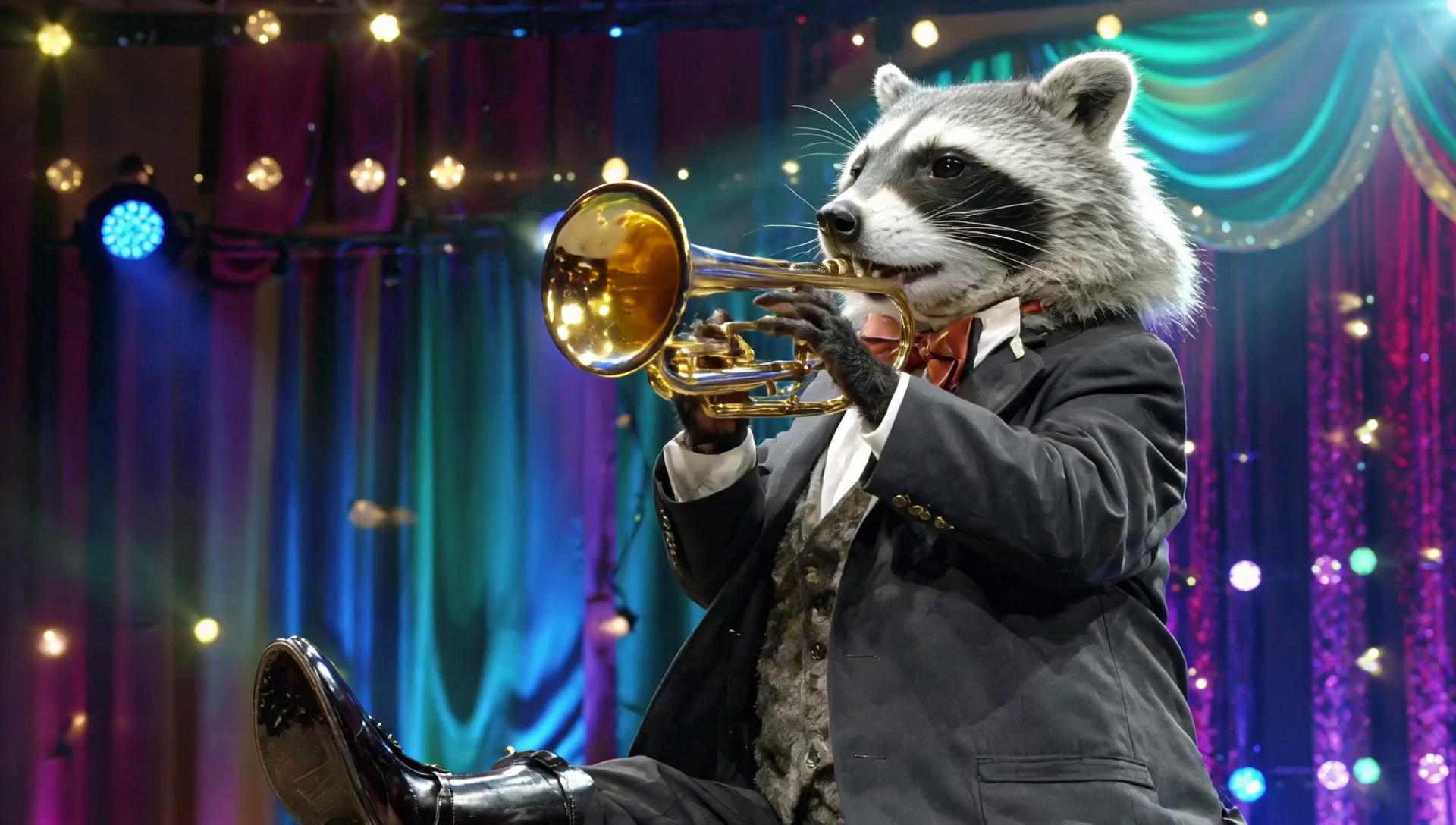} & 
  \includegraphics[width=\linewidth]{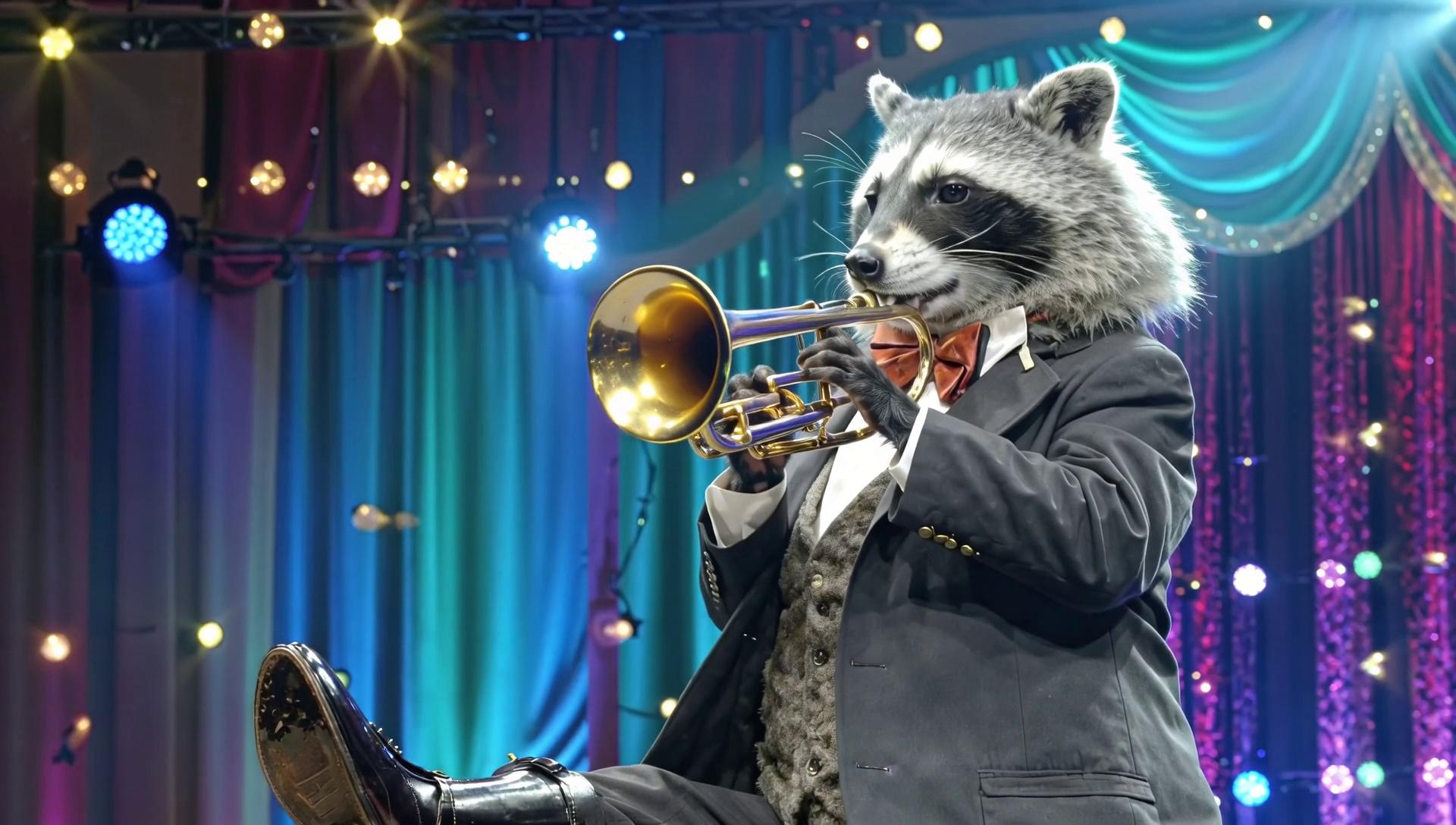} & 
  \includegraphics[width=\linewidth]{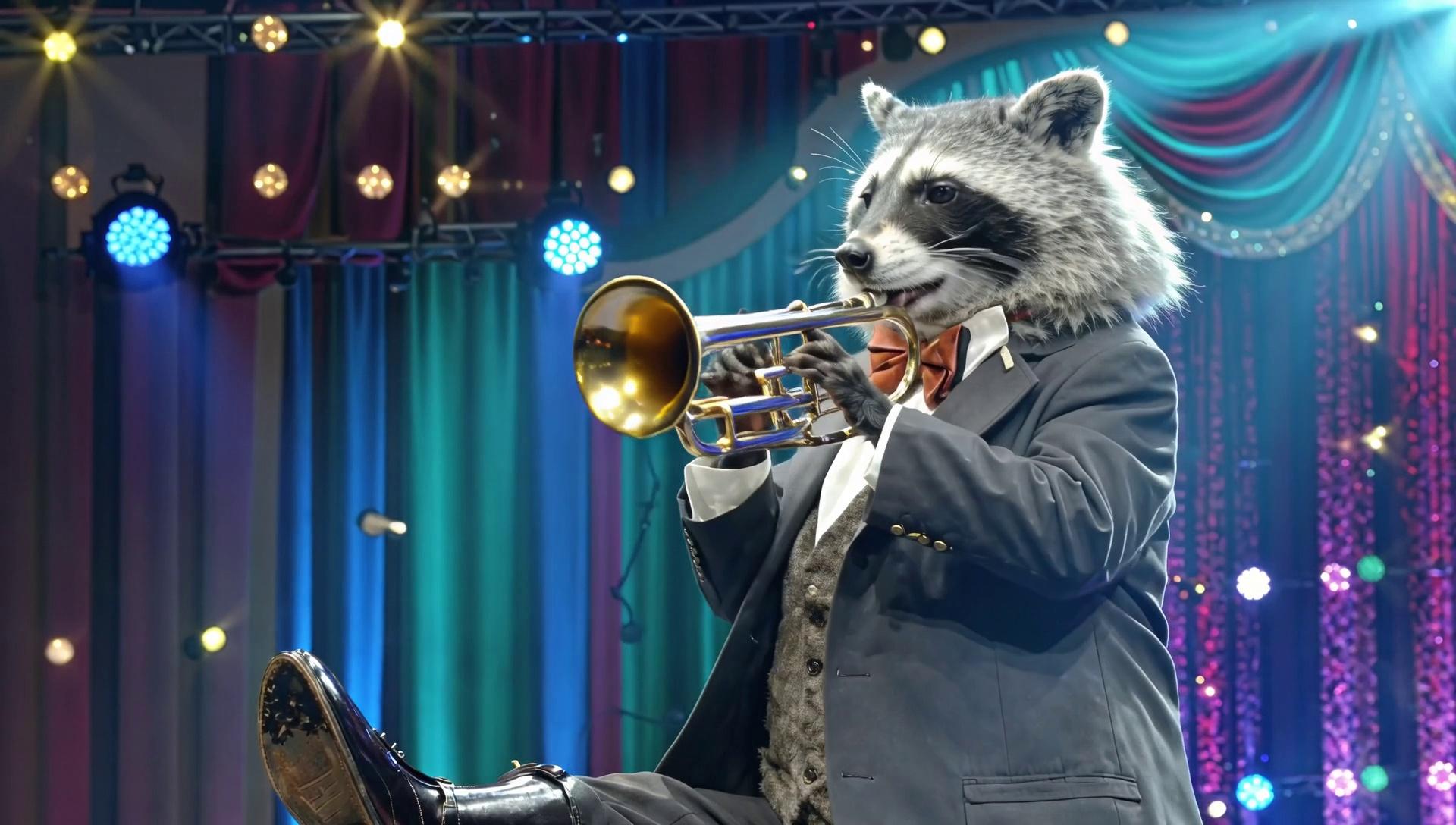} \\  
\end{tabular}
\caption{\textbf{Qualitative ablations for timesteps.} We perform controlled comparisons of HiStream with alternative variants in different timestep strategies.}
\vspace{-1.0em}
\label{fig:stepstudy}
\end{figure*}

\begin{table}[t]
\centering
% \vspace{-2mm}
\caption{\textbf{User study for Video Generation.} 
Users are required to pick the best one among our proposed HiStream and the other baseline methods in terms of video quality, semantic alignment, and detail fidelity.}
% \vspace{-2mm}
\label{tab:user}
\scalebox{0.99}{\begin{tabular}{@{}l|ccc@{}}
\toprule
 \textbf{Method}  & \makecell{Video \\ Quality}  & \makecell{Semantic \\ Alignment}  & \makecell{Detail \\ Fidelity} \\ \midrule
Wan2.1~\citep{wan2025}  &  2.78\%    &  5.95\%    &  3.17\%     \\  
Self Forcing~\citep{huang2025self}  &  10.71\%    &  7.54\%    &  8.73\%     \\ 
LTX~\citep{HaCohen2024LTXVideo}  &  0.79\%    &  1.59\%    &  0.79\%     \\ 
FlashVideo~\citep{zhang2025flashvideo}  &  12.30\%    &  14.68\%    &  11.51\%     \\ 
HiStream (Ours)  &  \textbf{73.41\%}    &  \textbf{70.24\%}    & \textbf{75.79\%}    \\ \bottomrule
\end{tabular}}
% \vspace{-2mm}
\end{table}

\subsection{Qualitative Ablations for Timesteps.}

We further investigate the impact of the denoising step count, comparing the robust 4-step setting against accelerated 2-step variants. As depicted in Figure~\ref{fig:stepstudy}, the \emph{uniform 2-step} approach suffers from catastrophic failure: the initial chunk exhibits severe blur and artifacts, manifesting as pronounced ghosting on the trumpet and faint details on the motorcycle’s windshield. In this autoregressive system, these early-stage errors propagate through the cache, critically degrading the temporal and visual fidelity of all subsequent chunks.

In stark contrast, \textbf{HiStream+} mitigates this failure by investing computation strategically. By dedicating the full 4 steps to the initial chunk, it establishes a robust, high-fidelity anchor cache. This high-quality initialization prevents error accumulation, resulting in minimal visual degradation across the entire video. This confirms that our asymmetric strategy is the superior acceleration approach, drastically reducing the computational load (effectively cutting steps in half for later chunks) with only a minimal and acceptable sacrifice in visual fidelity.

\begin{figure}[t!]
\centering
\includegraphics[width=0.99\linewidth]{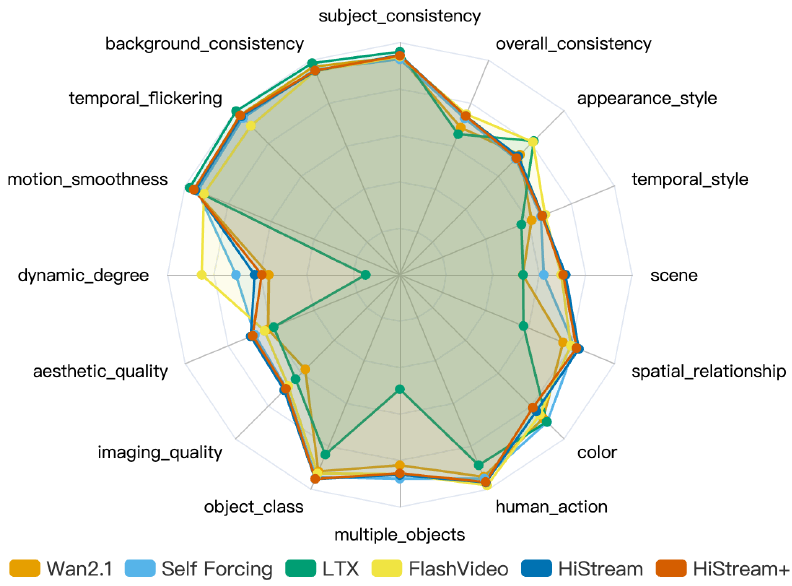}
% \vspace{-0.5em}
\caption{\textbf{VBench scores visualization.} We compare our two variants (HiStream and HiStream+) with Wan2.1~\citep{wan2025}, Self Forcing~\citep{huang2025self}, LTX~\citep{HaCohen2024LTXVideo}, and FlashVideo~\citep{zhang2025flashvideo} using all 16 VBench metrics.
}
\vspace{-1.0em}
\label{fig:vbench}
\end{figure}

\subsection{User Study}

In addition, we conducted a user study to assess the perceptual quality of our generated results. Participants were asked to view videos produced by all comparison methods, with each example presented in a randomized order to minimize potential bias. For each case, users selected the best result based on three evaluation criteria: video quality, semantic alignment, and detail fidelity. A total of $21$ participants took part in the study. As shown in Table~\ref{tab:user}, our method received the highest number of votes across all evaluation aspects, significantly outperforming the baseline approaches.

\subsection{VBench Scores Across All Dimensions}

Figure~\ref{fig:vbench} presents a comprehensive evaluation of our two variants (HiStream and HiStream+) against representative models using all 16 VBench metrics. Both variants generally outperform competitors in semantic alignment, achieving top scores in dimensions such as object class, spatial relationship, and scene. They also demonstrate robust frame-wise quality, scoring high in aesthetic quality and imaging quality.

\end{document}